\documentclass[twoside,11pt]{article}

\usepackage{graphicx, subcaption}
\usepackage[margin=1in, footnotesep=0.5in]{geometry}
\usepackage{makecell}
\usepackage[sort&compress, numbers]{natbib} 
\usepackage{amsfonts, amsmath, amssymb, amsthm}
\usepackage[shortlabels]{enumitem}
\usepackage{tabularx}
\usepackage{booktabs}
\usepackage{siunitx}
\usepackage{mathtools}
\usepackage{placeins}
\mathtoolsset{showonlyrefs}

\usepackage{xcolor}

\usepackage[final]{hyperref}
\definecolor{customdarkred}{RGB}{150,0,0}
\definecolor{customdarkgreen}{RGB}{0,150,0}
\definecolor{customdarkblue}{RGB}{0,0,150}
\hypersetup{colorlinks=true, linkcolor=customdarkred, citecolor=customdarkgreen, urlcolor=customdarkblue}
\usepackage{url}
\usepackage[linesnumbered, ruled]{algorithm2e}
\usepackage{authblk}

\def\K{\kappa}

\newcommand{\R}{\mathbb{R}}

\newtheorem{theorem}{Theorem}[section]

\theoremstyle{definition}

\theoremstyle{remark}
\newtheorem{remark}[theorem]{Remark}

\numberwithin{equation}{section}
\numberwithin{figure}{section}

\def\beq{\begin{equation}} % \setcounter{equation}{1}}
\def\eeq{\end{equation}}
\def\beqn{\begin{eqnarray*}}
\def\eeqn{\end{eqnarray*}}
\def\Bitem{\begin{itemize}\setlength{\itemsep}{.2in}}
\def\bitem{\begin{itemize}\setlength{\itemsep}{.05in}}
\def\eitem{\end{itemize}}
\def\Benum{\begin{enumerate}\setlength{\itemsep}{.2in}}
\def\benum{\begin{enumerate}\setlength{\itemsep}{.05in}}
\def\eenum{\end{enumerate}}
\def\bmult{\begin{multline*}}
\def\emult{\end{multline*}}
\def\bcenter{\begin{center}}
\def\ecenter{\end{center}}
\def\bframe{\begin{frame}}
\def\eframe{\end{frame}}

\newcommand{\secref}[1]{Section~\ref{sec:#1}}
\newcommand{\appref}[1]{Appendix~\ref{sec:#1}}
\newcommand{\figref}[1]{Figure~\ref{fig:#1}}
\newcommand{\tabref}[1]{Table~\ref{tab:#1}}

\def\cC{\mathcal{C}}

\def\cR{\mathcal{R}}

\def\bbR{\mathbb{R}}

\def\1{\mathbbm{1}}

\title{Cluster and then Embed: A Modular Approach for Visualization}
\author[1]{Elizabeth Coda} 
\author[1,2]{Ery Arias-Castro}
\author[2]{Gal Mishne} 
\affil[1]{\small Department of Mathematics, University of California, San Diego} 
\affil[2]{\small Halıcıoğlu Data Science Institute, University of California, San Diego}
\date{}

\begin{document}
\maketitle
\thispagestyle{empty}

\begin{abstract}
Dimensionality reduction methods such as t-SNE and UMAP are popular methods for visualizing data with a potential (latent) clustered structure. They  are known to group data points at the same time as they embed them, resulting in visualizations with well-separated clusters that preserve local information well. However, t-SNE and UMAP also tend to distort the global geometry of the underlying data. We propose a more transparent modular approach that first clusters the data, then embeds each cluster, and finally aligns the clusters to obtain a global embedding. We demonstrate this approach on several synthetic and real-world datasets and show that it is competitive with existing methods, while being much more transparent.
\end{abstract}

\section{Introduction}
\label{sec:intro}
Visualization is one of the most commonly used tools in exploratory data analysis (EDA)~\cite{de2025low}. However, when the data is high-dimensional, direct visualization of the data is not possible and dimensionality reduction techniques must be used to obtain a low-dimensional visualization.

In this paper, we assume the data are of the form $ \{ x_i \}_{i=1}^{n} \subset \bbR^d$. We seek an embedding $\{ y_i \}_{i=1}^{n} \subset \bbR^m$ with $m \leq d$ so that if a clustered structure is present in the data, the embedding uncovers the structure and provides visualization of that structure. Typically, $m=2$ and we will assume this is the case for the remainder of the paper, though results can easily be extended to $m > 2$. As stated, the problem is ill-defined as it is not clear what exactly is to be optimized.

The visualization of clustered data is certainly not a new problem. Early work includes that of \citet{shepard1972}, who proposed the practice of imposing the clusters obtained from a hierarchical clustering algorithm onto an embedding. Shepard explains that imposing hierarchical clustering onto the spatial representation provides more information than a hierarchical cluster tree alone (e.g., that two clusters are closer to one another than a third cluster). While his proposal is based on small datasets ($n = 16$ in his motivating example), the practice of imposing clustering on an embedding is very common. The visualization of single-cell RNA sequencing data in Bioinformatics is a case in point, as data of that sort are typically expected to be clustered~\citep{bohm2022}. Such data are often visualized by first producing an embedding using, e.g., t-SNE~\citep{van2008} or UMAP~\citep{mcinnes2018}, and the points are then colored according to known labels or the labeling of some clustering method, e.g., Louvain~\citep{blondel2008}, Leiden~\citep{traag2019}, or DBSCAN~\citep{ester1996}. Similarly, in machine learning in the task of classification, it is common to visualize high-dimensional data such as images or the hidden layers of a neural network with the embedded points colored according to the class label.

Many classical approaches for constructing an embedding seek to preserve some measure of dissimilarity between points, often the Euclidean distance. Examples include classical scaling (CS), also sometimes referred to as multidimensional scaling (MDS), which coincides with principal component analysis (PCA) when the dissimilarity measure is the Euclidean distance; and Isomap~\citep{tenenbaum2000}, which approximates the geodesic distances between points under the assumption that they lie on a smooth submanifold, and then embeds the points using CS with these distances. However, these methods for dimension reduction were not explicitly designed for the visualization of clustered data, and tend not to separate the clusters well. In particular, when embedding a configuration of points in $\bbR^d$ into $\bbR^2$ or $\bbR^3$ there may not be enough space in the low-dimensional embedding space to accommodate all points at a given distance scale. As a result of this ``crowding problem"~\citep{van2008}, the embedding may collapse distant points together and distinct clusters in the high-dimensional space may overlap in the visualization.

More recently, there has been a class of methods for nonlinear dimensionality reduction that seek to preserve local neighborhoods. Examples include Laplacian eigenmaps \cite{belkin2003}, as well as the highly popular t-SNE \cite{van2008} and UMAP \cite{mcinnes2018}.
Laplacian eigenmaps is presented as an algorithm that both clusters and embeds the data at the same time \cite{belkin2003}, as is clear through the strong connection between the method and spectral clustering. However, force-directed methods such as t-SNE and UMAP which aim to balance attractive and repulsive forces between points, are generally recognized as the most successful methods for the visualization of clustered data, in the sense that they tend to separate existing clusters rather well in the embedding.
\citet{bohm2022} demonstrates that all three of these methods are closely related, showing that varying the early exaggeration parameter in t-SNE yields a spectrum of embeddings that includes UMAP and Laplacian eigenmaps. In particular, when the t-SNE early exaggeration parameter is small, the resulting visualization contains a more discrete cluster structure and has higher $k$-nearest neighbor ($k$NN) recall, and when the parameter is larger, attractive forces are increased and the visualization emphasizes more connections between clusters. 
For another take, \citet{cai2022theoretical} show that during the initial early exaggeration stage of t-SNE, $\{y_i\}_{i=1}^n$ are clustered according to the cluster membership of $\{x_i\}_{i=1}^n$, and during the next stage, the embedding stage, each $y_i$ is initially updated in a manner determined by a repulsive force from each of the other clusters. \citet{arora2018} and \citet{linderman2019clustering} provide alternate conditions under which t-SNE will cluster well-separated data so that for each cluster in the original space, there exists a well-separated corresponding cluster in the t-SNE visualization. \citet{saidi2025embedor} also presents a variant of t-SNE with clustering guarantees under less stringent conditions.

However, while these theoretical guarantees on t-SNE lend credibility to it as a method that can separate clusters, we argue that a good visualization method should also preserve some geometric information, such as the relative positions, distances, and sizes of different clusters. Others have similarly advised that despite their popularity in Bioinformatics, t-SNE and UMAP should be used with caution, citing large distortions in distances and global structure \cite{chari2023}, as the relative positions and sizes of clusters do not necessarily reflect such information in the original space \cite{wattenberg2016}. Proper PCA initialization of these methods may help mitigate some of these issues \cite{kobak2019art, kobak2021initialization}, and more recently, \citet{kury2026} has even proposed a PCA-regularization for t-SNE to construct an embedding that balances global and local structure preservation. Empirically, however, there remains a trade-off between dimensionality reduction methods that: {\em (i)} preserve local structure and separate distinct clusters; and {\em (ii)} preserve global distances \cite{bohm2022, damrich2024}.

Related to our approach is a class of bottom-up manifold learning methods, which first construct low-dimensional, low-distortion embeddings of local neighborhoods of the data and then align these local embeddings to produce a single global embedding~\cite{Donoho2003-op,zhang2004principal, kohli2021, kohli2024rats}. However, these methods are not focused on clustered data, and indeed, their global alignment step can result in points from different clusters overlapping in the global embedding. 

\paragraph{Contribution}
In this paper, we propose a general framework that aims to strike a compromise between a faithful embedding at the cluster-level and a faithful global embedding of all the data that preserves distances. As explained in \cite{cai2022theoretical}, t-SNE can be viewed as a method that first clusters the data while embedding it and then aligns the clustered data. We propose an alternate method that consists of first clustering the data, then embedding each cluster individually, and finally finding rigid transformations to align the clusters in a global embedding so that there is meaningful organization of all clusters together. By embedding each cluster individually, we aim to produce an embedding that suffers from less distortion than if we were to embed all of the data together. Additionally, the method includes a tuning parameter to increase separation between clusters in a way that preserves relative distances between clusters. Our approach is modular in that the clustering and embedding methods may be chosen by the user based on what information they aim to obtain from the visualization.  We show that the method is competitive with existing methods, while offering more flexibility and transparency. 

\begin{remark}
So that there is no confusion, we mention of a very recent article \cite{peterfreund2025}, where the authors employ a partition-then-embed strategy (clearly stated in the title of the article), but they partition or group features (i.e., coordinates or dimensions) not observations, and produce an embedding based on each group of features, their intention being the visualization the same data from different `perspectives', as it were.
\end{remark}

\section{The {\sf Cluster+Embed} approach}

\begin{figure}[t]
\centering
\begin{tabular}{cccc}
\includegraphics[width=0.22\textwidth]{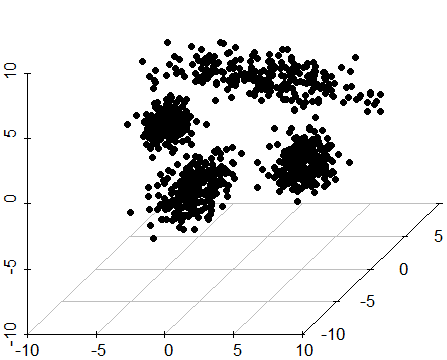} &
\includegraphics[width=0.22\textwidth]{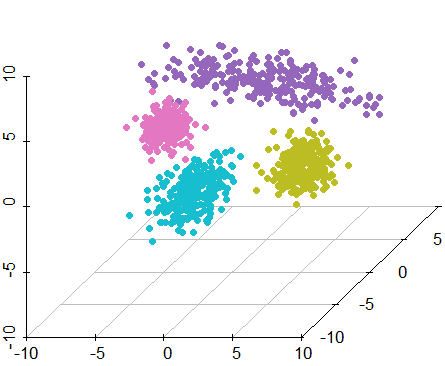} &
\includegraphics[width=0.18\textwidth]{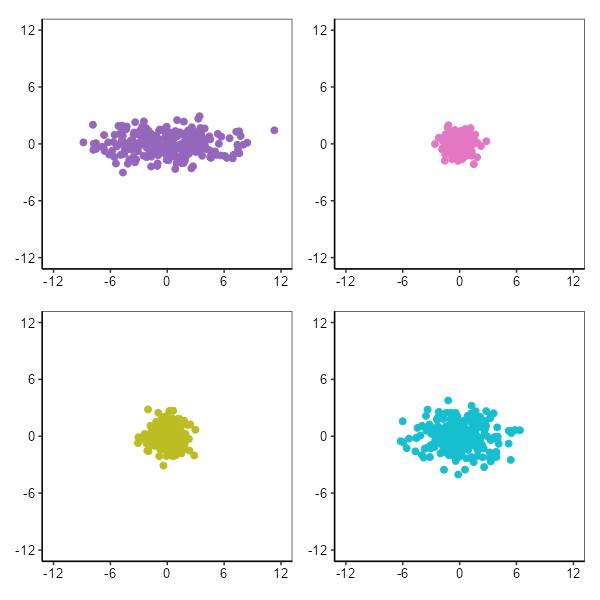} &
        \includegraphics[width=0.18\textwidth]{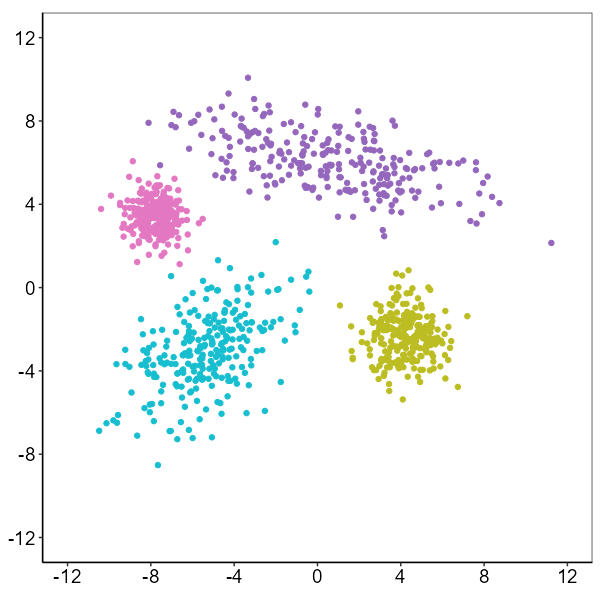} \\
Data & Step 1: Clustering & Step 2: Embed clusters & Step 3: Align clusters 
\end{tabular}
    \caption{Schematic diagram depicting the steps of the  {\sf Cluster+Embed} approach .}
    \label{fig:schematic}
\end{figure}

Among existing dimensionality reduction methods, there appears to be a trade-off between {\em (i)} preserving local structure and separate distinct clusters, and {\em (ii)} preserving global distances. Broadly stated, we study the general approach, which we simply refer to as {\em {\sf Cluster+Embed}} (abbreviated {\em C+E}), that consists in finding an embedding that highlights the cluster structure (when present) with the goal of striking a good balance between {\em (i)} and {\em (ii)} above. By first clustering the original data, we aim to preserve the cluster structure in the data, and by embedding each cluster individually, we aim to produce an embedding with less distortion than if we were to embed all of the data together. More concretely, we propose an approach consisting of three steps outlined in \figref{schematic}:
\begin{enumerate}
\item {\em Cluster the data.} This can be done using any clustering method the analyst chooses. The intention here is, of course, to reveal any cluster structure in the data. 
\item {\em Embed each cluster separately.} This can be done using any embedding method the analyst chooses. The embedding method is applied to each cluster of points identified in the previous step. The intention here is to obtain a high-quality embedding of each cluster as compared to an embedding of all the data together. 
\item {\em Align the embedded clusters.} The last step consists of aligning the embedded clusters using only rigid transformations. There are multiple ways of doing so and here too, the analyst has a choice. The intention is to obtain an embedding of the entire dataset that preserves the high-quality embeddings produced in the previous step while also reproducing, as faithfully as possible, the positioning of the clusters relative to each other. We propose the use of a tuning parameter that controls the separation between clusters, which can then be exaggerated to minimize overlap in order to better visualize the clusters. 
\end{enumerate}
The C+E approach is both modular and straightforward, allowing the user the flexibility to produce a visualization for their purpose while being much more transparent in terms of what is being preserved at the cluster level and at the global level compared to other methods such as t-SNE or UMAP that are known to produce clusters when embedding data.

\paragraph{Notation}
We take $\delta_{ij}$ as a measure of dissimilarity between $x_i$ and $x_j$. For example, $\delta_{ij}$ could be the Euclidean distance, $\| x_i - x_j \|$. The ``Swiss roll'' is an often used example of the importance of the choice of the dissimilarity measure, where the Euclidean distance between two points does not capture their intrinsic similarity well, and rather the geodesic distance would be a much more appropriate measure of dissimilarity \cite{tenenbaum2000, roweis2000}. Another example is the MNIST dataset of handwritten digits in $\bbR^{28 \times 28}$, where Euclidean distance is strongly affected by pixel position (e.g., shifting a digit to the left would result in an image very dissimilar from the original image) \cite{lecun2002gradient}. The C+E approach can be applied with any choice of dissimilarity measure. 

\subsection{Step 1: Cluster the data}
For the first step, any clustering method can be used to obtain $\K$ clusters $\cC_1, \cdots, \cC_\K$ which partition $[n] = \{1, \cdots, n\}$. In Biology, the Louvain and Leiden algorithms are popular choices that work well in practice and can efficiently handle large single-cell RNA sequencing datasets \cite{blondel2008, traag2019}. DBSCAN and spectral clustering do not make strong assumptions on the clusters and are also effective on a wide variety of datasets \cite{ester1996, vonluxburg2007, ng2001}. Some methods such as DBSCAN label some points as ``noise" not belonging to any cluster. These points could be discarded at this step and excluded from the visualization. 

t-SNE (described in \appref{t-SNE} for the reader's convenience) is known to produce an embedding in which the data is clustered. \citet{cai2022theoretical} provide some theoretical explanation, establishing that during the early exaggeration stage, t-SNE is approximately performing spectral clustering to both create an embedding and cluster the data. More specifically, during the early exaggeration stage, t-SNE converges to the projection of the initial state onto the null space of the Laplacian of the adjacency matrix with entries $p_{ij}$ \eqref{eq:p_matrix}. If the data is strongly clustered, this representation separates the data into tight clusters. Points in the same cluster will be close to one another, but in this representation the distance between points in the same cluster is not necessarily meaningful, and with random initialization, the relative positions of different clusters is meaningless. Moreover, if the data is only weakly clustered, early stopping of this stage becomes critical to avoid the formation of false clusters. In contrast, compared to t-SNE, the C+E approach is more explicit and also allows for more direct control of how the clustering is performed. 

In \appref{sensitivity} we show how the clustering quality (e.g., over- or under-clustering) affects the final embedding.

% For example, \citet{shepard1972} uses some form of hierarchical clustering. 

\subsection{Step 2: Embed each cluster separately}

For the second step, any embedding method can be used to separately embed each of the clusters obtained in Step~1. This gives us an embedding $\{y_i\}_{i=1}^n$. In general, any method will incur some distortion when embedding high-dimensional data into two dimensions, but by embedding each cluster individually we aim to produce an embedding that suffers from less distortion than if we were to embed all of the data together. An added advantage is that, when applying a separation between clusters as described in Step~3, we can more intentionally minimize or even avoid overlap of different clusters. 

The choice of embedding method can be guided by what the practitioner hopes to gain from the visualization at the cluster level. If distance preservation is desired, PCA or Isomap may be an appropriate choice. Landmark methods could be used to reduce the computation time if clusters contain a large number of points \cite{silva2002global, de2004sparse}. If preservation of local structure is important, a method that directly aims to embed a local neighborhood graph of the cluster could be used. For example, if the user wants to optimize for $k$NN recall for a particular $k$, a method for embedding nearest neighbor graphs could be used. Methods based on triplet comparisons (e.g. point $i$ is closer to point $j$ than point $l$) are outlined in \citet{vankadara2023insights}. In practice, we found that the TriMap algorithm \citep{amid2019trimap} is both effective and efficient at handling large clusters.

Note that if the embedding method is based only on ordinal information, synchronization of the scales of the cluster embeddings is necessary before the next step. To do this, we propose to scale all $y_i \in \mathcal{C}_l$ by
\[\gamma_l = \frac{\sum_{i,j \in \mathcal{C}_l} \delta_{ij} \|y_i - y_j\|}{\sum_{i,j \in \mathcal{C}_l} \|y_i - y_j\|^2},\]
where $\gamma_l$ comes from minimizing the quadratic loss between the embedded distances and original dissimilarities 
\[\sum\limits_{i,j \in \mathcal{C}_l}(\gamma_l\|y_j-y_j\| - \delta_{ij})^2.\]

\subsection{Step 3: Align the embedded clusters}
As the final step, we globally align the cluster embeddings using only rigid transformations (translations, reflections, and rotations) so as to position the clusters relative to each other with consideration for their relative positions before embedding, while strictly preserving the pairwise distances within each cluster-level embedding. Our intention is to find some rigid transformations $T_1, \dots, T_\K$  such that 
       \begin{equation}
       \label{eq:global_alignment}
           \|T_i(y_l) - T_j(y_m) \| \approx \delta_{lm} \quad \text{for all (or most) $l\in \cC_i, m \in \cC_j$, for all pairs $1 \le i < j \le \K$}.
       \end{equation}
That is, we seek rigid transformations 
that best preserve the pairwise dissimilarities between points in different clusters without updating the cluster-level embeddings. To formalize this, we use a definition of stress, for concreteness Kruskal's definition \cite{kruskal1964a,kruskal1964b} chosen among others \cite{takane1977, heiser1988, mcgee1966, sammon2006}, and write \eqref{eq:global_alignment} as an optimization problem 
\begin{gather}
\label{eq:stress}
   \underset{T_1, \cdots T_\K \in \cR} {\text{minimize}} \quad \mathop{\sum\sum}_{1 \le i < j \le \K}\ \sum\limits_{l \in \cC_i}\, \sum\limits_{m \in \cC_j} \big(\delta_{lm} - \|T_i(y_l) - T_j(y_m)\|\big)^2, %\\
    %\text{over}\quad T_1, \cdots T_\K \in \cR,
\end{gather}
where $\cR$ is the class of rigid transformations on $\bbR^2$. 
In this way, we attempt to enforce preservation of distances between points in different clusters, with the constraint that the distances between embedded points in the same cluster cannot be updated to do so. 

\subsubsection{Scaling to avoid cluster overlap}       
Note that the objective \eqref{eq:stress} does not preclude the possibility of overlap between the embedded clusters, and overlapping clusters may be undesirable when visualization is the end goal. We address this issue by effectively forcing a separation between clusters. The basic idea is to introduce a scaling in \eqref{eq:global_alignment} controlled by a tuning parameter $\alpha \geq 1$, specifically 
       \begin{equation}
       \label{eq:global_alignment_scaled}
           \|T_i(y_l) - T_j(y_m) \| \approx \alpha\, \delta_{lm} \quad \text{for all (or most) $l\in \cC_i, m \in \cC_j$, for all pairs $1 \le i < j \le \K$}.
       \end{equation}
This uniformly increases the distances between points in different clusters, which encourages separation between the embedded clusters and avoids the crowding problem (see \secref{crowding}). 
In this case we, the optimization problem \eqref{eq:stress} becomes 
\begin{gather}
\label{eq:scaled_stress}
    \underset{T_1, \cdots T_\K \in \cR} {\text{minimize}} \quad \mathop{\sum\sum}_{1 \le i < j \le \K}\ \sum\limits_{l \in \cC_i}\, \sum\limits_{m \in \cC_j} \big(\alpha\, \delta_{lm} - \|T_i(y_l) - T_j(y_m)\|\big)^2. %\\
    %\text{over}\quad T_1, \cdots T_\K \in \cR,
\end{gather}

Numerically, to search over rigid transformations, we use the fact that any rigid transformation of $\R^2$ is either a rotation followed by translation or a reflection followed by a translation. Thus, we parameterize $T_i$ by $(\theta_i, \pi_i, v_i) \in [0, 2\pi) \times \{0,1\} \times \R^2$, where
\begin{equation}
\label{eq:rigid_transform}
T_i(x) =  \begin{bmatrix}
1 & 0 \\
0 & (-1)^{\pi_i}
\end{bmatrix} \begin{bmatrix}
\cos\theta_i & \sin\theta_i \\
-\sin\theta_i & \cos\theta_i
\end{bmatrix} x + v_i.
\end{equation}
Then the optimization in \eqref{eq:scaled_stress} is carried out over $\{ (\theta_i, \pi_i, v_i)\}_{i=1}^\K$. After some initialization, we adopt an alternate minimization going over the $\K \times 3$ parameters in turn, each time with the intention of minimizing the stress \eqref{eq:scaled_stress}: minimization over $\theta$ is done by grid search over $[0, 2\pi)$; minimization over $\pi$ is exhaustive and trivial as it only takes two values; and minimization over $v$ is done with the use of the BFGS algorithm. In our implementation, we start by embedding the largest cluster, which is then held fixed, serving as a reference while the other clusters are iteratively positioned. 

Compared to t-SNE, which uses attractive and repulsive forces to produce a global embedding that effectively aligns the clusters (this is done simultaneously in t-SNE), here too the C+E approach is more transparent and more modular, as the analyst is free to align the clusters as they wish. Our choice is based on strictly preserving the cluster embeddings obtained in Step~2.

\subsubsection{The crowding problem}
\label{sec:crowding}

When embedding high-dimensional data, it is possible there will not be enough space in the low-dimensional embedding space to accommodate all points at a given distance scale. This is referred to as ``the crowding problem” by \citet{van2008}. 

\begin{figure}[t]
\centering
\begin{tabular}{cccc}
\includegraphics[width=0.225\textwidth]{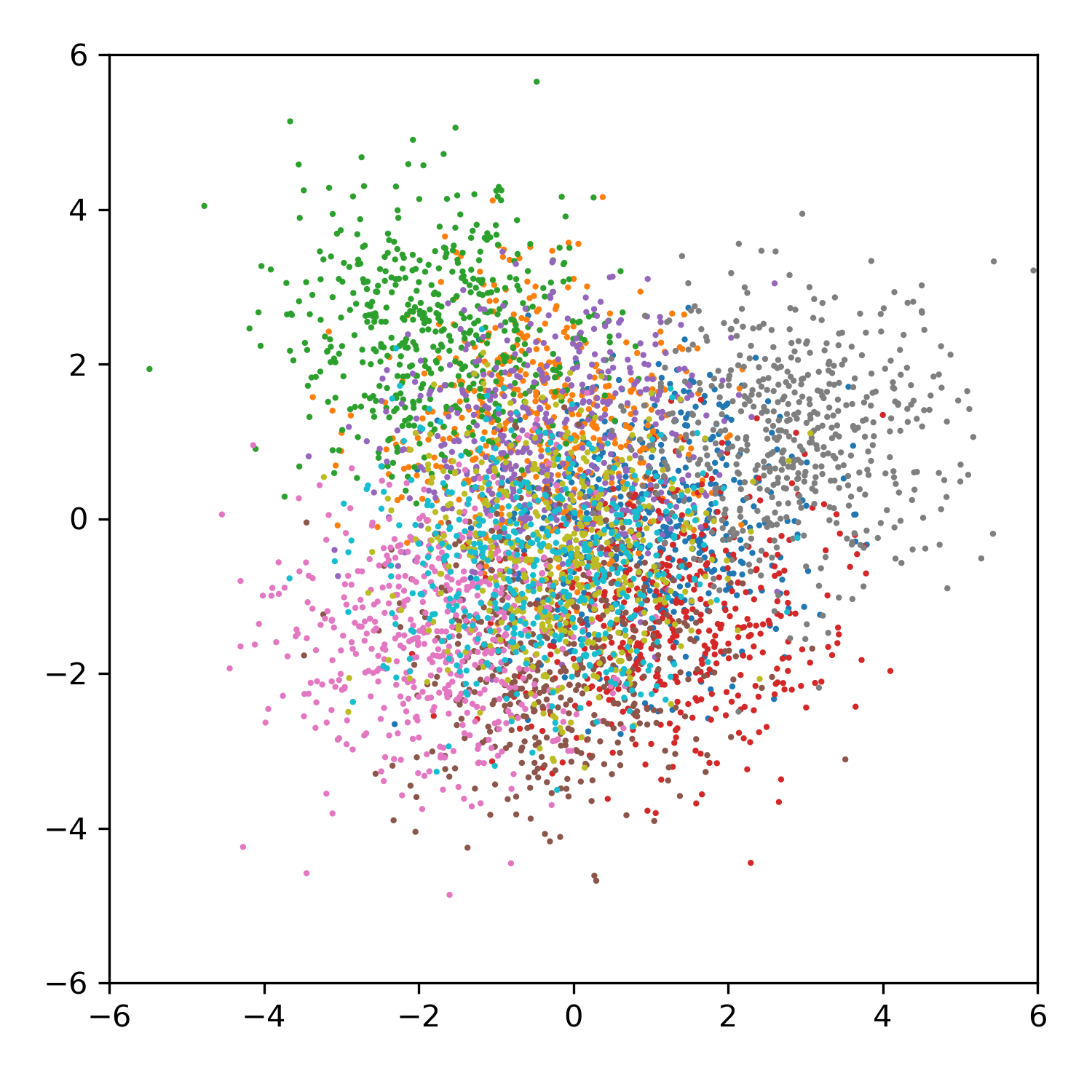} &
\includegraphics[width=0.225\textwidth]{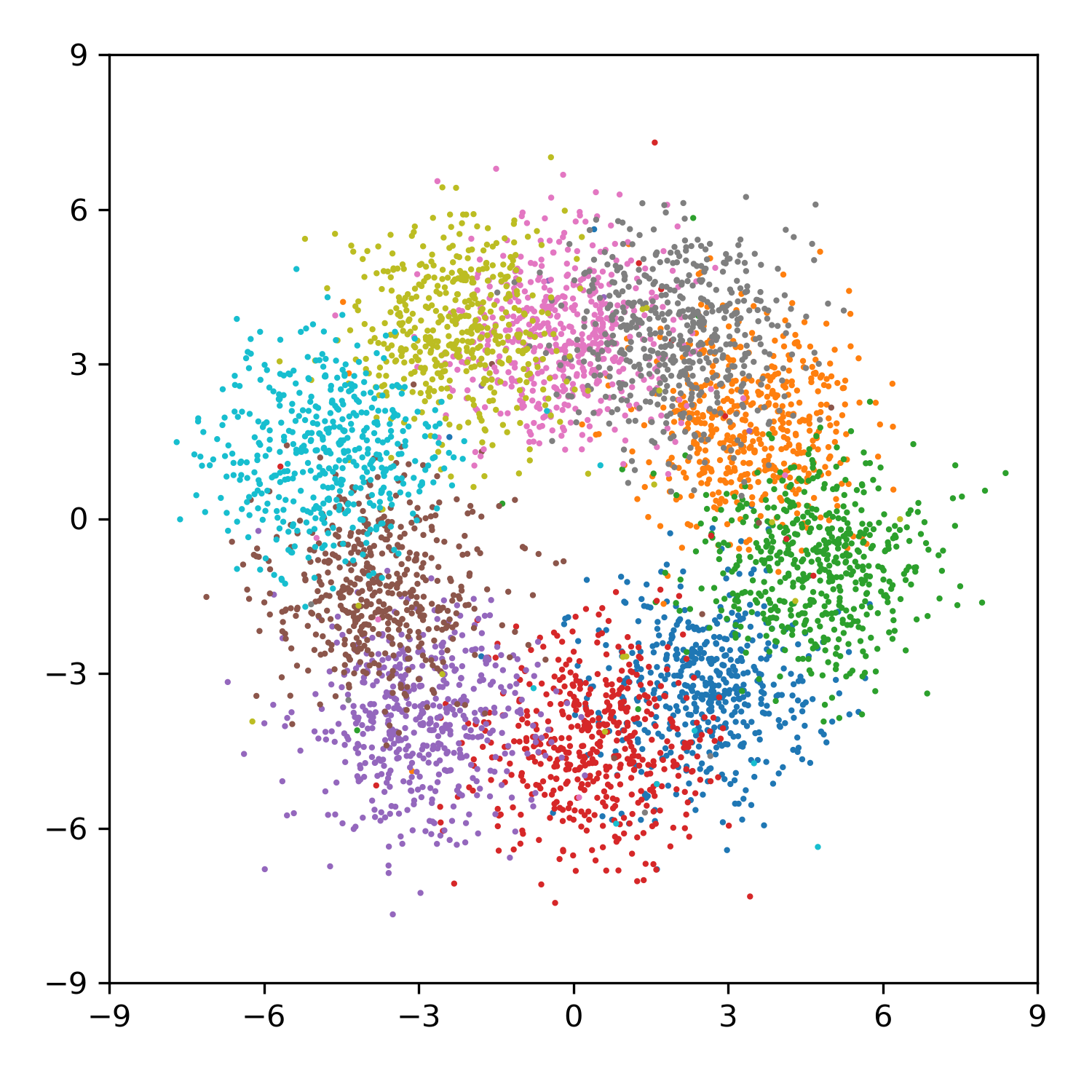} &
        \includegraphics[width=0.225\textwidth]{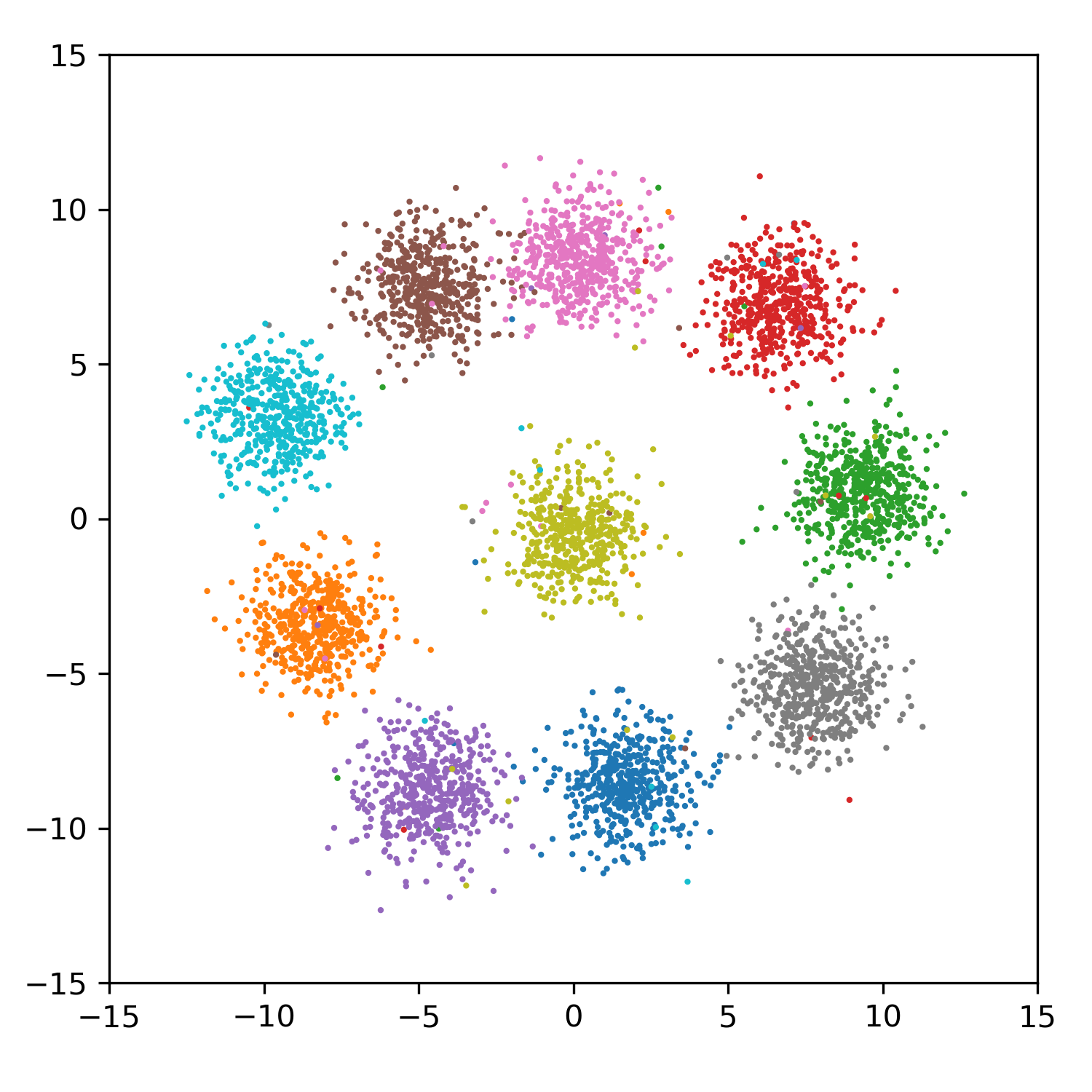} &
        \includegraphics[width=0.225\textwidth]{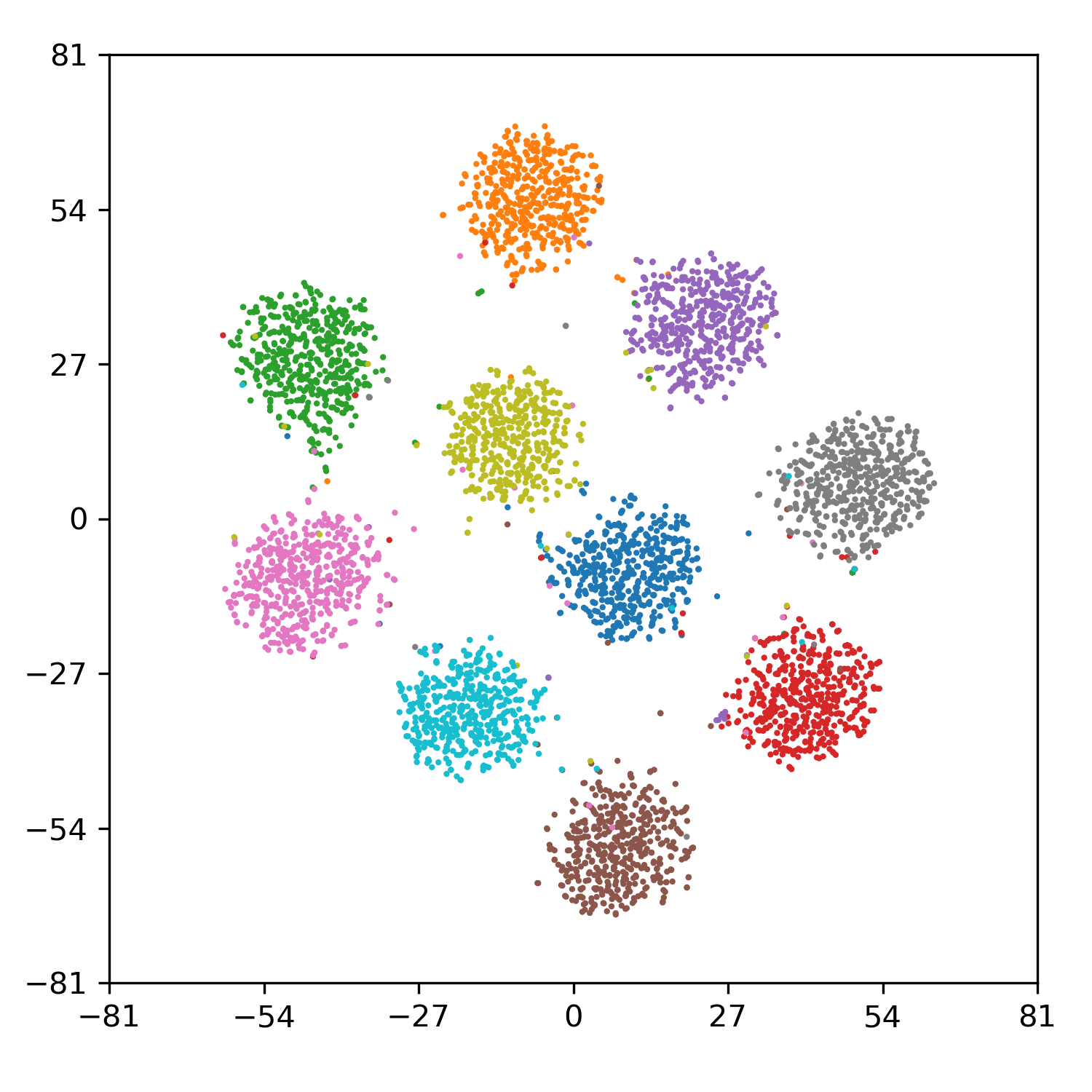} \\
PCA & C+E ($\alpha=1$) & C+E ($\alpha=2$) & t-SNE ($u = 30$)   
\end{tabular}
    \caption{Data generated from a $10$-dimensional isotropic Gaussian mixture model ($n = 5,000$) with equal distance between every pair of clusters. Points are colored by the mixture component they were generated from. These labels were not used by the embedding methods. When $\alpha = 1$, C+E produces an embedding with overlapping clusters, whereas $\alpha = 2$ scales the distances enough to avoid the crowding problem.
    }
    \label{fig:crowding_problem}
\end{figure}

We illustrate this with an example in \figref{crowding_problem} where we consider a $d$-dimensional isotropic Gaussian mixture model (GMM) with $d$ components where the means of each component have equal pairwise separation and large enough pairwise separation so that the components are well-separated clusters. We take $d=10$ in this example. At the far left of \figref{crowding_problem}, we observe that PCA on this dataset gives a high amount of overlap between the clusters. Second from left is the result of our method, with clustering via $k$-means, embedding each cluster using PCA, and then alignment to optimize \eqref{eq:stress} (or \eqref{eq:scaled_stress} with $\alpha = 1$). We observe a moderate amount of overlap between the clusters, which form a ring of radius $r \approx 4.40$.  However, in a ring of radius $r$, there is roughly only room for at most  $2\pi r/\tau$ clusters without overlap, where $\tau$ is the average diameter of an embedded cluster. In this example, we find  $\tau\approx 7.32$, so that $2\pi r/ \tau \approx 4$, and indeed we observe that the clusters overlap.

Third from the left is the result of our method with $\alpha = 2$, which is seen in this example to be enough scaling to avoid almost any overlap. For reference, we also provide on the far right the embedding produced by t-SNE. 
Qualitatively, both methods produce similar embeddings, with clusters in a ring around other clusters. We also note that the relative positioning of clusters in this example is not particularly meaningful, as in the underlying model all clusters have the same distance from one another. Despite the similarity of the embeddings of our method with $\alpha=2$ and t-SNE, the crowding problem is addressed differently, and more opaquely by t-SNE, which is based on using the Cauchy kernel, instead of the Gaussian kernel, to model the similarity between points in the embedding. The justification given by  \citet{van2008} for using the Cauchy kernel is that the heavy tails of this kernel ``allows a moderate distance in the high-dimensional space to be faithfully modeled by a much larger distance in the map''. See \appref{additional_metrics} for a more quantitative comparison between our method and t-SNE.

In general, a value of $\alpha$ can be selected in an automatic or data-driven manner by optimizing (over $\alpha$) some criterion of interest, such as the $k$NN recall for a particular value of $k$. However, this tuning can be computationally expensive and we also suggest an alternative choice which does not require any optimization over $\alpha$. 
The basic idea is that we wish to accommodate $2\pi r/\tau > \K$ clusters within a distance $r$ from a fixed cluster. Setting $r = \alpha \Delta$, where $\Delta$ is the average distance between two different clusters defined as
\begin{gather}
\label{eq:cluster_distance}
    \Delta = \frac{1}{\K\cdot(\K-1)} \mathop{\sum\sum}_{1 \le i < j \le \K} \Delta_{ij}, \qquad \text{with } \Delta_{ij} = \frac{1}{|\cC_i|} \frac{1}{|\cC_j|} \sum\limits_{l \in \cC_i}\sum\limits_{m \in \cC_j} \delta_{lm},
\end{gather}
we solve for $\alpha$ to obtain
\begin{gather}
    \label{eq:alpha}
    \alpha = \max\left(1, \frac{ \K \tau }{2 \pi \Delta} \right).
\end{gather}
This yields $\alpha = 1.65$ in the GMM example of \figref{crowding_problem}. Again, see \appref{additional_metrics} for a quantitative performance evaluation of this choice of $\alpha$. We also show in \appref{sensitivity} the final embedding obtained for a range of $\alpha$ values on real-world datasets of interest.

\subsubsection{Runtime}
\label{sec:runtime_main}
The computational complexity of the alignment step scales linearly with the number of clusters and quadratically with the number of data points, which limits its practicality to real use cases. To overcome this limitation, we use subsampling to evaluate \eqref{eq:scaled_stress}\textemdash and only consider a random fraction of points from each cluster at each iteration for alignment. This strategy is also compatible with an incomplete dissimilarity matrix of the original data, as the subsampling can be restricted to the pairs of points for which complete dissimilarity information is available. In all of our experiments, we randomly sample $50$ points from each cluster at each iteration to evaluate \eqref{eq:scaled_stress}. All runtimes are reported in \appref{runtime}.

\section{Visualizing Meaningful Clusters}
\label{sec:meaningful_clusters}

In this section, we demonstrate our method, {\sf Cluster+Embed} (C+E), on a variety of datasets. We include baseline methods of t-SNE, UMAP, PCA, Isomap, TriMap, and PHATE \cite{moon2019visualizing}, a popular method for visualization of single-cell data. For large datasets,  ($n \geq 20k$), we use Landmark Isomap (L-Isomap) in place of Isomap \cite{silva2002global}. TriMap parameters are set to $n_\text{inliers} = 5, n_\text{outliers} = 10 , n_\text{random} = 10 $ throughout for better $k$NN recall for small values of $k$. For all other methods, we use the default parameters unless otherwise noted, which are also reported in the performance tables. 

For our simplest examples, we set the dissimilarities used by C+E to $\delta_{ij} = \|x_i-x_j\|$ though for some examples where we do not expect the Euclidean distance to capture dissimilarity well, we approximate the geodesic distance using shortest paths in $k$NN graph as in Isomap~\cite{tenenbaum2000}. 
In particular, for computational efficiency, we only compute the pairwise dissimilarities for $5k$ points, using a subgraph of the full $k$NN graph ($q=15$) with $20k$ vertices. The subsampling used during alignment (described in \secref{runtime_main}) is then restricted to these $5k$ points for which complete dissimilarity information is available. 
% In \appref{additional_metrics}, we also include C+E results using Euclidean distances for the datasets using geodesic distances in the main text.

We include metrics to evaluate  preservation of local, cluster-level structure in terms of neighborhood recall and distance preservation, as well as metrics to evaluate preservation of global structure in terms of distance preservation and the relative positioning of clusters; see \tabref{metrics_def} for details. Complete metrics are in \appref{additional_metrics}. Experiments depicting the sensitivity of the C+E to the clustering step and the parameter $\alpha$ are included in \appref{sensitivity}. Python code to reproduce all figures and results is available at \url{https://github.com/lizzycoda/ClusterEmbed_v2}.

\begin{table}[htbp!]
    \centering
    \begin{tabularx}{\textwidth}{p{4.5cm}X}
        \toprule
        \textbf{Performance metric} & \textbf{Description} \\
        \midrule
        $k$NN recall & The fraction of $k$-nearest neighbors ($k$NN) of a point in the original space that are also $k$NN in the embedding. We report the average $k$NN recall over all points for $k=30$ in the main text, and over a range of $k$ values in \appref{additional_metrics}. 
%        $$\frac{1}{n}\sum\limits_{i=1}^n \frac{1}{|k|} |\cN_i \cap k\text{NN}(y_i)| $$
        \\ \midrule
 \parbox[t]{4.5cm}{Spearman correlation} &  Spearman correlation of the pairwise dissimilarities in the original space and the pairwise distances in the embedding. We report the Spearman correlation considering all points, as well as the average for within class and between class pairs of points.
 \\ \midrule
 \parbox[t]{4.5cm}{Normalized stress} & Given the aligned embeddings $\{y_i\}_{i=1}^n$, Kruskal's \cite{kruskal1964a,kruskal1964b} normalized stress is  $$
\left( \frac{\sum_{(i,j)} (\delta_{ij} - \|y_i - y_j\|)^2}{\sum_{(i,j)} \delta_{ij}^2} \right)^{1/2}.
$$ We report the normalized stress over all pairs of points, as well as the average for within class and between class pairs of points. \\ \midrule
%$$ \left( \sum\limits_{i\neq j}(\delta_{ij} - \|y_i-y_j\|)^2 /\sum\limits_{i\neq j}\delta_{ij}^2 \right) ^{1/2} $$ \\

\parbox[t]{4.5cm}{Scale-normalized stress} & Given the aligned embeddings $\{y_i\}_{i=1}^n$, the scale-normalized stress \cite{smelser2024normalized} first rescales the embeddings so that the normalized stress is minimized. It is  $$ \min\limits_{\beta}
\left( \frac{\sum_{(i,j)} (\delta_{ij} - \beta \|y_i - y_j\|)^2}{\sum_{(i,j)} \delta_{ij}^2} \right)^{1/2}.
$$ 
We report the scale-normalized stress over all pairs of points, as well as the average for within class and between class pairs of points. \\ \midrule

  \parbox[t]{4.5cm}{Class preservation} &   Spearman correlation of the distance between clusters in the original space and the embedded space. (We take the distance between a pair of clusters as $\Delta_{ij}$, defined in \eqref{eq:cluster_distance}, in the original space, and the analog in the embedded space.)  \\
        \bottomrule
    \end{tabularx}
    \caption{Description of metrics used in evaluation. Here we use class to refer to a ground truth cluster label rather than ones obtained by a clustering algorithm. 
}
    \label{tab:metrics_def}
\end{table}

\subsection{Toy Dataset: Shapes}

\begin{figure}[h!]
    \centering
    % ----- Row 1 -----
    \begin{subfigure}[b]{0.18\textwidth}
        \centering
        \includegraphics[width=\textwidth]{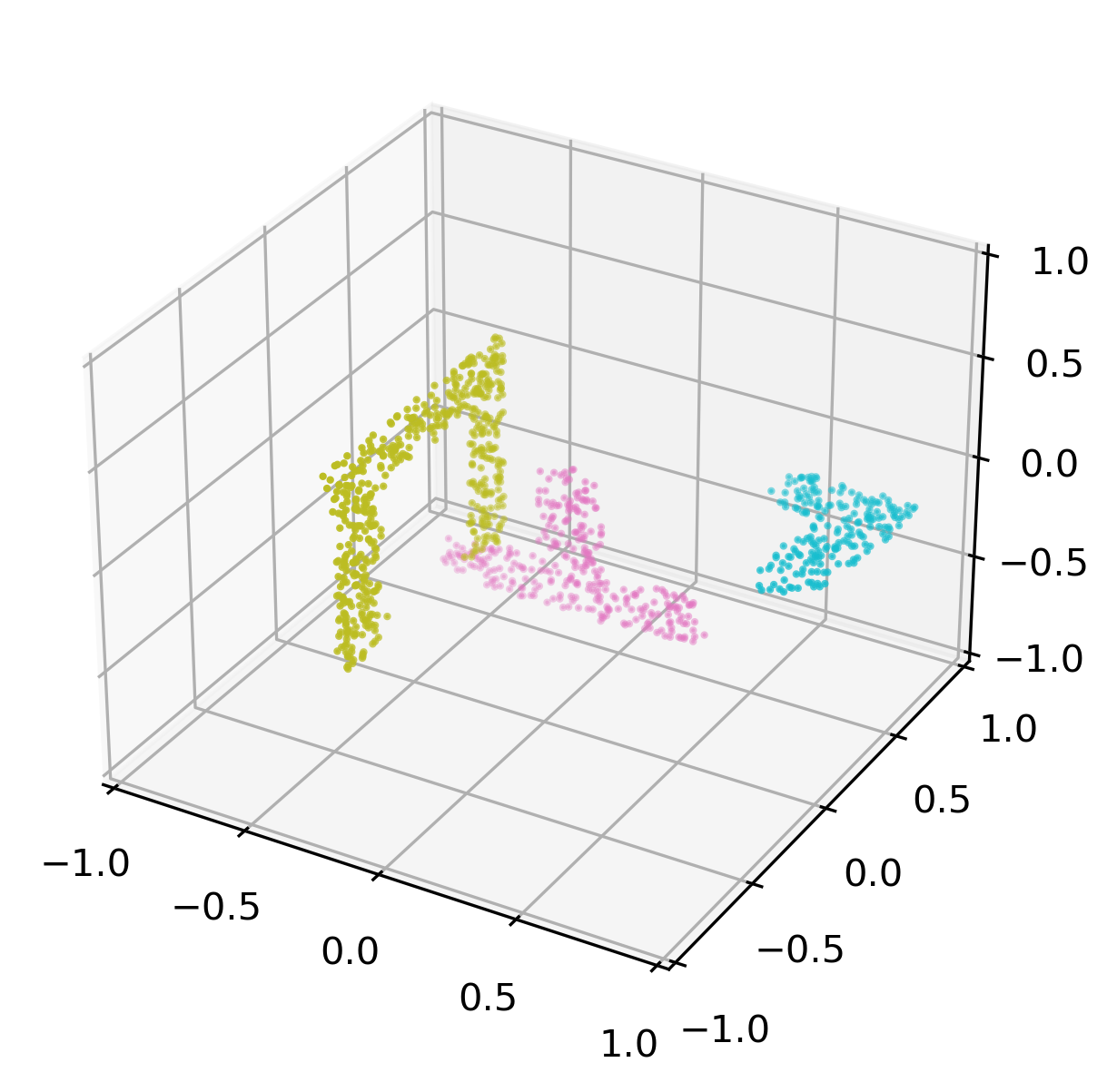}
        \caption*{Original Data}
    \end{subfigure}\hfill
    \begin{subfigure}[b]{0.18\textwidth}
        \centering
        \includegraphics[width=\textwidth]{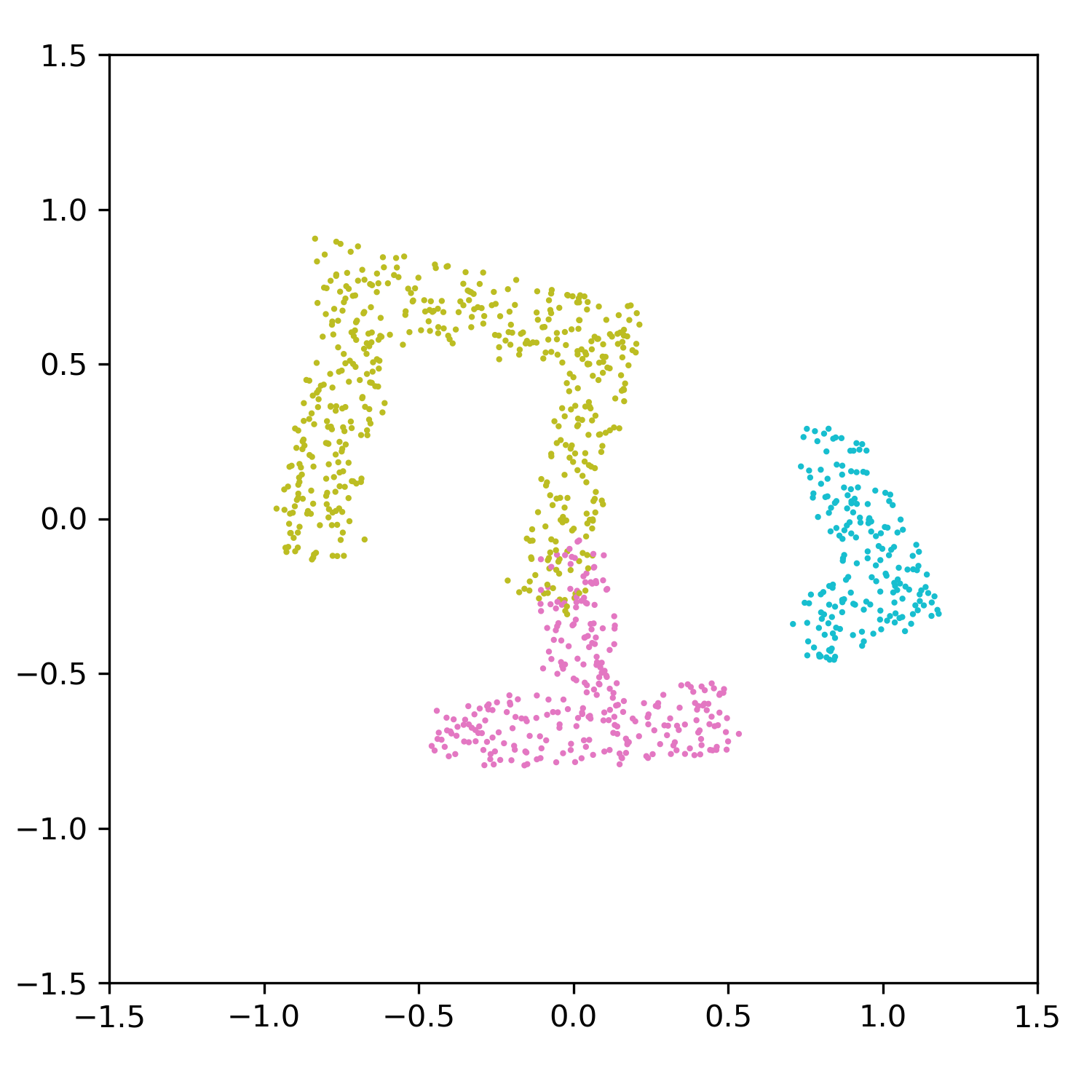}
        \caption*{C+E ($\alpha = 1$)}
    \end{subfigure}\hfill
    \begin{subfigure}[b]{0.18\textwidth}
        \centering
        \includegraphics[width=\textwidth]{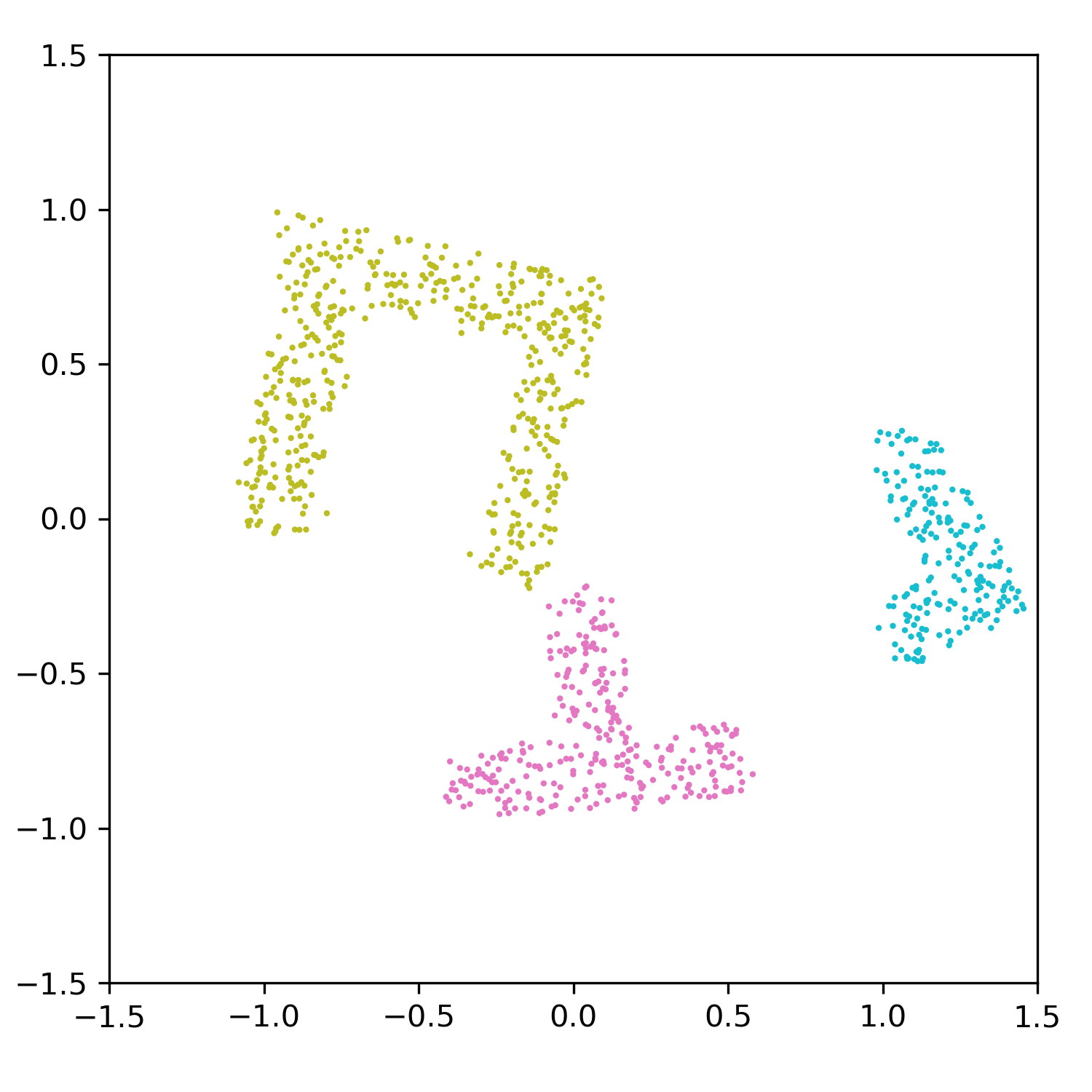}
        \caption*{C+E ($\alpha = 1.25$)}
    \end{subfigure}\hfill
    \begin{subfigure}[b]{0.18\textwidth}
        \centering
        \includegraphics[width=\textwidth]{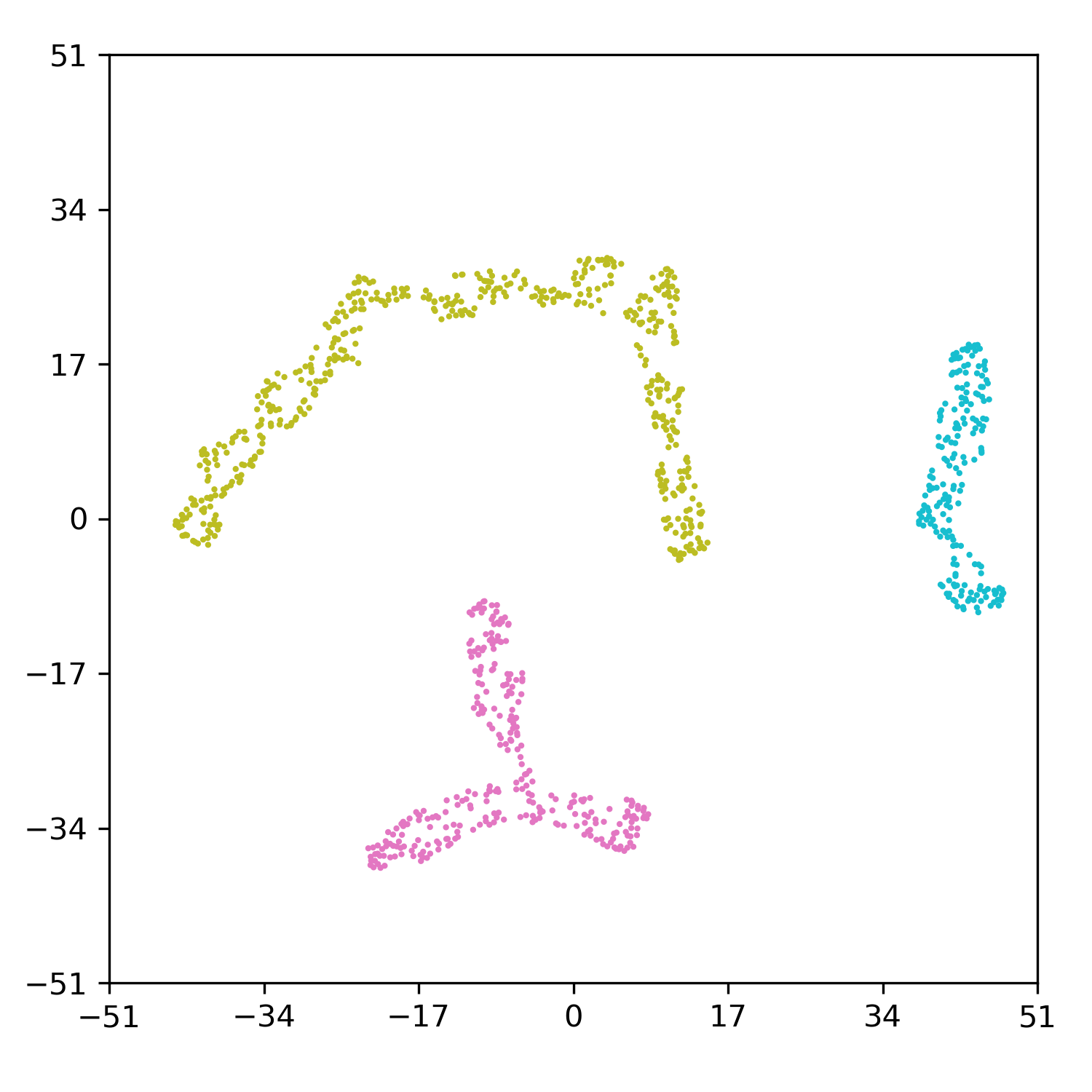}
        \caption*{t-SNE}
    \end{subfigure}\hfill
    \begin{subfigure}[b]{0.18\textwidth}
        \centering
        \includegraphics[width=\textwidth]{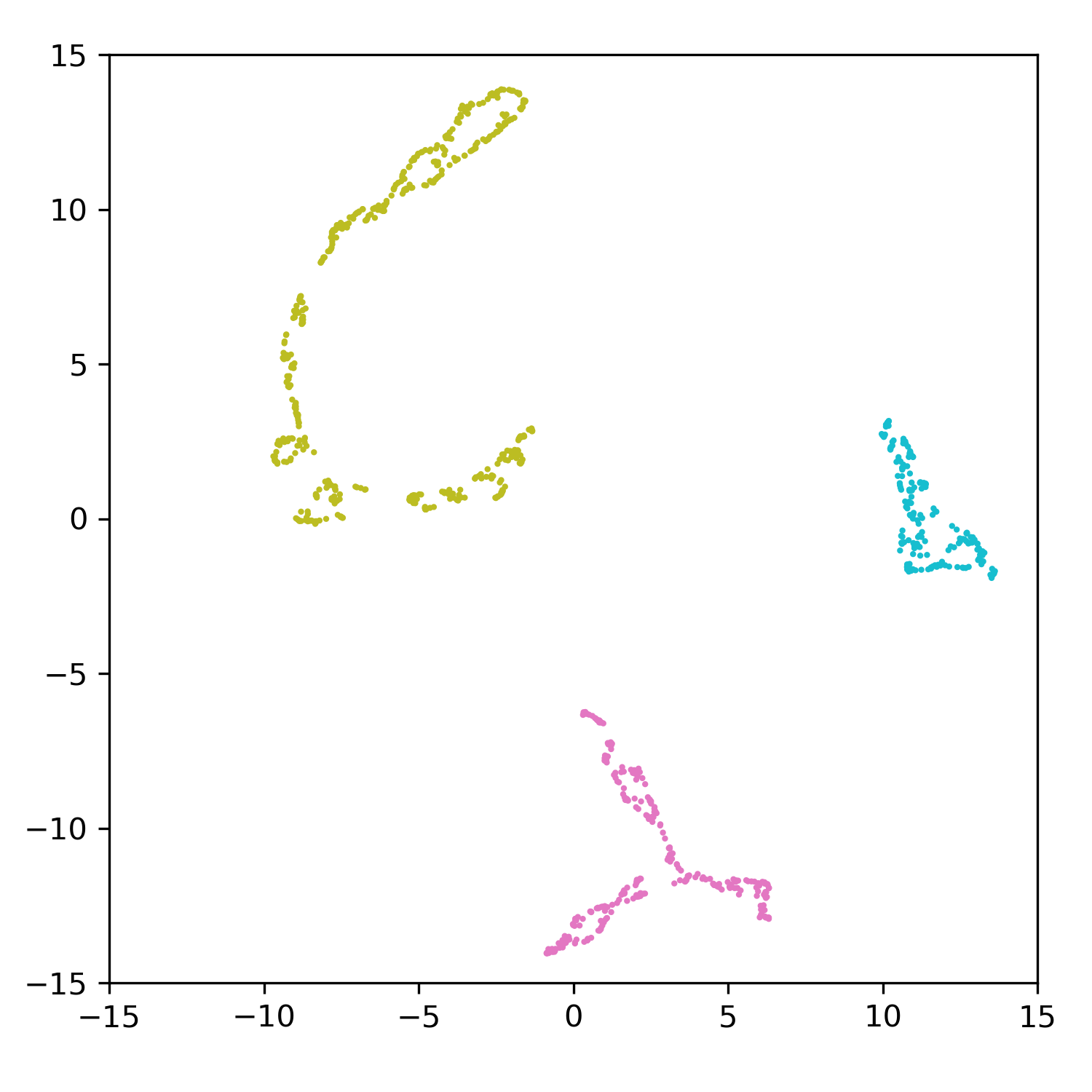}
        \caption*{UMAP}
    \end{subfigure}

    \vspace{0.4cm}

    % ----- Row 2 -----
    \begin{subfigure}[b]{0.18\textwidth}
        \centering
        \includegraphics[width=\textwidth]{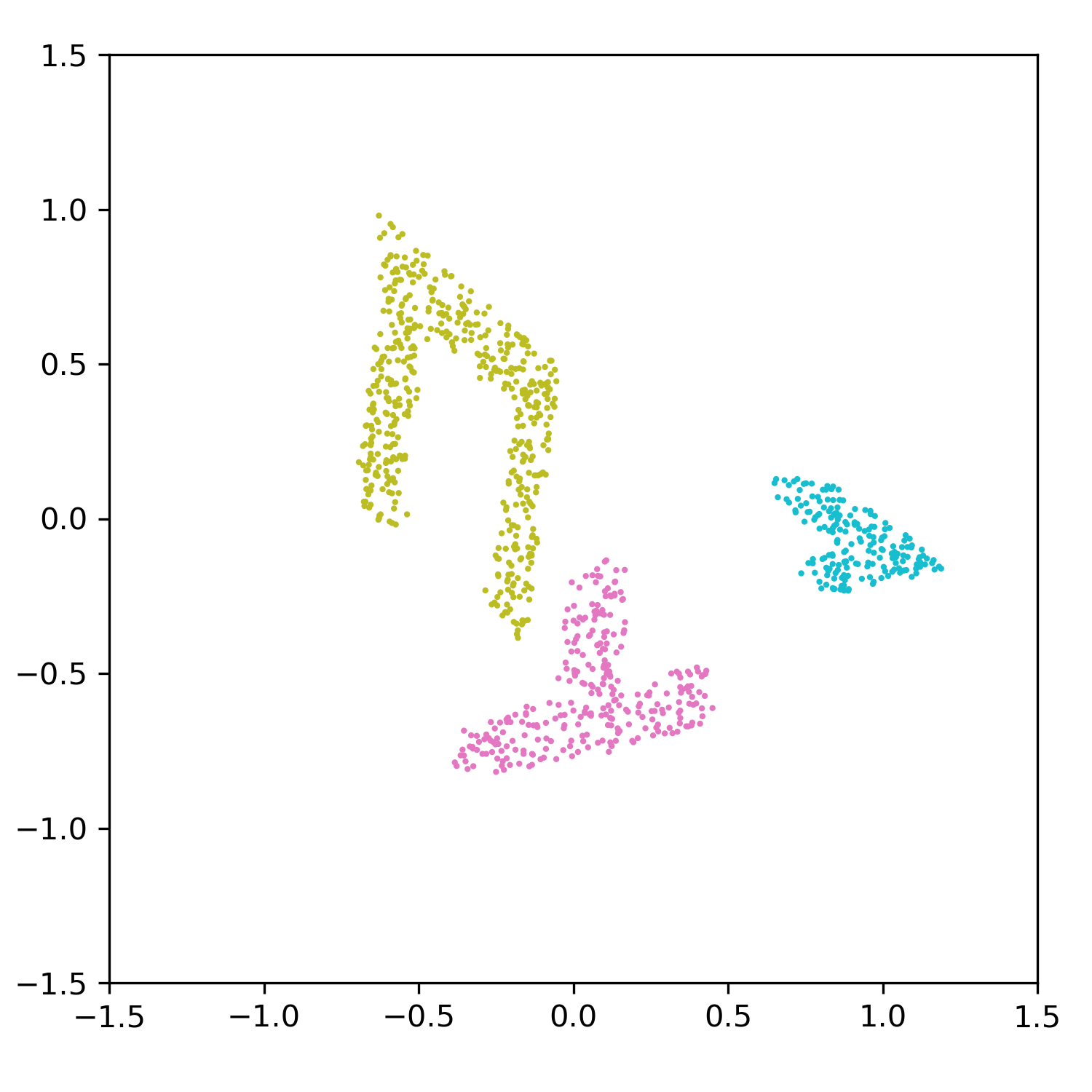}
        \caption*{PCA}
    \end{subfigure}
   \hspace{0.02\textwidth}
    \begin{subfigure}[b]{0.18\textwidth}
        \centering
        \includegraphics[width=\textwidth]{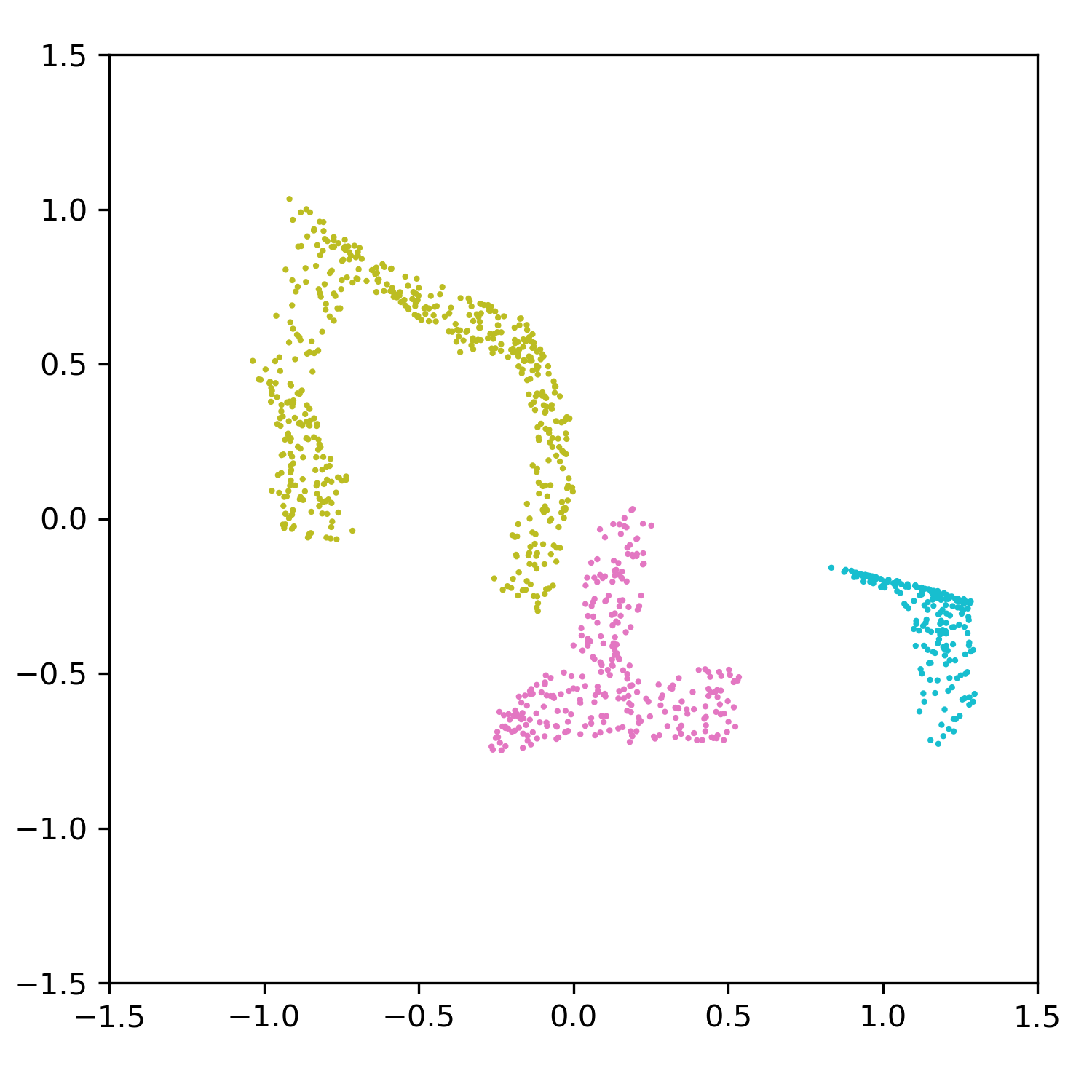}
        \caption*{Isomap}
    \end{subfigure}
   \hspace{0.02\textwidth}
    \begin{subfigure}[b]{0.18\textwidth}
        \centering
        \includegraphics[width=\textwidth]{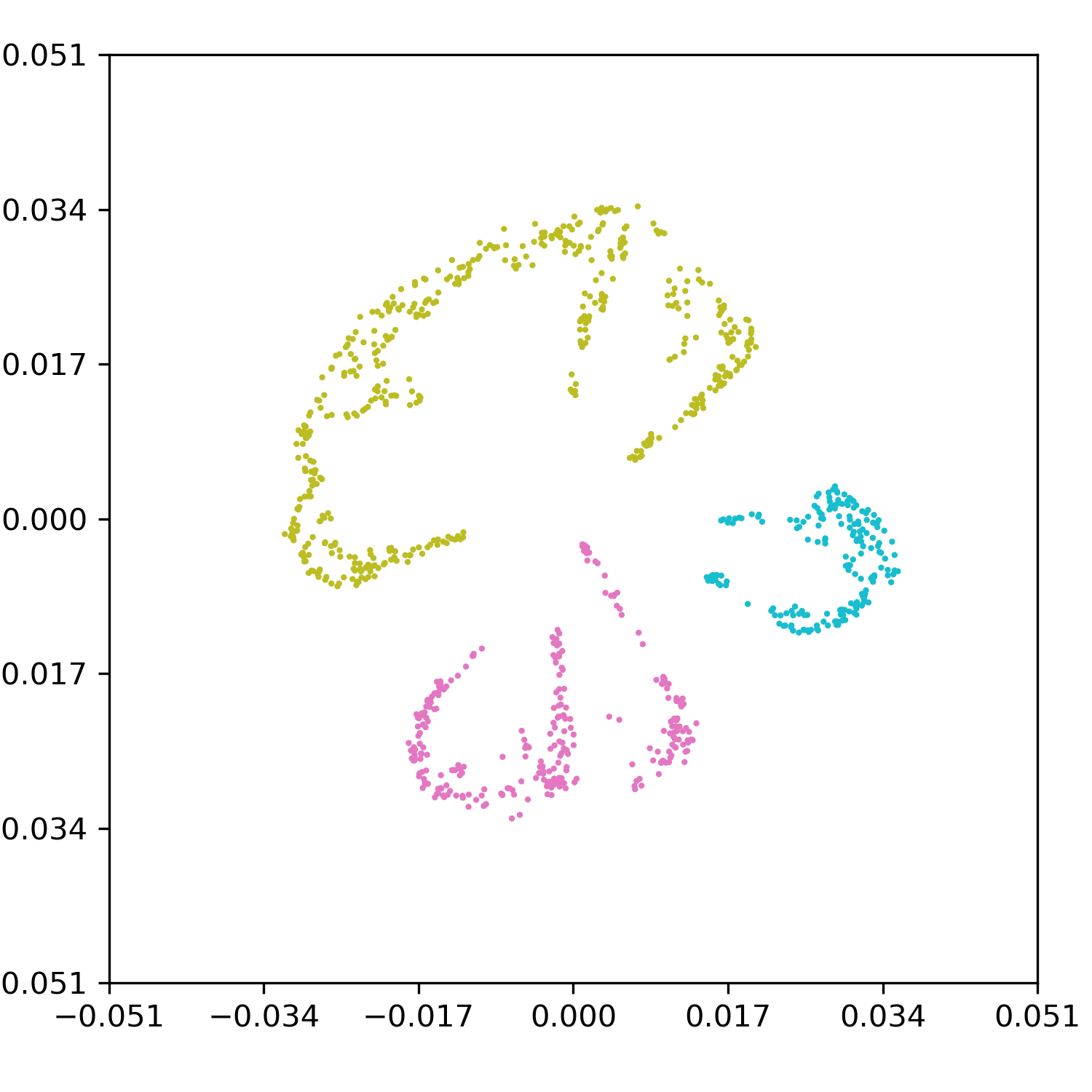}
        \caption*{PHATE}
    \end{subfigure}
    \hspace{0.02\textwidth}
    \begin{subfigure}[b]{0.18\textwidth}
        \centering
        \includegraphics[width=\textwidth]{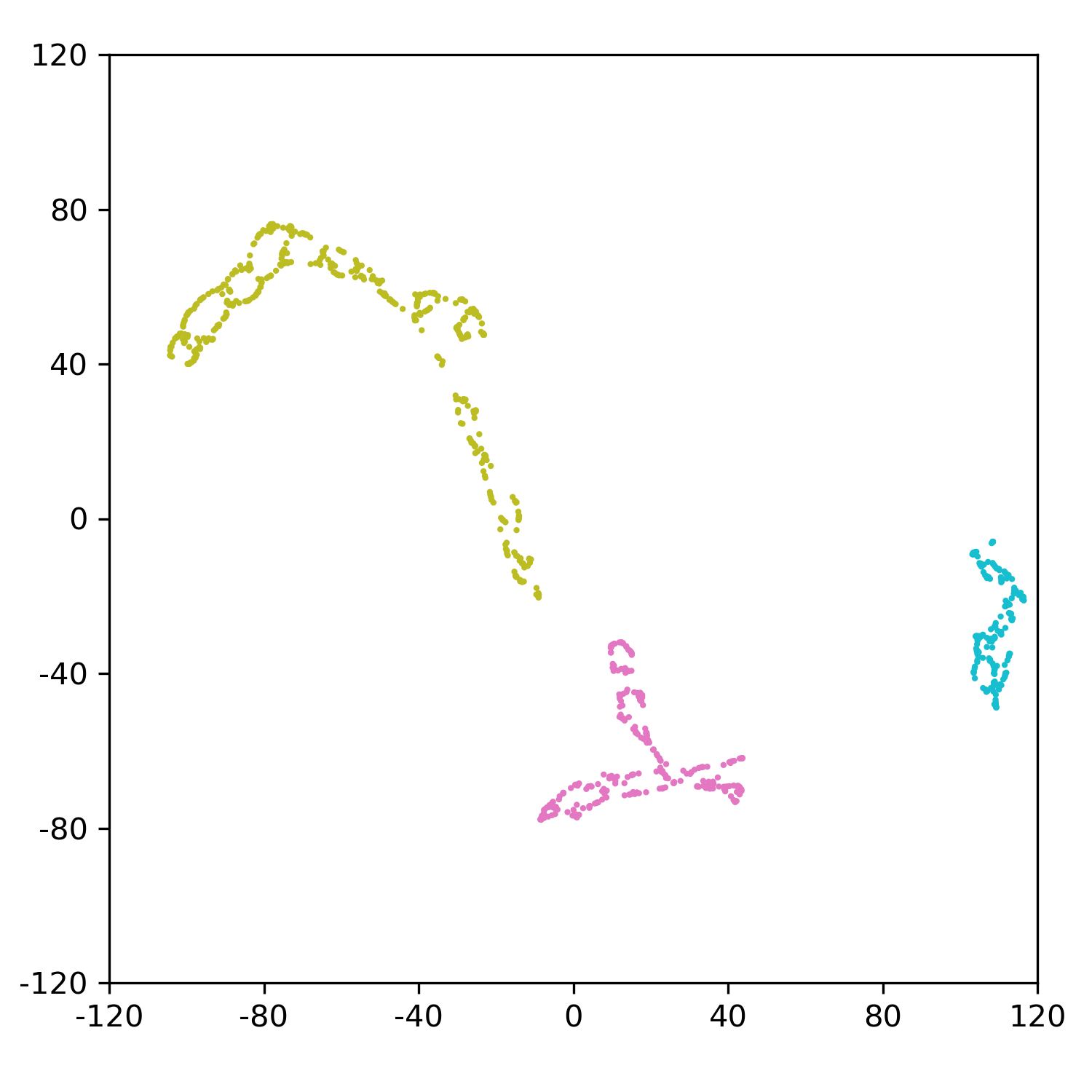}
        \caption*{TriMap}
    \end{subfigure}

    \caption{Shapes dataset ($n=1,000$) and embeddings. Points are colored by original cluster labels which were not used by embedding methods.}
    \label{fig:shapes3d}
\end{figure}

\begin{table}[!ht]
\centering
\footnotesize
\setlength{\tabcolsep}{3pt}
\begin{tabular}{llccccccccccc}
\toprule
\textbf{Method} & \textbf{Param} 
& \multicolumn{4}{c}{\textbf{Local}} 
& \multicolumn{7}{c}{\textbf{Global}} \\
\cmidrule(lr){3-6} \cmidrule(lr){7-13}
& 
& W-S ${\scriptstyle \uparrow}$ 
& W-NS ${\scriptstyle \downarrow}$ 
& W-SNS ${\scriptstyle \downarrow}$ 
& KNN ${\scriptstyle \uparrow}$
& B-S ${\scriptstyle \uparrow}$  
& T-S ${\scriptstyle \uparrow}$  
& B-NS ${\scriptstyle \downarrow}$ 
& T-NS ${\scriptstyle \downarrow}$ 
& B-SNS ${\scriptstyle \downarrow}$ 
& T-SNS ${\scriptstyle \downarrow}$  
& CP ${\scriptstyle \uparrow}$ \\
\midrule

C+E & $\alpha = 1$ &
$\mathbf{1.00}$ & $\mathbf{0.00}$ & $\mathbf{0.00}$ & $0.96$ & $0.96$ & $\mathbf{0.98}$ & $0.09$ & $\mathbf{0.09}$ & $0.09$ & $\mathbf{0.09}$ & $\mathbf{1.00}$\\

C+E & $\alpha = 1.25$ &
$\mathbf{1.00}$ & $\mathbf{0.00}$ & $\mathbf{0.00}$ & $\mathbf{1.00}$ & $0.96$ & $\mathbf{0.98}$& $0.25$ & $0.24$ & $0.07$ & $0.10$ & $\mathbf{1.00}$ \\

t-SNE & $u = 30$ &
$0.93$ & $41.11$ & $0.20$ & $0.87$ & $0.71$ & $0.84$ & $46.17$ & $43.50$ & $0.19$ & $0.23$ & $-0.33$ \\

% t-SNE & $u = 260$ &
% $0.95$ & $4.02$ & $0.18$ & $0.88$ & $0.70$ & $0.93$ & $7.65$ & $7.99$ & $0.18$ & $0.26$ & $\mathbf{1.00}$ \\

UMAP & $q = 15$ &
$0.93$ & $9.03$ & $0.17$ & $0.81$ & $0.60$ & $0.88$ & $13.75$ & $13.70$ & $0.16$ & $0.23$ & $0.33$ \\

PCA & - &
$0.92$ & $0.28$ & $0.19$ & $0.84$ & $\mathbf{0.98}$ & $\mathbf{0.98}$ & $\mathbf{0.06}$ & $0.12$ & $\mathbf{0.05}$ & $0.11$ & $\mathbf{1.00}$ \\

Isomap & $q = 200$ &
$0.94$ & $0.19$ & $0.16$ & $0.83$ & $0.90$ & $\mathbf{0.98}$ & $0.15$ & $0.21$ & $0.08$ & $0.15$ & $\mathbf{1.00}$ \\

PHATE & $q = 5$ &
$0.73$ & $0.96$ & $0.35$ & $0.65$ & $0.22$ & $0.68$ & $0.97$ & $0.97$ & $0.31$ & $0.32$ & $\mathbf{1.00}$\\

TriMap & - & $0.93$ & $66.38$ & $0.21$ & $0.79$ & $0.82$ & $0.95$ & $112.28$ & $114.00$ & $0.11$ & $0.19$ & $\mathbf{1.00}$\\

\bottomrule
\end{tabular}
\caption{Evaluation of the shapes dataset against Euclidean distances. Metrics are reported for within class (W), between class (B), and total (T). S denotes Spearman correlation, NS normalized stress, SNS scale-normalized stress, CP class preservation, and KNN the $30$-NN recall. Arrows indicate whether higher ($\uparrow$) or lower ($\downarrow$) values are better.}
\label{tab:shapes}
\end{table}

We begin with a simple, synthetic example where $\{x_i\}_{i=1}^n   \subset \bbR^3$, consists of clusters of geometric shapes taken from \cite{hahsler2019}.  We position $\{x_i\}_{i=1}^n $ so that it is not possible to obtain an embedding in $\bbR^2$ that preserves all pairwise distances. In this case, there is not a unique, well-defined correct embedding in  $\bbR^2$. For our method, we cluster using DBSCAN, embed each cluster using PCA, and align the clusters to preserve Euclidean distances. In \figref{shapes3d}, we include the results of our method using $\alpha = 1$ and $\alpha = 1.25$. For Isomap, we set the number of neighbors ($q = 200$) large enough so that the $k$NN graph is connected.

Note that with this dataset, DBSCAN clusters the data without error into three clusters, and because each cluster is flat, PCA applied to each cluster individually exactly recovers the configuration of each cluster. Thus, C+E achieves perfect local metrics, with the exception of $k$NN recall in \tabref{shapes}. The alignment step also orients the clusters consistently with their orientation and their positioning in the original space, as seen both visually and quantitatively. In contrast, while t-SNE ($u = 30$) clusters the data well and positions the clusters relative to one another reasonably well, the individual clusters are distorted, as well as the overall scale of the embedding. All other methods similarly recover the global geometry well, but incur significant distortion at the local, cluster level.

\subsection{MNIST}
\begin{figure}[h!]
    \centering
    \captionsetup{justification=centering, singlelinecheck=false, font = footnotesize}
    % ----- Row 1 -----
    \begin{subfigure}[c]{0.18\textwidth}
        \centering
        \includegraphics[width=\textwidth]{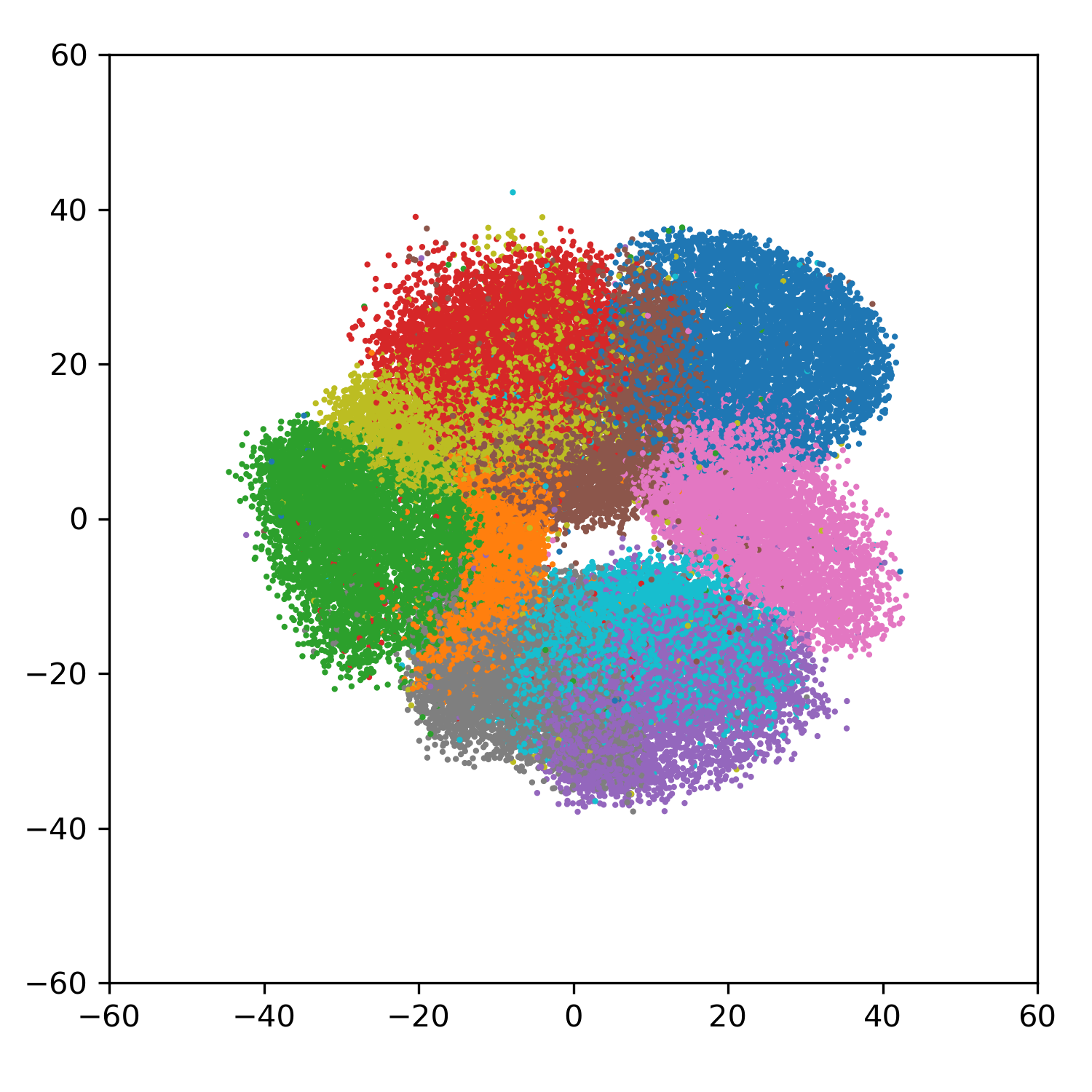}
        \caption*{C+E \\ (L-Isomap, $\alpha=1$)}
    \end{subfigure}\hfill
    \begin{subfigure}[c]{0.18\textwidth}
        \centering
        \includegraphics[width=\textwidth]{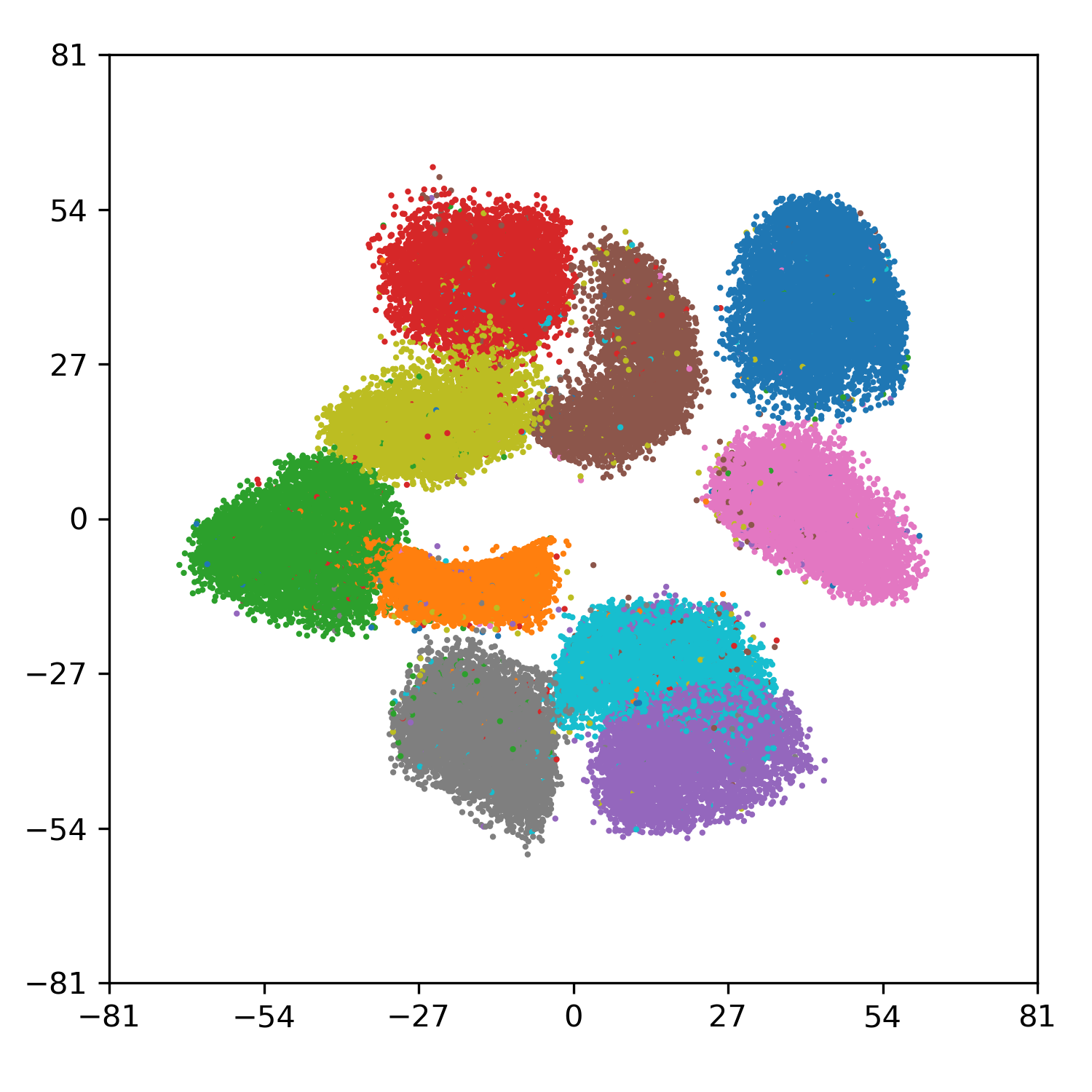}
        \caption*{C+E \\ (L-Isomap, $\alpha=1.75$)}
    \end{subfigure}\hfill
    \begin{subfigure}[c]{0.18\textwidth}
        \centering
        \includegraphics[width=\textwidth]{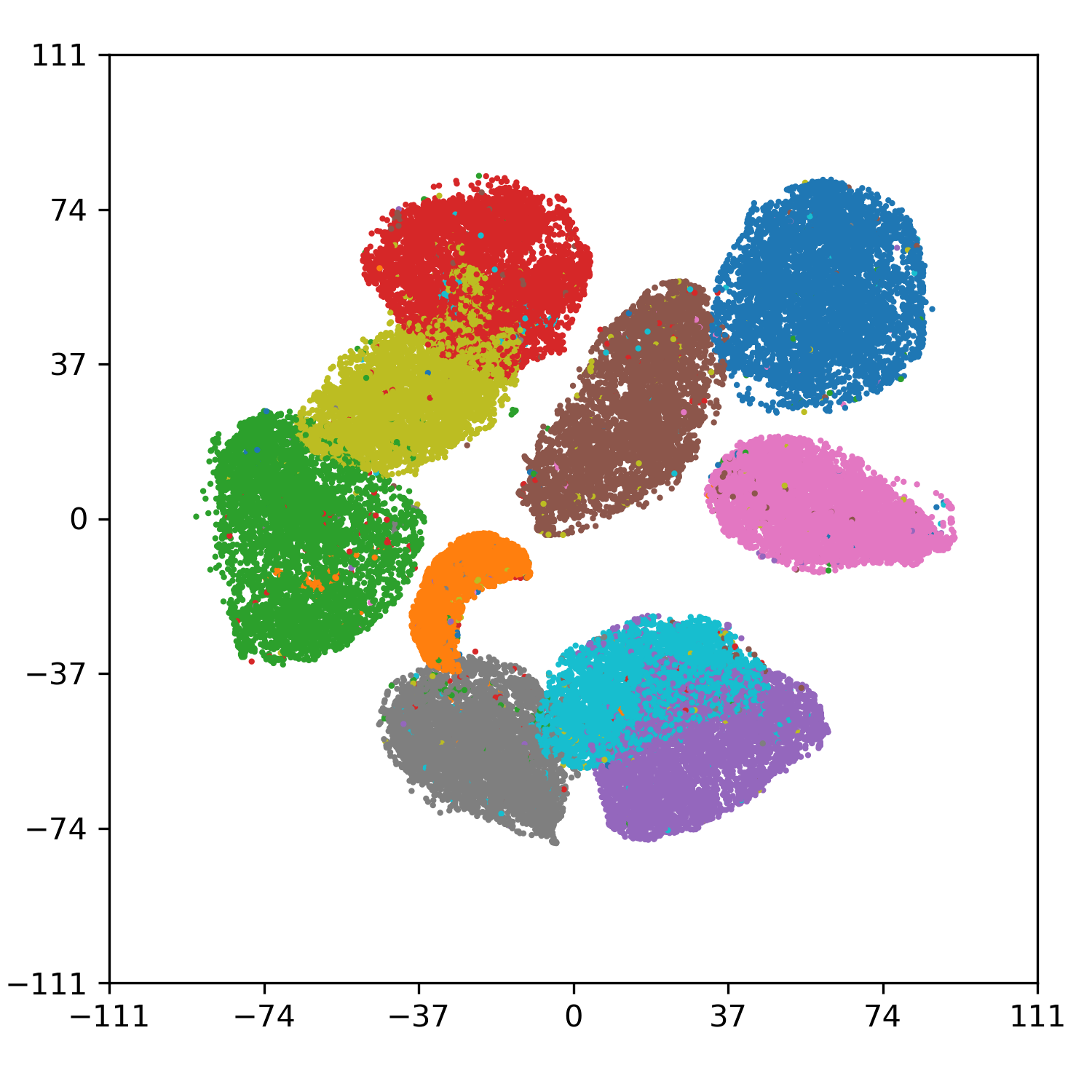}
        \caption*{C+E \\ (TriMap, $\alpha = 2.53$)}
    \end{subfigure}\hfill
    \begin{subfigure}[c]{0.18\textwidth}
        \centering
        \includegraphics[width=\textwidth]{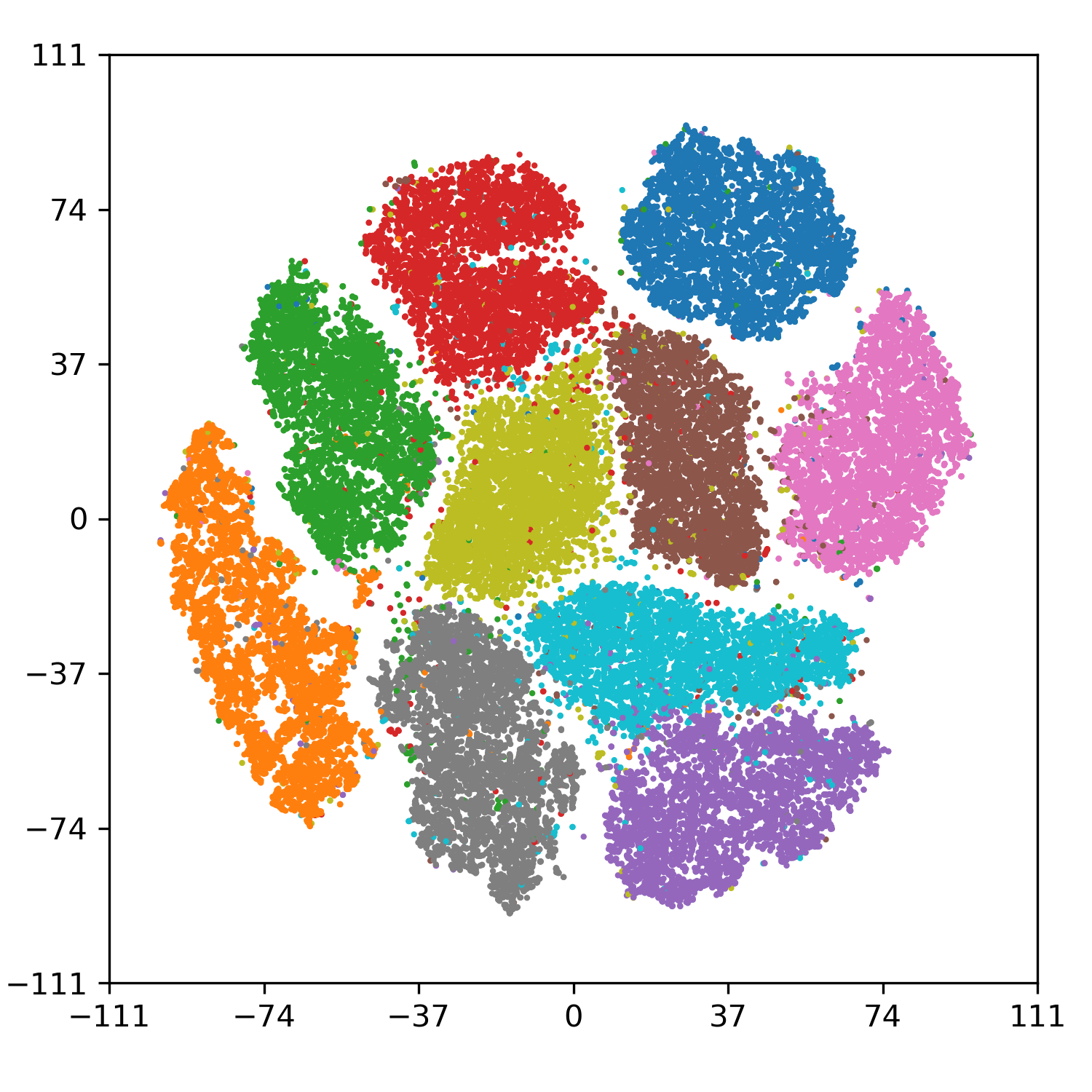}
        \caption*{t-SNE \\  \strut} 
    \end{subfigure}\hfill
    \begin{subfigure}[c]{0.18\textwidth}
        \centering
        \includegraphics[width=\textwidth]{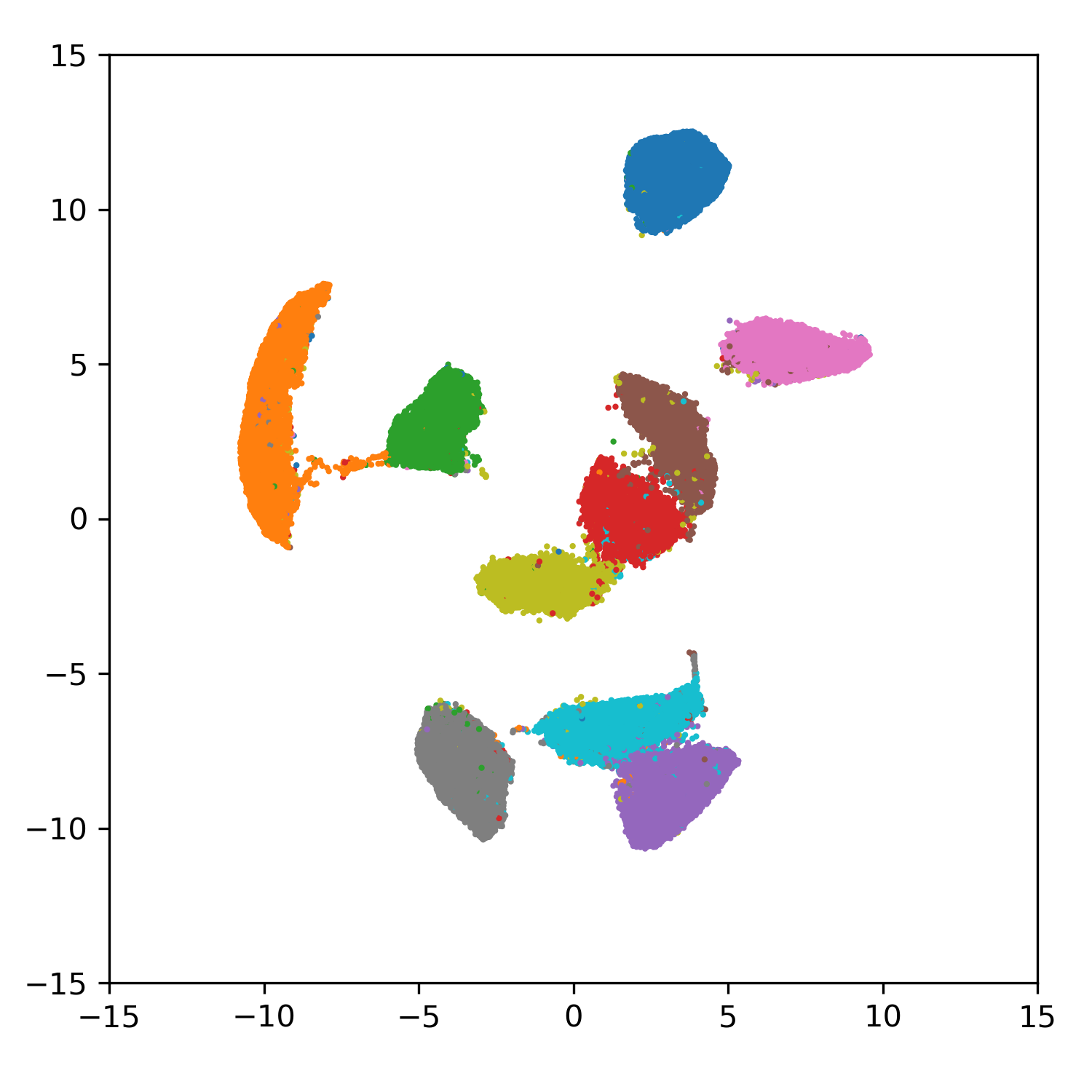}
        \caption*{UMAP \\  \strut}
    \end{subfigure}
    
% ----- Row 2 -----
    \begin{subfigure}[c]{0.18\textwidth}
        \centering
        \includegraphics[width=\textwidth]{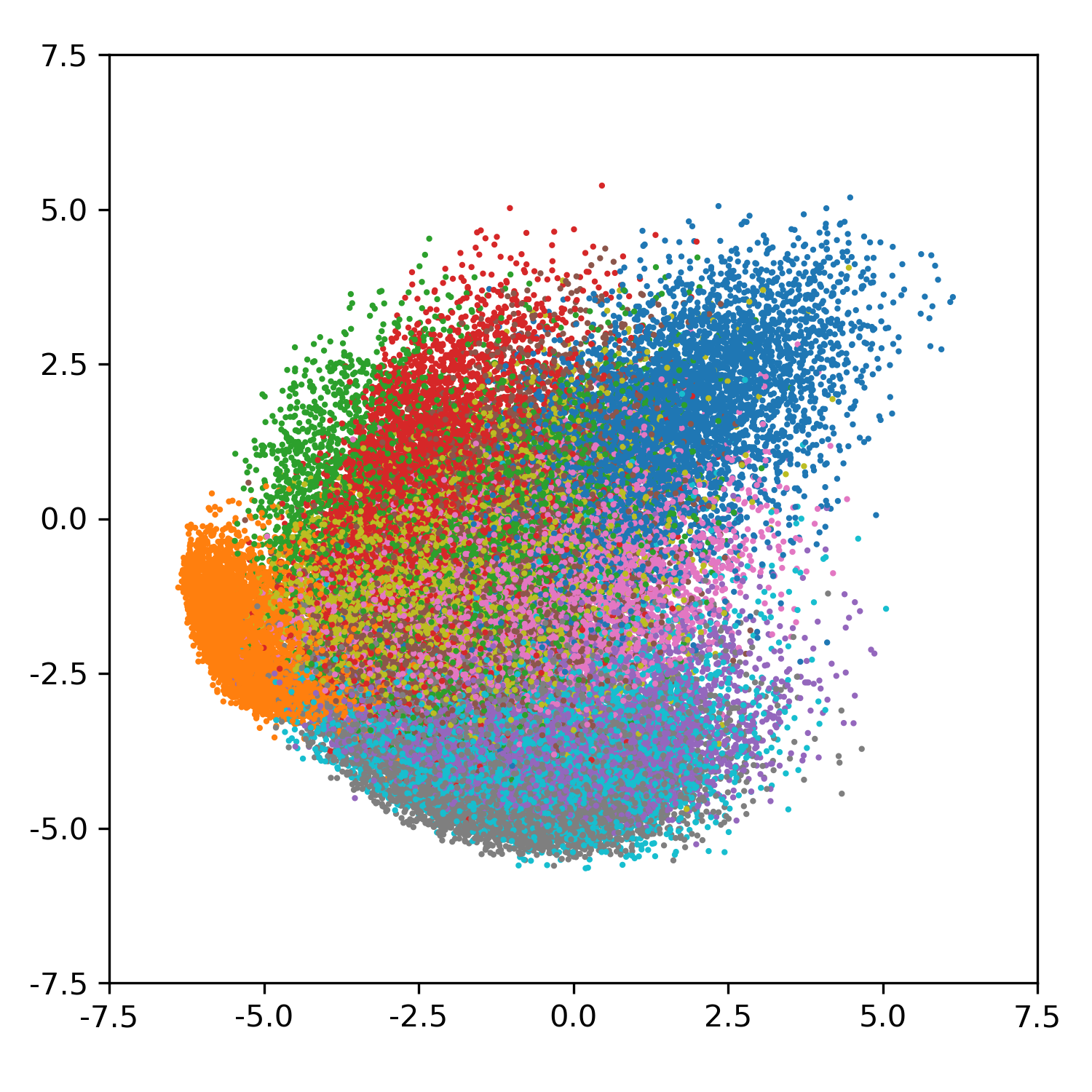}
        \caption*{PCA}
    \end{subfigure}
   \hfill
    \begin{subfigure}[c]{0.18\textwidth}
        \centering
        \includegraphics[width=\textwidth]{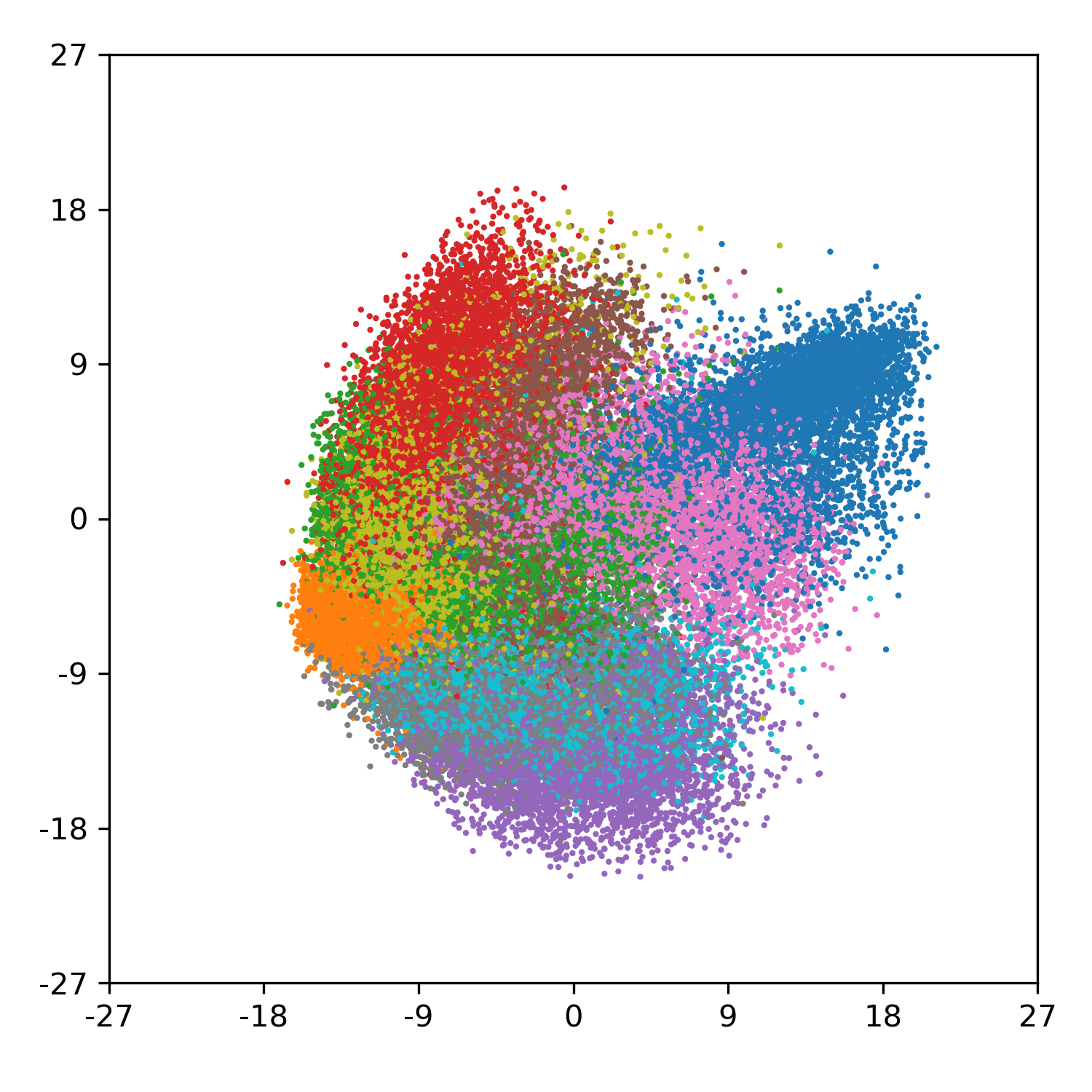}
        \caption*{L-Isomap}
    \end{subfigure}
    \hfill
    \begin{subfigure}[c]{0.18\textwidth}
        \centering
        \includegraphics[width=\textwidth]{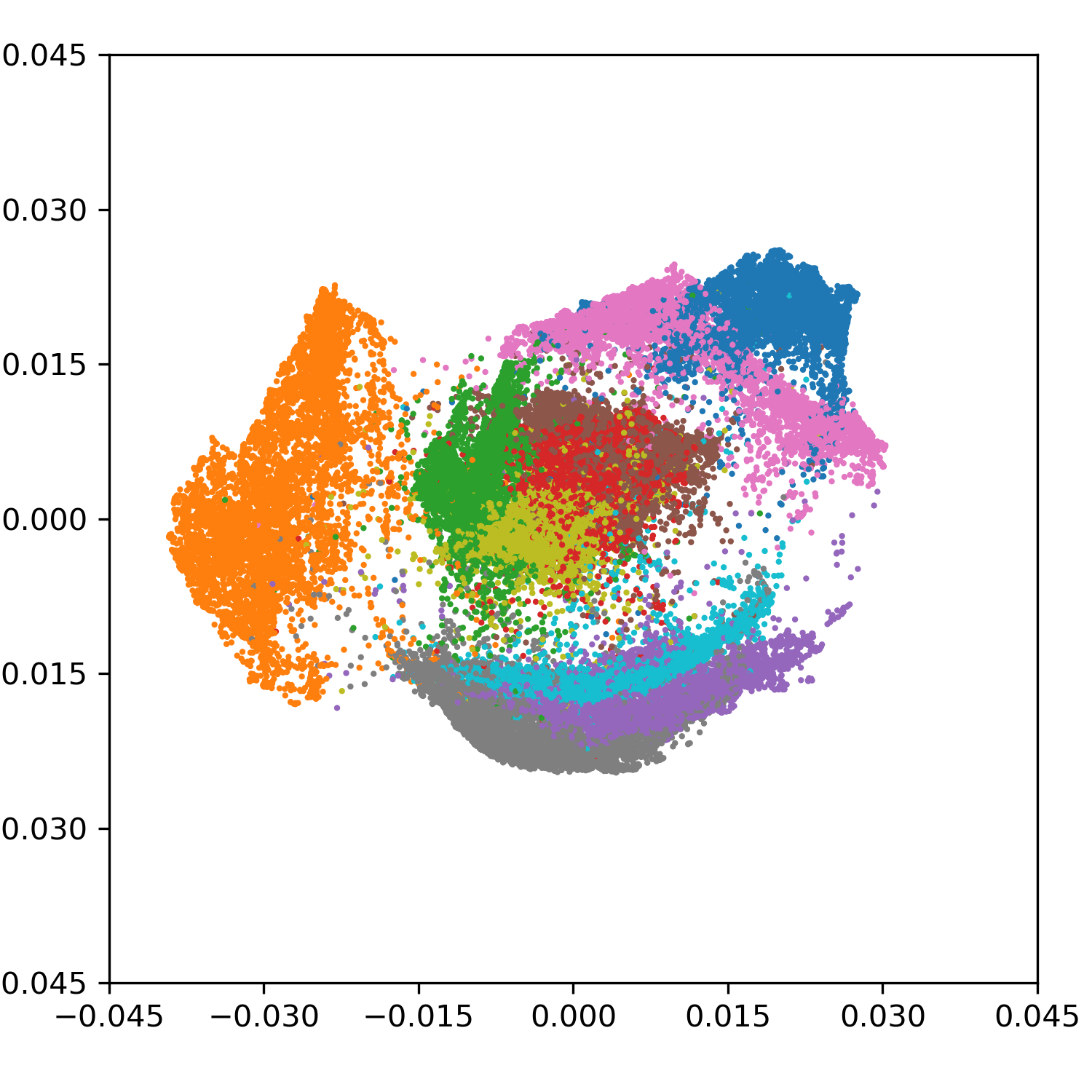}
        \caption*{PHATE}
    \end{subfigure}
    \hfill
    \begin{subfigure}[c]{0.18\textwidth}
        \centering
        \includegraphics[width=\textwidth]{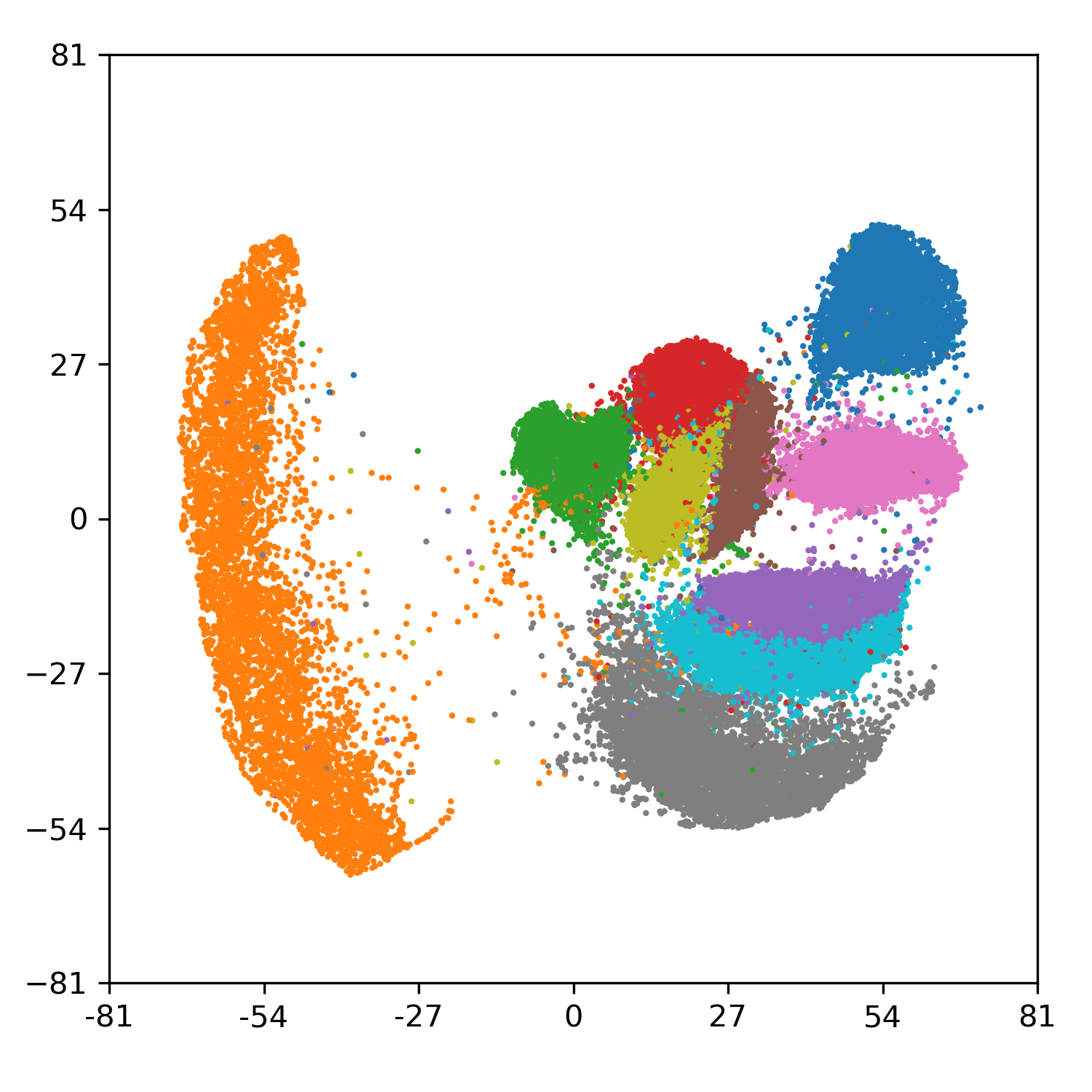}
        \caption*{TriMap}
    \end{subfigure}
     \hfill
    \begin{subfigure}[c]{0.18\textwidth}
        \centering
        \includegraphics[width=.25\textwidth]{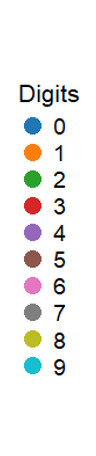}
         \caption*{}
    \end{subfigure}
    \caption{ Embedding of MNIST dataset ($n = 60,000$). Points are colored according to their original digit label. These labels were not used by the embedding methods.}
    \label{fig:mnist}
\end{figure}

\begin{table}[!ht]
\centering
\footnotesize
\setlength{\tabcolsep}{2pt}
\begin{tabular}{llccccccccccc}
\toprule
\textbf{Method} & \textbf{Params} 
& \multicolumn{4}{c}{\textbf{Local}} 
& \multicolumn{7}{c}{\textbf{Global}} \\
\cmidrule(lr){3-6} \cmidrule(lr){7-13}
& 
& W-S ${\scriptstyle \uparrow}$ 
& W-NS ${\scriptstyle \downarrow}$ 
& W-SNS ${\scriptstyle \downarrow}$ 
& KNN ${\scriptstyle \uparrow}$
& B-S ${\scriptstyle \uparrow}$  
& T-S ${\scriptstyle \uparrow}$  
& B-NS ${\scriptstyle \downarrow}$ 
& T-NS ${\scriptstyle \downarrow}$ 
& B-SNS ${\scriptstyle \downarrow}$ 
& T-SNS ${\scriptstyle \downarrow}$  
& CP ${\scriptstyle \uparrow}$\\
\midrule

C+E (L-Isomap) & $\alpha = 1$ &
$\mathbf{0.81}$ & $0.52$ & $\mathbf{0.31}$ & $0.26$ & $0.43$ & $\mathbf{0.63}$ & $\mathbf{0.35}$ & $\mathbf{0.36}$ & $0.28$ & $\mathbf{0.34}$ & $\mathbf{0.74}$ \\

C+E  (L-Isomap) & $\alpha = 1.75$ &
$0.80$ & $0.53$ & $0.44$ & $0.37$ & $0.42$ & $0.61$ & $0.51$ & $0.58$ & $\mathbf{0.23}$ & $\mathbf{0.34}$ & $0.73$ \\

C+E  (TriMap) & $\alpha = 2.53$ &
$0.79$ & $\mathbf{0.42}$ & $0.40$ & $0.38$ & $0.38$ & $0.59$ & $1.03$ & $1.14$ & $0.24$ & $\mathbf{0.34}$ & $0.72$ \\

t-SNE & $u = 30$ &
$0.76$ & $0.51$ & $0.39$ & $\mathbf{0.41}$ & $0.33$ & $0.49$ & $1.25$ & $1.39$ & $0.26$ & $0.37$ & $0.59$
 \\
 
 UMAP & $q = 15$ &
$\mathbf{0.81}$ & $0.92$ & $0.57$ & $0.40$ & $0.29$ & $0.45$ & $0.76$ & $0.76$ & $\mathbf{0.23}$ & $0.41$ & $0.57$\\

PCA & - &
$0.40$ & $0.92$ & $0.46$ & $0.09$ & $0.35$ & $0.45$ & $0.91$ & $0.91$ & $0.36$ & $0.41$ & $0.45$ \\

L-Isomap & $q = 15$  &
$0.49$ & $0.76$ & $0.44$ & $0.12$ & $\mathbf{0.45}$ & $0.61$ & $0.68$ & $0.68$ & $0.32$ & $0.37$ & $0.70$
 \\

PHATE & $q = 5$ &
$0.65$ & $1.00$ & $0.42$ & $0.23$ & $0.38$ & $0.46$ & $1.00$ & $1.00$ & $0.25$ & $0.41$ & $0.53$ \\

TriMap & $-$ &
$0.75$ & $0.69$ & $0.38$ & $0.35$ & $0.28$ & $0.28$ & $0.58$ & $0.72$ & $0.25$ & $0.52$ & $0.37$\\

\bottomrule
\end{tabular}
\caption{Evaluation of the MNIST dataset against geodesic distances for $5000$ points (selected independently of the points used in alignment for C+E). Metrics are reported for within class (W), between class (B), and total (T). S denotes Spearman correlation, NS normalized stress, SNS scale-normalized stress, CP class preservation, and KNN the $30$-NN recall. Arrows indicate whether higher ($\uparrow$) or lower ($\downarrow$) values are better.}
\label{tab:mnist}
\end{table}

\begin{figure}[h!]
    \centering
    \begin{subfigure}[c]{0.45\textwidth}
        \centering
        \includegraphics[width=\textwidth]{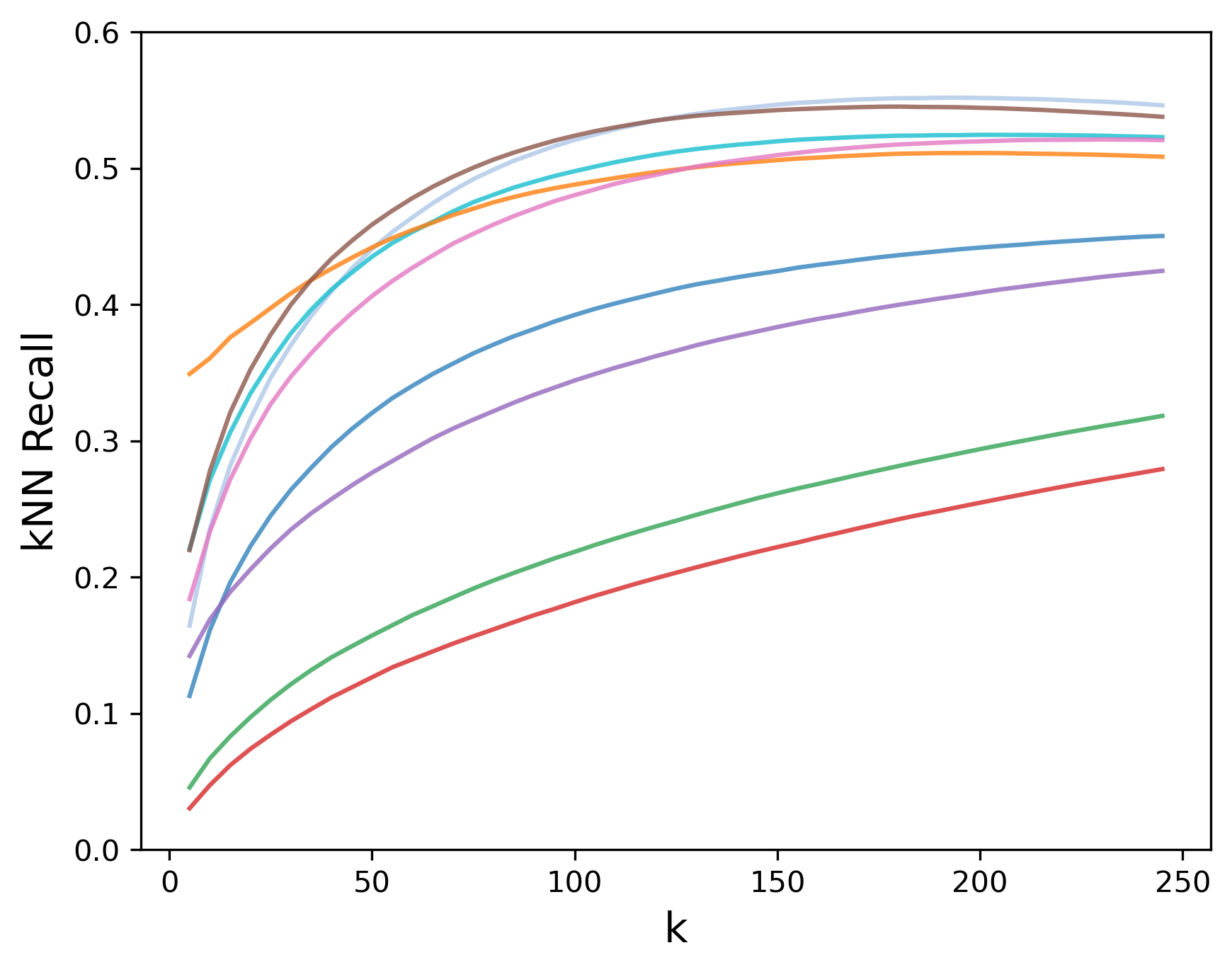}
    \end{subfigure}
    \hspace{0.03\textwidth}
    \begin{subfigure}[c]{0.18\textwidth}
        \centering
        \includegraphics[width=\textwidth]{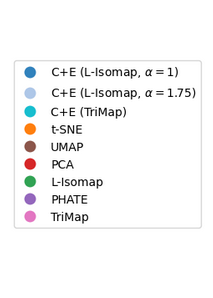}
    \end{subfigure}
    \caption{$k$NN recall for the MNIST dataset for various values of $k$ for all methods compared.}
    \label{fig:mnist_knn}
    \end{figure}

For our first real-world example we test our method on the MNIST dataset \cite{lecun2002gradient}. We first preprocess the data by flattening the images and apply PCA to project to dimension $50$. We cluster the data via the Leiden algorithm ($q = 15$, $\gamma = 0.3$) with the resolution parameter $\gamma$ selected to obtain ten clusters. The clustering has a Rand index of $0.989$ with the digit labels. We include results for two embedding methods,  L-Isomap ($q = 15 , n_\text{landmarks} = 300$) and TriMap. Clusters are aligned to preserve geodesic distances. The scaling parameter $\alpha$ is obtained using \eqref{eq:alpha}, resulting in $\alpha = 1.75$ for the L-Isomap embeddings and $\alpha = 2.53$ for the Trimap embeddings. For reference, we also include the results with $\alpha = 1$ for the L-Isomap embeddings. We report metrics in \tabref{mnist} against geodesic distances. In \appref{mnist_app}, we include results where clusters are embedded and aligned to preserve Euclidean distances. 

In  \figref{mnist}, we demonstrate that the C+E (L-Isomap, $\alpha = 1.75$) embedding separates the clusters and organizes the clusters into larger groupings of related clusters (i.e. fours and nines are close to one another). UMAP finds a similar arrangement of clusters, though the scale of the embedding is small compared to the scale of the pairwise distances, and thus the scale-dependent distance preservation metrics are poor for UMAP, though some scale-invariant metrics like the Spearman correlation are also poor. The t-SNE embedding suggests approximately equal spacing between each cluster\footnote{This can be improved by increasing the early exaggeration parameter of t-SNE \cite{bohm2022}.}, and while there is some global organization of the clusters present in the embedding, C+E outperforms t-SNE on all global metrics. Locally, the methods have comparable metrics, with the exceptions of UMAP having poor local distance preservation metrics, and t-SNE having dominant $k$NN recall for small values of $k$. 

C+E (L-Isomap, $\alpha = 1$), PCA, L-Isomap, and PHATE also find a similar arrangement of clusters, but fail to completely separate the clusters. C+E (L-Isomap, $\alpha = 1$) has better metrics than  C+E (L-Isomap, $\alpha = 1.75$), though does not create a visualization with distinct clusters. However, even with $\alpha = 1.75$, C+E with L-Isomap embeddings achieves better local metrics and comparable or better global metrics than embedding all of the data at once with L-Isomap. Similarly, compared to applying TriMap to all of the data at once, C+E (TriMap) achieves better local and global performance on most metrics. In particular, C+E (TriMap) is meant to target $k$NN recall for small $k$, and successfully improves this metric compared to C+E (L-Isomap) (\figref{mnist_knn}). While C+E (TriMap) does not reach the performance of t-SNE and UMAP on $k$NN recall for any value of $k$, it generally has better global metrics. In this sense the choice of embedding method used in the C+E approach can be used to emphasize certain information, while also enforcing a faithful global visualization.

\subsection{Human brain organoid data}
\begin{figure}[h!]
    \centering
    \captionsetup{justification=centering, singlelinecheck=false, font = footnotesize}
    % ----- Row 1 -----
    \begin{subfigure}[b]{0.18\textwidth}
        \centering
        \includegraphics[width=\textwidth]{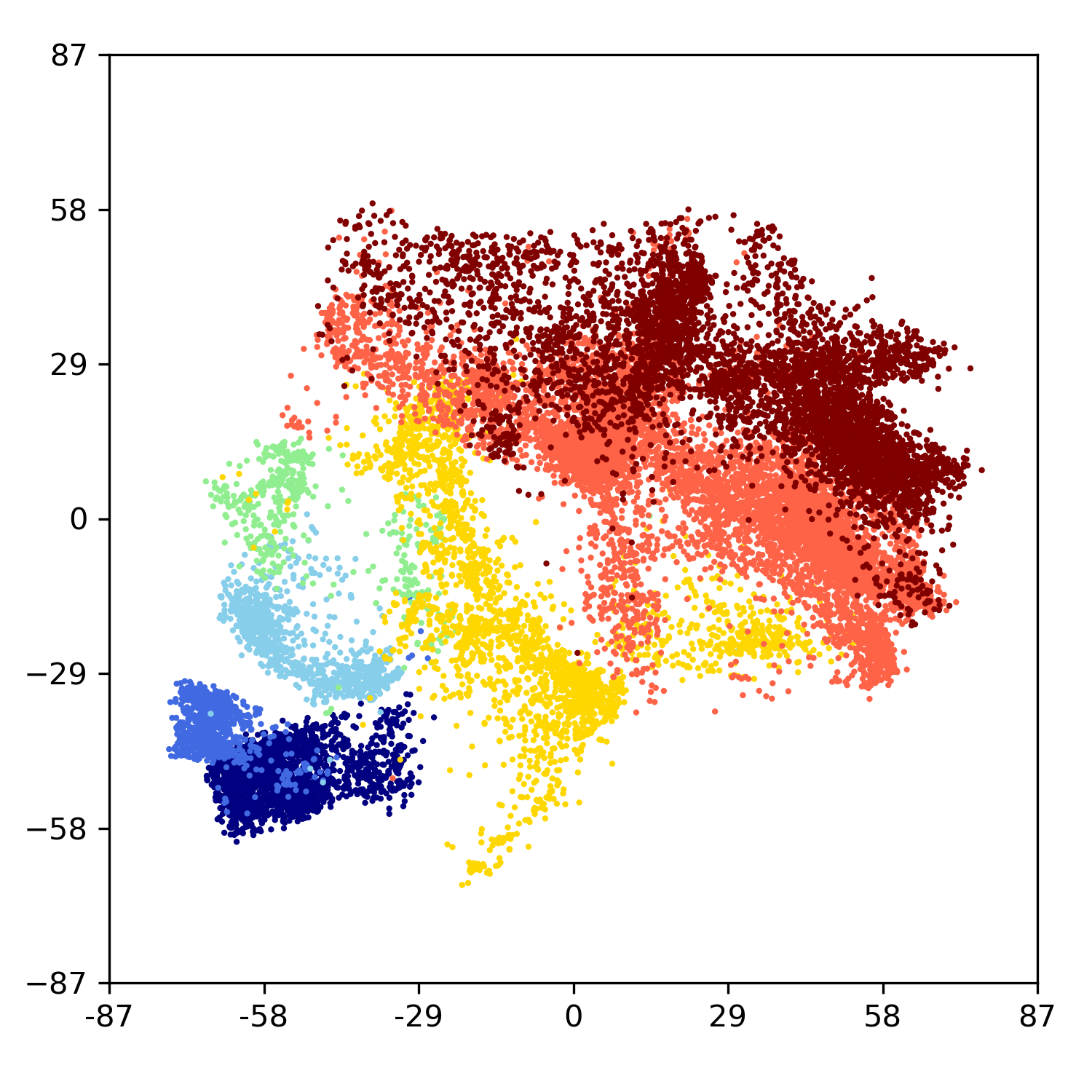}
        \caption*{C+E \\ (L-Isomap, $\alpha=1$)}
    \end{subfigure}\hfill
    \begin{subfigure}[b]{0.18\textwidth}
        \centering
        \includegraphics[width=\textwidth]{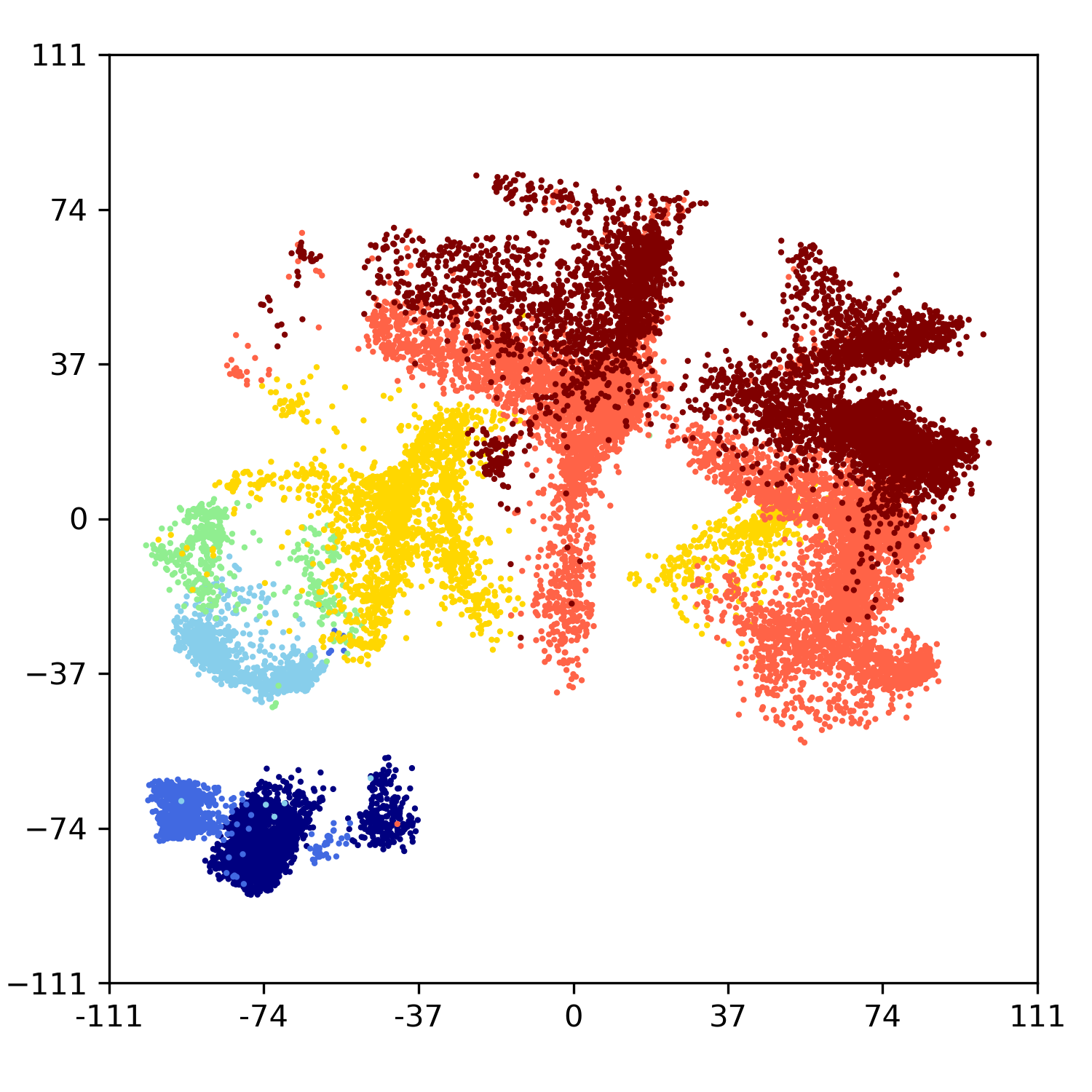}
        \caption*{C+E \\ (L-Isomap, $\alpha=1.45$)}
    \end{subfigure}\hfill
    \begin{subfigure}[b]{0.18\textwidth}
        \centering
        \includegraphics[width=\textwidth]{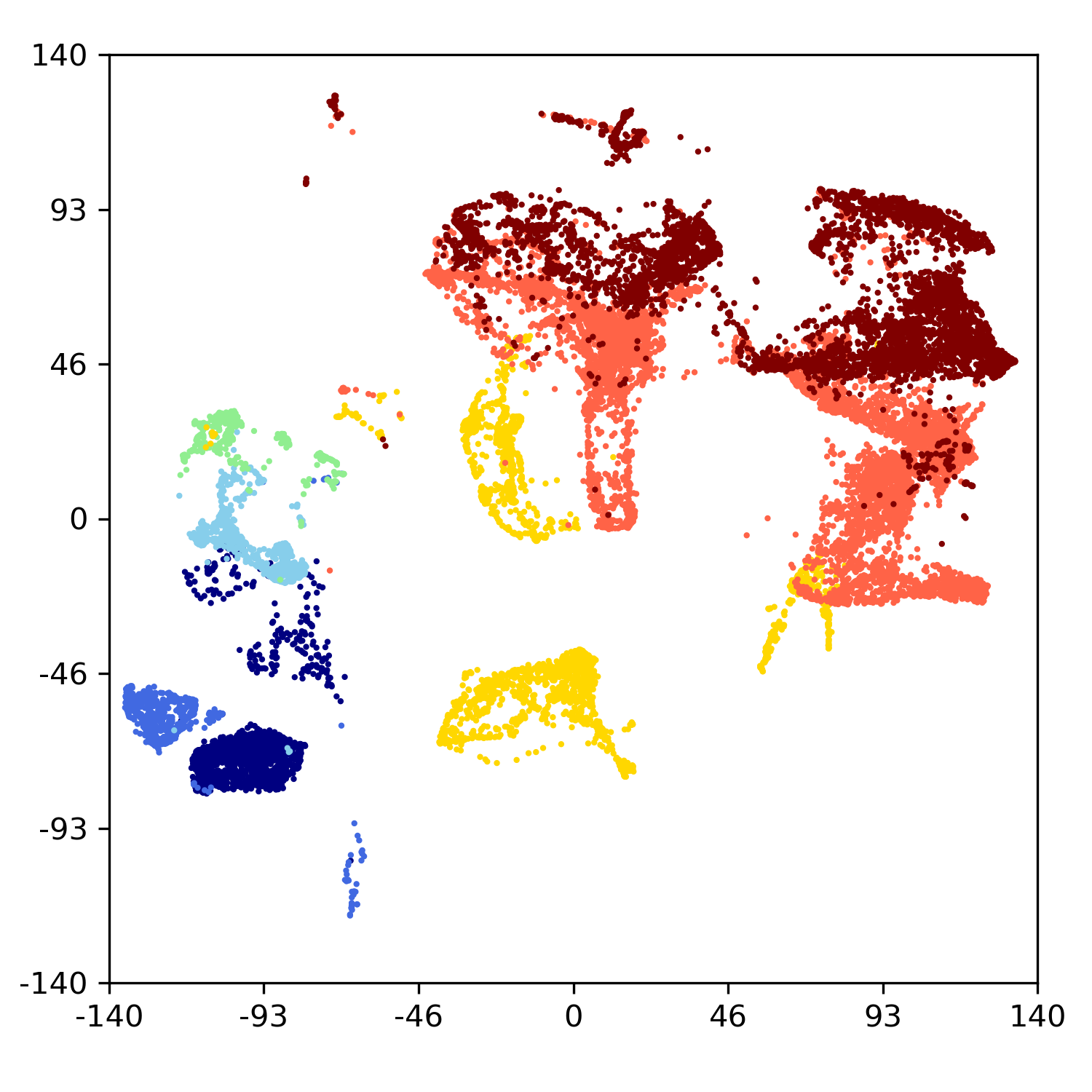}
        \caption*{C+E \\ (TriMap, $\alpha = 1.95$)}
    \end{subfigure}\hfill
    \begin{subfigure}[b]{0.18\textwidth}
        \centering
        \includegraphics[width=\textwidth]{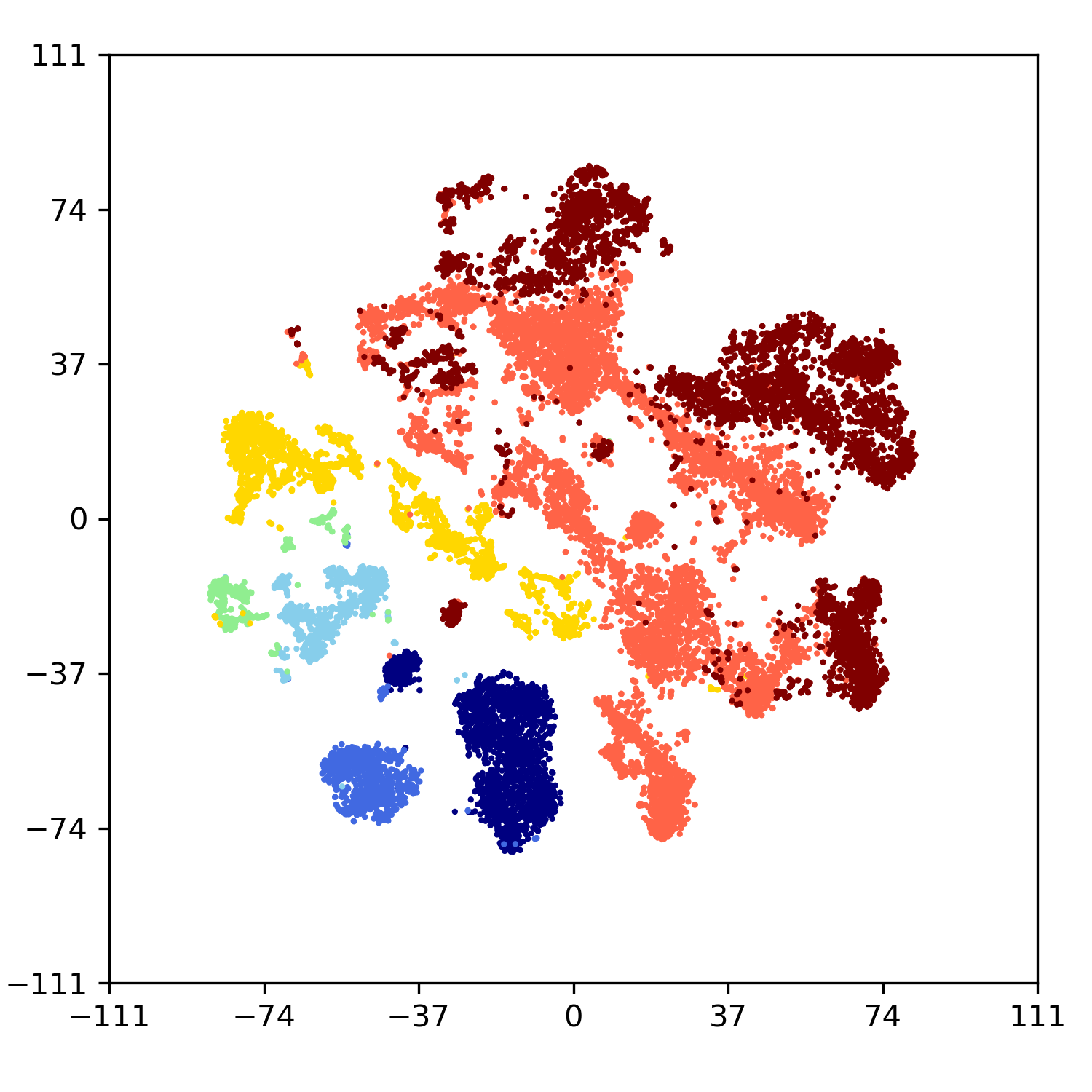}
        \caption*{t-SNE  \\  \strut}
    \end{subfigure}\hfill
    \begin{subfigure}[b]{0.18\textwidth}
        \centering
        \includegraphics[width=\textwidth]{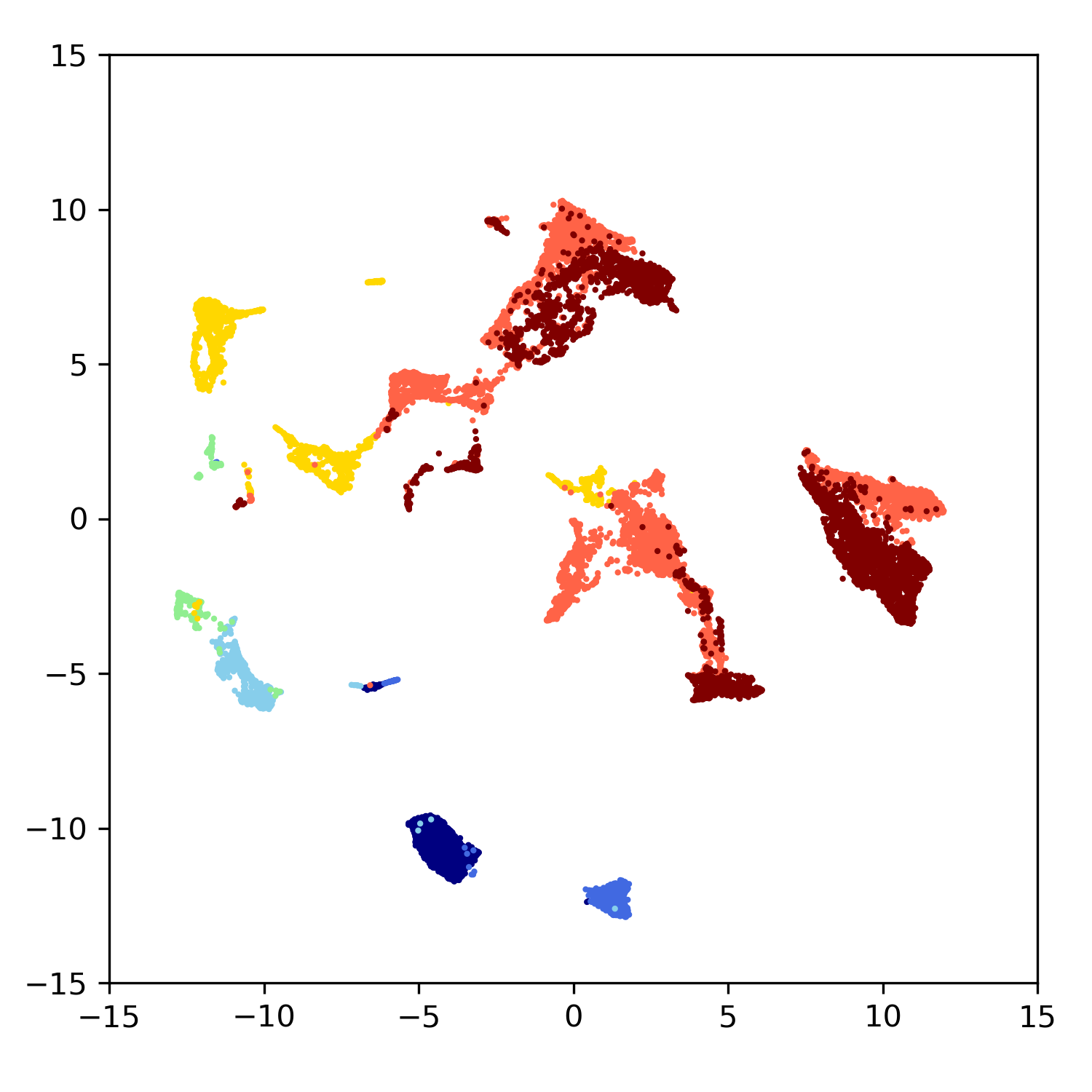}
        \caption*{UMAP  \\  \strut}
    \end{subfigure}

% ----- Row 2 -----
    \begin{subfigure}[b]{0.18\textwidth}
        \centering
        \includegraphics[width=\textwidth]{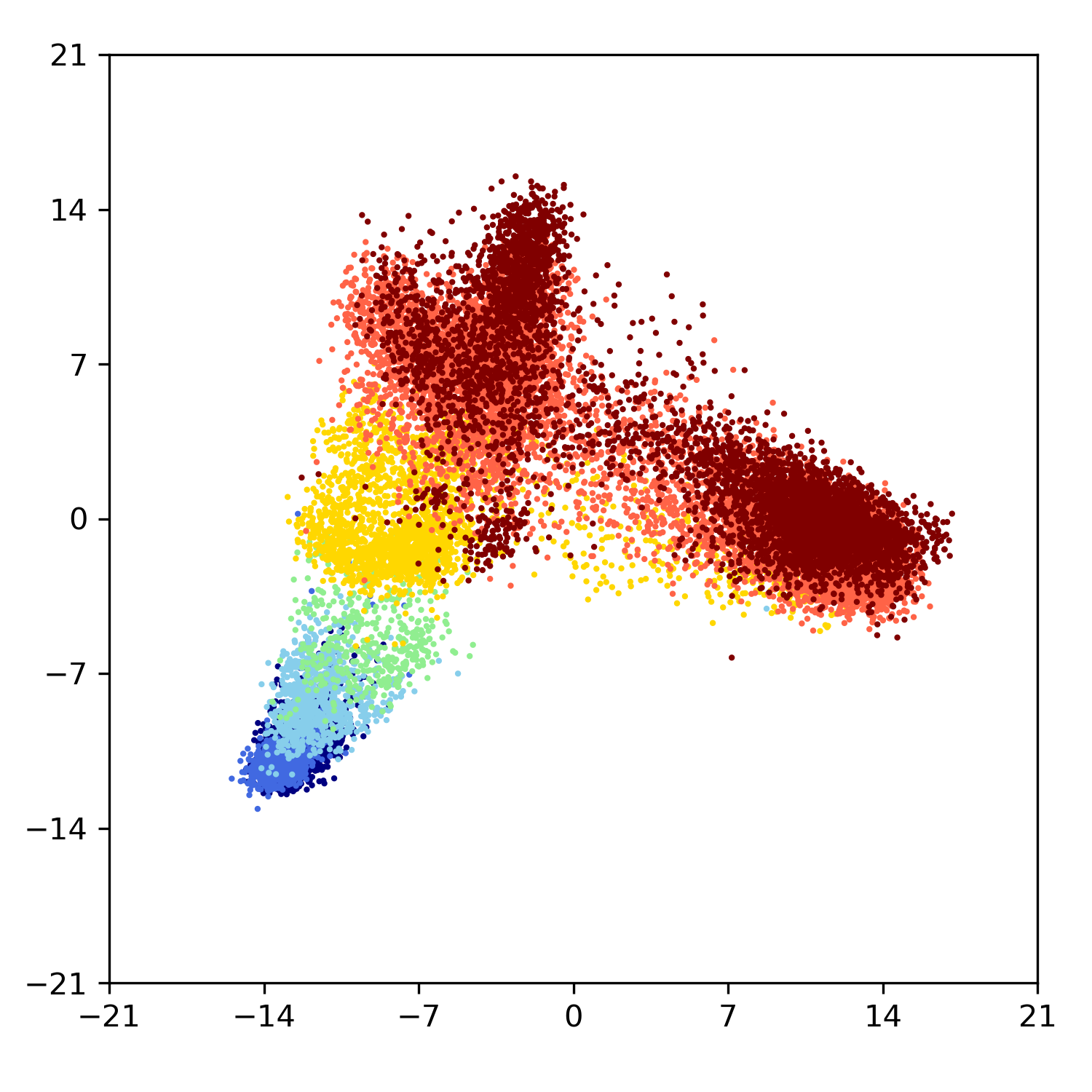}
        \caption*{PCA}
    \end{subfigure}
        \hfill
    \begin{subfigure}[b]{0.18\textwidth}
        \centering
        \includegraphics[width=\textwidth]{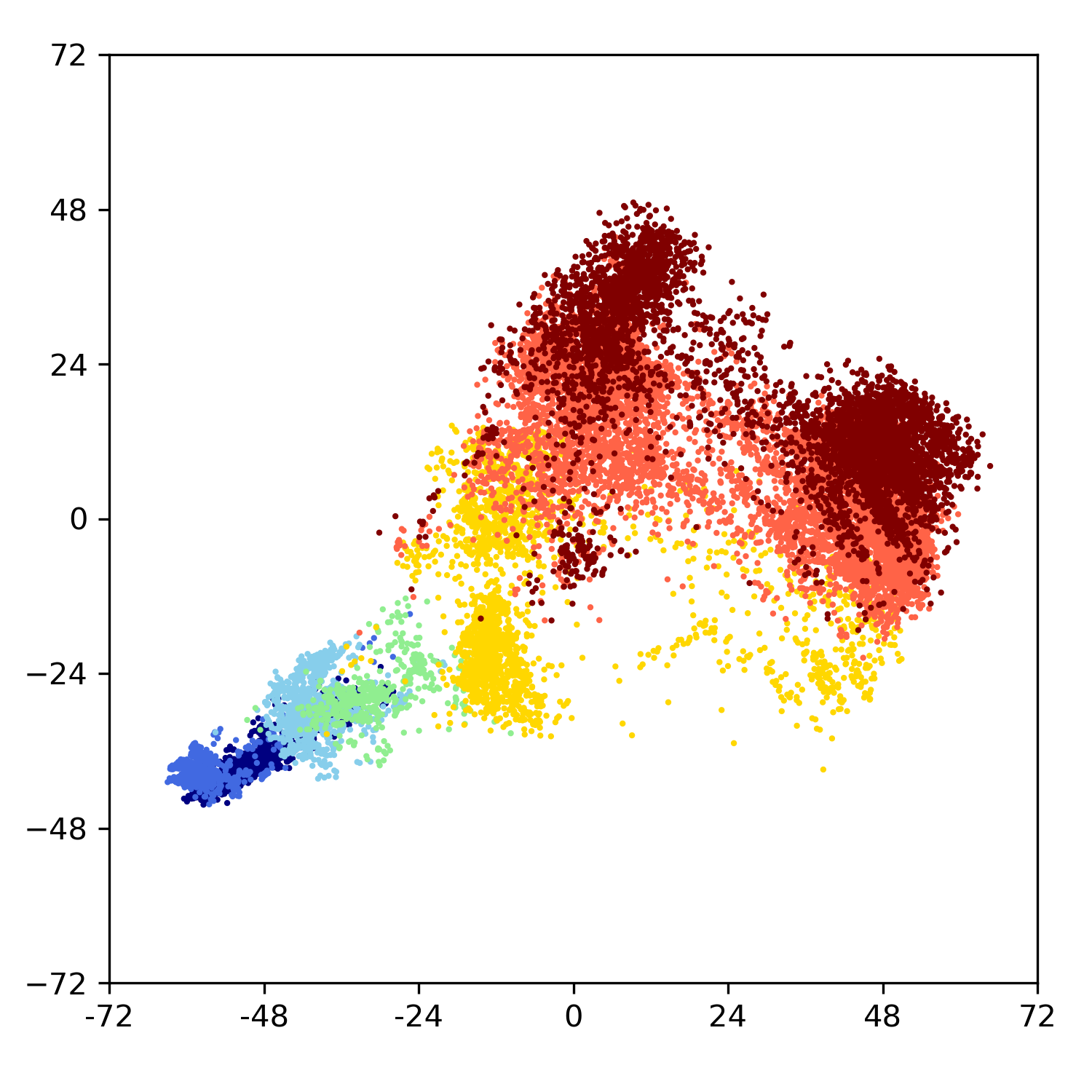}
        \caption*{L-Isomap}
    \end{subfigure}
    \hfill
    \begin{subfigure}[b]{0.18\textwidth}
        \centering
        \includegraphics[width=\textwidth]{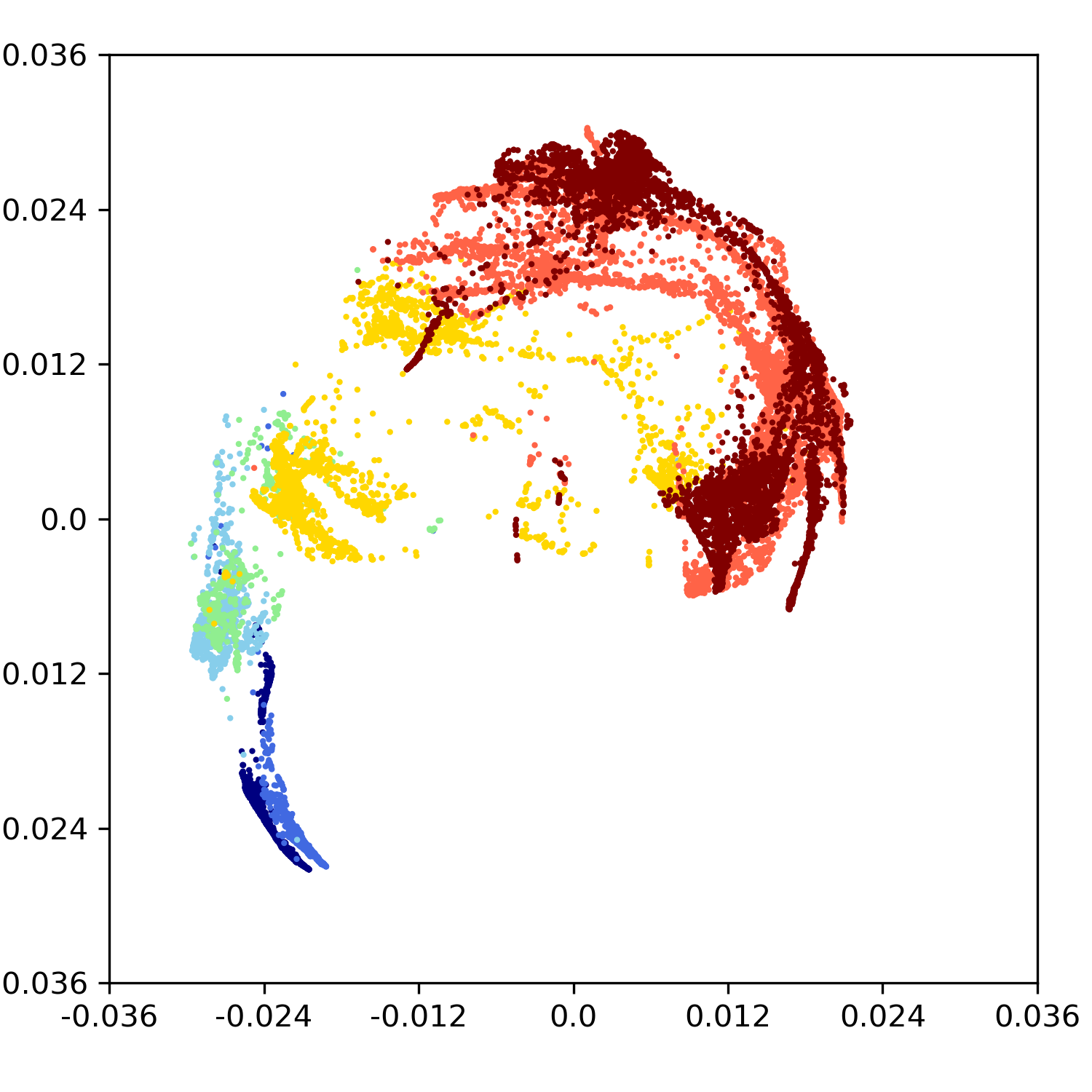}
        \caption*{PHATE}
    \end{subfigure}
    \hfill
    \begin{subfigure}[b]{0.18\textwidth}
        \centering
        \includegraphics[width=\textwidth]{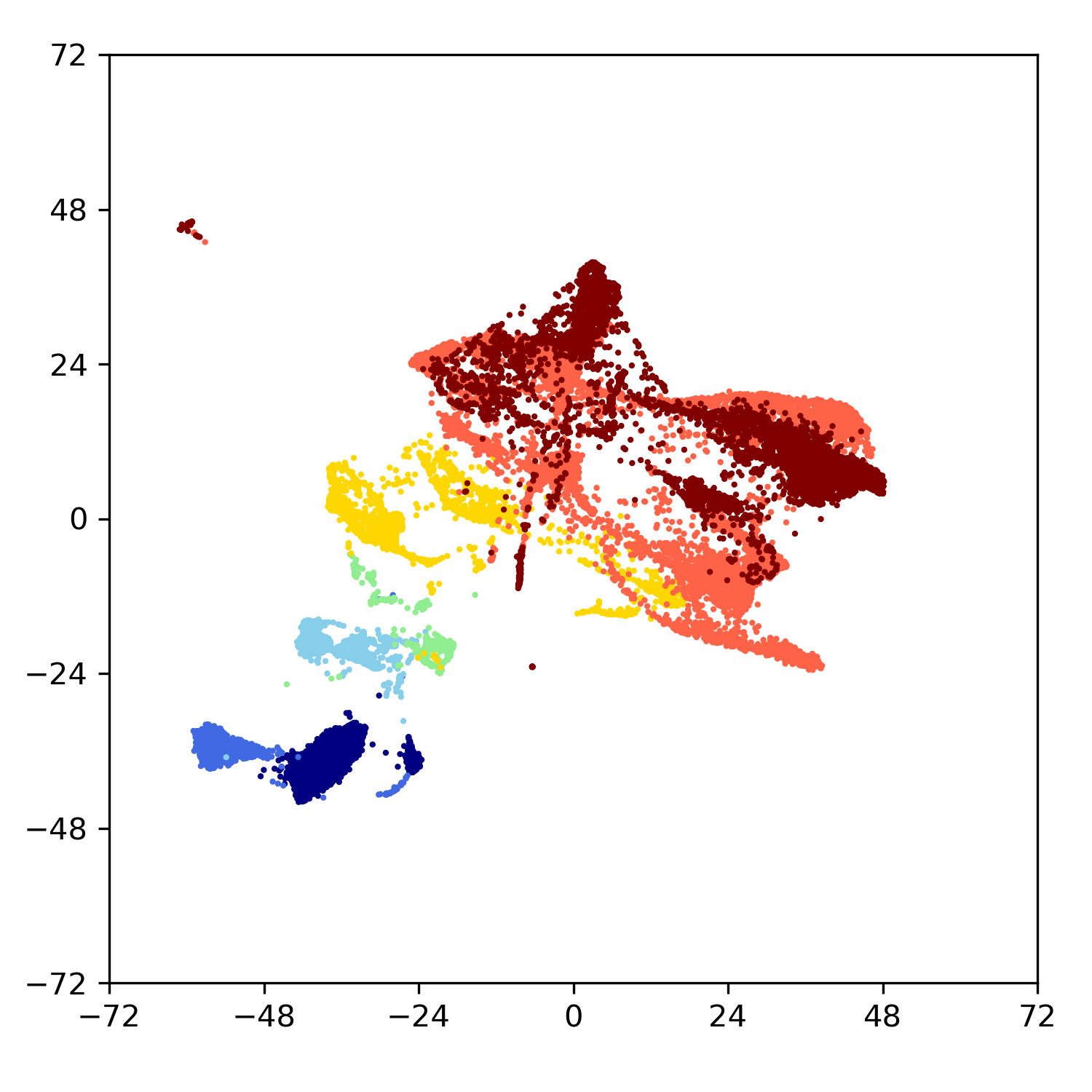}
        \caption*{TriMap}
    \end{subfigure}
    \hfill
    \begin{subfigure}[b]{0.18\textwidth}
        \centering
    \includegraphics[width=.3\textwidth]{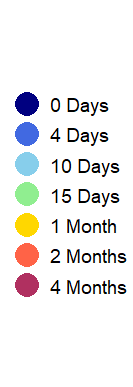}
         \caption*{}
    \end{subfigure}
    \caption{Embedding of human brain organoid dataset ($n = 20,272$).  Points are colored by the time points they were sampled at. These labels were not used by the embedding methods.}
    \label{fig:brain}
\end{figure}

\begin{figure}[h!]
    \centering
    \captionsetup{justification=centering, singlelinecheck=false, font = footnotesize}
    % ----- Row 1 -----
    \begin{subfigure}[b]{0.18\textwidth}
        \centering
        \includegraphics[width=\textwidth]{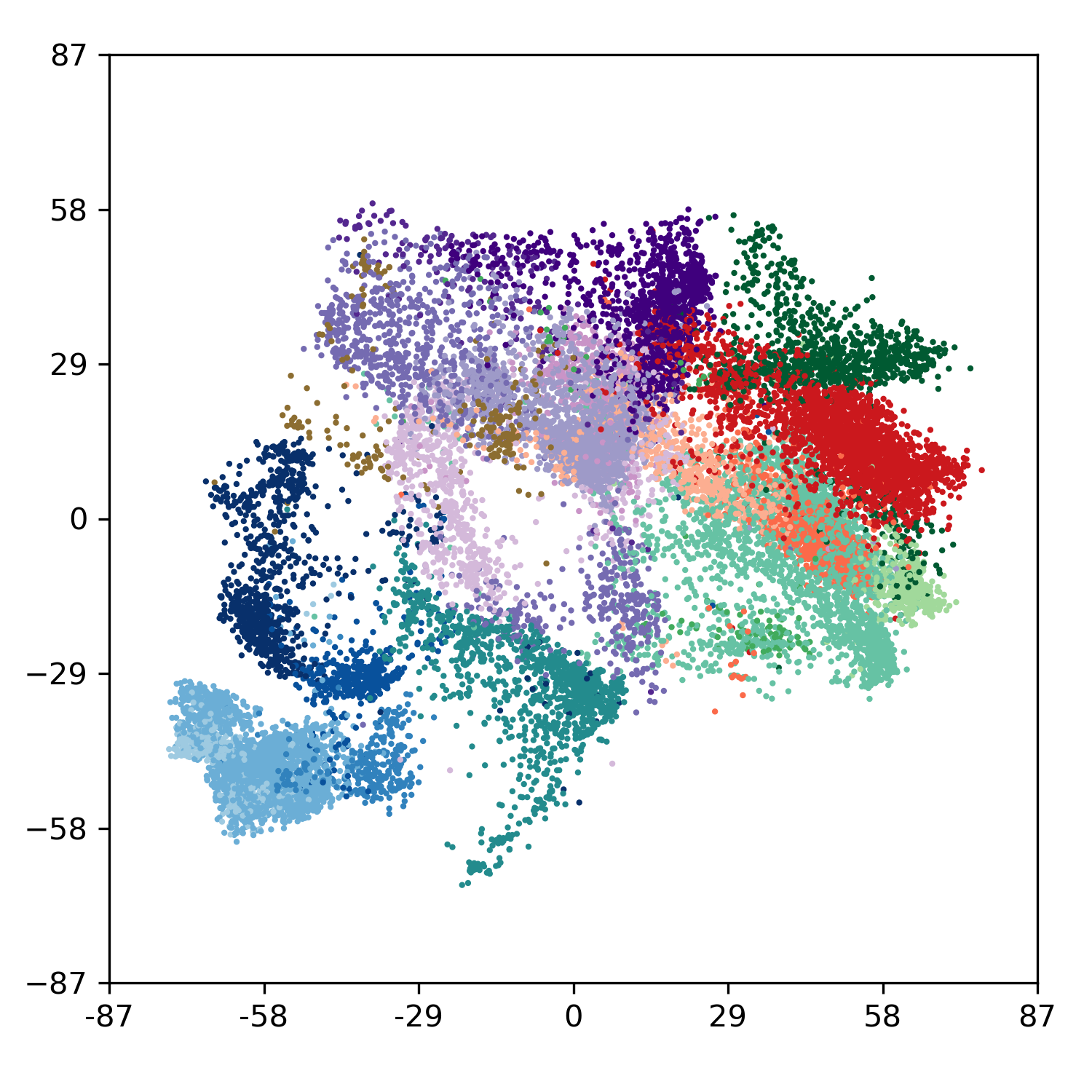}
        \caption*{C+E \\ (L-Isomap, $\alpha=1$)}
    \end{subfigure}\hfill
    \begin{subfigure}[b]{0.18\textwidth}
        \centering
        \includegraphics[width=\textwidth]{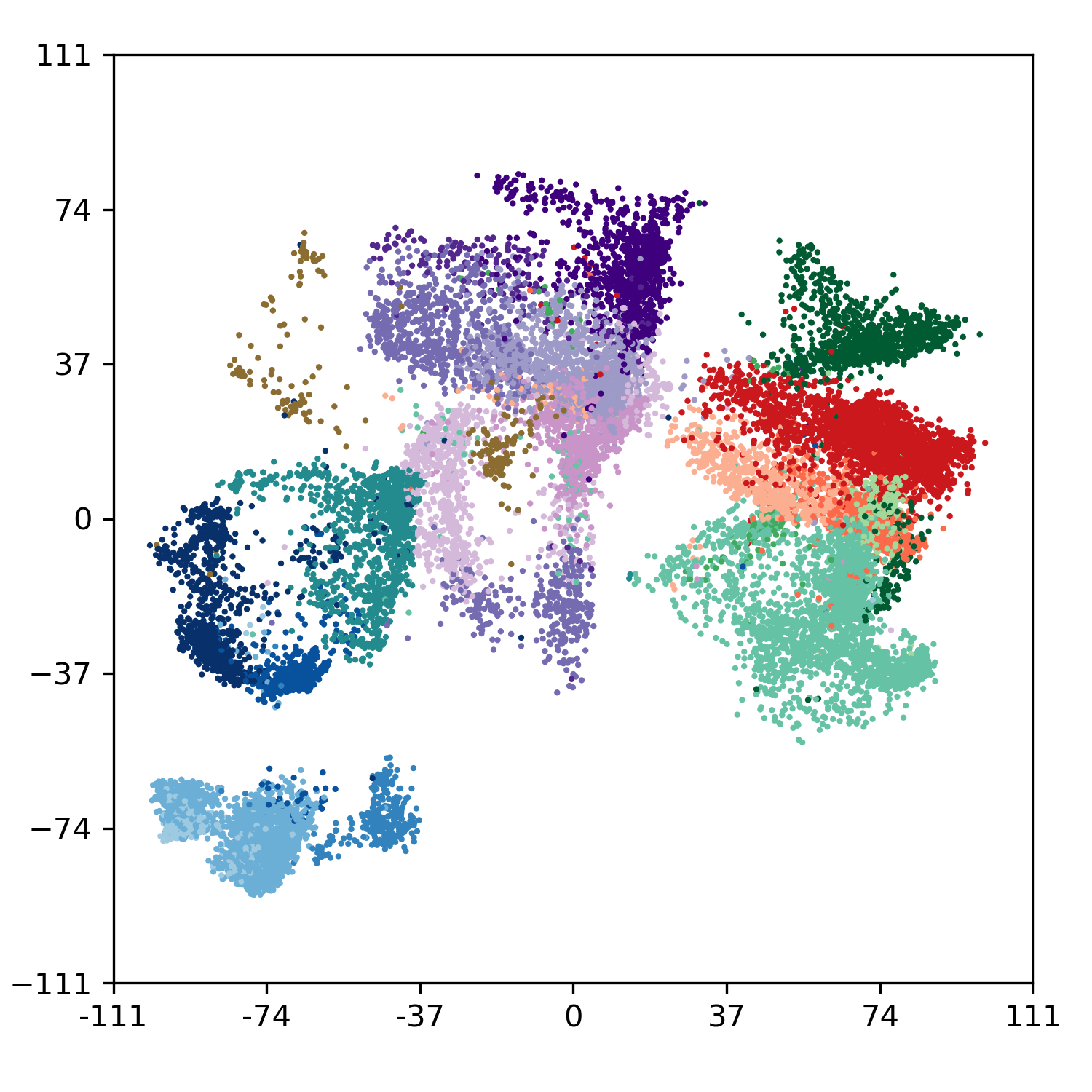}
        \caption*{C+E \\ (L-Isomap, $\alpha=1.45$)}
    \end{subfigure}\hfill
    \begin{subfigure}[b]{0.18\textwidth}
        \centering
        \includegraphics[width=\textwidth]{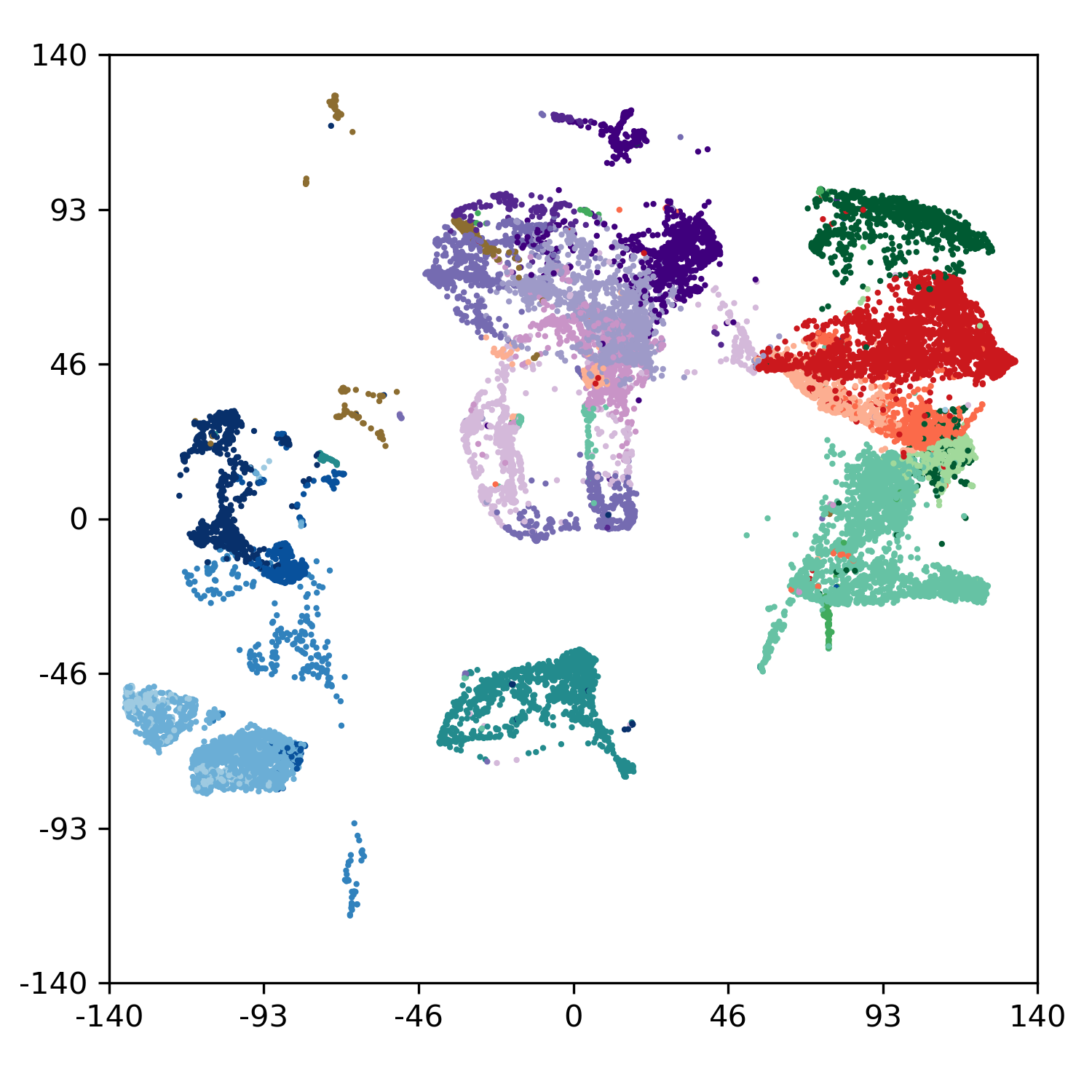}
        \caption*{C+E \\ (TriMap, $\alpha = 1.95$)}
    \end{subfigure}\hfill
    \begin{subfigure}[b]{0.18\textwidth}
        \centering
        \includegraphics[width=\textwidth]{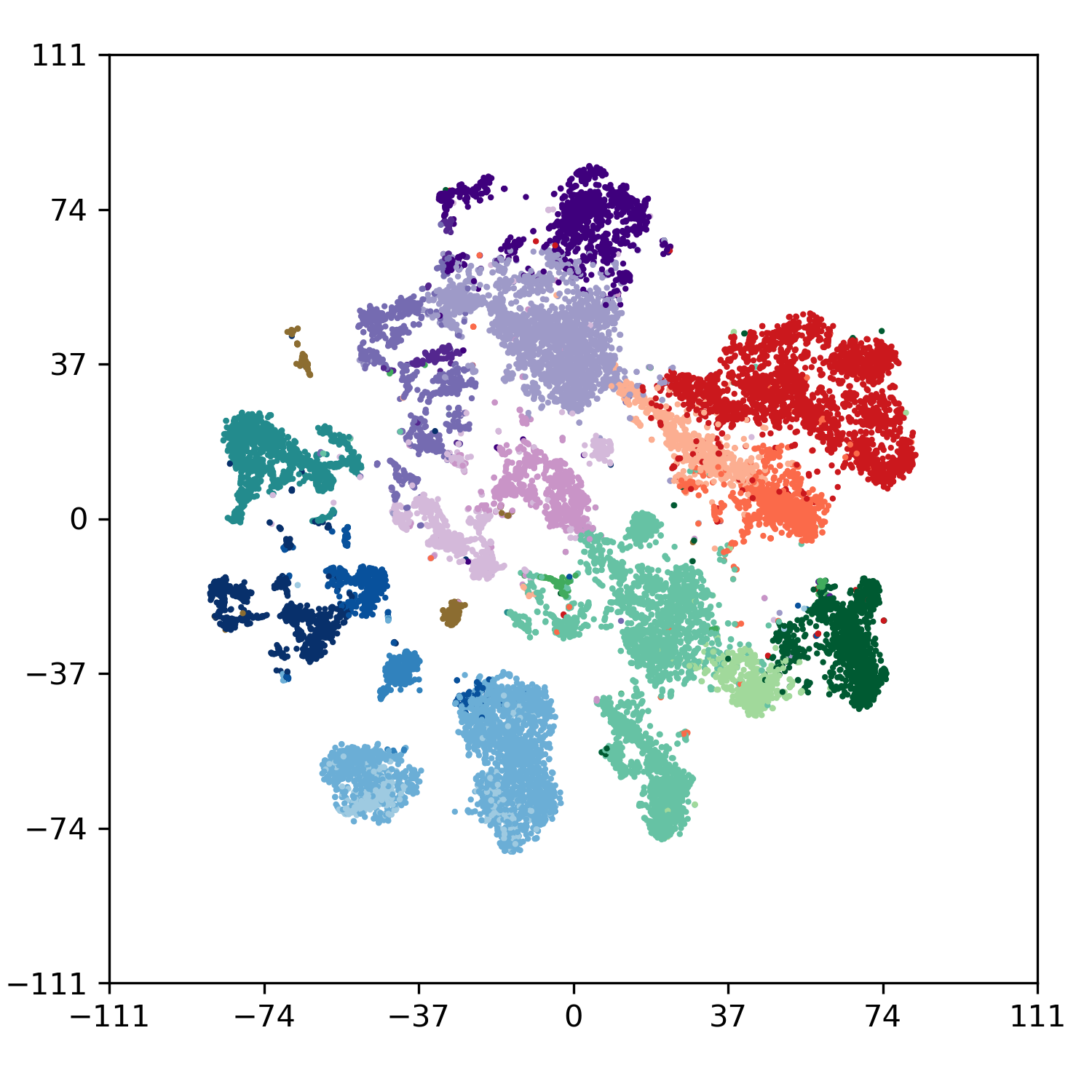}
        \caption*{t-SNE  \\  \strut}
    \end{subfigure}\hfill
    \begin{subfigure}[b]{0.18\textwidth}
        \centering
        \includegraphics[width=\textwidth]{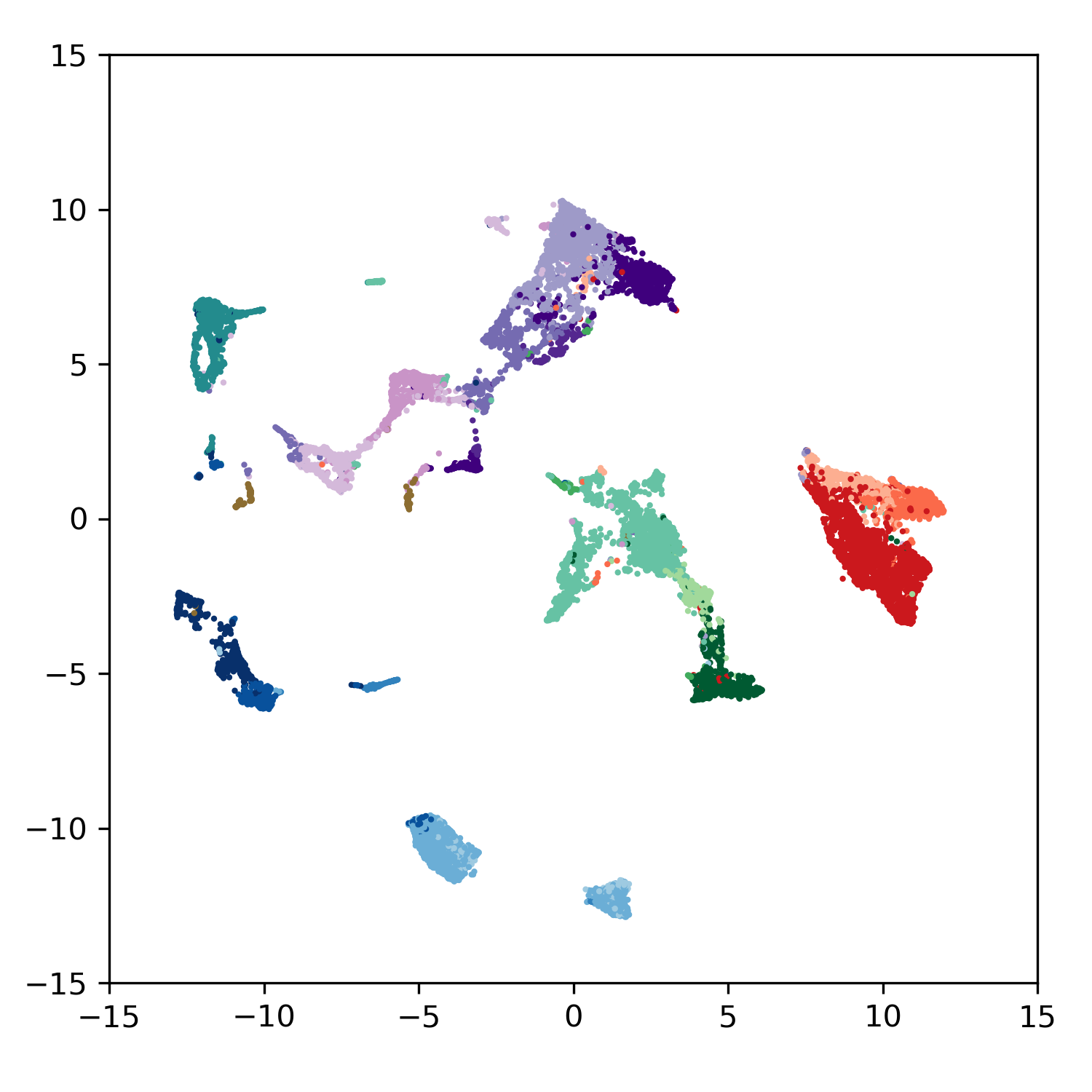}
        \caption*{UMAP  \\  \strut}
    \end{subfigure}

% ----- Row 2 -----
    \begin{subfigure}[b]{0.18\textwidth}
        \centering
        \includegraphics[width=\textwidth]{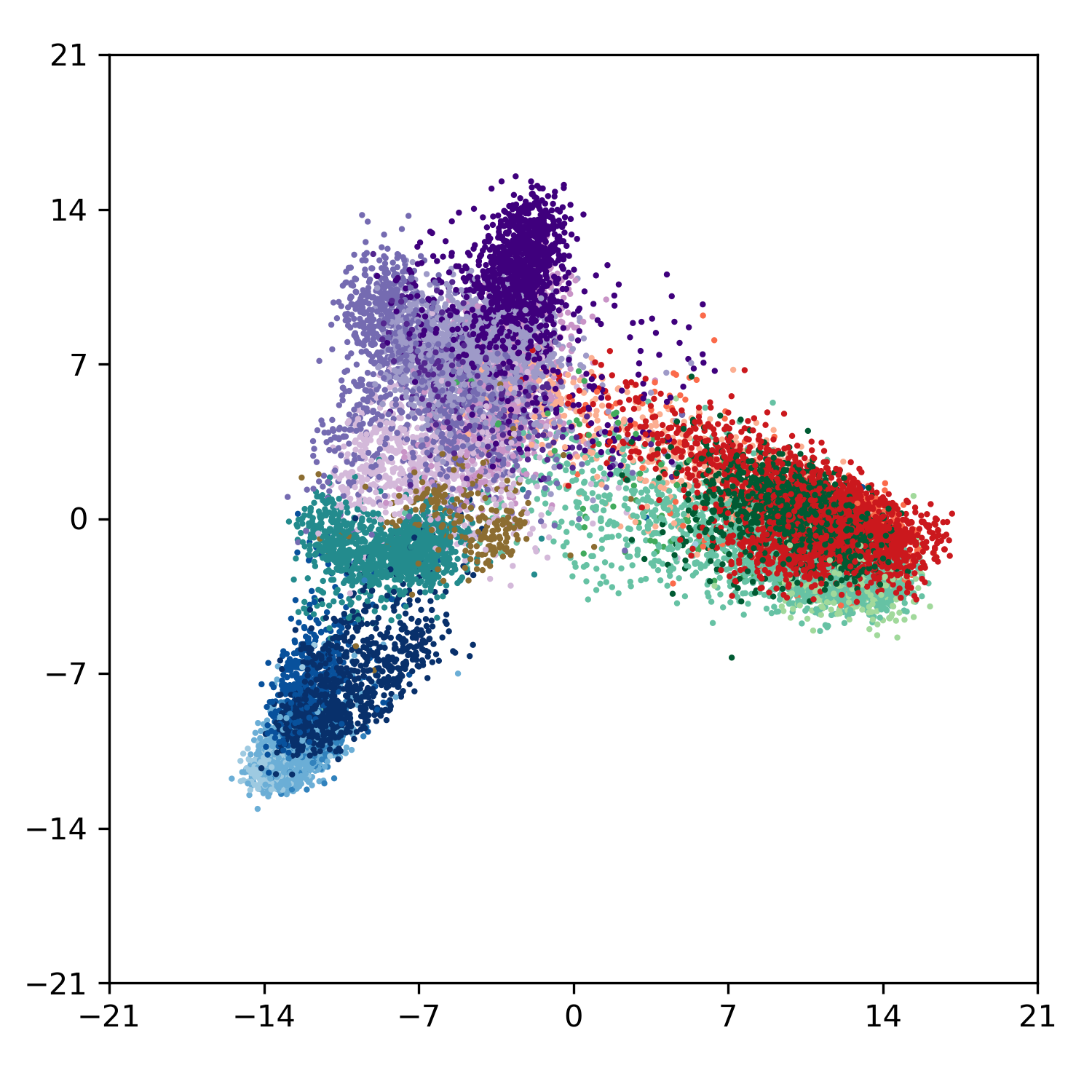}
        \caption*{PCA}
    \end{subfigure}
       \hfill
    \begin{subfigure}[b]{0.18\textwidth}
        \centering
        \includegraphics[width=\textwidth]{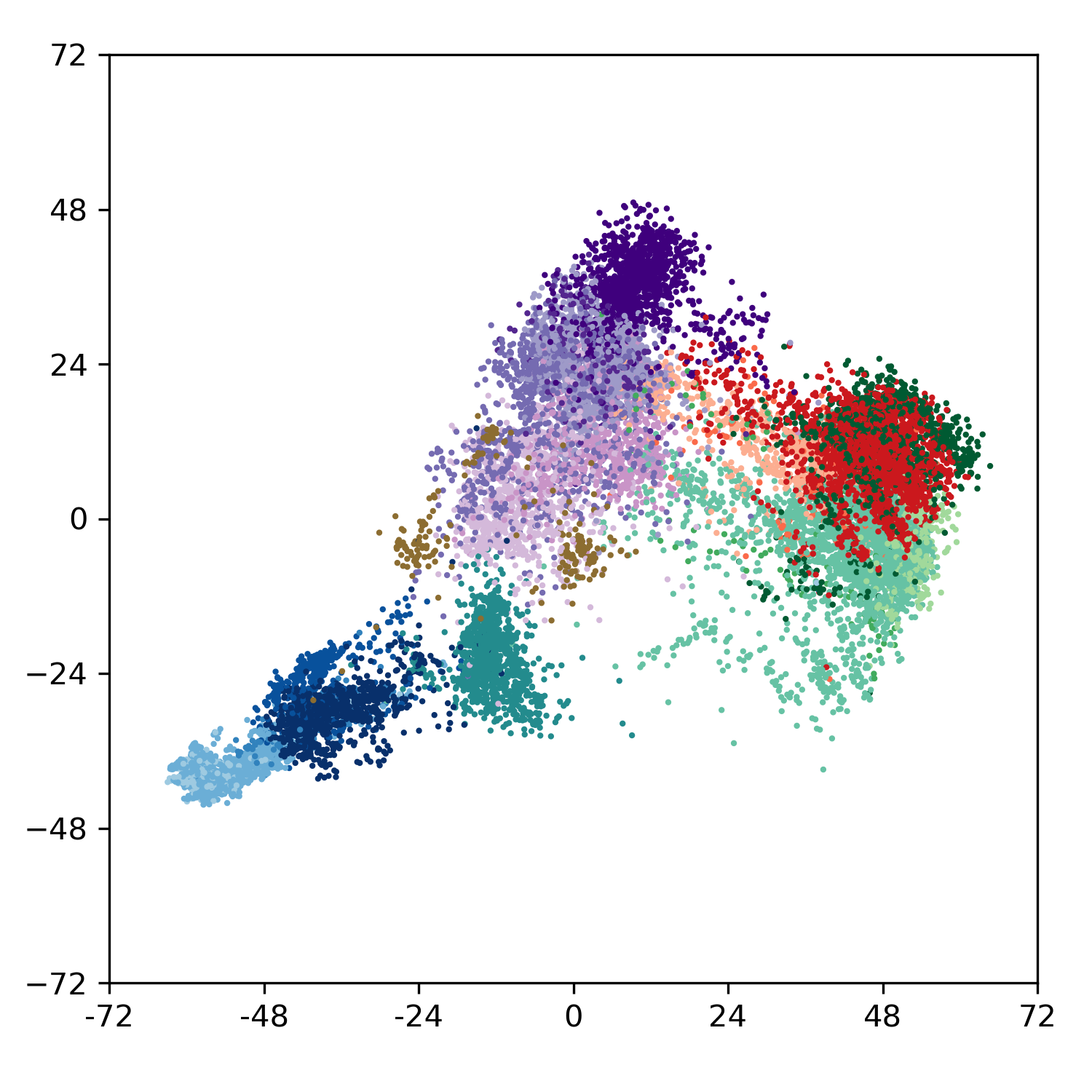}
        \caption*{L-Isomap}
    \end{subfigure}
    \hfill
    \begin{subfigure}[b]{0.18\textwidth}
        \centering
        \includegraphics[width=\textwidth]{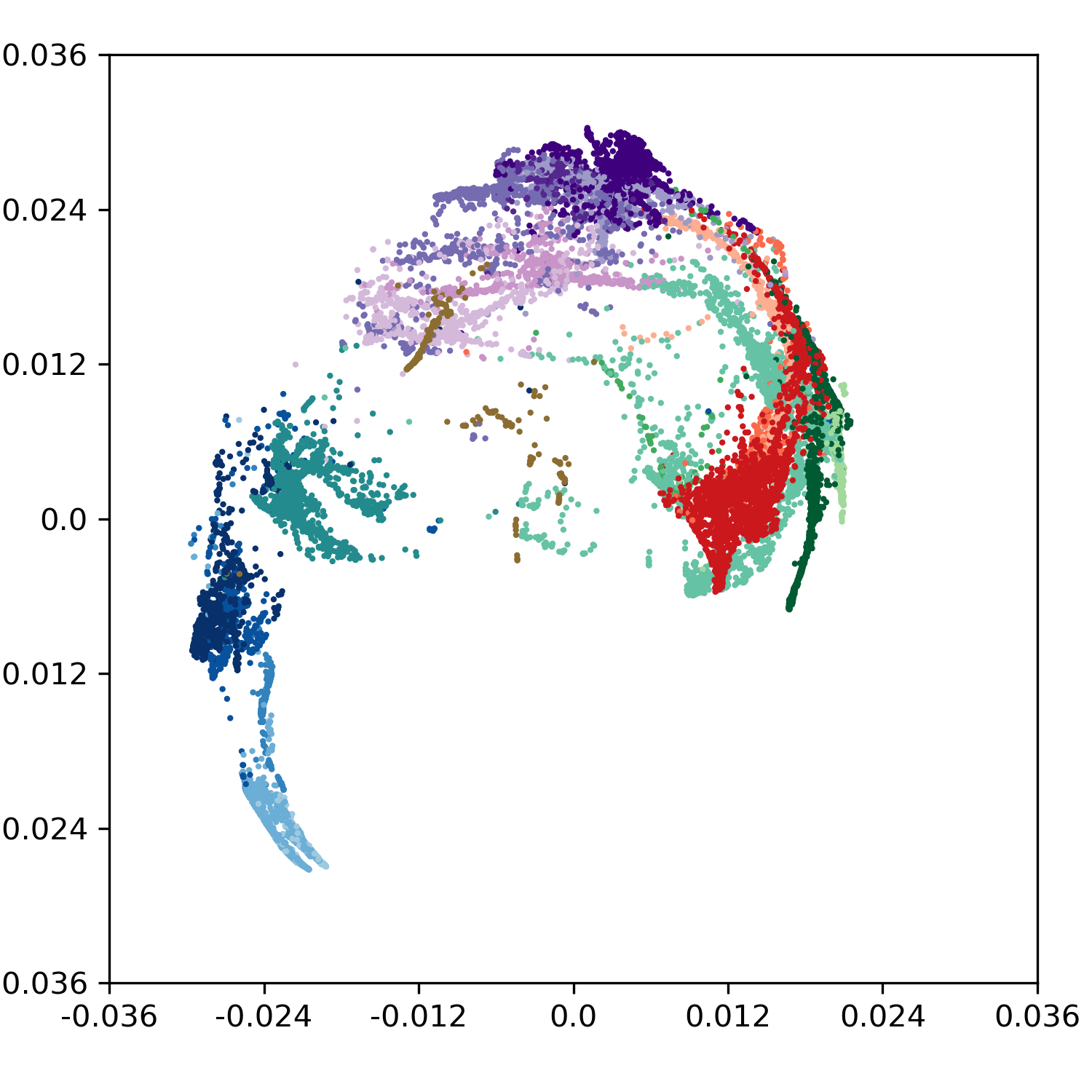}
        \caption*{PHATE}
    \end{subfigure}
    \hfill
    \begin{subfigure}[b]{0.18\textwidth}
        \centering
        \includegraphics[width=\textwidth]{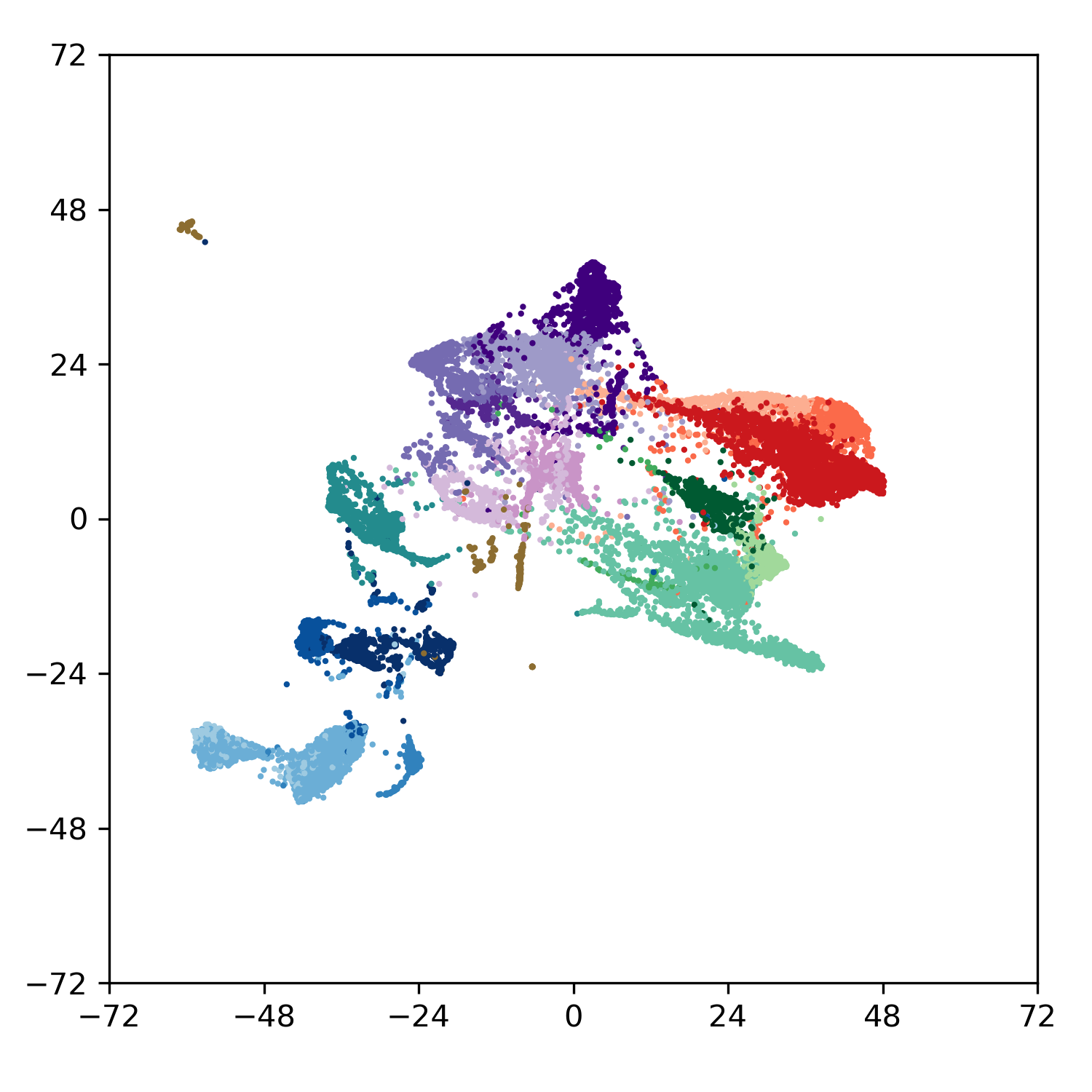}
        \caption*{TriMap}
    \end{subfigure}
     \hfill
    \begin{subfigure}[b]{0.18\textwidth}
        \centering
        \includegraphics[width=.8\textwidth]{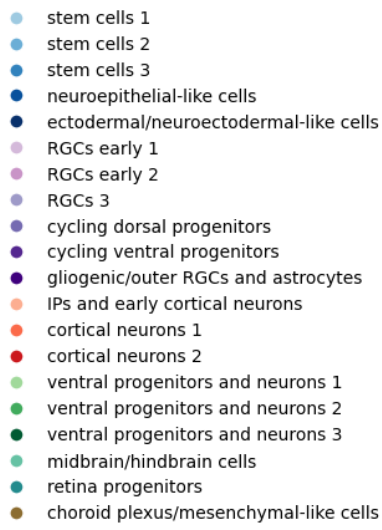}
         \caption*{}
    \end{subfigure}

    \caption{Embedding of human brain organoid dataset ($n = 20,272$).  Points are colored by cell types. These labels were not used by the embedding methods.}
    \label{fig:brain_cell}
\end{figure}

\begin{table}[!ht]
\centering
\footnotesize
\setlength{\tabcolsep}{2pt}
\begin{tabular}{llccccccccccc}
\toprule
\textbf{Method} & \textbf{Params} 
& \multicolumn{4}{c}{\textbf{Local}} 
& \multicolumn{7}{c}{\textbf{Global}} \\
\cmidrule(lr){3-6} \cmidrule(lr){7-13}
& 
& W-S ${\scriptstyle \uparrow}$ 
& W-NS ${\scriptstyle \downarrow}$ 
& W-SNS ${\scriptstyle \downarrow}$ 
& KNN ${\scriptstyle \uparrow}$
& B-S ${\scriptstyle \uparrow}$  
& T-S ${\scriptstyle \uparrow}$  
& B-NS ${\scriptstyle \downarrow}$ 
& T-NS ${\scriptstyle \downarrow}$ 
& B-SNS ${\scriptstyle \downarrow}$ 
& T-SNS ${\scriptstyle \downarrow}$  
& CP ${\scriptstyle \uparrow}$
 \\
\midrule

C+E (L-Isomap) & $\alpha = 1$ &
$0.76$ & $0.50$ & $\mathbf{0.36}$ & $0.30$ & $0.65$ & $0.88$ & $0.37$ & $0.38$ & $\mathbf{0.20}$ & $0.24$ & $\mathbf{0.95}$\\

C+E  (L-Isomap) & $\alpha = 1.45$ &
$\mathbf{0.80}$ & $0.44$ & $0.38$ & $0.35$ & $\mathbf{0.68}$ & $0.90$ & $\mathbf{0.23}$ & $\mathbf{0.23}$ & $\mathbf{0.20}$ & $\mathbf{0.23}$ & $0.93$ \\

C+E  (TriMap) & $\alpha = 1.95$ &
$0.79$ & $0.50$ & $0.43$ & $0.43$ & $0.63$ & $\mathbf{0.91}$ & $0.40$ & $0.40$ & $\mathbf{0.20}$ & $\mathbf{0.23}$ & $0.91$ \\

t-SNE & $u = 30$ &
$0.77$ & $\mathbf{0.43}$ & $0.38$ & $\mathbf{0.51}$ & $0.30$ & $0.68$ & $0.34$ & $0.35$ & $0.26$ & $0.32$ & $0.84$
 \\
 
 UMAP & $q = 15$ &
$0.78$ & $0.91$ & $0.46$ & $0.47$ & $0.12$ & $0.67$ & $0.86$ & $0.88$ & $0.29$ & $0.34$ & $0.63$\\

PCA & - &
$0.59$ & $0.89$ & $0.45$ & $0.14$ & $0.53$ & $0.85$ & $0.87$ & $0.85$ & $0.27$ & $0.28$ & $0.91$ \\

L-Isomap & $q = 15$  &
$0.68$ & $0.66$ & $0.41$ & $0.18$ & $0.55$ & $\mathbf{0.91}$ & $0.51$ & $0.46$ & $0.24$ & $0.27$ & $0.93$
 \\

PHATE & $q = 5$ &
$0.71$ & $1.00$ & $0.44$ & $0.29$ & $0.11$ & $0.86$ & $1.00$ & $1.00$ & $0.30$ & $0.29$ & $0.87$ \\

TriMap & - &
$\mathbf{0.80}$ & $0.69$ & $\mathbf{0.36}$ & $0.43$ & $0.44$ & $0.87$ & $0.56$ & $0.55$ & $0.22$ & $0.26$ & $0.91$\\

\bottomrule
\end{tabular}
\caption{Evaluation of the human brain organoid dataset against geodesic distances for $5000$ points selected independently of the points used in alignment. Time points are taken as the ground truth class labels. Metrics are reported for within class (W), between class (B), and total (T). S denotes Spearman correlation, NS normalized stress, SNS scale-normalized stress, CP class preservation, and KNN the $30$-NN recall. Arrows indicate whether higher ($\uparrow$) or lower ($\downarrow$) values are better.}
\label{tab:brain}
\end{table}

\begin{figure}[ht!]
    \centering
    \begin{subfigure}[c]{0.45\textwidth}
        \centering
        \includegraphics[width=\textwidth]{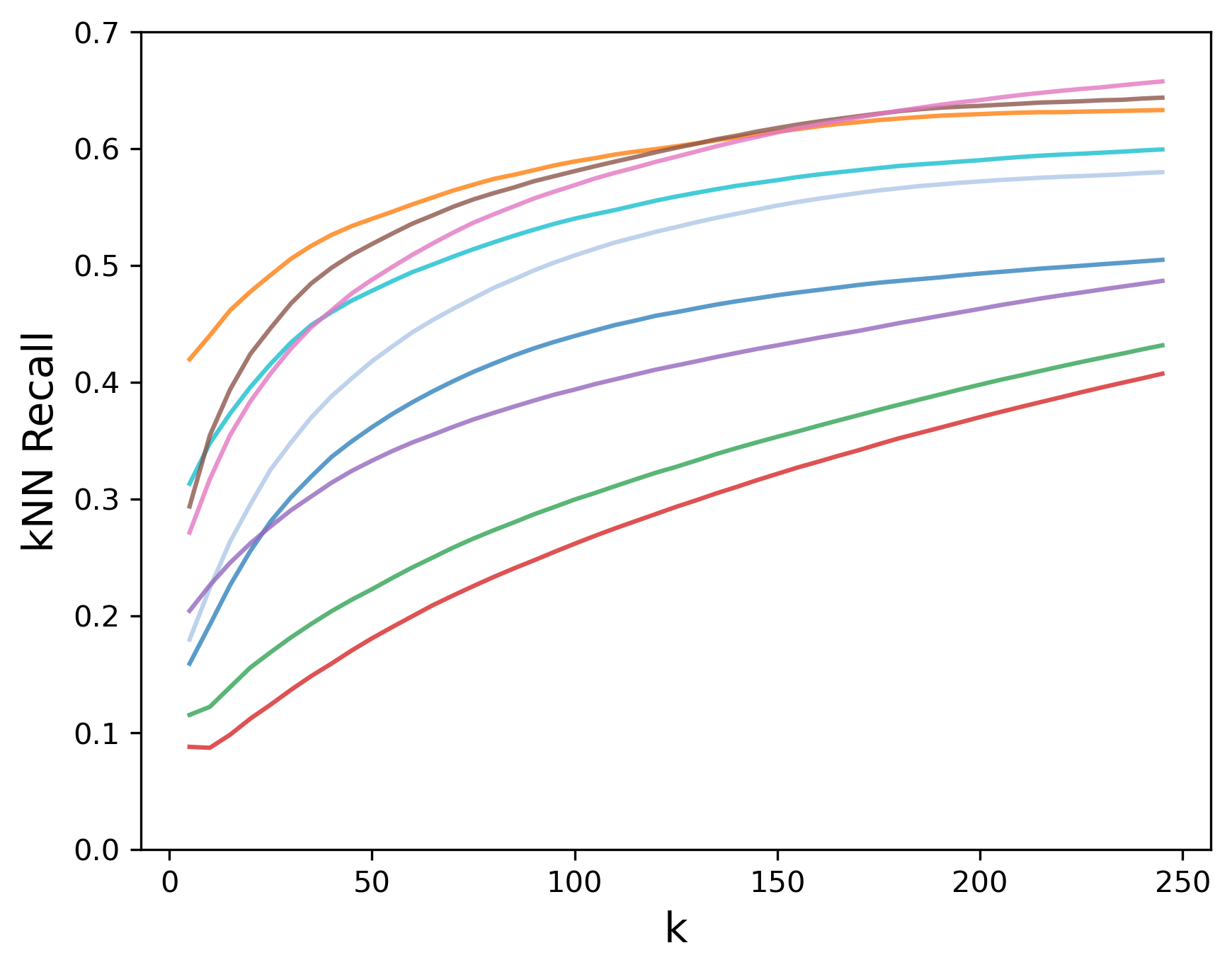}
    \end{subfigure}
    \hspace{0.03\textwidth}
    \begin{subfigure}[c]{0.2\textwidth}
        \centering
        \includegraphics[width=\textwidth]{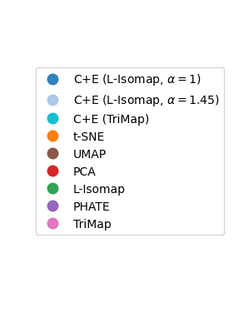}
    \end{subfigure}
    \caption{$k$NN recall for the human brain organoid dataset for various values of $k$ for all methods compared.}
    \label{fig:brain_knn}
\end{figure}

As another real-world example, we use the developmental human brain organoid data from \citet{kanton2019organoid}, using the same preprocessing as \cite{bohm2022, damrich2024}. The dataset has two sets of class labels, corresponding to seven time points the cells were sampled at, and twenty cell-type labels. We include plots colored by each set of labels. Here we evaluate the class-dependent metrics using the time labels as the ground truth classes, and in \appref{brain_cell} we use the cell-type labels to evaluate these metrics. For the C+E approach, we cluster the data using the Leiden algorithm ($k = 15$, $\gamma = 0.1$) and obtain thirteen clusters. We embed each cluster using either L-Isomap ($k = 15 , n_\text{landmarks} = 300$) or TriMap, and align the clusters to preserve geodesic distances with $\alpha$ set to $1.45$ for the L-Isomap embeddings and $1.95$ for the TriMap embeddings, obtained using \eqref{eq:alpha}. 
% The results are in \figref{brain} and evaluation against geodesic distances is in \tabref{brain}. 
% Euclidean-based embeddings and evaluation are in \appref{brain_app}. 

Qualitatively, in \figref{brain}, coloring the embedding by time points, the C+E (L-Isomap) embedding follows a natural clockwise progression across days and exhibits some branching into two groups each containing both two and four-month-old cells. When coloring by cell type in \figref{brain_cell}, we observe that one of these branches contains RGCs and cycling progenitors, and the other ventral progenitors and neurons, cortical neurons, and midbrain/hindbrain cells. Globally, the C+E embedding preserves the temporal developmental trajectory observed in the PCA and L-Isomap embeddings, and achieves global metrics comparable or better than PCA and L-Isomap. However, the C+E embedding attains better local metrics and exhibits improved local structure, with less collapse in the embedding that allows cell-type organization to emerge. 

The temporal developmental trajectory is less clear in the t-SNE and UMAP embeddings, though is still somewhat present. Both methods produce distinct clusters corresponding to the earlier time points, resulting in a progression that appears more discrete than continuous. t-SNE appears to organize the data by cell-type much better than it does by developmental time, while UMAP organizes the data by both developmental time and cell type. In particular, both the t-SNE and UMAP embeddings clearly separate the ventral progenitors and neurons and midbrain/hindbrain cells from the cortical neurons, which are much closer to each other in the C+E embedding, with the cortical neurons somewhat collapsed on other cells.  Embedding in three dimensions may help overcome this in the C+E embedding, though UMAP may also be somewhat overclustering and exaggerating the distances between clusters. The metrics suggest that C+E (L-Isomap, $\alpha = 1.45$) is more faithful to the true distances between cell types, as the between class global metrics and class preservation score all strongly favor the C+E embedding. This is true using the time labels as the ground truth labels in \tabref{brain}, and using the cell-types as the ground truth labels in \appref{brain_cell}.

The time progression is also evident in the PHATE and TriMap embeddings, with the metrics again favoring the C+E approach. Like PCA and L-Isomap, the different cell types are collapsed in the PHATE embedding. The TriMap embedding does separate the different cell types and also has strong local metrics, though as intended, C+E (TriMap) improves the $k$NN recall for small values of $k$ compared to TriMap (\figref{brain_knn}), while generally achieving comparable or better global metrics.

\subsection{Mouse cortex data}

\begin{figure}[h!]
    \centering
    \captionsetup{justification=centering, singlelinecheck=false, font = footnotesize}
    % ----- Row 1 -----
    \begin{subfigure}[b]{0.18\textwidth}
        \centering
        \includegraphics[width=\textwidth]{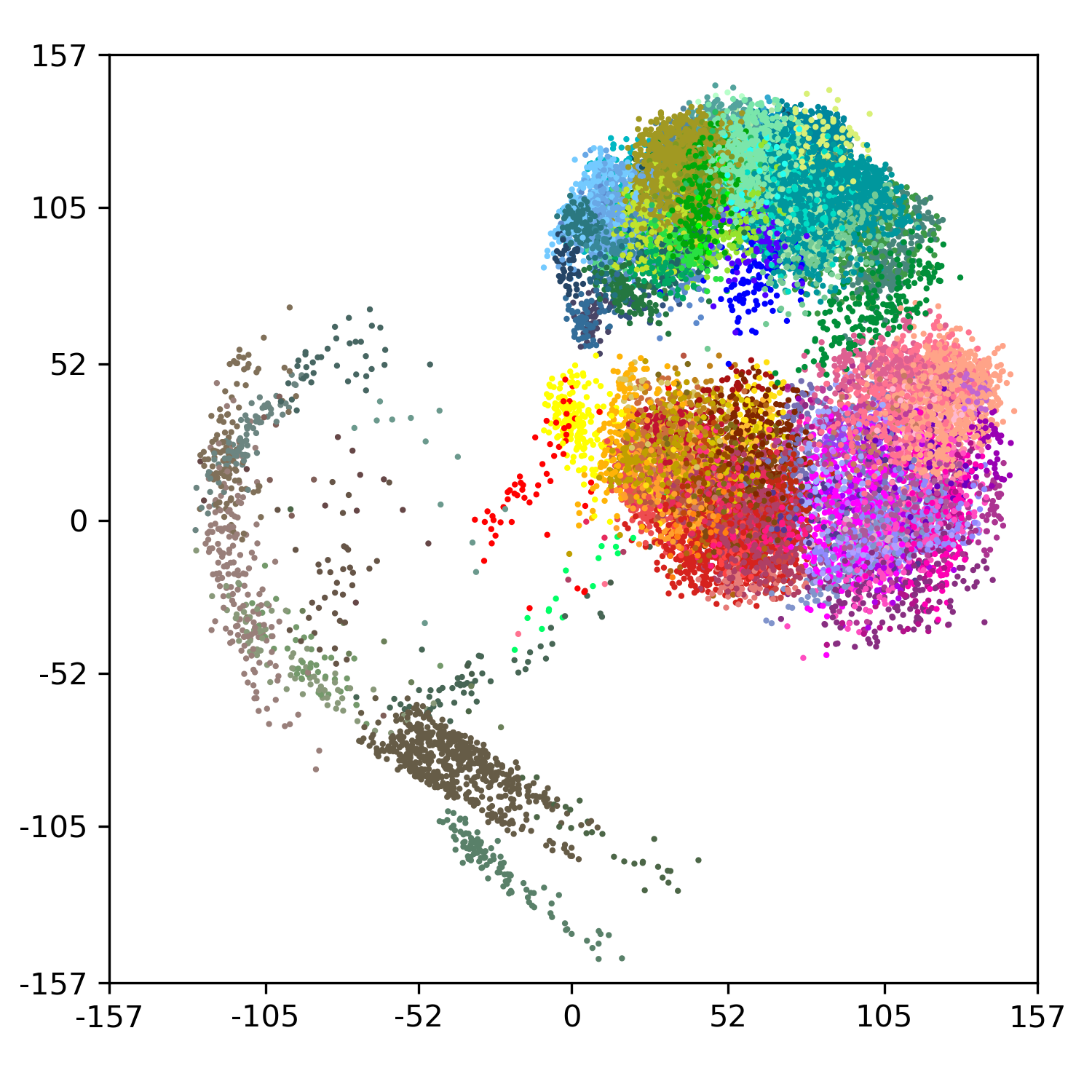}
        \caption*{C+E \\ (PCA, $\alpha=1$)}
    \end{subfigure}\hfill
    \begin{subfigure}[b]{0.18\textwidth}
        \centering
        \includegraphics[width=\textwidth]{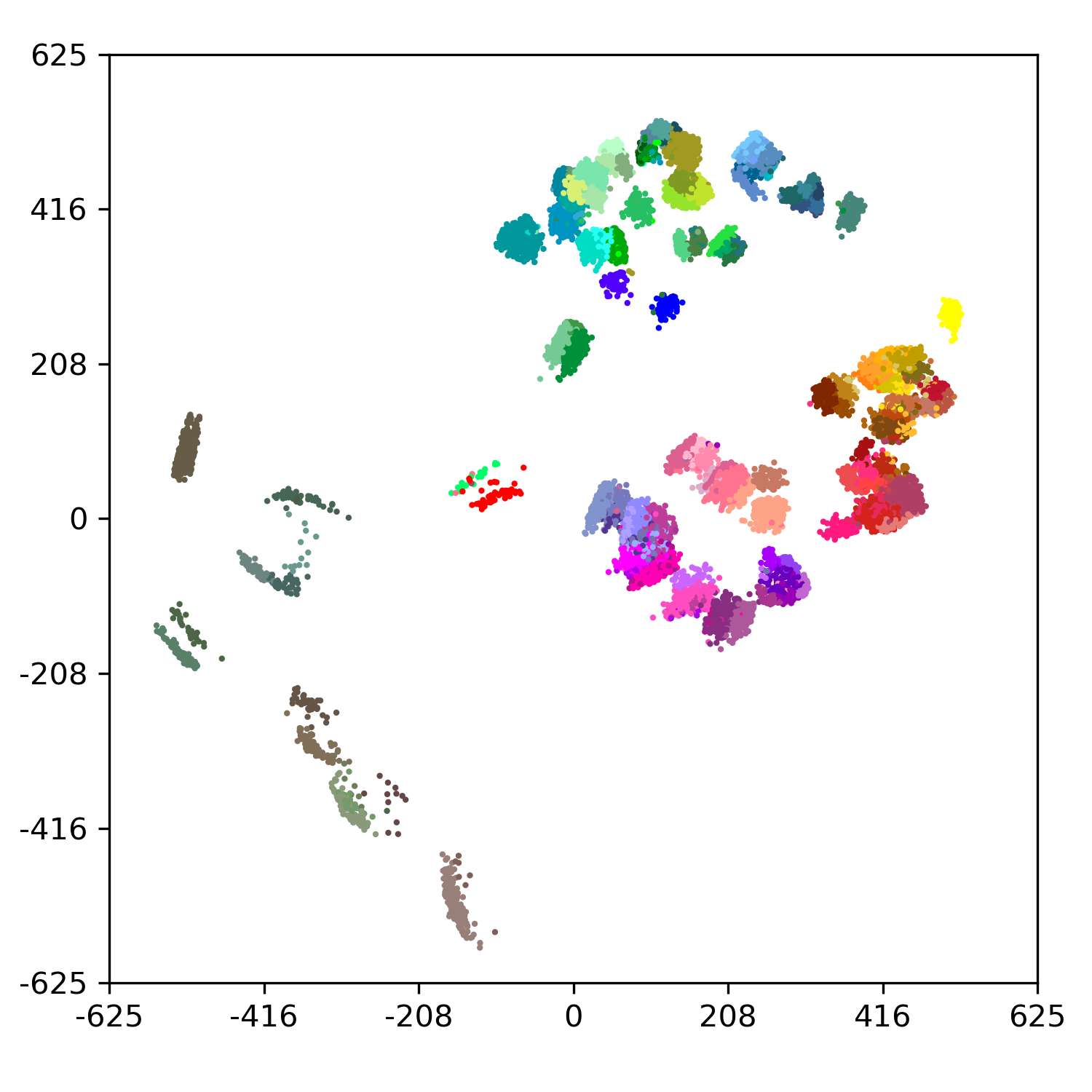}
        \caption*{C+E \\ (PCA, $\alpha=4.27$)}
    \end{subfigure}\hfill
    \begin{subfigure}[b]{0.18\textwidth}
        \centering
        \includegraphics[width=\textwidth]{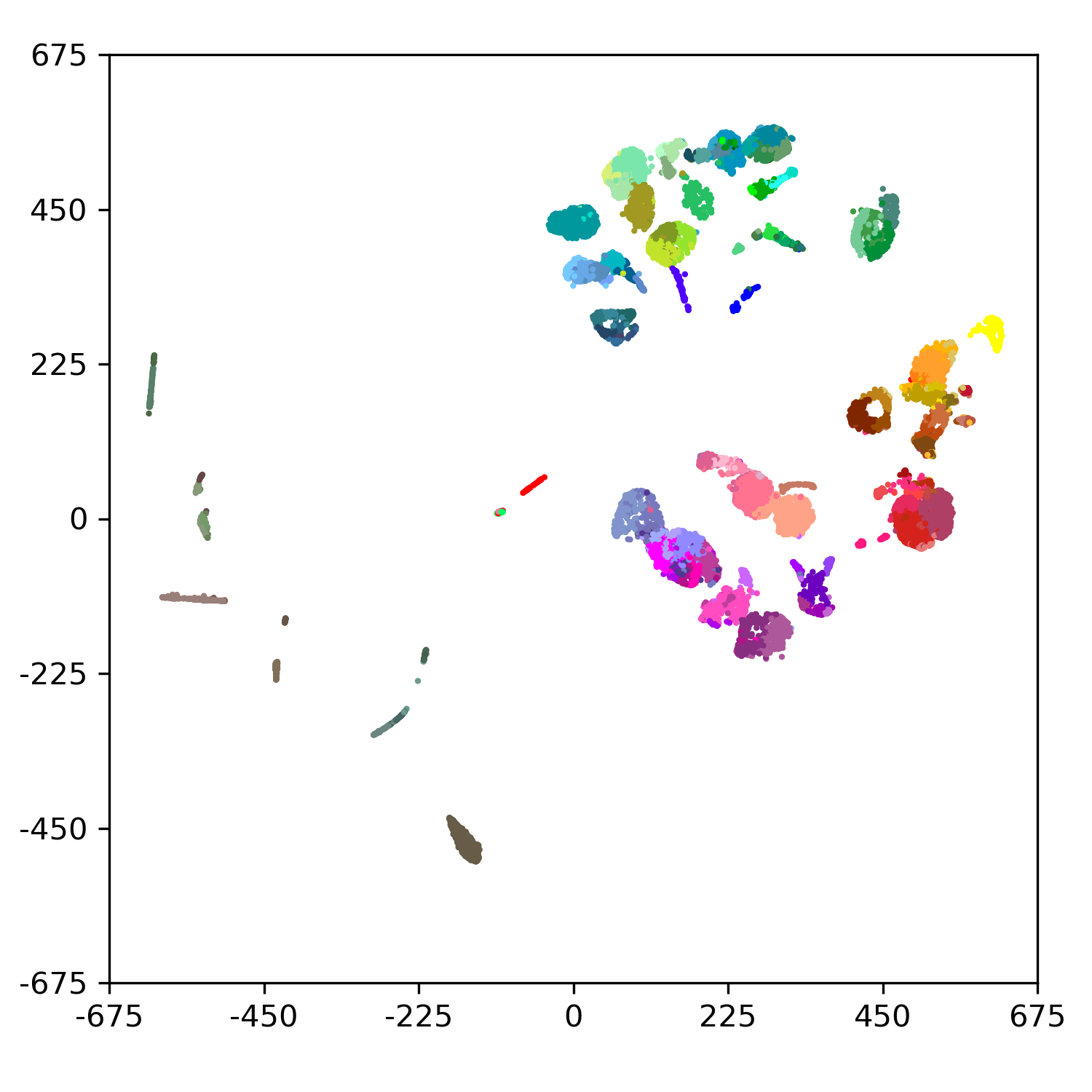}
        \caption*{C+E \\ (TriMap, $\alpha = 4.63$)}
    \end{subfigure}\hfill
    \begin{subfigure}[b]{0.18\textwidth}
        \centering
        \includegraphics[width=\textwidth]{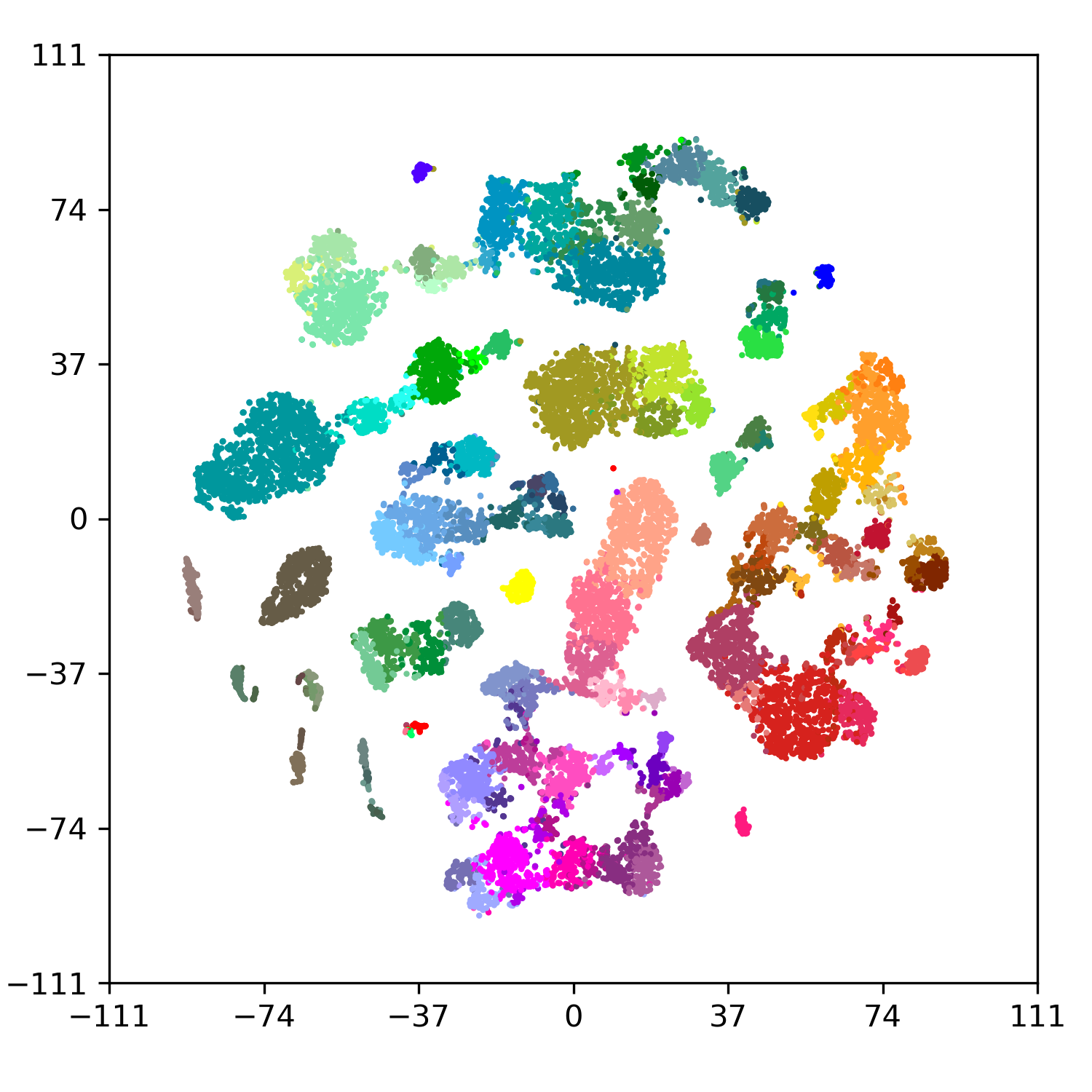}
        \caption*{t-SNE \\  \strut}
    \end{subfigure}\hfill
    \begin{subfigure}[b]{0.18\textwidth}
        \centering
         \includegraphics[width=\textwidth]{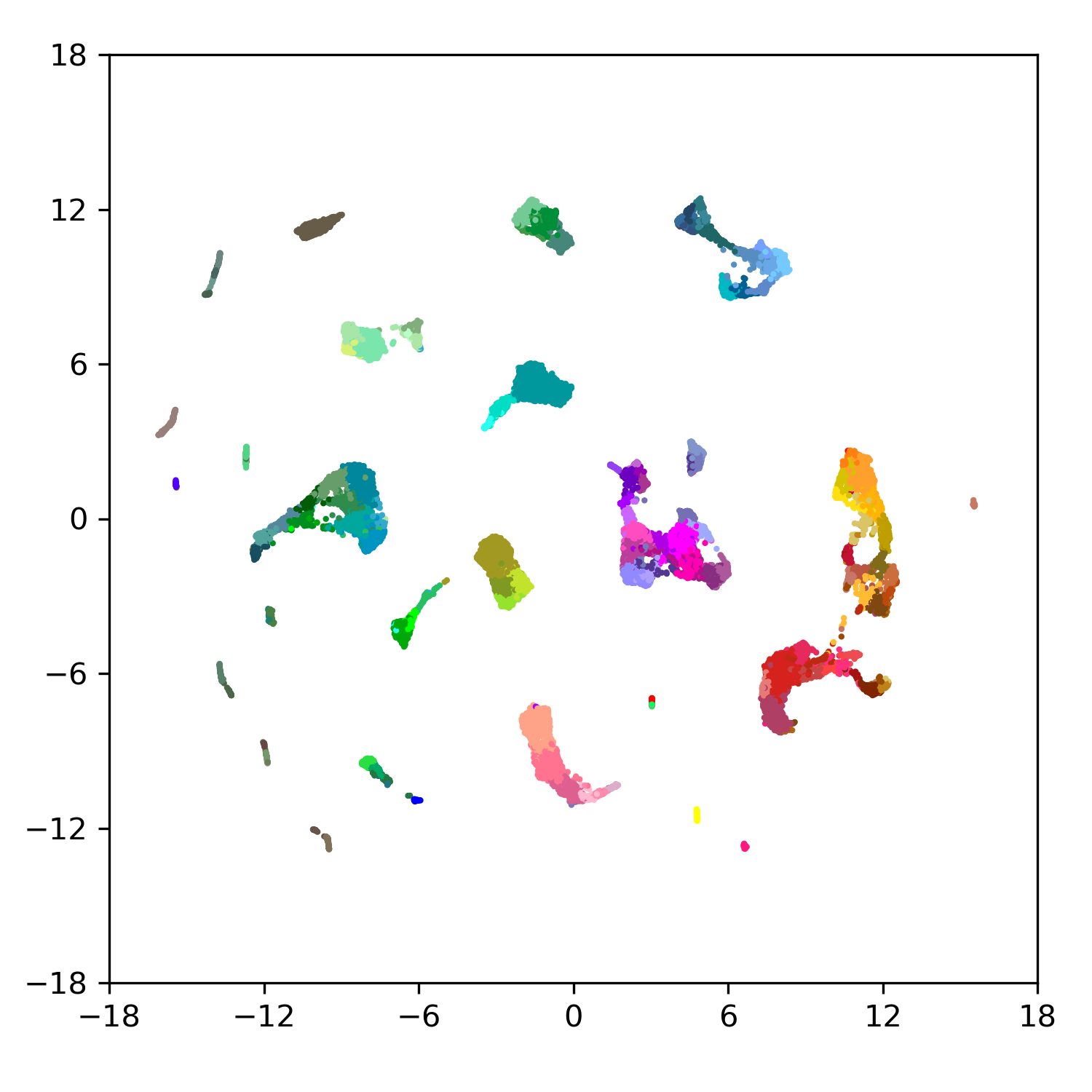}
        \caption*{UMAP \\  \strut}
    \end{subfigure}
    
% ----- Row 2 -----
    \begin{subfigure}[c]{0.18\textwidth}
        \centering
        \includegraphics[width=\textwidth]{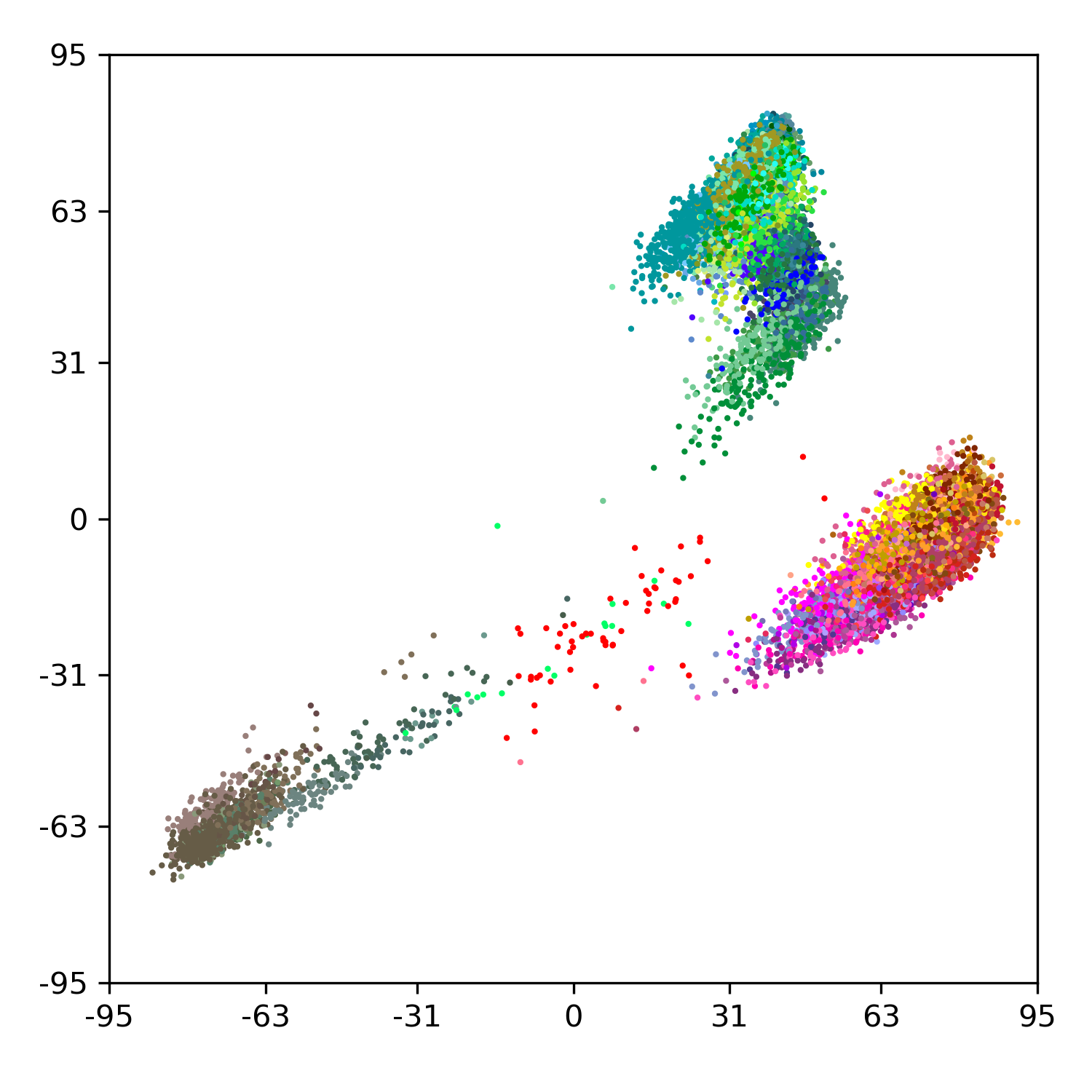}
        \caption*{PCA}
    \end{subfigure}
    \hfill
    \begin{subfigure}[c]{0.18\textwidth}
        \centering
        \includegraphics[width=\textwidth]{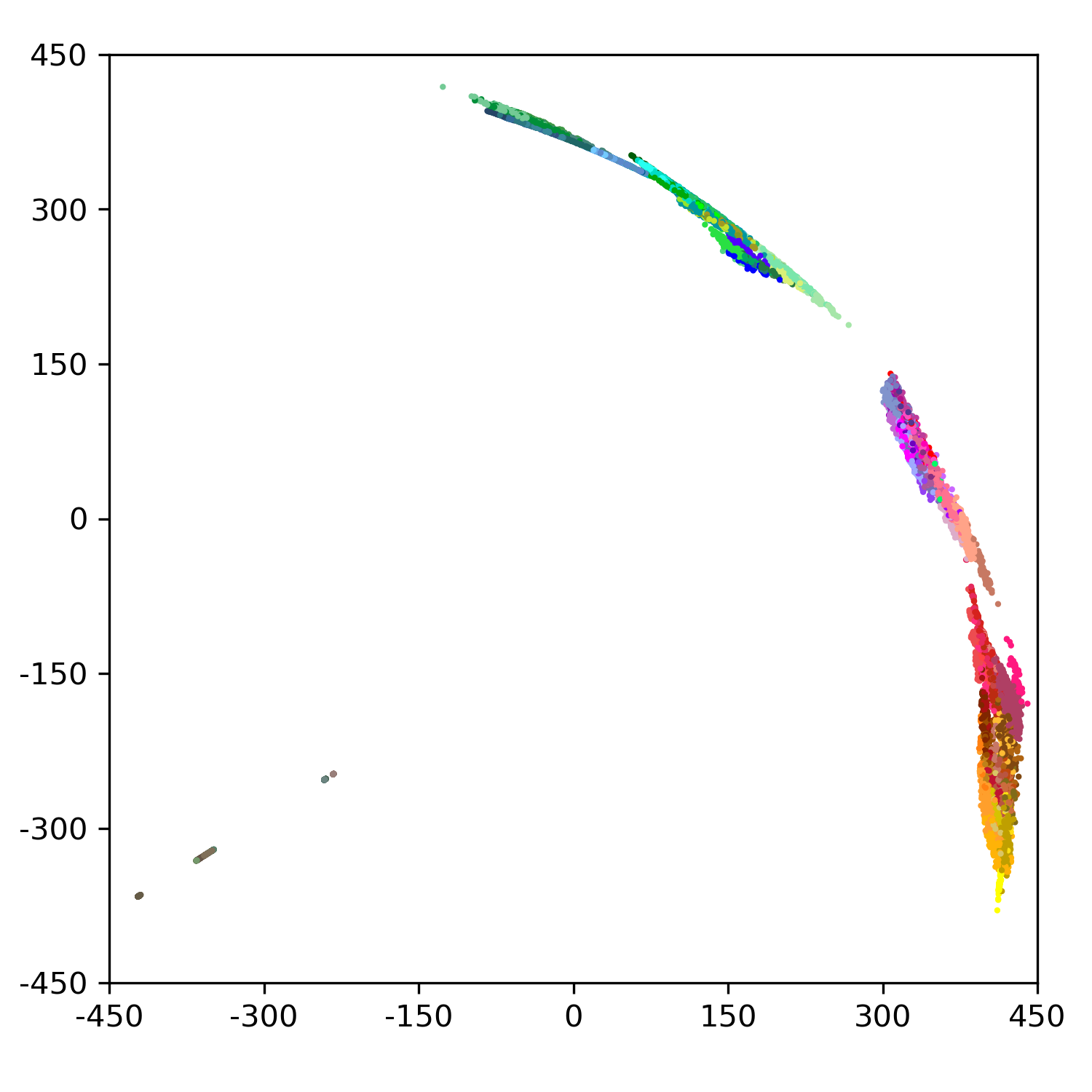}
        \caption*{L-Isomap}
    \end{subfigure}
    \hfill
    \begin{subfigure}[c]{0.18\textwidth}
        \centering
        \includegraphics[width=\textwidth]{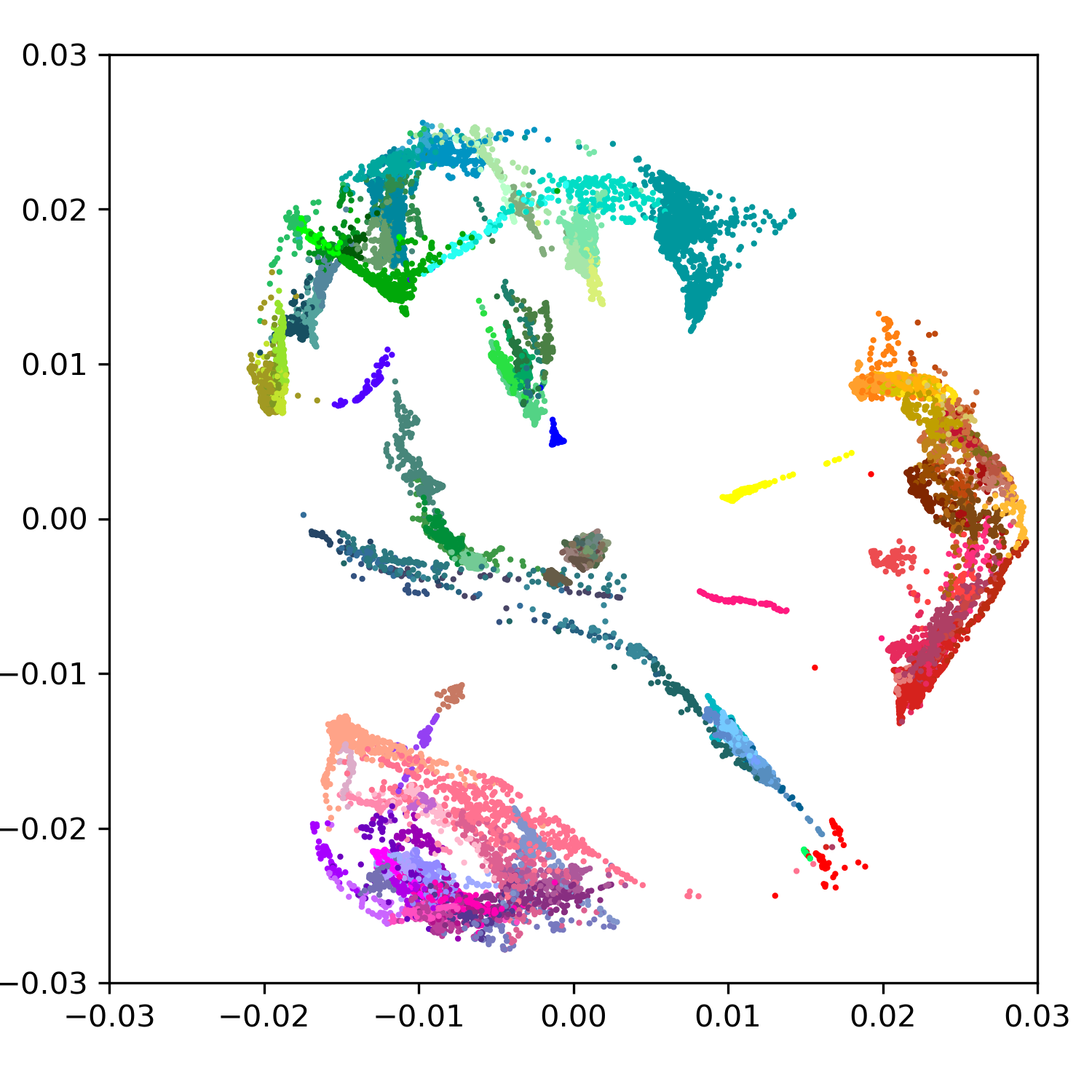}
        \caption*{PHATE}
    \end{subfigure}
    \hfill
    \begin{subfigure}[c]{0.18\textwidth}
        \centering
        \includegraphics[width=\textwidth]{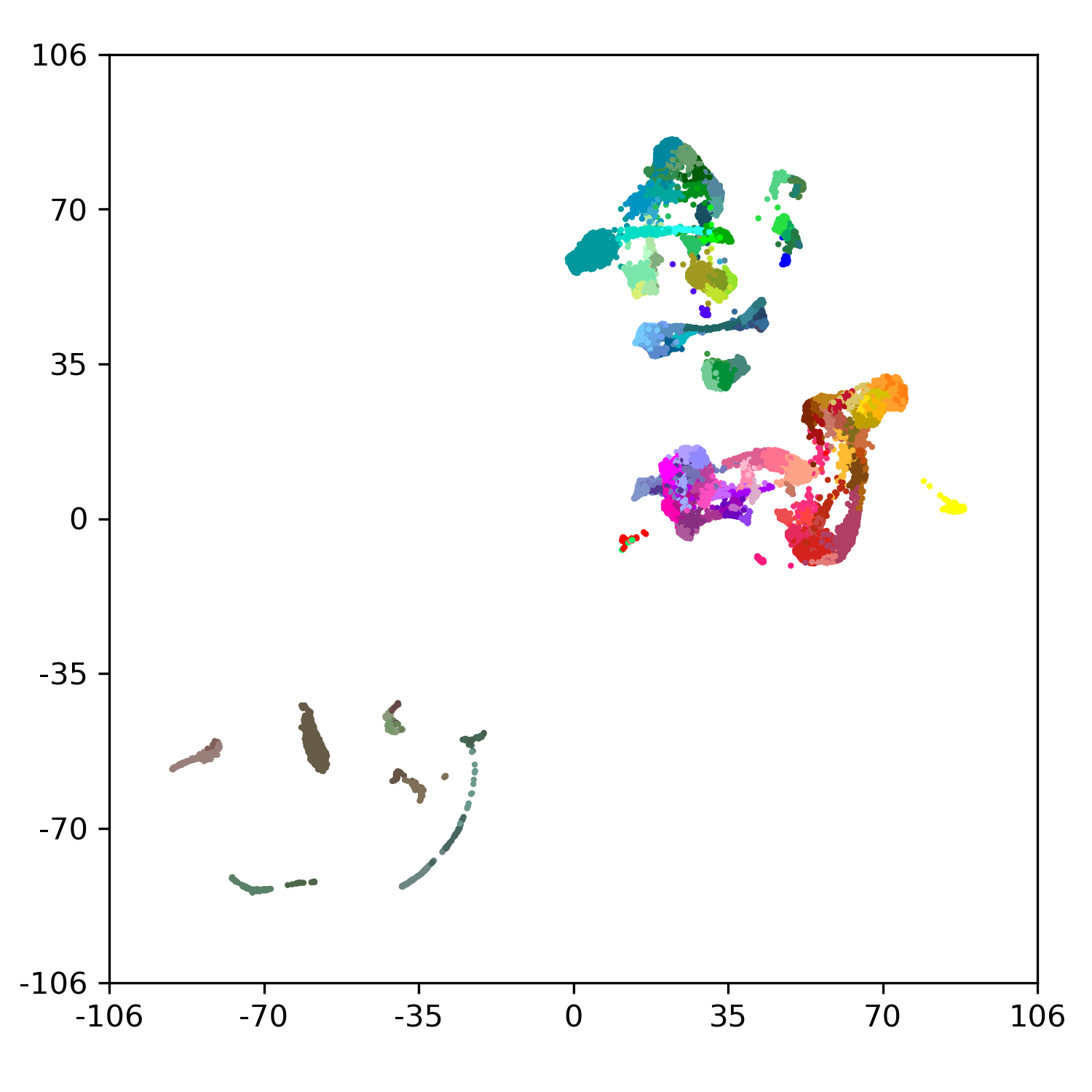}
        \caption*{TriMap}
    \end{subfigure}
    \begin{subfigure}[c]{0.18\textwidth}
        \centering
\includegraphics[width=.7\textwidth]{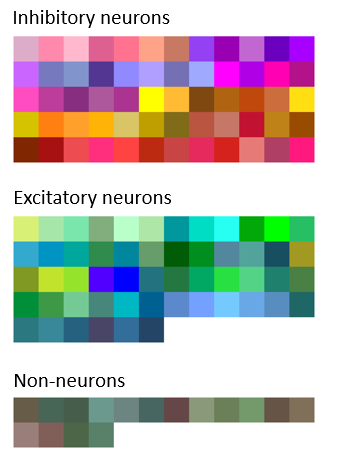}
         \caption*{}
    \end{subfigure}
    \caption{Embedding of the mouse cortex dataset ($n = 23,822$). Points are colored by original cluster labels which were not used by embedding methods.}
    \label{fig:mouse}
\end{figure}

\begin{table}[!ht]
\centering
\footnotesize
\setlength{\tabcolsep}{2pt}
\begin{tabular}{llccccccccccc}
\toprule
\textbf{Method} & \textbf{Params} 
& \multicolumn{4}{c}{\textbf{Local}} 
& \multicolumn{7}{c}{\textbf{Global}} \\
\cmidrule(lr){3-6} \cmidrule(lr){7-13}
& 
& W-S ${\scriptstyle \uparrow}$ 
& W-NS ${\scriptstyle \downarrow}$ 
& W-SNS ${\scriptstyle \downarrow}$ 
& KNN ${\scriptstyle \uparrow}$
& B-S ${\scriptstyle \uparrow}$  
& T-S ${\scriptstyle \uparrow}$  
& B-NS ${\scriptstyle \downarrow}$ 
& T-NS ${\scriptstyle \downarrow}$ 
& B-SNS ${\scriptstyle \downarrow}$ 
& T-SNS ${\scriptstyle \downarrow}$  
& CP ${\scriptstyle \uparrow}$
 \\
\midrule

C+E (PCA) & $\alpha = 1$ &
$\mathbf{0.68}$ & $\mathbf{0.50}$ & $\mathbf{0.35}$ & $0.22$ & $0.35$ & $0.88$ & $\mathbf{0.28}$ & $\mathbf{0.27}$ & $0.16$ & $0.27$ & $0.83$\\

C+E  (PCA) & $\alpha = 4.27$ &
$\mathbf{0.68}$ & $0.63$ & $0.43$ & $0.56$ & $0.36$ & $\mathbf{0.91}$ & $2.62$ & $3.16$ & $\mathbf{0.08}$ & $\mathbf{0.26}$ & $0.86$
\\

C+E  (TriMap) & $\alpha = 4.63$ &
$0.65$ & $0.71$ & $0.49$ & $0.58$ & $0.24$ & $0.90$ & $2.90$ & $3.53$ & $\mathbf{0.08}$ & $\mathbf{0.26}$ & $0.84$ \\

t-SNE & $u = 30$ &
$\mathbf{0.68}$ & $0.84$ & $0.42$ & $\mathbf{0.60}$ & $0.02$ & $0.55$ & $0.35$ & $0.41$ & $0.09$ & $0.40$ & $0.59$ \\

UMAP & $q = 15$ &
$0.57$ & $0.98$ & $0.51$ & $0.58$ & $0.02$ & $0.49$ & $0.87$ & $0.88$ & $\mathbf{0.08}$ & $0.40$ & $0.43$\\

PCA & - &
$0.58$ & $0.73$ & $0.41$ & $0.11$ & $\mathbf{0.60}$ & $\mathbf{0.91}$ & $0.47$ & $0.43$ & $0.18$ & $0.37$ & $\mathbf{0.94}$
 \\

L-Isomap & $q = 15$  &
$0.37$ & $0.63$ & $0.52$ & $0.24$ & $0.08$ & $0.80$ & $2.53$ & $3.14$ & $0.12$ & $0.39$ & $0.73$
 \\

PHATE & $q = 5$ &
$0.46$ & $1.00$ & $0.50$ & $0.36$ & $0.05$ & $0.39$ & $1.00$ & $1.00$ & $0.10$ & $0.48$ & $0.28$ \\

TriMap & - &
$0.58$ & $0.92$ & $0.44$ & $0.54$ & $0.25$ & $0.87$ & $0.53$ & $0.48$ & $0.07$ & $0.32$ & $0.86$\\

\bottomrule
\end{tabular}
\caption{Evaluation of the mouse cortex dataset against Euclidean distances for $5000$ points selected independently of the points used in alignment. Metrics are reported for within class (W), between class (B), and total (T). S denotes Spearman correlation, NS normalized stress, SNS scale-normalized stress, CP class preservation, and KNN the $30$-NN recall. Arrows indicate whether higher ($\uparrow$) or lower ($\downarrow$) values are better.}
\label{tab:mouse}
\end{table}

\begin{figure}[ht!]
    \centering
    \begin{subfigure}[c]{0.4\textwidth}
        \centering
        \includegraphics[width=\textwidth]{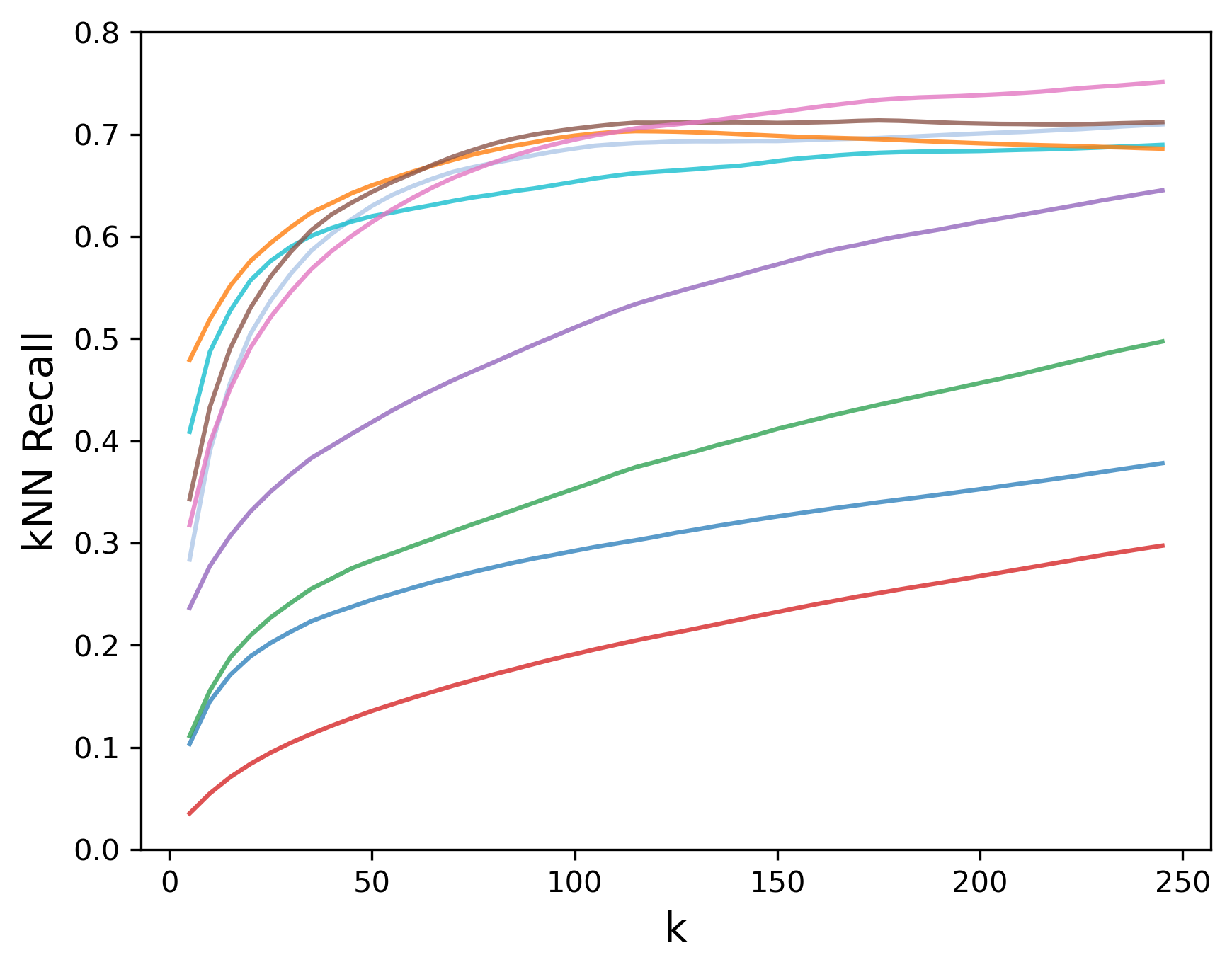}
    \end{subfigure}
    \hspace{0.03\textwidth}
    \begin{subfigure}[c]{0.2\textwidth}
        \centering
        \includegraphics[width=\textwidth]{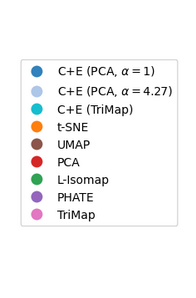}
    \end{subfigure}
    \caption{$k$NN recall for the mouse cortex dataset for various values of $k$ for all methods compared.}
    \label{fig:knn_mouse}
\end{figure}

We use the adult mouse cortex data from \cite{tasic2018shared}, with the same preprocessing as in \cite{kobak2019art}. The original dataset contains $133$ clusters labeled by the authors. Using Leiden clustering ($q=15, \gamma = 1.5 $), we obtain $46$ clusters with a Rand index of $0.983$ with the original labels. We include results for two embedding methods, PCA and TriMap, and align the cluster-level embeddings to preserve Euclidean distances, using $\alpha = 4.27$ and $\alpha = 4.63$, respectively, determined using \eqref{eq:alpha}. We were unable to approximate pairwise geodesic distances in a reasonable runtime on this dataset due to the $k$NN graph being disconnected for values of $k$ small enough to keep the overall computation time comparable to other methods. For L-Isomap, distances between points in disconnected components of the $k$NN graph were set to the maximum pairwise distance observed within any connected component.   

In \figref{mouse}, we observe a separation of inhibitory neurons, excitatory neurons, and non-neurons in the C+E embeddings. The overall layout is similar to PCA, so globally the C+E organization is aligned with maximizing variance, but locally, when $\alpha \approx 4$, the additional space between clusters allows them to spread out, revealing finer structure and organization. In contrast, while t-SNE and UMAP separate the excitatory neurons (upper left) and inhibitory neurons (lower right), the coarse-to-fine organization revealed in the C+E embedding is not present.  

The cost of the space for the finer cluster structure to emerge in the C+E embedding is that the scale of the embedding is significantly inflated. In \tabref{mouse}, the global normalized stress of C+E is quite poor. However, the scale-invariant global metrics are generally much better for C+E compared to UMAP and t-SNE, and the local metrics generally also favor C+E, with the exception of the $k$NN recall. The TriMAP embedding also appears to organize the clusters similarly to C+E, but has worse local metrics, improved by C+E (TriMap). PHATE divides inhibitory neurons into two groups with a large separation in the embedding. The subgrouping of the inhibitory neurons should be on a finer scale, and consequently, PHATE is outperformed by C+E on all metrics except the global normalized stress.

% \gal{again add some details. We can see separation of excitatory and inhibitory and non neurons. The overall layout is similar to PCA, so globally our organization is aligned with maximizing variance, but locally the clusters are spread out so we see more structure and organization. In tSNE and UMAP there is also excitatory on one side and inhibitory on the other side, but you don't actually see a coarse to fine organization like C+E reveals. I think we should actually make this a main figure.}

\section{Discussion}

We presented a general framework for visualization that aims to preserve both local cluster structure and global distances and structures. The approach we propose consists of clustering the data, embedding each cluster individually, and finally aligning the cluster-level embeddings to preserve (possibly inflated) distances between pairs of points in different clusters. By embedding each cluster individually, we showed that we can create an embedding with better local metrics than embedding all of the data at once, while generally maintaining comparable or better global metrics. The inflation of between-cluster distances at the alignment step is our approach to the crowding problem. By increasing the target distances between clusters, more space becomes available at a given distance scale in the low-dimensional embedding space, consequently avoiding overlap between distinct clusters while still preserving their relative distances. Indeed, we demonstrated on multiple datasets that with large enough scaling parameter, C+E successfully separates clusters while preserving a meaningful global organization. In the human brain dataset, the C+E embedding revealed both a developmental trajectory, and some grouping into different cell-types along this trajectory. In the mouse cortex dataset, the C+E embedding revealed a coarse-to-fine structure not present in most other embedding methods.

The C+E approach also allows the user the flexibility to choose an embedding method that targets a specific metric. We showed that embedding the individual clusters with PCA or L-Isomap results in a global embedding with strong local distance preservation metrics, while embedding the individual clusters with TriMap, can improve the $k$NN recall performance for small values of $k$. Nonetheless, t-SNE still achieves superior performance to any other method for small values of $k$. This exceptional performance of t-SNE on $k$NN recall compared to other embedding methods is somewhat of a mystery to us. Previous works have argued that t-SNE optimizes the $k$NN recall \cite{venna2010,im2018}, though we find that the argument strays from t-SNE as originally formulated, and assumes a simplified setting where $q_{ij}$ in \eqref{eq:kl_div} takes on a binary form. This is exemplified by simply running t-SNE on data in $\bbR^2$ with initialization as is, so that the $k$NN recall is $=1$ for all $k$ at initialization. We find that t-SNE updates the points in a way that decreases the $k$NN recall. We also note that we can somewhat improve the $k$NN recall of the C+E approach by increasing $\alpha$ at the cost of other distance preservation metrics, or by breaking up the data into more clusters at the clustering step, though this can introduce false clusters in the visualization. This approach of breaking up the data into clusters that are not necessarily meaningful appears to somewhat mimic how t-SNE appears to break the data up into small patches and arrange these patches when the perplexity is small, even when the data is not clustered, though we leave further investigation of this to future work. This remains a limitation of the proposed method and if the user wishes to obtain a visualization with high $k$NN recall for small $k$, t-SNE may still be the state of the art. Nonetheless, high $k$NN recall is only one of many potentially desirable properties for data visualization. With that in mind, our method is competitive and allows for more directly controllable and interpretable visualizations.
 
\subsection*{Acknowledgments} 
We would like to thank Amelia Kawasaki and Tianyao Xu for their initial contributions during early stages of this work. We would also like to thank Eran Mukamel and Elizabeth Purdom for helpful discussions.

\small
\bibliographystyle{chicago}
\bibliography{ref}

\newpage
\appendix
\label{appendix:Appendix}

\section{A description of t-SNE}
\label{sec:t-SNE}
For each pair of points $x_i$ and $x_j$, t-SNE first computes the normalized affinities
\begin{align}
    \label{eq:p_matrix}
    p_{ij} = \frac{p_{i|j} + p_{j|i}}{2n} \quad  \text{where} \quad p_{j|i} = \frac{K(\|x_i - x_j\|/\sigma_i)}{\sum\limits_{k \neq i}K(\|x_i - x_k\|/\sigma_i)},
\end{align}
where $K$ is the Gaussian kernel, $K(t) = e^{-t^2/2}$, and $\sigma_i$ is chosen according to the perplexity parameter $u$, so that 
\[2^{\mathcal{H}(p_{|i})} = u, \quad \forall i = 1, \dots, n,\] 
where $\mathcal{H}(p_{|i}) = -\sum_j p_{j|i} \log_2 p_{j|i}$ is the entropy. The perplexity can be thought of as controlling the neighborhood size considered in computing the affinities. Most implementations of t-SNE set $p_{j|i} = 0$ if $j$ is not in the $\lfloor 3u \rfloor$ nearest neighbors of $x_i$. The perplexity is typically set in the range of $5 \le u \le 50$ \cite{van2008}. 

The t-SNE algorithm, introduced by \citet{van2008} as a variant of the SNE algorithm proposed earlier by \citet{hinton2002stochastic}, aims to find $(y_i)_{i = 1}^n$ in $\bbR^2$ that minimizes the KL divergence
\begin{align}
\label{eq:kl_div}
\sum\limits_{i\neq j} p_{ij}\log\left( \frac{p_{ij}}{q_{ij}} \right),
\end{align}
where $q_{ij}$ is the normalized similarity between $y_i$ and $y_j$ computed using the Cauchy kernel $L(t) = (1+t^2)^{-1}$,
\[q_{ij} = \frac{L(\|y_i - y_j\|)}{\sum\limits_{l \neq m}L(\|y_l - y_m\|)}.\]

In practice, the KL divergence \eqref{eq:kl_div} is minimized using gradient descent with a momentum term. Implementations often use either the Barnes-Hut variant of t-SNE \cite{van2014accelerating} or  Fast Fourier Transform accelerated Interpolation-based t-SNE \cite{linderman2019fast} to accelerate gradient computation. An initial early exaggeration stage is also used during which the $p_{ij}$ in \eqref{eq:kl_div} are replaced with $\rho p_{ij}$ for the first few iterations of gradient descent, with $\rho >0$ being a tuning parameter. 

\section{Sensitivity of C+E}
\label{sec:sensitivity}
\subsection{Number of clusters}

\begin{figure}[h!]
    \centering
    \captionsetup{justification=centering, singlelinecheck=false, font = footnotesize}
    % ----- Row 1 -----
    \begin{subfigure}[b]{0.18\textwidth}
        \centering
        \includegraphics[width=\textwidth]{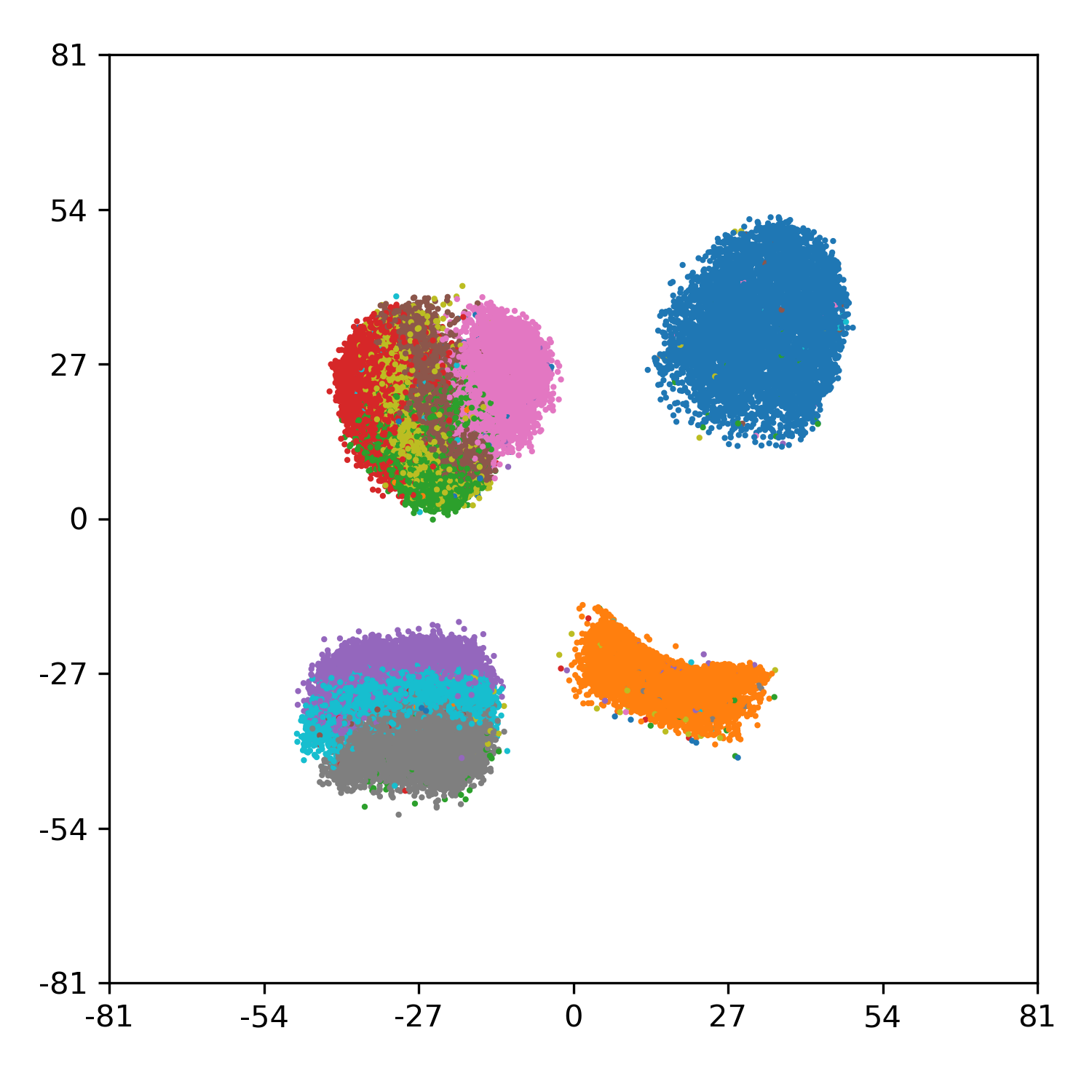}
        \caption*{4 clusters}
    \end{subfigure}\hfill
    \begin{subfigure}[b]{0.18\textwidth}
        \centering
        \includegraphics[width=\textwidth]{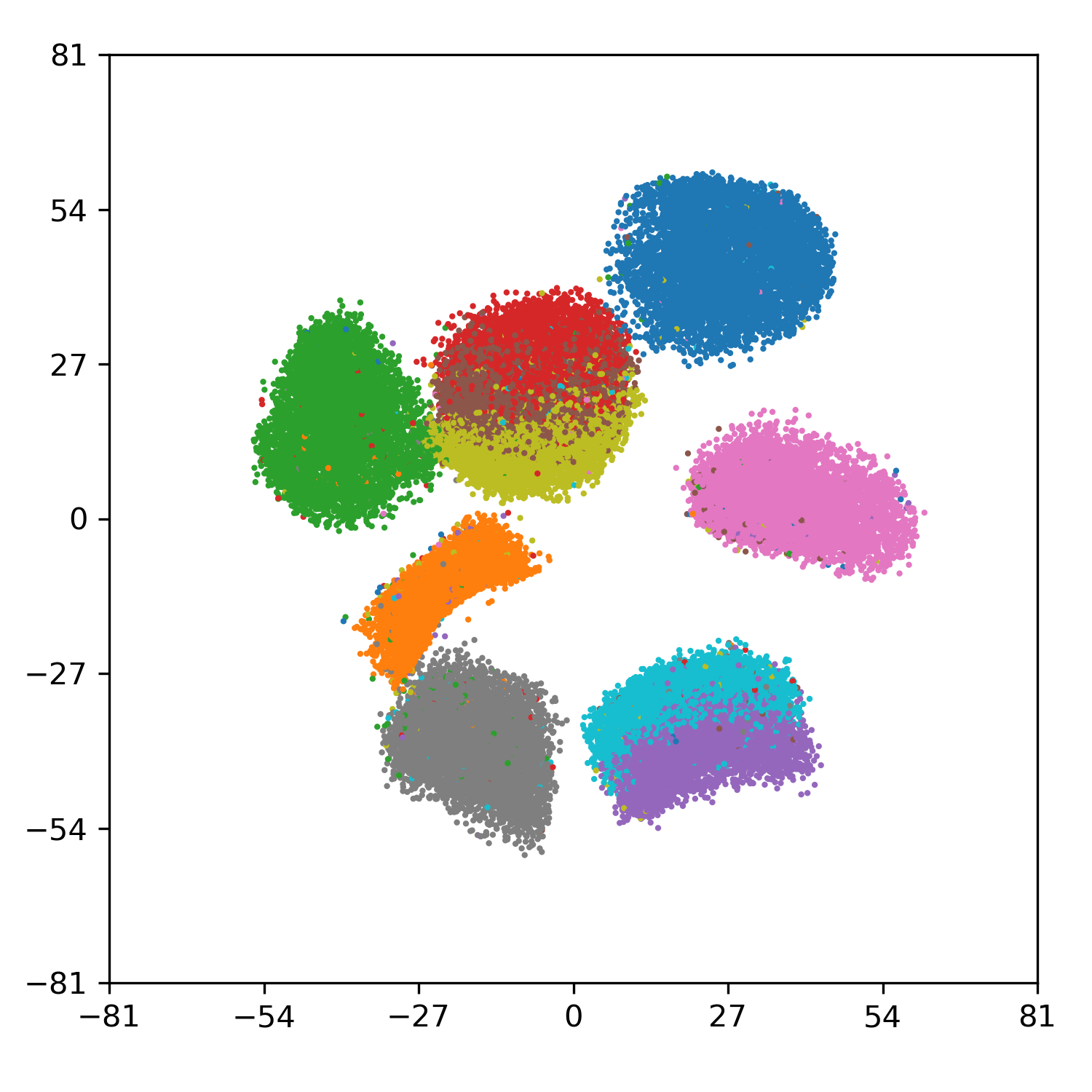}
        \caption*{7 clusters}
    \end{subfigure}\hfill
    \begin{subfigure}[b]{0.18\textwidth}
        \centering
        \includegraphics[width=\textwidth]{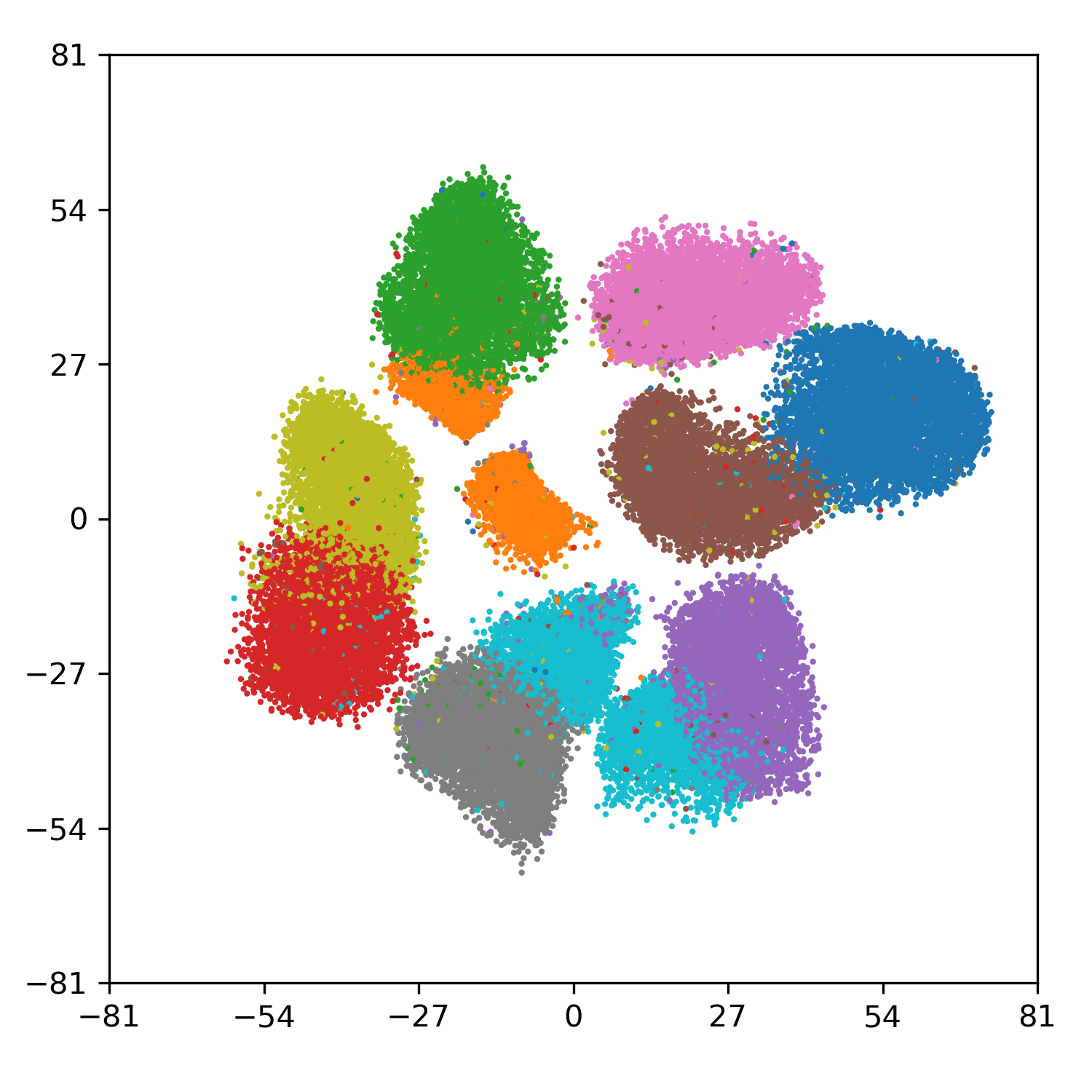}
        \caption*{12 clusters}
    \end{subfigure}\hfill
    \begin{subfigure}[b]{0.18\textwidth}
        \centering
        \includegraphics[width=\textwidth]{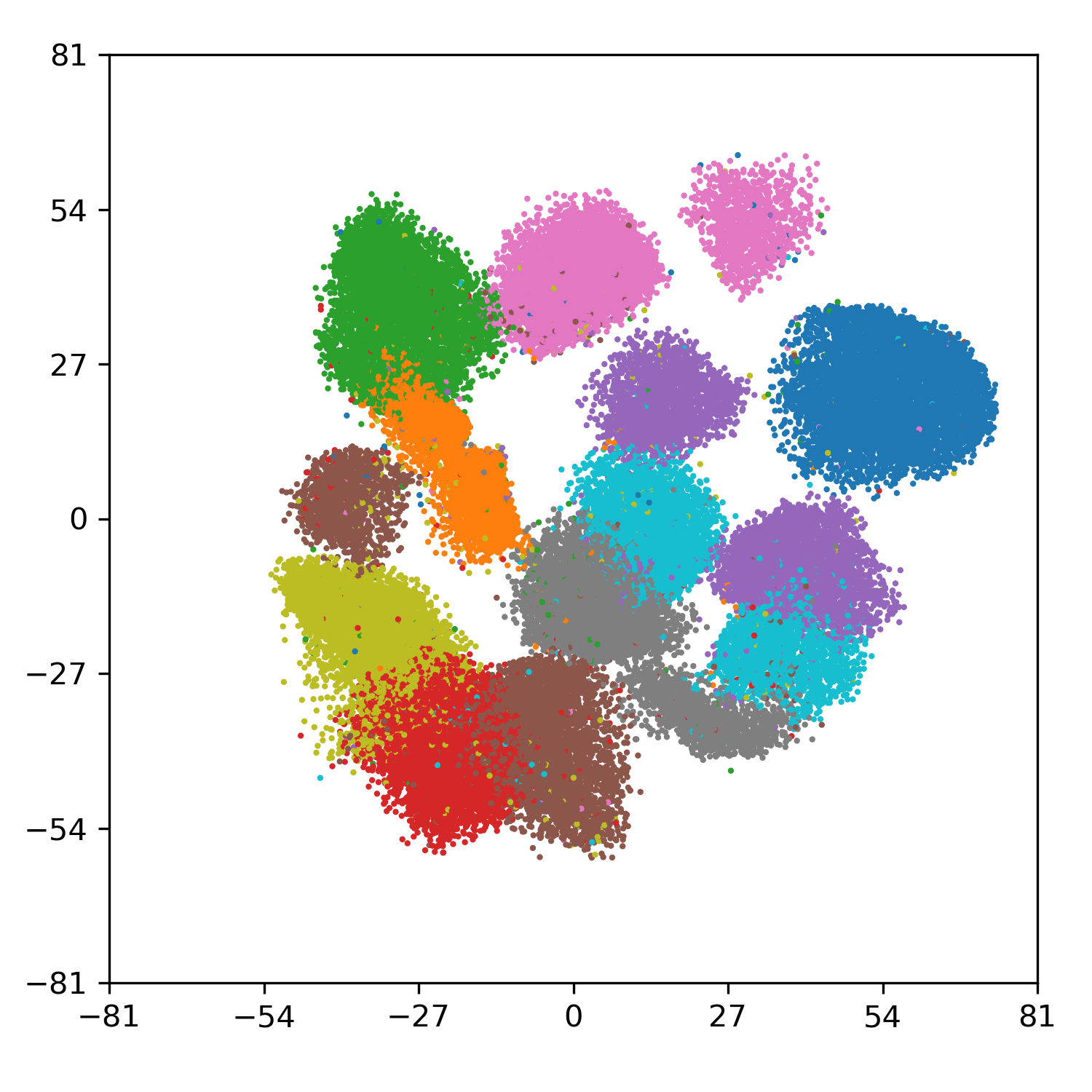}
        \caption*{16 clusters}
    \end{subfigure}\hfill
    \begin{subfigure}[b]{0.18\textwidth}
        \centering
        \includegraphics[width=\textwidth]{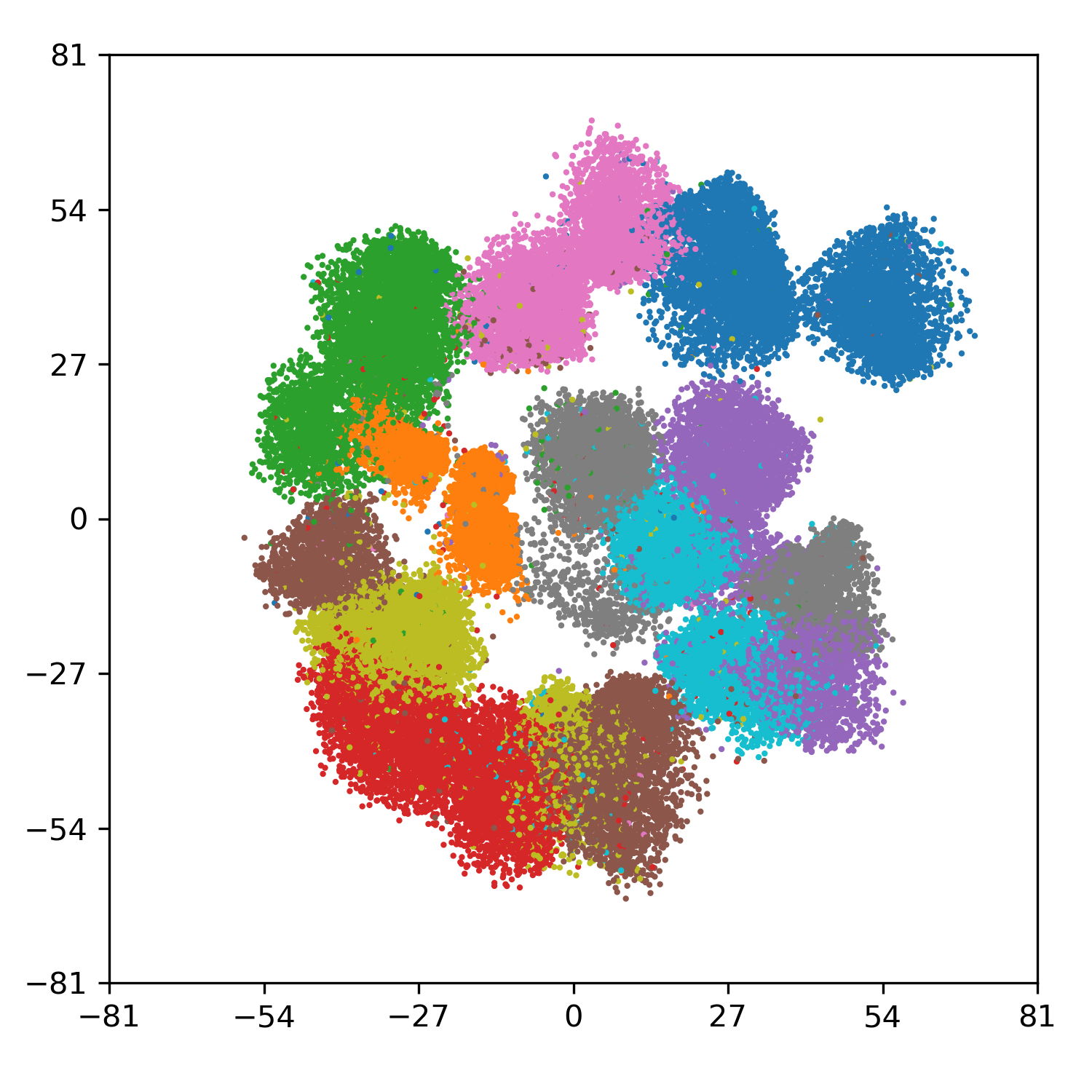}
        \caption*{24 clusters}
    \end{subfigure}
    
% ----- Row 2 -----
\begin{center}
\includegraphics[width=0.3\textwidth]{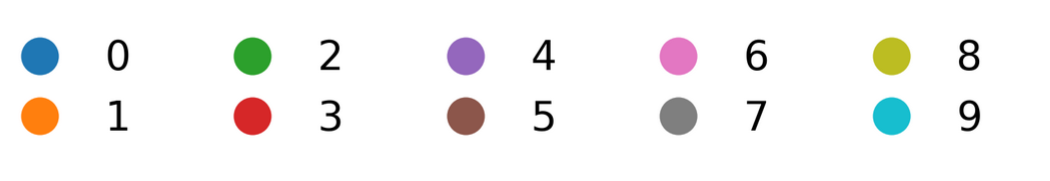}
\end{center}

    \caption{Embedding of the MNIST dataset with C+E (L-Isomap) as the number of clusters varies and $\alpha = 1.75 $ is held constant.}
    \label{fig:mnist_clustering_effect}
\end{figure}

\begin{figure}[h!]
    \centering
    \captionsetup{justification=centering, singlelinecheck=false, font = footnotesize}
    % ----- Row 1 -----
    \begin{subfigure}[b]{0.18\textwidth}
        \centering
        \includegraphics[width=\textwidth]{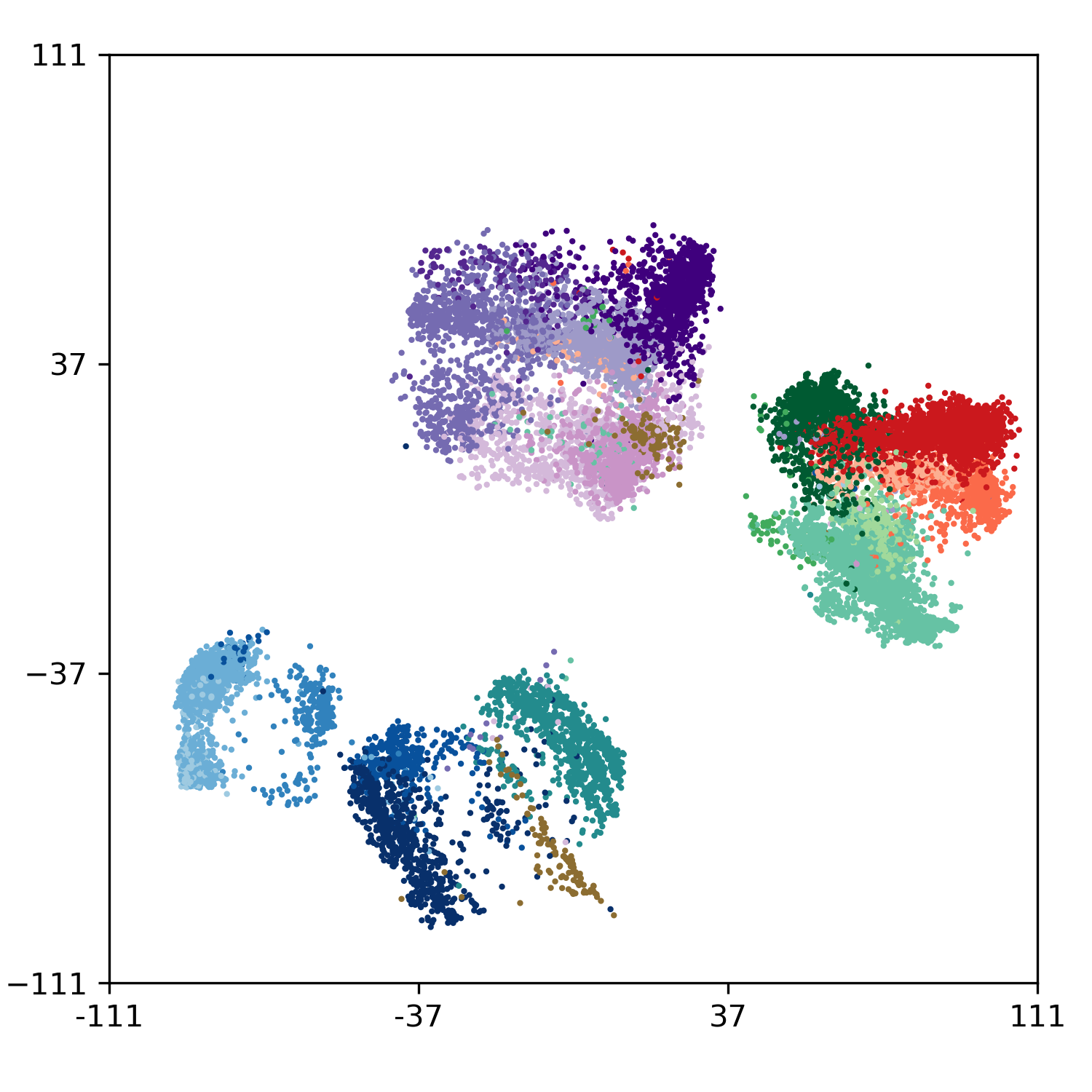}
        \caption*{4 clusters}
    \end{subfigure}\hfill
    \begin{subfigure}[b]{0.18\textwidth}
        \centering
        \includegraphics[width=\textwidth]{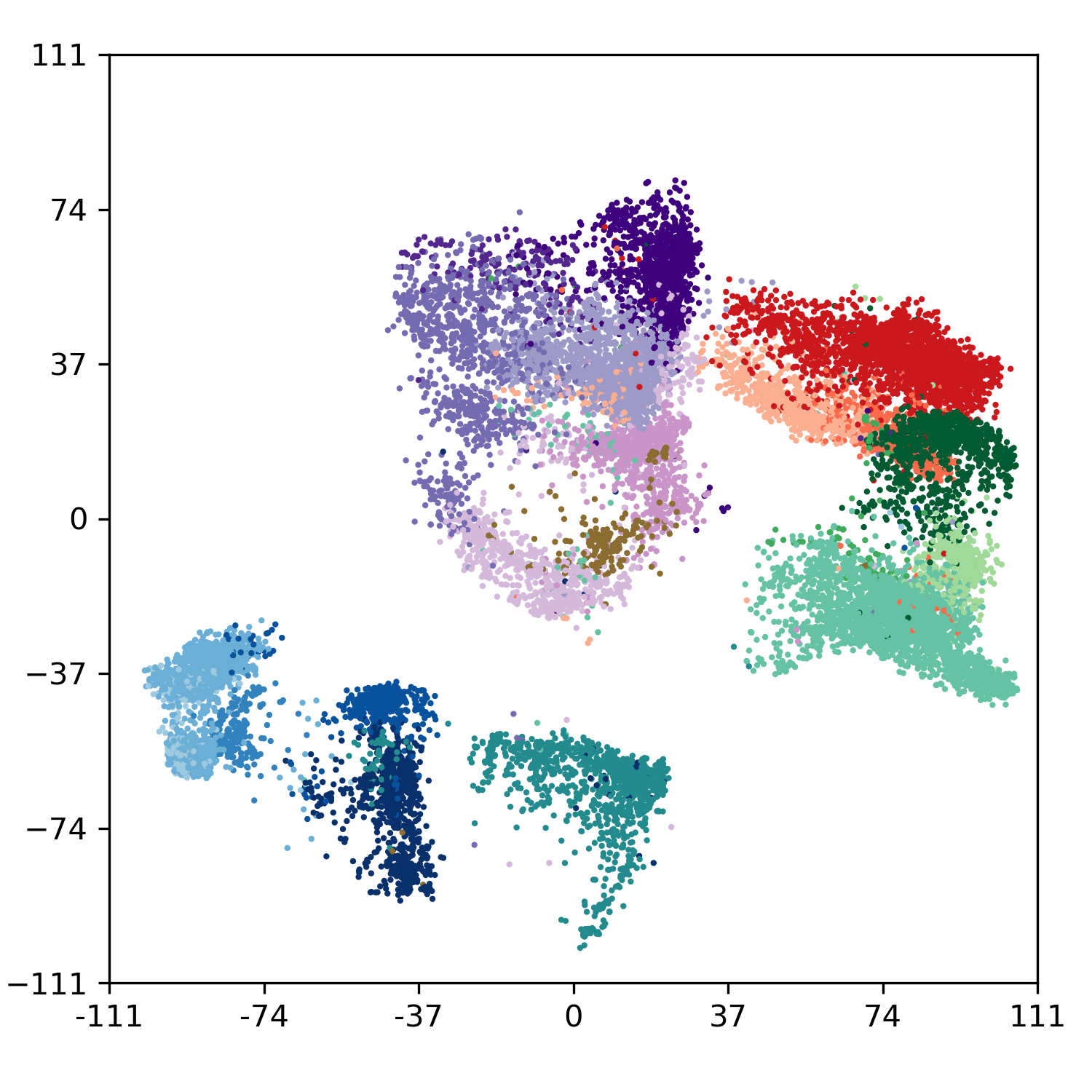}
        \caption*{7 clusters}
    \end{subfigure}\hfill
    \begin{subfigure}[b]{0.18\textwidth}
        \centering
        \includegraphics[width=\textwidth]{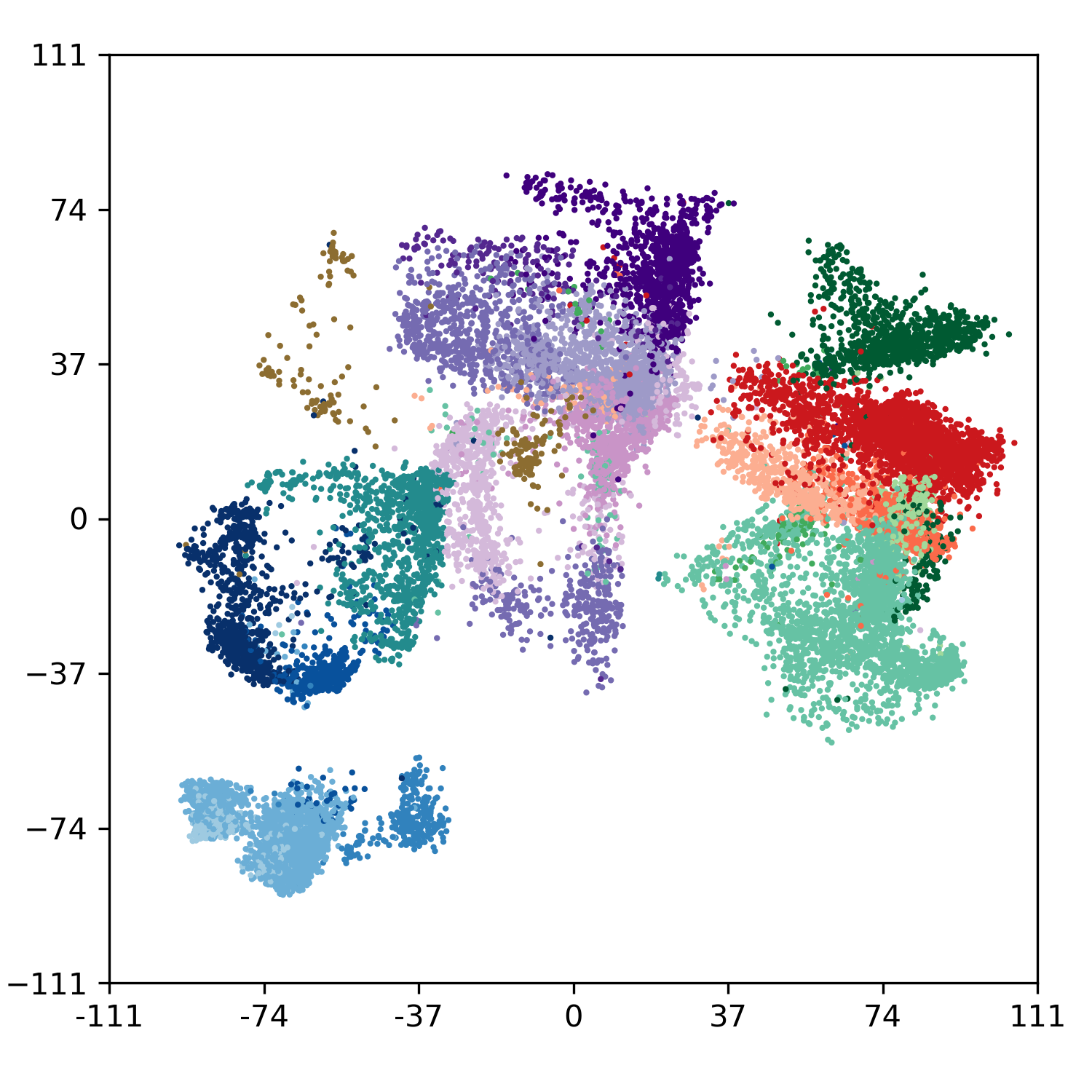}
        \caption*{13 clusters}
    \end{subfigure}\hfill
    \begin{subfigure}[b]{0.18\textwidth}
        \centering
        \includegraphics[width=\textwidth]{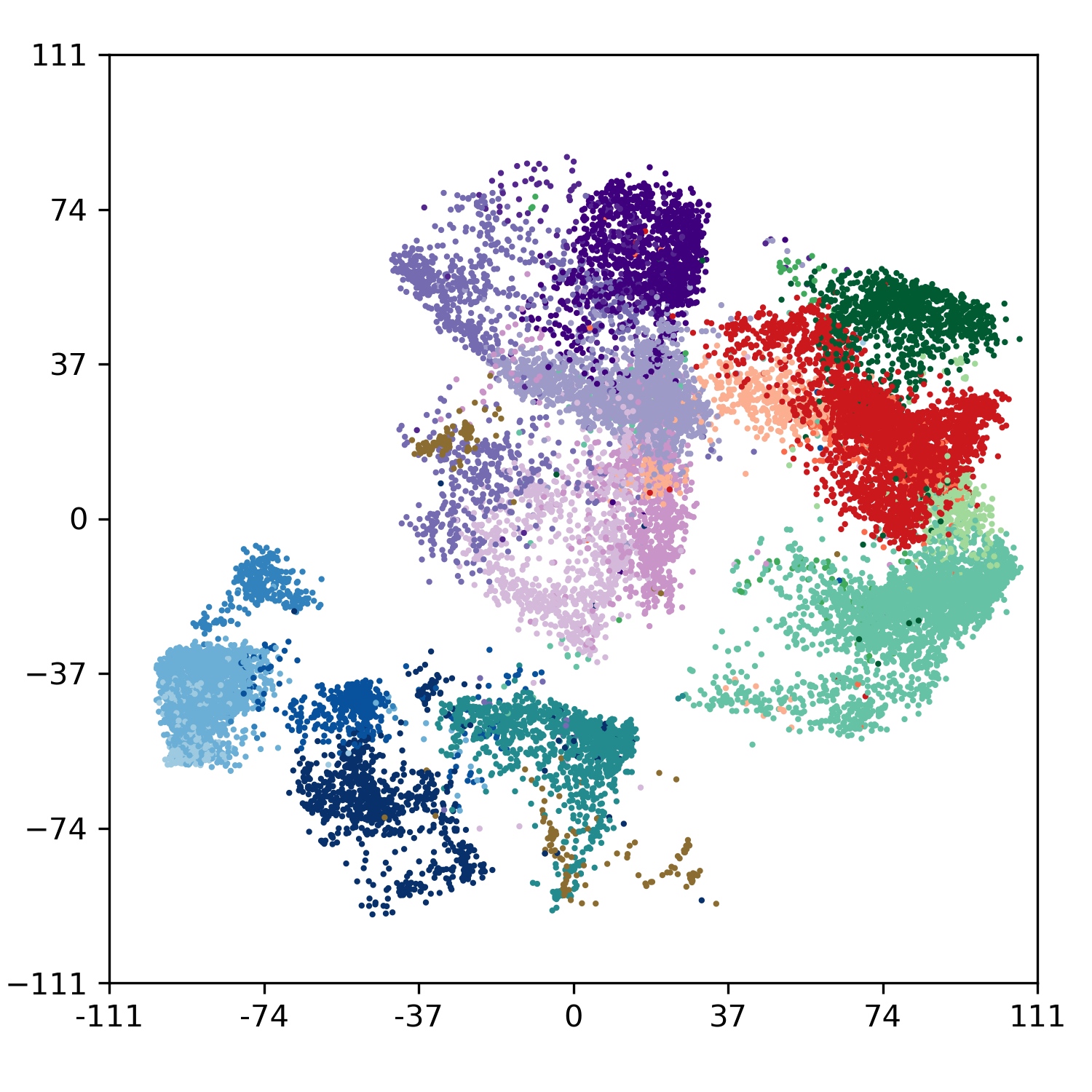}
        \caption*{20 clusters}
    \end{subfigure}\hfill
    \begin{subfigure}[b]{0.18\textwidth}
        \centering
        \includegraphics[width=\textwidth]{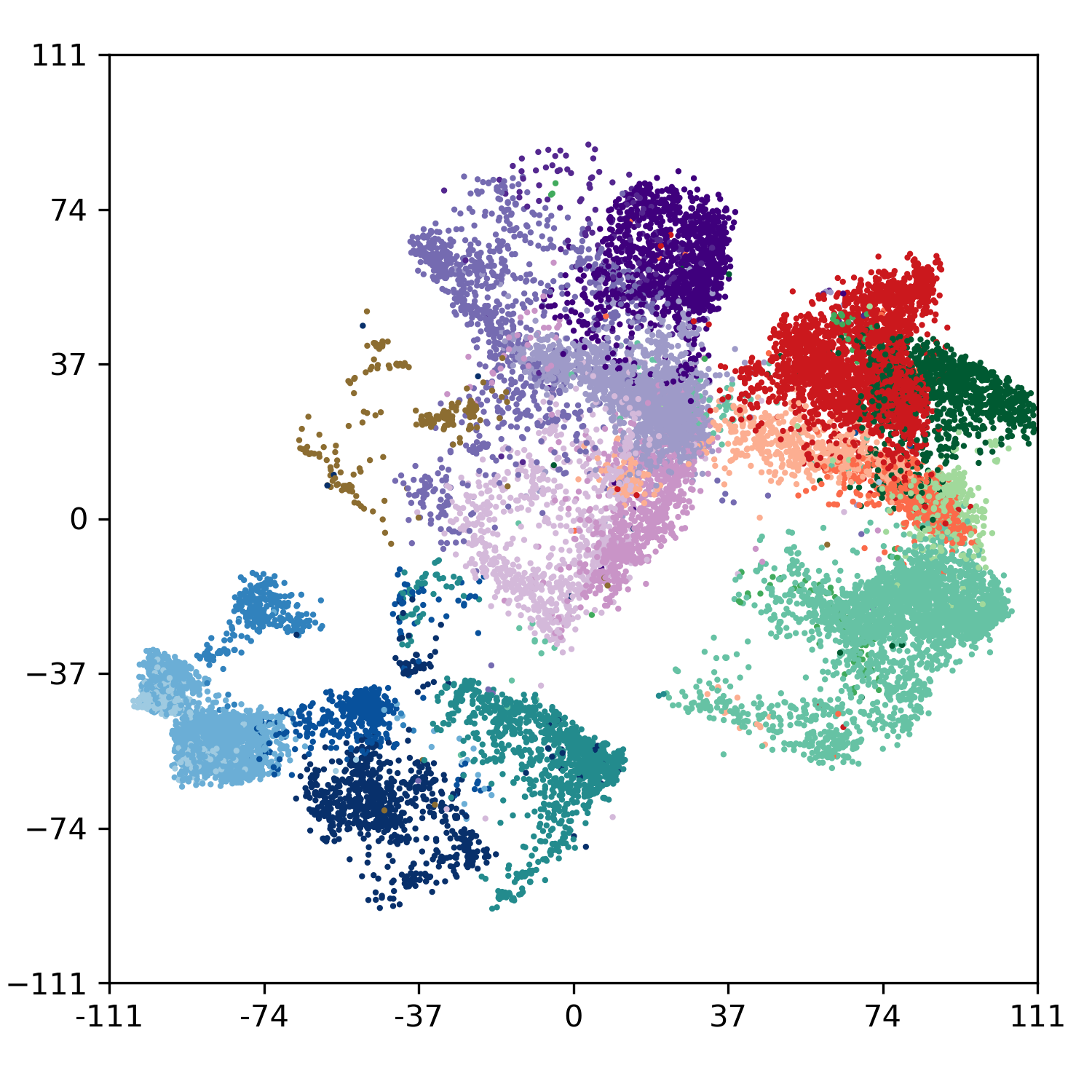}
        \caption*{28 clusters}
    \end{subfigure}
    
% ----- Row 2 -----
\begin{center}
\includegraphics[width=0.45\textwidth]{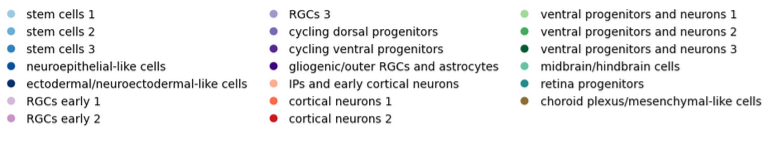}
\end{center}

    \caption{Embedding of the human brain organoid dataset with C+E (L-Isomap) as the number of clusters varies and $\alpha = 1.45 $ is held constant. Coloring is by cell-type.}
    \label{fig:brain_clustering_effect}
\end{figure}

\begin{table}[!ht]
\centering
\footnotesize
\setlength{\tabcolsep}{3pt}
\begin{tabular}{llccccccccccc}
\toprule
\textbf{Dataset} & \textbf{Clusters}
& \multicolumn{4}{c}{\textbf{Local}} 
& \multicolumn{7}{c}{\textbf{Global}} \\
\cmidrule(lr){3-6} \cmidrule(lr){7-13}
& 
& W-S ${\scriptstyle \uparrow}$ 
& W-NS ${\scriptstyle \downarrow}$ 
& W-SNS ${\scriptstyle \downarrow}$ 
& KNN ${\scriptstyle \uparrow}$
& B-S ${\scriptstyle \uparrow}$  
& T-S ${\scriptstyle \uparrow}$  
& B-NS ${\scriptstyle \downarrow}$ 
& T-NS ${\scriptstyle \downarrow}$ 
& B-SNS ${\scriptstyle \downarrow}$ 
& T-SNS ${\scriptstyle \downarrow}$  
& CP ${\scriptstyle \uparrow}$
 \\
\midrule

MNIST & $4$ &
$0.73$ & $0.63$ & $0.52$ & $0.25$ & $0.32$ & $0.44$ & $0.52$ & $0.55$ & $0.29$ & $0.45$ & $0.46$\\

 & $7$ &
$\mathbf{0.80}$ & $0.55$ & $0.44$ & $0.33$ & $\mathbf{0.43}$ & $0.58$ & $\mathbf{0.47}$ & $\mathbf{0.52}$ & $0.24$ & $0.36$ & $0.65$ \\

 & $10$ &
$\mathbf{0.80}$ & $0.53$ & $0.44$ & $\mathbf{0.37}$ & $0.42$ & $\mathbf{0.61}$ & $0.51$ & $0.58$ & $\mathbf{0.23}$ & $\mathbf{0.34}$ & $\mathbf{0.73}$ \\

 & $12$ &
$\mathbf{0.80}$ & $0.51$ & $0.43$ & $\mathbf{0.37}$ & $0.38$ & $0.56$ & $0.53$ & $0.60$ & $0.24$ & $0.35$ & $0.54$
 \\
 
  & $16$ &
$0.78$ & $0.48$ & $0.42$ & $0.36$ & $0.39$ & $0.54$ & $0.55$ & $0.60$ & $0.26$ & $0.37$ & $0.49$
 \\
  & $24$ &
$0.72$ & $\mathbf{0.46}$ & $\mathbf{0.40}$ & $\mathbf{0.37}$ & $0.39$ & $0.55$ & $0.55$ & $0.59$ & $0.27$ & $0.36$ & $0.58$
 \\
\midrule
Brain & $4$ &
$0.67$ & $0.59$ & $0.44$ & $0.28$ & $0\mathbf{.47}$ & $0.89$ & $0.29$ & $0.26$ & $0.20$ & $0.26$ & $0.86$
\\

& $7$ &
$0.70$ & $0.55$ & $0.43$ & $0.34$ & $0.46$ & $\mathbf{0.91}$ & $0.27$ & $0.23$ & $\mathbf{0.19}$ & $\mathbf{0.22}$ & $0.91$
 \\

 & $13$  &
$\mathbf{0.72}$ & $0.52$ & $\mathbf{0.41}$ & $\mathbf{0.35}$ & $0.46$ & $0.90$ & $0.28$ & $0.23$ & $0.21$ & $0.23$ & $\mathbf{0.92}$ 
 \\

 & $20$ &
$0.70$ & $\mathbf{0.49}$ & $\mathbf{0.41}$ & $0.34$ & $0.45$ & $\mathbf{0.91}$ & $\mathbf{0.26}$ & $\mathbf{0.22}$ & $0.20$ & $\mathbf{0.22}$ & $\mathbf{0.92}$
\\

 & $28$ &
$0.70$ & $0.50$ & $\mathbf{0.41}$ & $\mathbf{0.35}$ & $0.45$ & $0.90$ & $\mathbf{0.26}$ & $0.23$ & $0.20$ & $0.23$ & $\mathbf{0.92}$\\

\bottomrule
\end{tabular}
\caption{Evaluation of the MNIST and human brain organoid datasets against geodesic distances as the number of clusters in the embedding varies. Cell-types are used as the ground truth labels for the human brain organoid dataset. Metrics are reported for within class (W), between class (B), and total (T). S denotes Spearman correlation, NS normalized stress, SNS scale-normalized stress, CP class preservation, and KNN the $30$-NN recall. Arrows indicate whether higher ($\uparrow$) or lower ($\downarrow$) values are better.}
\label{tab:clustering}
\end{table}

In this section, we show the sensitivity of C+E to clustering choices on the MNIST and human brain organoid datasets. As in the main text, we cluster the data using the Leiden algorithm, adjusting the resolution parameter to obtain a different number of clusters. We select $\gamma \in [0.025, 1.5]$. Then, each cluster is embedded using L-Isomap, and aligned to preserve geodesic distances, holding $\alpha$ fixed for all clusterings. 

\figref{mnist_clustering_effect} shows the result on the MNIST dataset, and we see that the relative placement of clusters corresponding to each digit is somewhat robust to the clustering step. Even when a digit is oversegmented into multiple clusters, the clusters end up near each other, though some clusters are split (e.g., digit eight) when $\kappa = 16$ and $\kappa = 24$. With the exception of the severe under-clustering case of $\kappa = 4$, the local metrics in \tabref{clustering} are relatively stable and tend to improve as the number of clusters increases. Globally, there appears to be a U-shaped trend for each metric, with $\kappa = 7$ or $\kappa = 10$ obtaining the best global metrics. This reflects our observations that for larger number of clusters, some of the clusters are split into multiple sub-clusters.  

We observe similar trends on the human brain organoid dataset in \figref{brain_clustering_effect}, where the final embedding also appears generally robust to the clustering step. In this case, the local metrics tend to slightly improve as the number of clusters increases. The global metrics are generally stable across the number of clusters.

\subsection{Tuning $\alpha$}
\begin{figure}[h!]
    \centering
    \captionsetup{justification=centering, singlelinecheck=false, font = footnotesize}
    % ----- Row 1 -----
    \begin{subfigure}[b]{0.18\textwidth}
        \centering
        \includegraphics[width=\textwidth]{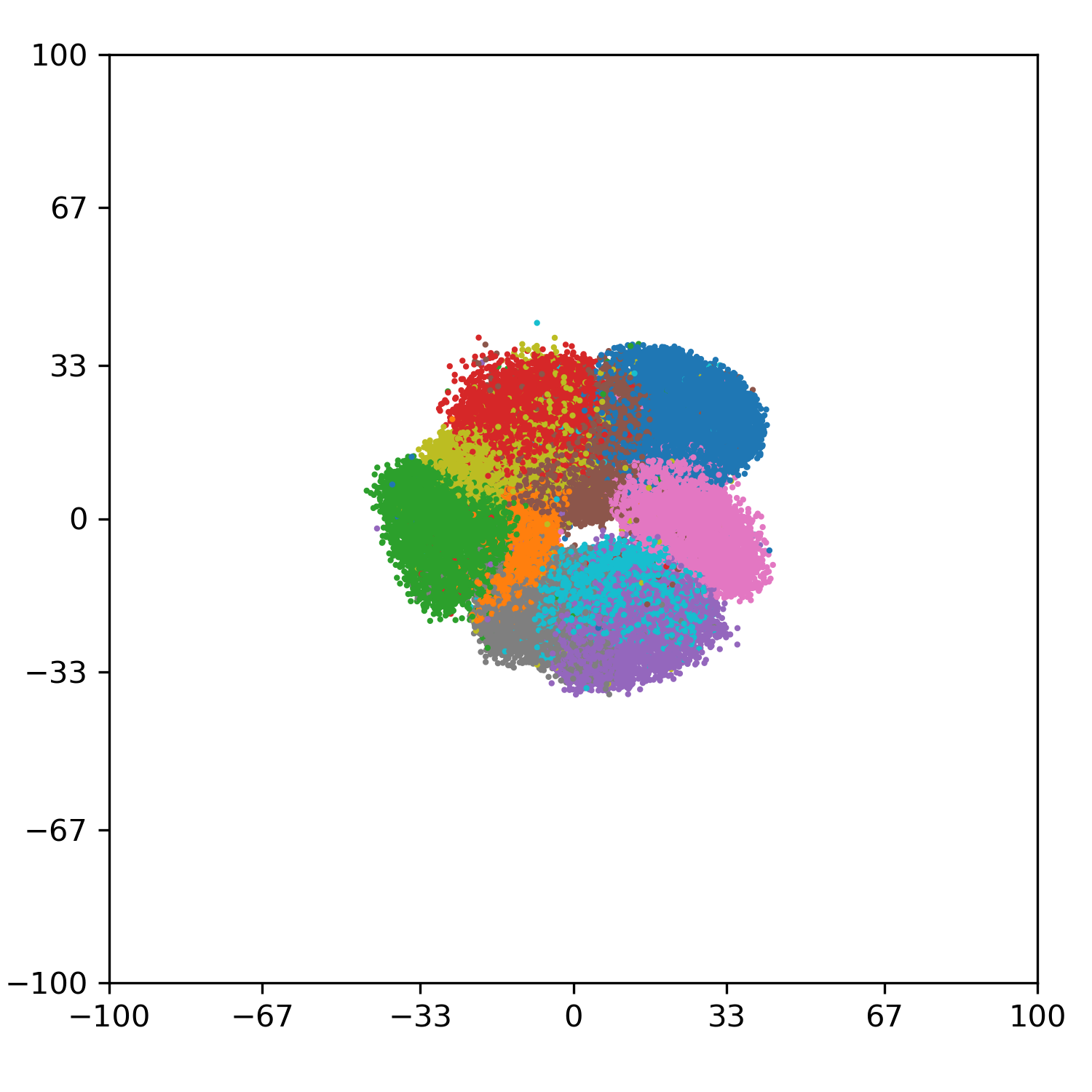}
        \caption*{$\alpha=1$}
    \end{subfigure}\hfill
    \begin{subfigure}[b]{0.18\textwidth}
        \centering
        \includegraphics[width=\textwidth]{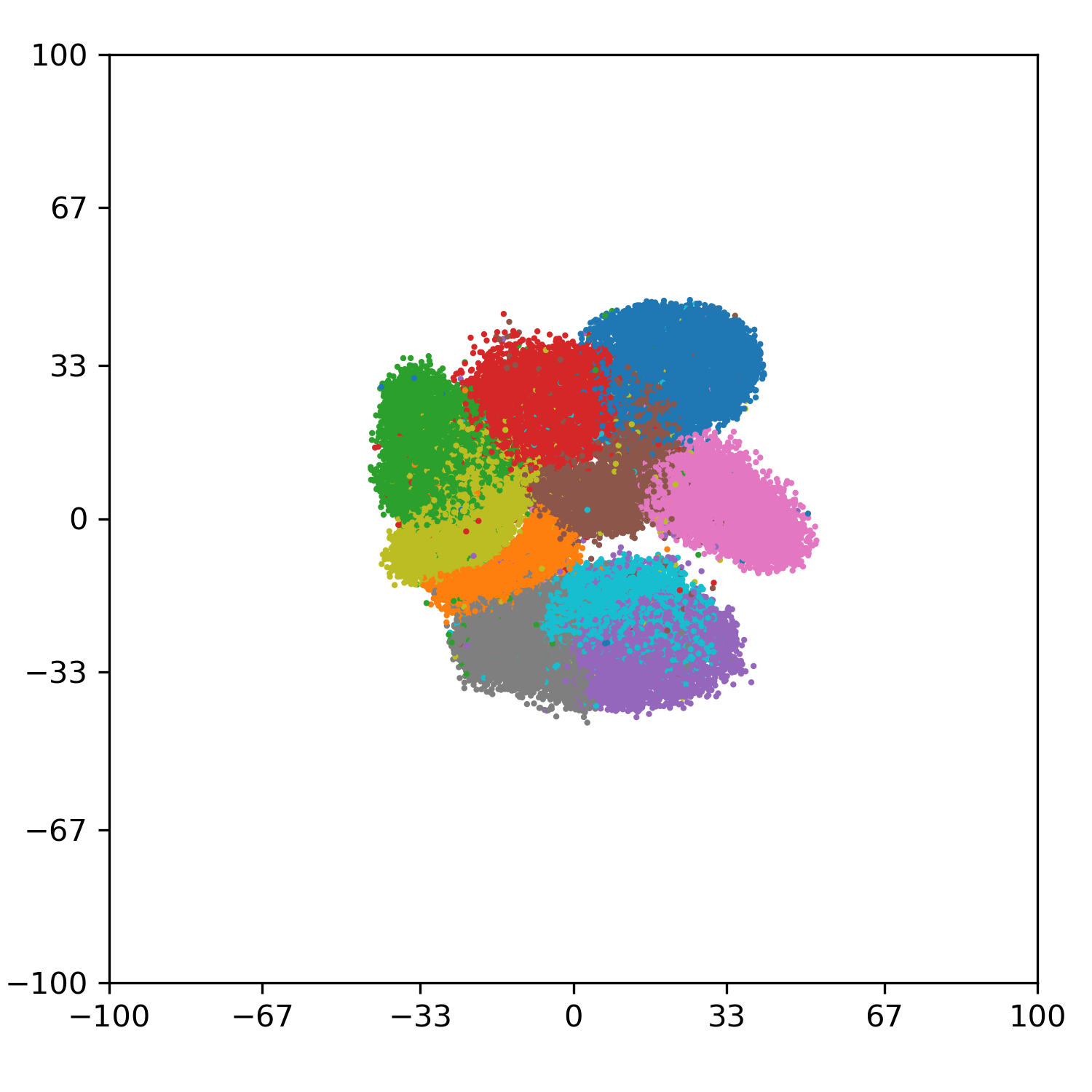}
        \caption*{$\alpha=1.25$}
    \end{subfigure}\hfill
    \begin{subfigure}[b]{0.18\textwidth}
        \centering
        \includegraphics[width=\textwidth]{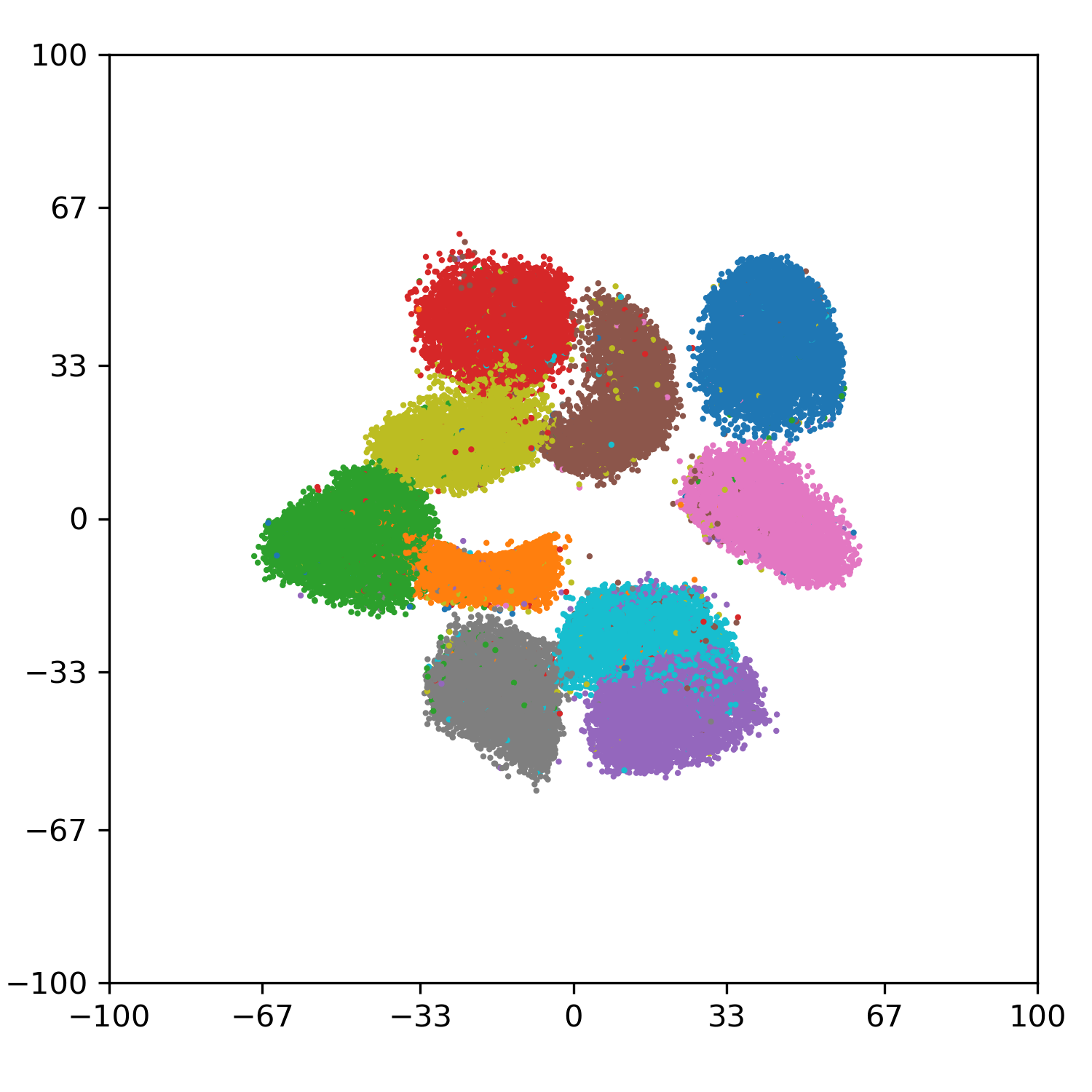}
        \caption*{$\alpha=1.75$}
    \end{subfigure}\hfill
    \begin{subfigure}[b]{0.18\textwidth}
        \centering
        \includegraphics[width=\textwidth]{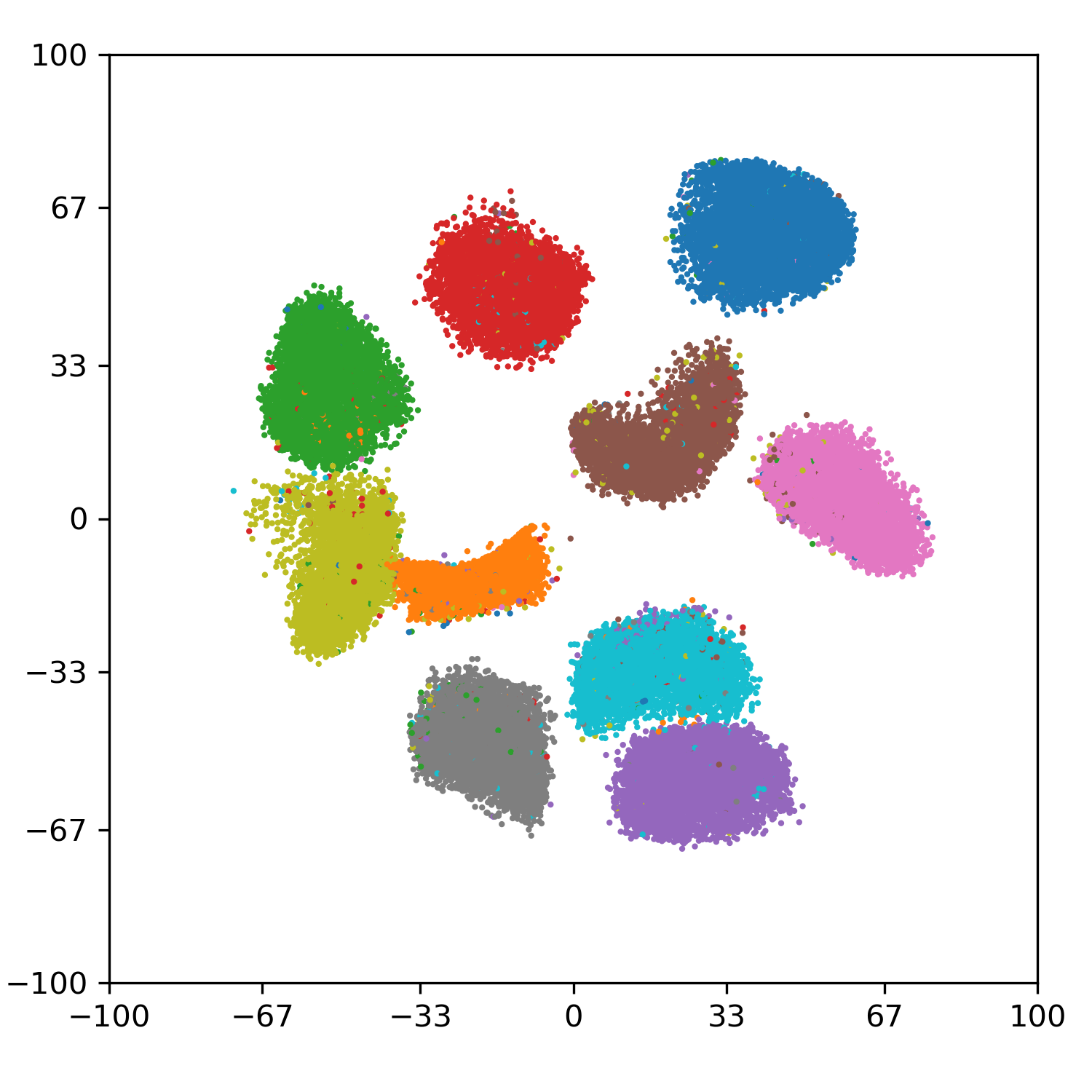}
        \caption*{$\alpha=2.25$}
    \end{subfigure}\hfill
    \begin{subfigure}[b]{0.18\textwidth}
        \centering
        \includegraphics[width=\textwidth]{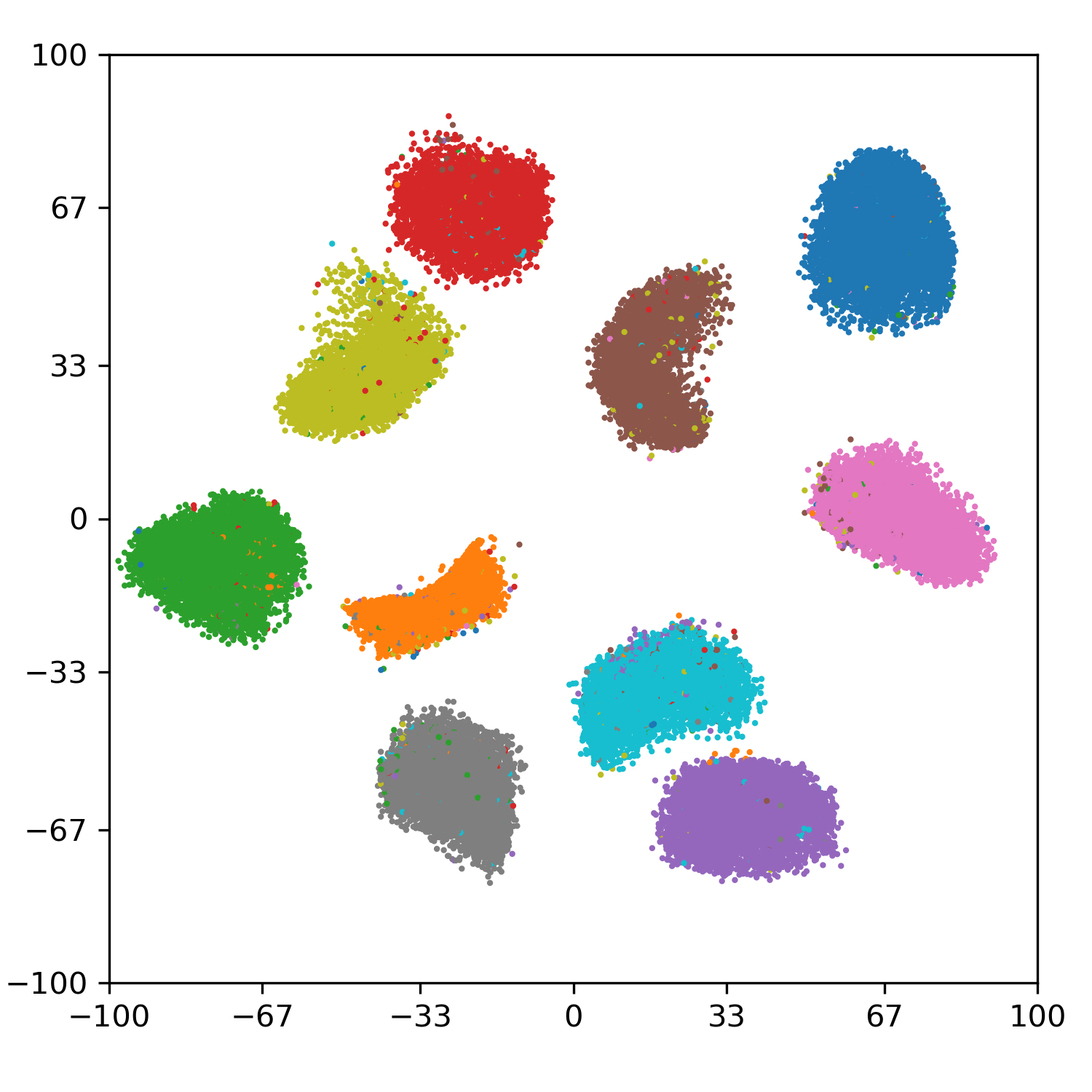}
        \caption*{$\alpha=2.75$}
    \end{subfigure}
    
% ----- Row 2 -----
\begin{center}
\includegraphics[width=0.3\textwidth]{figs/mnist_legend_horizontal.png}
\end{center}

    \caption{Embedding of the MNIST dataset with C+E (L-Isomap) as the $\alpha$ parameter is adjusted, increasing the separation between the ten embedded clusters.}
    \label{fig:mnist_alpha}
\end{figure}

\begin{figure}[h!]
    \centering
    \captionsetup{justification=centering, singlelinecheck=false, font = footnotesize}
    % ----- Row 1 -----
    \begin{subfigure}[b]{0.18\textwidth}
        \centering
        \includegraphics[width=\textwidth]{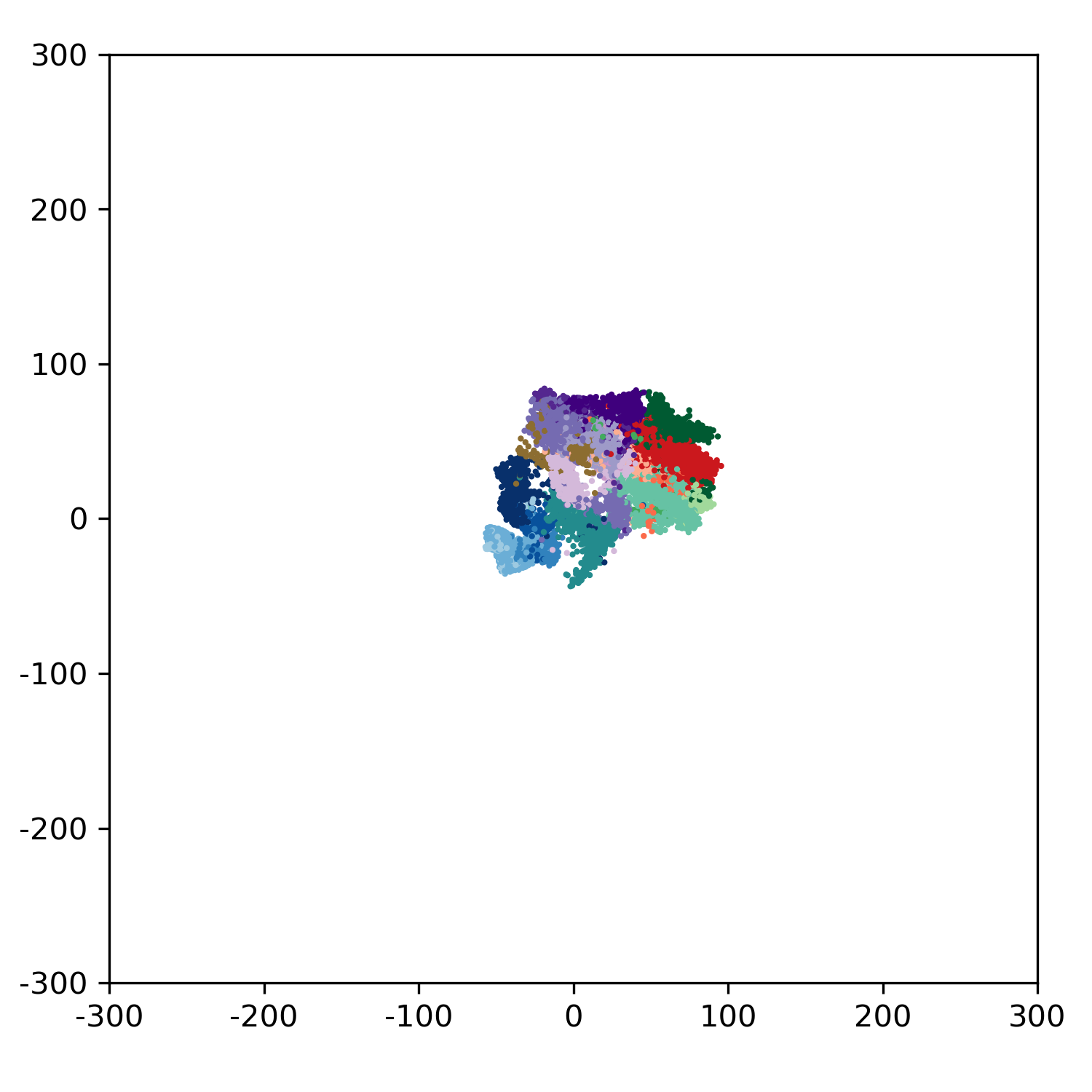}
        \caption*{$\alpha=1$}
    \end{subfigure}\hfill
    \begin{subfigure}[b]{0.18\textwidth}
        \centering
        \includegraphics[width=\textwidth]{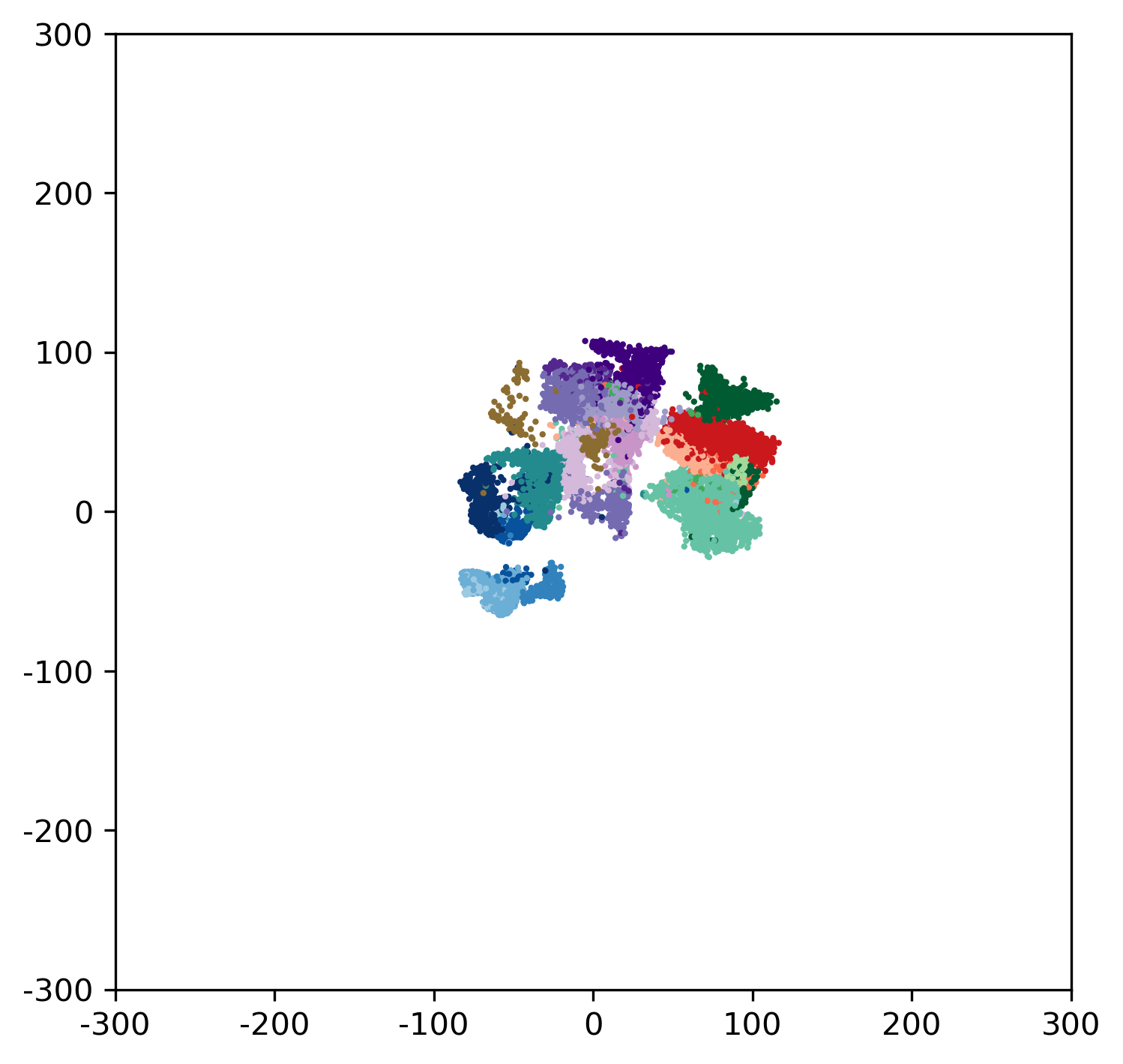}
        \caption*{$\alpha=1.45$}
    \end{subfigure}\hfill
      \begin{subfigure}[b]{0.18\textwidth}
        \centering
        \includegraphics[width=\textwidth]{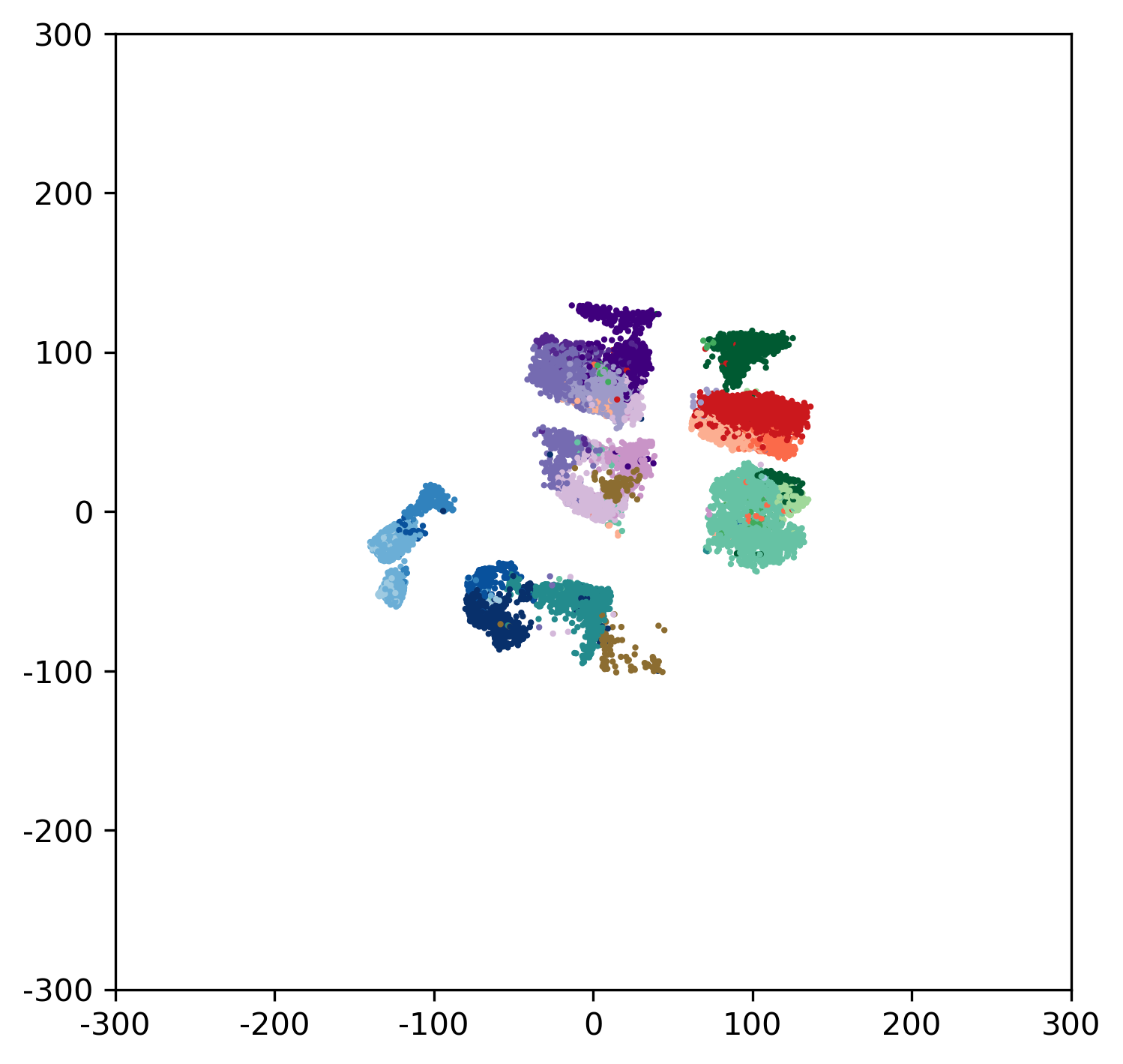}
        \caption*{$\alpha=2$}
    \end{subfigure}\hfill
    \begin{subfigure}[b]{0.18\textwidth}
        \centering
        \includegraphics[width=\textwidth]{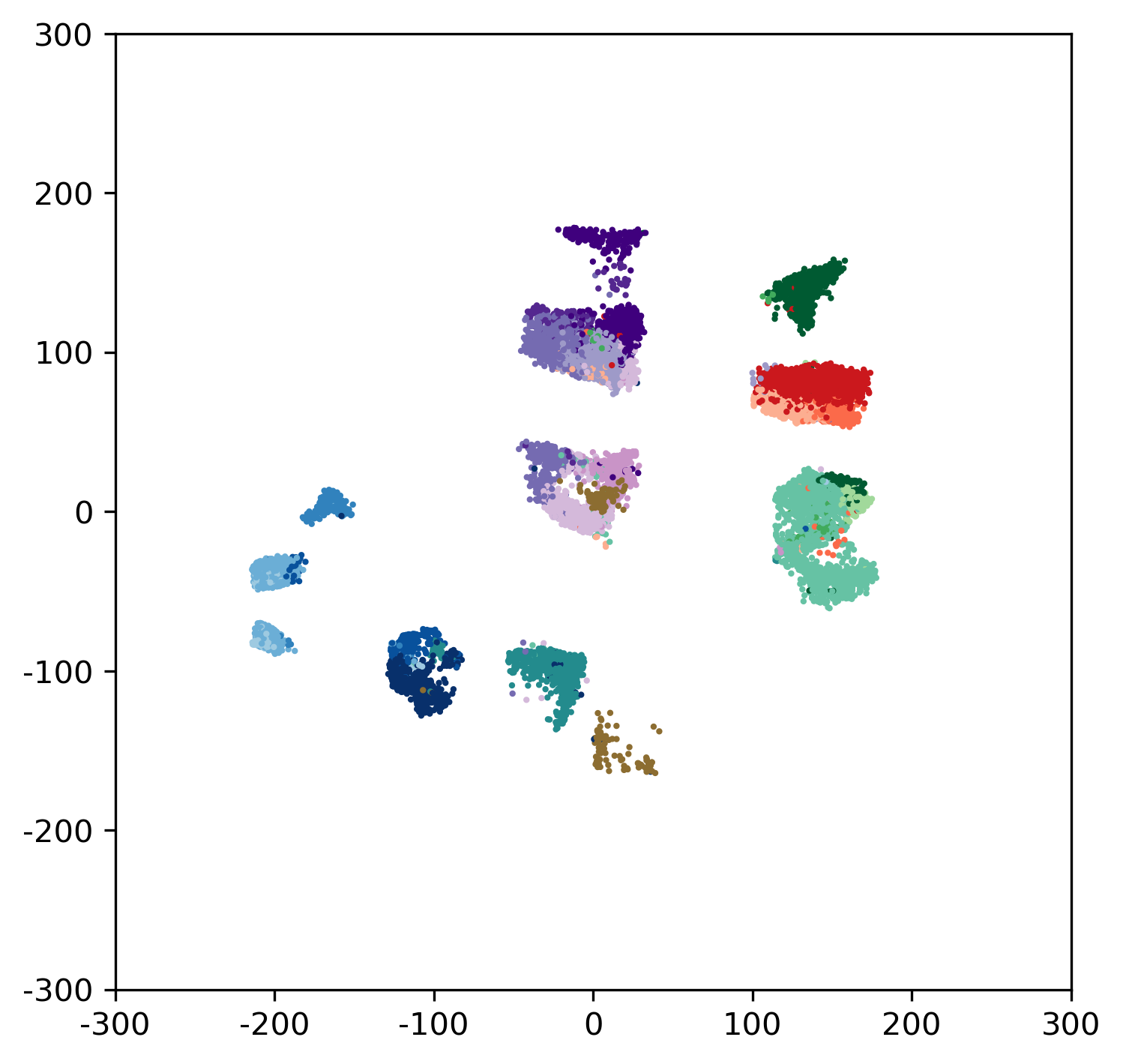}
        \caption*{$\alpha=3$}
      \end{subfigure}\hfill
      \begin{subfigure}[b]{0.18\textwidth}
        \centering
        \includegraphics[width=\textwidth]{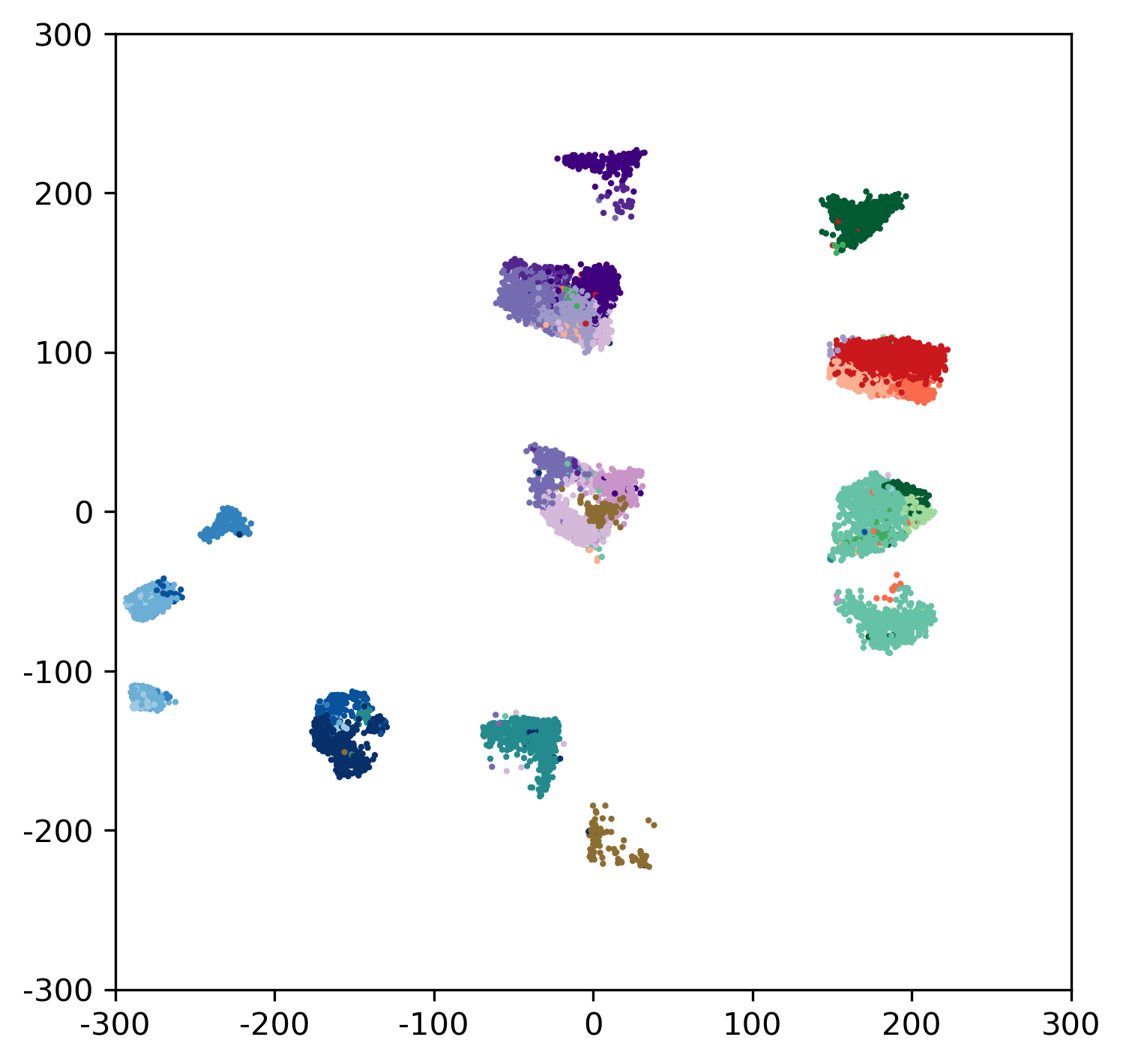}
        \caption*{$\alpha=4$}
    \end{subfigure}

% ----- Row 2 -----
\begin{center}
\includegraphics[width=0.45\textwidth]{figs/brain/bio_cell_horizontal.png}
\end{center}

    \caption{Embedding of the human brain dataset with C+E (L-Isomap) as the $\alpha$ parameter is adjusted, increasing the separation between the thirteen embedded clusters. Note that we keep the axes at the same scale for all $\alpha$ values. Coloring is by cell-type.}
    \label{fig:brain_alpha}
\end{figure}

\begin{table}[!ht]
\centering
\footnotesize
\setlength{\tabcolsep}{3pt}
\begin{tabular}{llccccccccccc}
\toprule
\textbf{Dataset} & \textbf{$\alpha$}
& \multicolumn{4}{c}{\textbf{Local}} 
& \multicolumn{7}{c}{\textbf{Global}} \\
\cmidrule(lr){3-6} \cmidrule(lr){7-13}
& 
& W-S ${\scriptstyle \uparrow}$ 
& W-NS ${\scriptstyle \downarrow}$ 
& W-SNS ${\scriptstyle \downarrow}$ 
& KNN ${\scriptstyle \uparrow}$
& B-S ${\scriptstyle \uparrow}$  
& T-S ${\scriptstyle \uparrow}$  
& B-NS ${\scriptstyle \downarrow}$ 
& T-NS ${\scriptstyle \downarrow}$ 
& B-SNS ${\scriptstyle \downarrow}$ 
& T-SNS ${\scriptstyle \downarrow}$  
& CP ${\scriptstyle \uparrow}$
 \\
\midrule

MNIST & $1.00$ &
$\mathbf{0.81}$ & $0.52$ & $0\mathbf{.31}$ & $0.26$ & $\mathbf{0.43}$ & $\mathbf{0.63}$ & $0.35$ & $0.36$ & $0.28$ & $\mathbf{0.34}$ & $\mathbf{0.74}$\\

 & $1.25$ &
$\mathbf{0.81}$ & $\mathbf{0.51}$ & $0.35$ & $0.31$ & $0.41$ & $0.62$ & $\mathbf{0.33}$ & $\mathbf{0.35}$ & $0.26$ & $\mathbf{0.34}$ & $0.68$ \\

 & $2.25$ &
$0.80$ & $0.57$ & $0.53$ & $\mathbf{0.38}$ & $0.39$ & $0.58$ & $0.80$ & $0.92$ & $0.22$ & $0.35$ & $0.64$
 \\
 
  & $2.75$ &
$0.80$ & $0.62$ & $0.59$ & $\mathbf{0.38}$ & $0.38$ & $0.52$ & $1.16$ & $1.30$ & $\mathbf{0.21}$ & $0.37$ & $0.54$
 \\
 
\midrule
Brain & $1.00$ &
$0.71$ & $0.53$ & $\mathbf{0.38}$ & $0.30$ & $0.44$ & $0.88$ & $0.41$ & $0.38$ & $0.23$ & $0.24$ & $0.90$

\\

& $1.45$ &
$0.72$ & $\mathbf{0.52}$ & $0.41$ & $0.35$ & $0.46$ & $0.90$ & $\mathbf{0.28}$ & $\mathbf{0.23}$ & $0.21$ & $\mathbf{0.23}$ & $0.92$
\\

%  & $1.75$  &
% $\mathbf{0.82}$ & $\mathbf{0.44}$ & $0.39$ & $0.37$ & $\mathbf{0.68}$ & $0.90$ & $0.30$ & $0.31$ & $0.20$ & $0.24$ & $0.94$
%  \\

 & $2.00$ &
$\mathbf{0.74}$ & $0.53$ & $0.44$ & $0.37$ & $\mathbf{0.48}$ & $\mathbf{0.91}$ & $0.39$ & $0.44$ & $\mathbf{0.19}$ & $\mathbf{0.23}$ & $\mathbf{0.93}$\\

 & $3.00$ &
$0.73$ & $0.74$ & $0.52$ & $\mathbf{0.38}$ & $0.46$ & $\mathbf{0.91}$ & $0.87$ & $1.05$ & $0.20$ & $0.24$ & $\mathbf{0.93}$
\\

 & $4.00$ &
$0.73$ & $1.07$ & $0.58$ & $\mathbf{0.38}$ & $0.46$ & $0.90$ & $1.45$ & $1.72$ & $0.21$ & $0.25$ & $0.92$\\

\bottomrule
\end{tabular}
\caption{Evaluation of the MNIST and human brain organoid datasets against geodesic distances as $\alpha$ varies.  Cell types are used as the ground truth labels for the human brain organoid dataset. Metrics are reported for within class (W), between class (B), and total (T). S denotes Spearman correlation, NS normalized stress, SNS scale-normalized stress, CP class preservation, and KNN the $30$-NN recall. Arrows indicate whether higher ($\uparrow$) or lower ($\downarrow$) values are better.}
\label{tab:alpha_tuning}
\end{table}

In this section, we fix the number of clusters and cluster embeddings (L-Isomap) as in the main text, and adjust $\alpha$ at the alignment step for both the MNIST and human brain organoid datasets. In \figref{mnist_alpha}, we show that the relative positioning of the clusters is generally stable as $\alpha$ varies, with the spacing between the clusters increasing as $\alpha$ increases. Observe that as the space between the clusters increases, the clusters corresponding to the digits four and nine, for example, still stay relatively close together.  In \tabref{alpha_tuning}, the local distance metrics generally worsen as $\alpha$ is increased, with the exception of $k$NN recall. As the local distance metrics are averaged over the class digit labels and some instances are misclustered, this is not too surprising. Poor $k$NN recall at smaller values of $\alpha$ is likely due to overlaps between clusters. At the global level, normalized stress favors smaller $\alpha$, as increasing $\alpha$ does scale the between-cluster distances to be larger than in the original data. In contrast, the scale-invariant global metrics, Spearman correlation and scale-normalized stress, are less sensitive to $\alpha$, though generally favor smaller values of $\alpha$. 

On the human brain organoid dataset in \figref{brain_alpha}, not until $\alpha = 4$ is the separation between all thirteen clusters clear. For the smaller values of $\alpha$, the embedding suggests a more connected trajectory between the clusters, whereas for larger values of $\alpha$ the clusters appear as more discrete entities. The metrics exhibit similar trends to the MNIST dataset. 

\section{Runtime}
\label{sec:runtime}

In \tabref{all_method_runtime}, we record the runtime of each method on each dataset. We used the {\sf scanpy} implementation of the Leiden algorithm \cite{wolf2018scanpy} and {\sf openTSNE} \cite{Policar2024}, with  FFT-accelerated gradient computation for $n \geq 20k$ to obtain t-SNE embeddings. The runtime of C+E is generally comparable to that of t-SNE and UMAP. It is faster than both on small datasets $(n \leq 5k)$, faster than t-SNE on the MNIST and FMIST datasets, though is slower than the other methods on the Human Brain and Mouse Cortex datasets.

In \tabref{ce_runtime}, we record the runtime of each step of C+E. Distance calculation refers to the calculation of a subset of all pairwise distances which are then used in the alignment step. The calculation of geodesic distances is much slower than calculation of Euclidean distances, as this includes the construction of a $k$NN graph, and finding shortest paths. The runtime of the clustering and embedding steps increases as dataset size increases, and because we use a subsampling procedure to align the cluster-level embeddings, the runtime of the alignment step primarily depends on the number of clusters. Thus, even though the mouse cortex dataset is smaller than MNIST and FMNIST, because the clustering step of C+E identifies a larger number of clusters, the alignment step, and consequently the overall runtime, is slower.

% \gal{add a bit more detail on comparing the methods and how the number of points / number of clusters impacts runtime performance, for example, compared to t-SNE}.

\begin{table}[h!]
    \centering
    \small
    \begin{tabular}{ccccccc}
    \toprule
        {\bf Method} & {\bf Shapes} &  {\bf GMM} & {\bf MNIST}  & {\bf Human Brain}  & {\bf Mouse Cortex} & {\bf FMNIST} \\
         & $n = 1k$ &  $n = 5k$  &  $n = 60k$  & $n = 20k$   &  $n = 24k$ & $n=60k$ \\
        
             \midrule
        C + E ($\alpha = 1$) & $0.64$  & $8.21$  & $111.57$ & $159.88$ & $290.70$ & $121.46$ \\
        C + E ($\alpha > 1$) & $0.66$  & $9.27$  & $110.17$ & $159.84$ & $232.52$ & $123.56$ \\
        t-SNE & $4.04$ & $18.34$ & $225.46$ & $99.00$ & $133.27$ & $248.81$ \\
        PCA & $0.00$ & $0.00$ & $0.03$ & $0.01$ & $0.02$ & $0.02$ \\
        Isomap & $2.32$ & $9.12$ & - & - & - & - \\
        L-Isomap & - & - & $97.56$ & $52.52$ & $51.18$ & $98.27$ \\
        UMAP & $18.29$ & $41.30$ & $36.77$ & $29.38$ & $32.06$ & $38.82$ \\
        PHATE & $5.86$ & $13.85$ & $33.82$ & $19.67$ & $20.05$ & $35.47$ \\
        TriMAP & $1.05$ & $4.78$ & $59.04$ & $22.93$ & $28.50$ & $68.52$ \\
         \bottomrule
    \end{tabular}
    \caption{Runtime (in seconds) of each method.}
    \label{tab:all_method_runtime}
\end{table}

\begin{table}
    \centering
    \small
    \setlength{\tabcolsep}{3pt}
    \begin{tabular}{l r r c c c c c c}
    \toprule
    {\bf Dataset} & {\bf Size} & {\bf $\#$ Clusters} & {\bf $\alpha$} &
    \shortstack{\bf Embedding\\\bf Method} &
    {\bf Cluster} &
    {\bf Embed} &
    \shortstack{\bf Distance\\\bf Calculation} &
    {\bf Align} \\
    \midrule
    Shapes        & $1k$   & $3$   & 1    & PCA       & $0.02$  & $0.01$ & $0.00$ & $0.61$  \\
    Shapes        & $1k$   & $3$   & 1.25 & PCA       & $0.02$  & $0.01$  &$0.00$ & $0.63$  \\
    
    GMM           & $5k$   & $10$  & 1    & PCA       & $1.72$  & $0.01$  &$0.07$& $6.41$  \\
    GMM           & $5k$   & $10$  & 1.65 & PCA       & $1.72$  & $0.01$  &$0.07$& $7.47$  \\
    GMM           & $5k$   & $10$  & 2    & PCA       & $1.72$  & $0.01$  & $0.07$& $6.46$  \\
    
    MNIST         & $60k$  & $10$  & 1    & L-Isomap  & $49.12$ & $8.81$  & $44.82$& $8.82$ \\
    MNIST         & $60k$  & $10$  & 1.75 & L-Isomap  & $49.12$ & $8.81$  &$44.82$& $7.42$ \\
    MNIST         & $60k$  & $10$  & 1    & TriMAP    & $49.12$ & $57.27$ &$44.82$& $13.32$ \\
    MNIST         & $60k$  & $10$  & 1    & PCA       & $49.12$ & $0.07$  &$0.20$& $8.20$  \\
    MNIST         & $60k$  & $10$  & 1.83 & PCA       & $49.12$ & $0.07$  &$0.20$& $7.07$  \\
    
    Human Brain   & $20k$  & $13$  & 1    & L-Isomap  & $41.59$ & $3.01$  &$102.27$& $13.01$ \\
    Human Brain   & $20k$  & $13$  & 1.45 & L-Isomap  & $41.59$ & $3.01$  &$102.27$& $12.97$ \\
    Human Brain   & $20k$  & $13$  & 1    & TriMAP    & $41.59$ & $83.82$ &$102.27$& $12.58$ \\
    % Human Brain   & $20k$  & $13$  & 1    & PCA       & $41.59$ & $0.06$  &$0.18$& $11.72$ \\
    % Human Brain   & $20k$  & $13$  & 1.86 & PCA       & $41.59$ & $0.06$  &$0.18$& $12.72$ \\

    Mouse Cortex  & $24k$  & $46$  & 1    & PCA       & $38.53$ & $0.20$  &$0.27$& $251.43$ \\
    Mouse Cortex  & $24k$  & $46$  & 4.27 & PCA       & $38.53$ & $0.20$  &$0.27$& $193.52$ \\
    Mouse Cortex  & $24k$  & $46$  & 4.63 & TriMAP    & $38.53$ & $103.93$ &$0.27$& $192.08$ \\

    FMNIST        & $60k$  & $10$  & 1    & L-Isomap  & $55.10$ & $14.69$ &$44.48$& $7.19$ \\
    FMNIST        & $60k$  & $10$  & 1.7  & L-Isomap  & $55.10$ & $14.69$ &$44.48$& $9.29$ \\
    FMNIST        & $60k$  & $10$  & 1    & TriMAP    & $55.10$ & $64.14$ &$44.48$& $6.37$ \\
    
    \bottomrule
    \end{tabular}
    \caption{Runtime (in seconds) for each step of the C+E approach.}
    \label{tab:ce_runtime}
\end{table}

\section{Additional Results}
\label{sec:additional_metrics}

\subsection{GMM}
 \captionsetup{justification=centering, singlelinecheck=false, font = footnotesize}
In \secref{crowding}, we depict a $d$-dimensional isotropic Gaussian mixture model (GMM) with $d$ components. Specifically, the mean vector of each component is $\frac{5}{4}\sqrt{d} e_i$ where the $i$th coordinate of $e_i$ is one and zero otherwise, and we set $d = 10$. The complete metrics and additional figures on this example follow. 

\begin{figure}[h!]
    \begin{subfigure}[b]{0.18\textwidth}
        \centering
        \includegraphics[width=\textwidth]{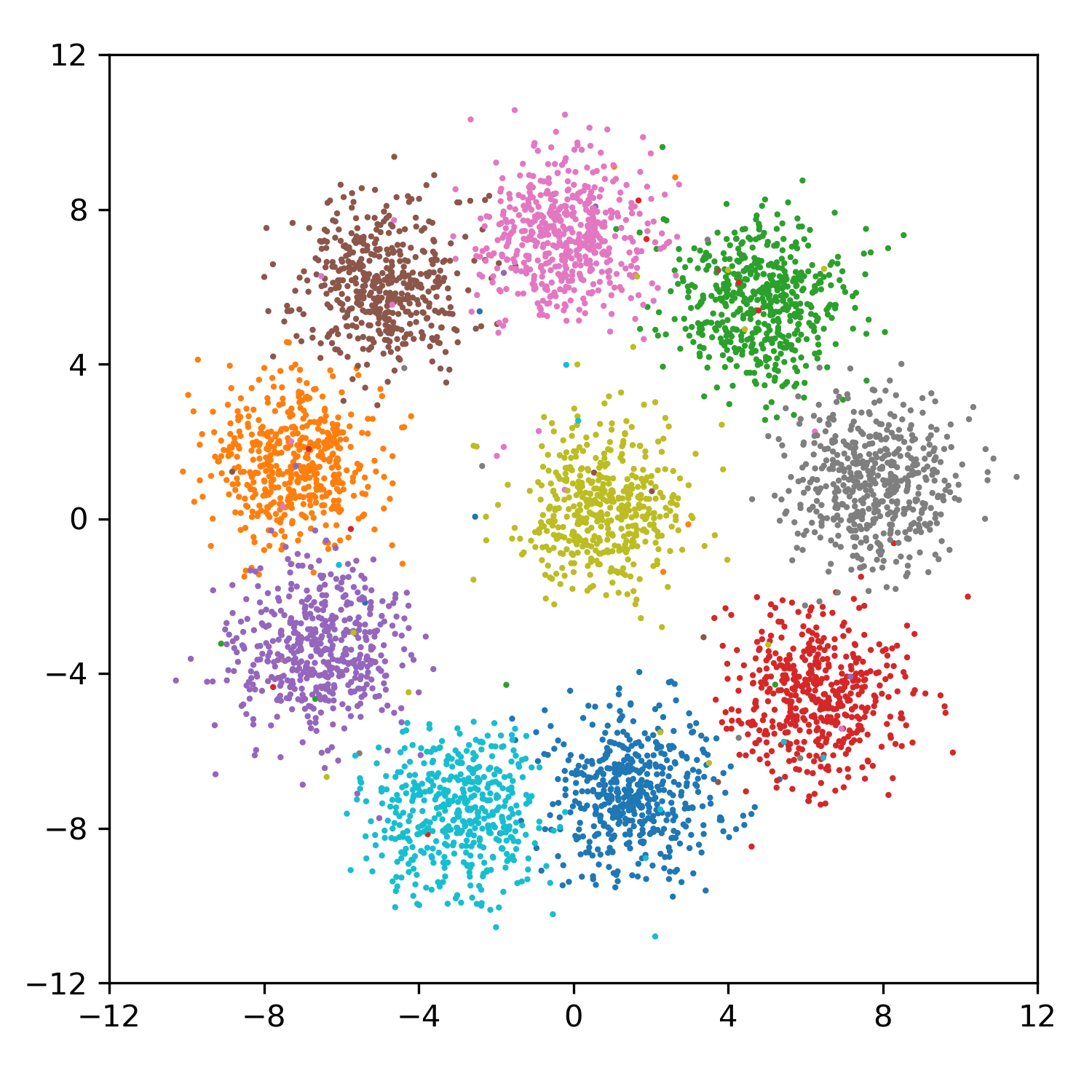}
        \caption*{C+E \\ (PCA, $\alpha=1.65$)}
    \end{subfigure}\hfill
        \begin{subfigure}[b]{0.18\textwidth}
        \centering
        \includegraphics[width=\textwidth]{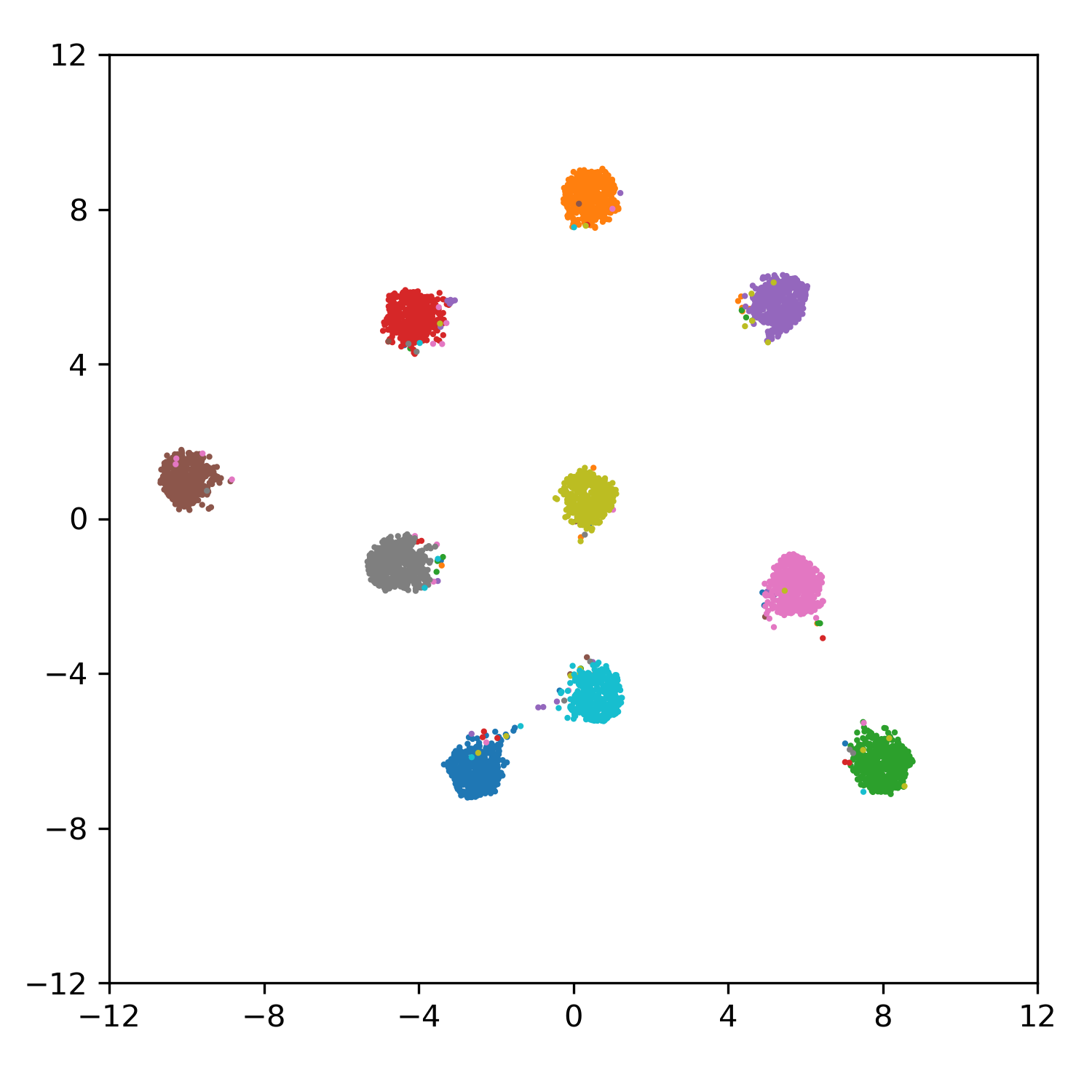}
        \caption*{UMAP \\  \strut}
    \end{subfigure}\hfill
    \begin{subfigure}[b]{0.18\textwidth}
        \centering
        \includegraphics[width=\textwidth]{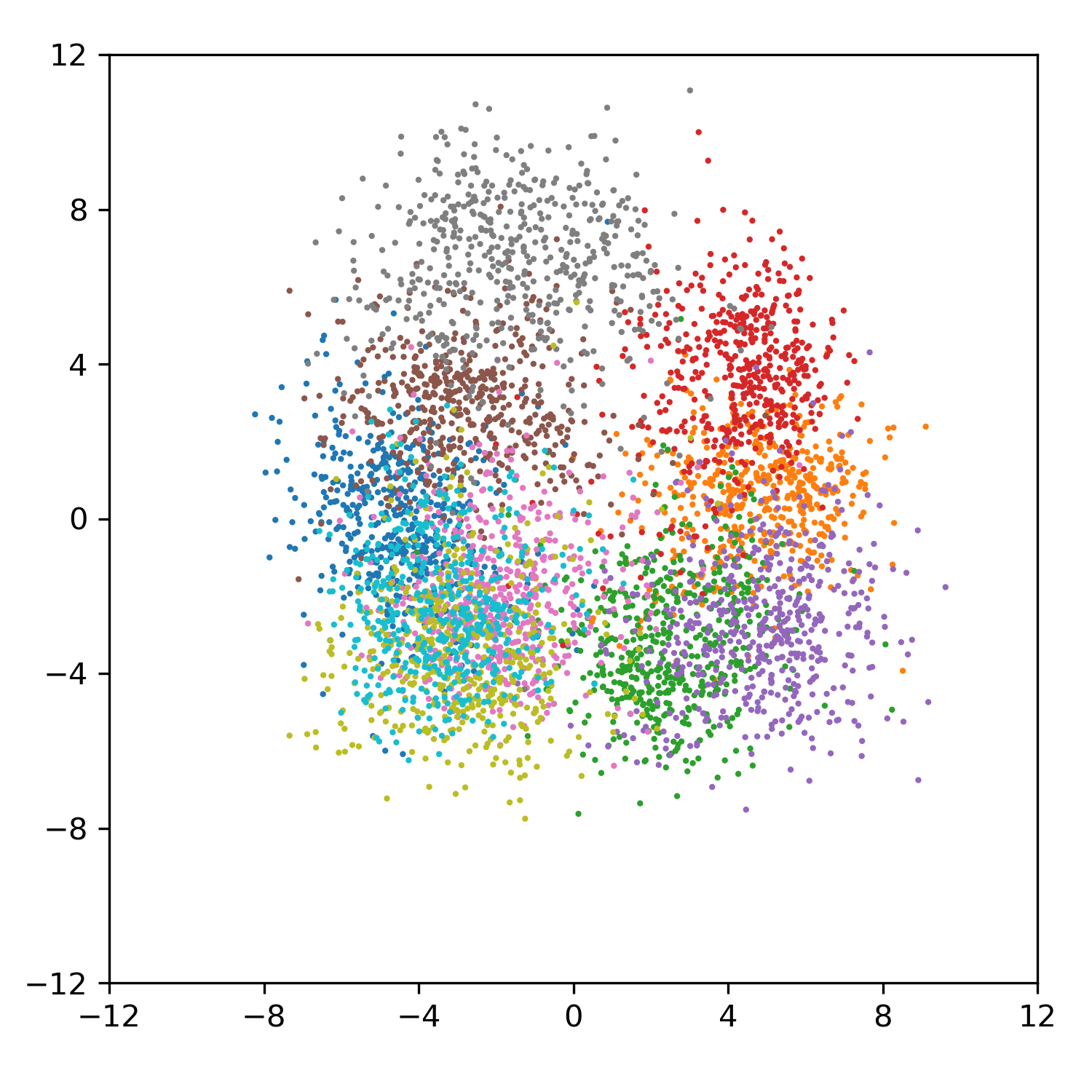}
        \caption*{Isomap \\  \strut}
    \end{subfigure}\hfill
    \begin{subfigure}[b]{0.18\textwidth}
        \centering
        \includegraphics[width=\textwidth]{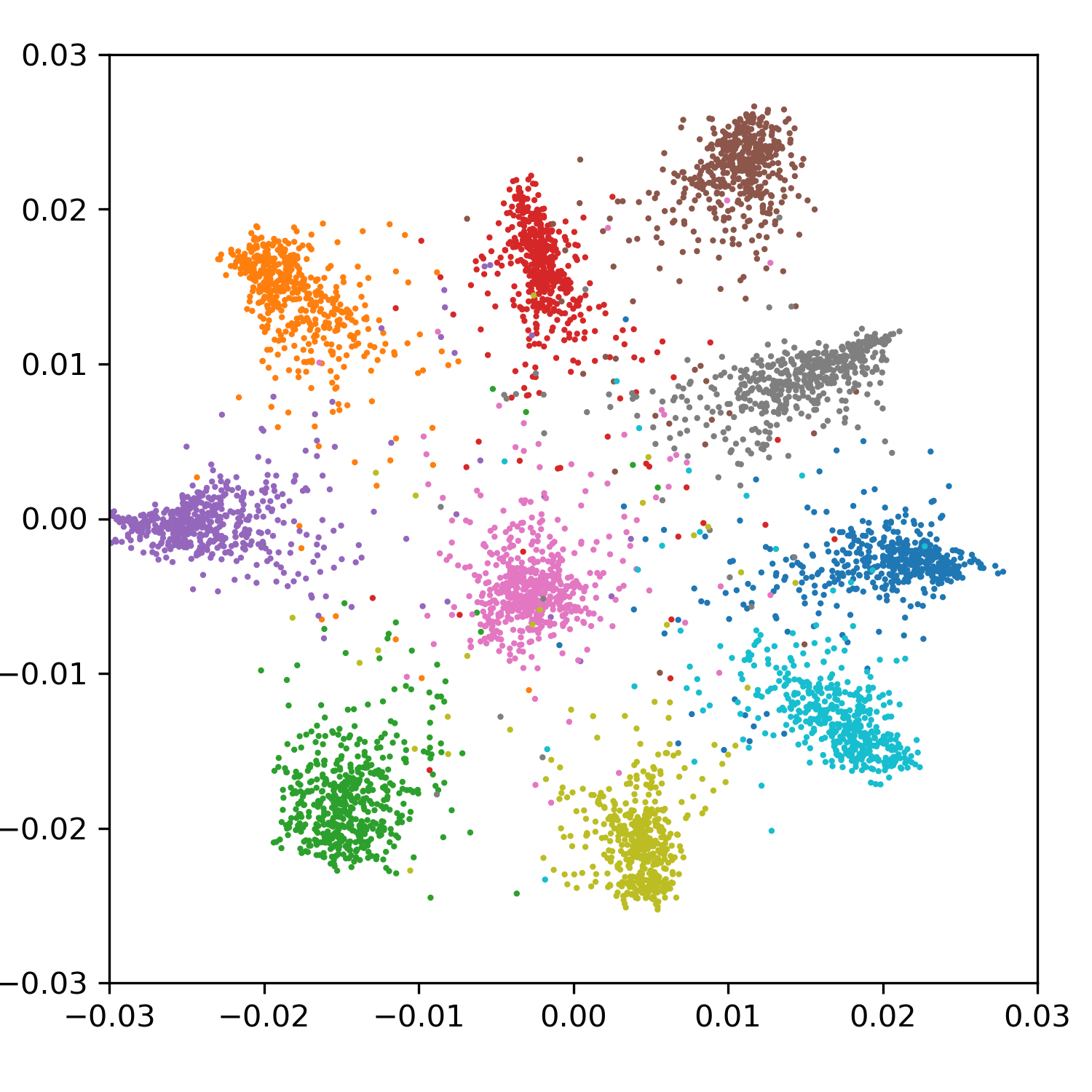}
        \caption*{PHATE \\  \strut}
    \end{subfigure}\hfill
        \begin{subfigure}[b]{0.18\textwidth}
        \centering
        \includegraphics[width=\textwidth]{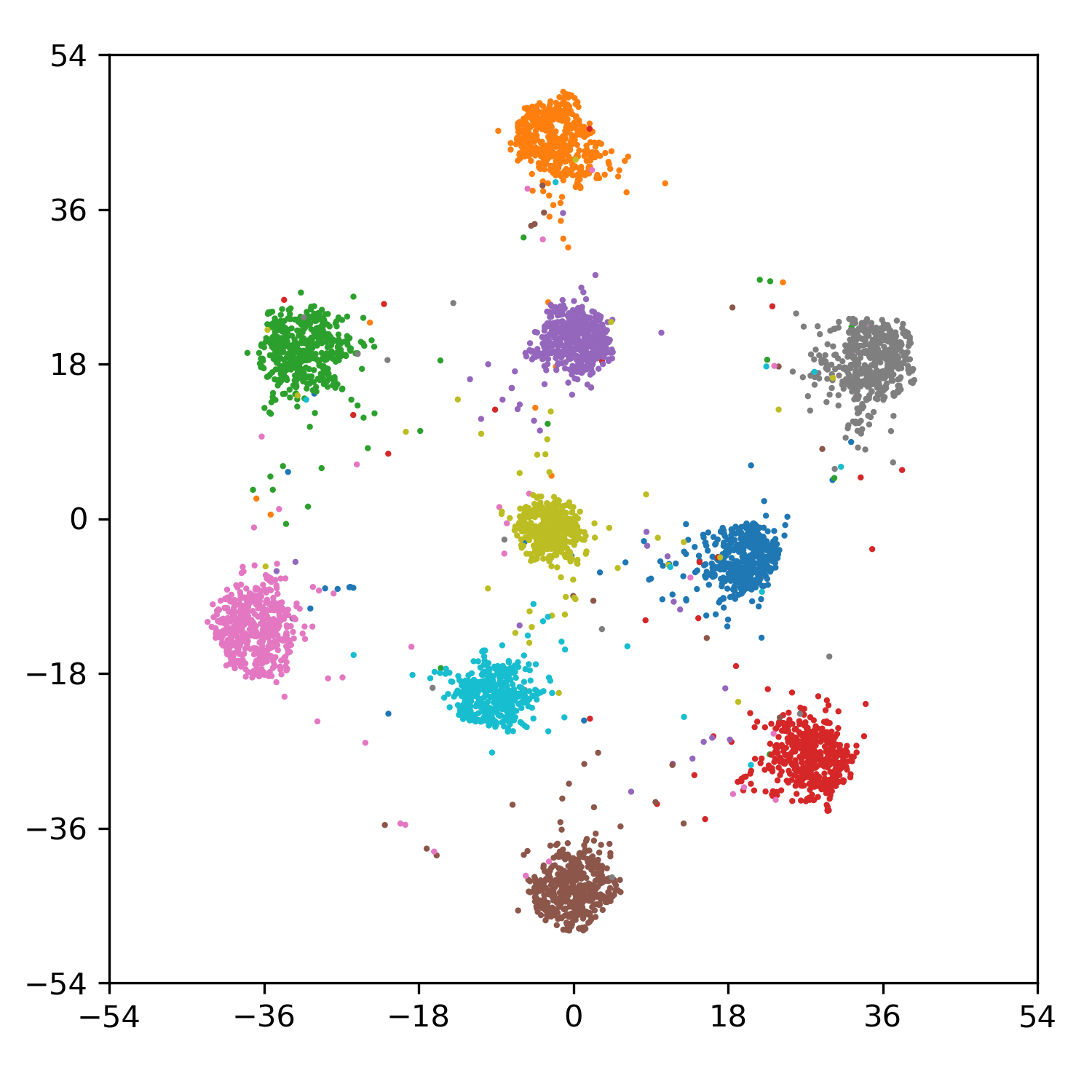}
        \caption*{TriMap \\  \strut}
    \end{subfigure}\hfill
    
    \caption{Additional embeddings for data generated from a $10$-dimensional isotropic Gaussian mixture model ($n = 5,000$). Points are colored by the mixture component from which they were generated. These labels were not used by the embedding methods.}
    \label{fig:crowding_problem_app}
\end{figure}

\begin{table}[!ht]
\centering
\footnotesize
\setlength{\tabcolsep}{2pt}
\begin{tabular}{llccccccccccc}
\toprule
\textbf{Method} & \textbf{Params} 
& \multicolumn{4}{c}{\textbf{Local}} 
& \multicolumn{7}{c}{\textbf{Global}} \\
\cmidrule(lr){3-6} \cmidrule(lr){7-13}
& 
& W-S ${\scriptstyle \uparrow}$ 
& W-NS ${\scriptstyle \downarrow}$ 
& W-SNS ${\scriptstyle \downarrow}$ 
& KNN ${\scriptstyle \uparrow}$
& B-S ${\scriptstyle \uparrow}$  
& T-S ${\scriptstyle \uparrow}$  
& B-NS ${\scriptstyle \downarrow}$ 
& T-NS ${\scriptstyle \downarrow}$ 
& B-SNS ${\scriptstyle \downarrow}$ 
& T-SNS ${\scriptstyle \downarrow}$  
& CP ${\scriptstyle \uparrow}$
 \\
\midrule

C+E (PCA) & $\alpha = 1$ &
$\mathbf{0.48}$ & $0.59$ & $0.51$ & $0.11$ & $0.24$ & $0.32$ & $\mathbf{0.37}$ & $\mathbf{0.40}$ & $0.30$ & $0.40$ & $-0.12$\\

C+E  (PCA) & $\alpha = 1.65$ &
$\mathbf{0.48}$ & $0.69$ & $0.67$ & $0.17$ & $0.21$ & $0.30$ & $0.88$ & $0.99$ & $0.23$ & $0.40$ & $0.08$ \\

C+E  (PCA) & $\alpha = 2$ &
$\mathbf{0.48}$ & $0.64$ & $0.62$ & $0.17$ & $0.24$ & $0.32$ & $0.64$ & $0.72$ & $0.24$ & $\mathbf{0.39}$ & $0.38$ \\

t-SNE & $u = 30$ &
$0.39$ & $3.26$ & $0.64$ & $\mathbf{0.33}$ & $0.30$ & $0.35$ & $8.08$ & $8.49$ & $0.22$ & $0.40$ & $0.45$
 \\
 
 UMAP & $q = 15$ &
$0.34$ & $0.86$ & $0.85$ & $0.23$ & $0.23$ & $0.29$ & $0.52$ & $0.62$ & $0.22$ & $0.42$ & $0.19$\\

PCA & - &
$0.42$ & $0.61$ & $\mathbf{0.41}$ & $0.05$ & $\mathbf{0.43}$ & $\mathbf{0.42}$ & $0.59$ & $0.60$ & $0.35$ & $0.41$ & $0.36$ \\

L-Isomap & $q = 15$  &
$0.26$ & $\mathbf{0.51}$ & $0.49$ & $0.08$ & $0.29$ & $0.33$ & $0.45$ & $0.48$ & $0.31$ & $0.44$ & $0.29$
 \\

PHATE & $q = 5$ &
$0.26$ & $1.00$ & $0.67$ & $0.16$ & $\mathbf{0.43}$ & $0.36$ & $1.00$ & $1.00$ & $\mathbf{0.20}$ & $\mathbf{0.39}$ & $0.12$ \\

TriMap & - &
$0.34$ & $2.07$ & $0.77$ & $0.22$ & $0.25$ & $0.31$ & $5.56$ & $5.96$ & $\mathbf{0.20}$ & $0.42$ & $\mathbf{0.60}$\\
\bottomrule
\end{tabular}
\caption{Evaluation of the GMM dataset against Euclidean distances. Metrics are reported for within class (W), between class (B), and total (T). S denotes Spearman correlation, NS normalized stress, SNS scale-normalized stress, CP class preservation, and KNN the $30$-NN recall. Arrows indicate whether higher ($\uparrow$) or lower ($\downarrow$) values are better.}
\label{tab:gmm}
\end{table}

Note that the class preservation metric is not particularly meaningful in this example because in the underlying model, the distance is the same between every pair of clusters. In other words, if in the original sample, cluster $i$ is closer to cluster $j$ than cluster $\kappa$, this is an artifact of the sample. 

\begin{figure}[h!]
    \centering
    \begin{subfigure}[c]{0.4\textwidth}
        \centering
        \includegraphics[width=\textwidth]{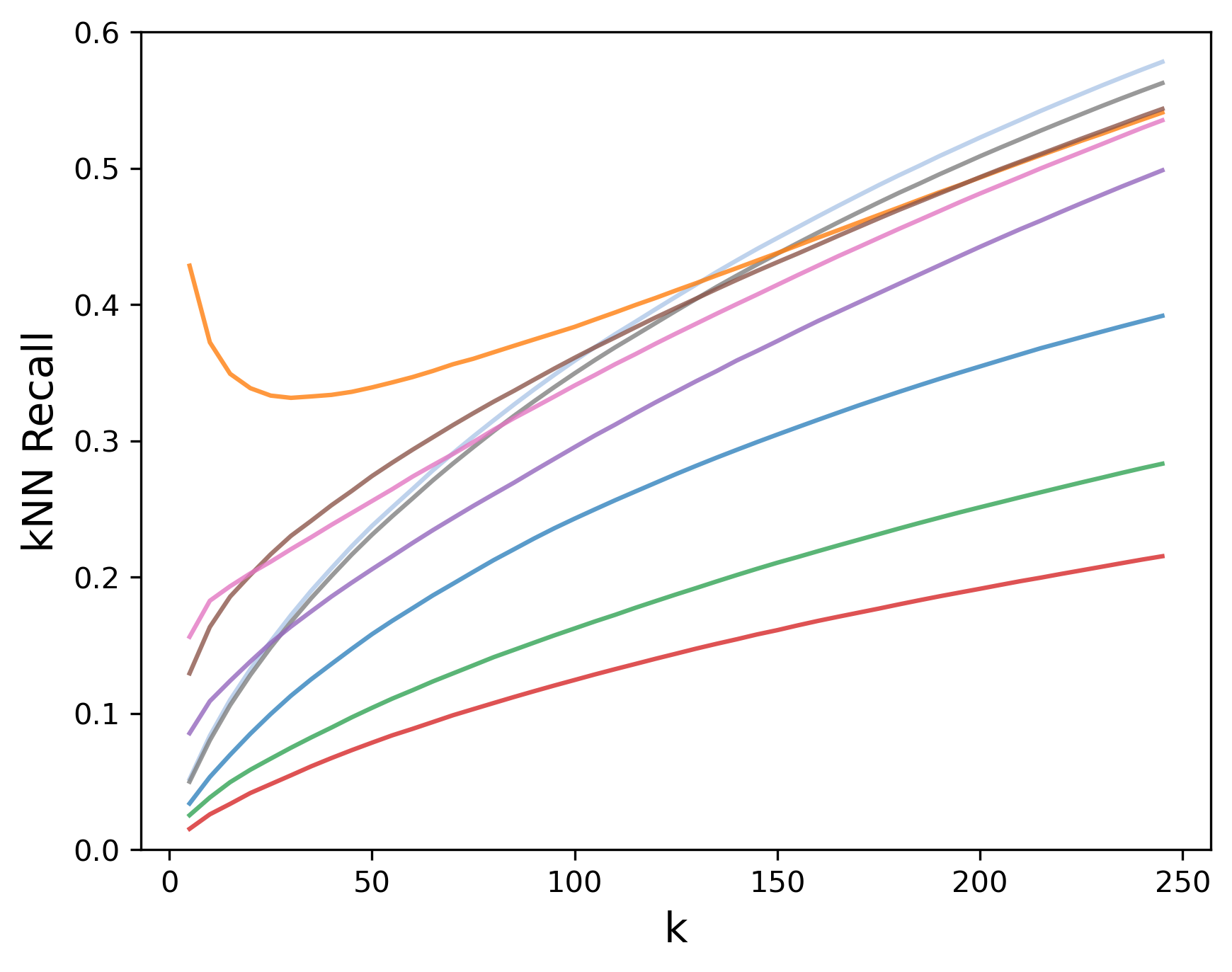}
    \end{subfigure}
    \hspace{0.03\textwidth}
    \begin{subfigure}[c]{0.18\textwidth}
        \centering
        \includegraphics[width=\textwidth]{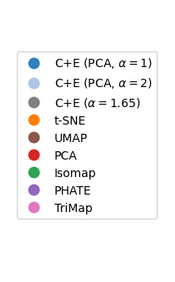}
    \end{subfigure}
    \caption{$k$NN recall for the GMM dataset for various values of $k$ for all methods compared.}
\end{figure}
\FloatBarrier

\subsection{MNIST: Euclidean-based embeddings and evaluation}
\label{sec:mnist_app}

\begin{figure}[h!]
\centering
    \begin{subfigure}[c]{0.225\textwidth}
        \centering
        \includegraphics[width=\textwidth]{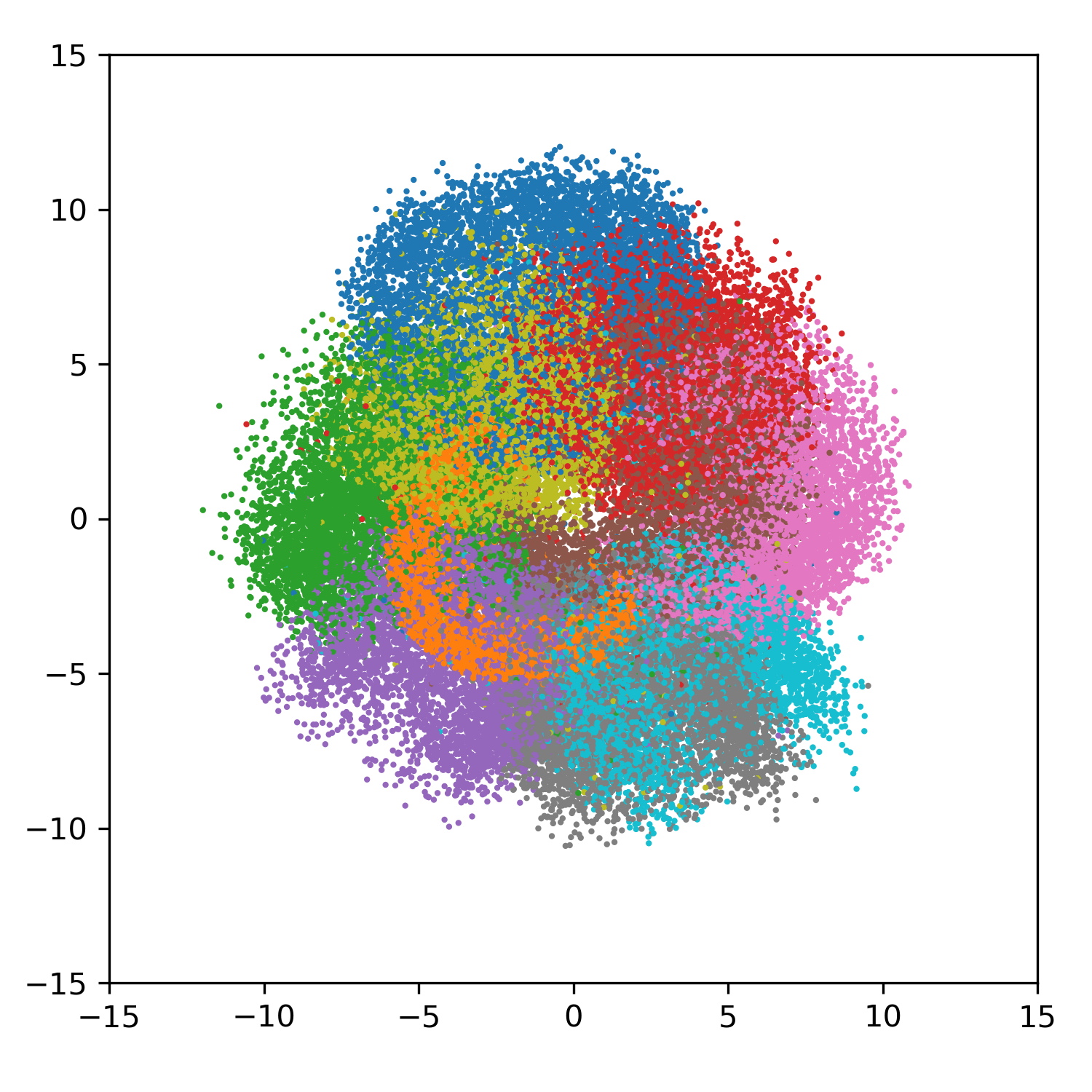}
        \caption*{C+E (PCA, $\alpha=1$)}
    \end{subfigure} \hspace{0.02\textwidth}
        \begin{subfigure}[c]{0.225\textwidth}
        \centering
        \includegraphics[width=\textwidth]{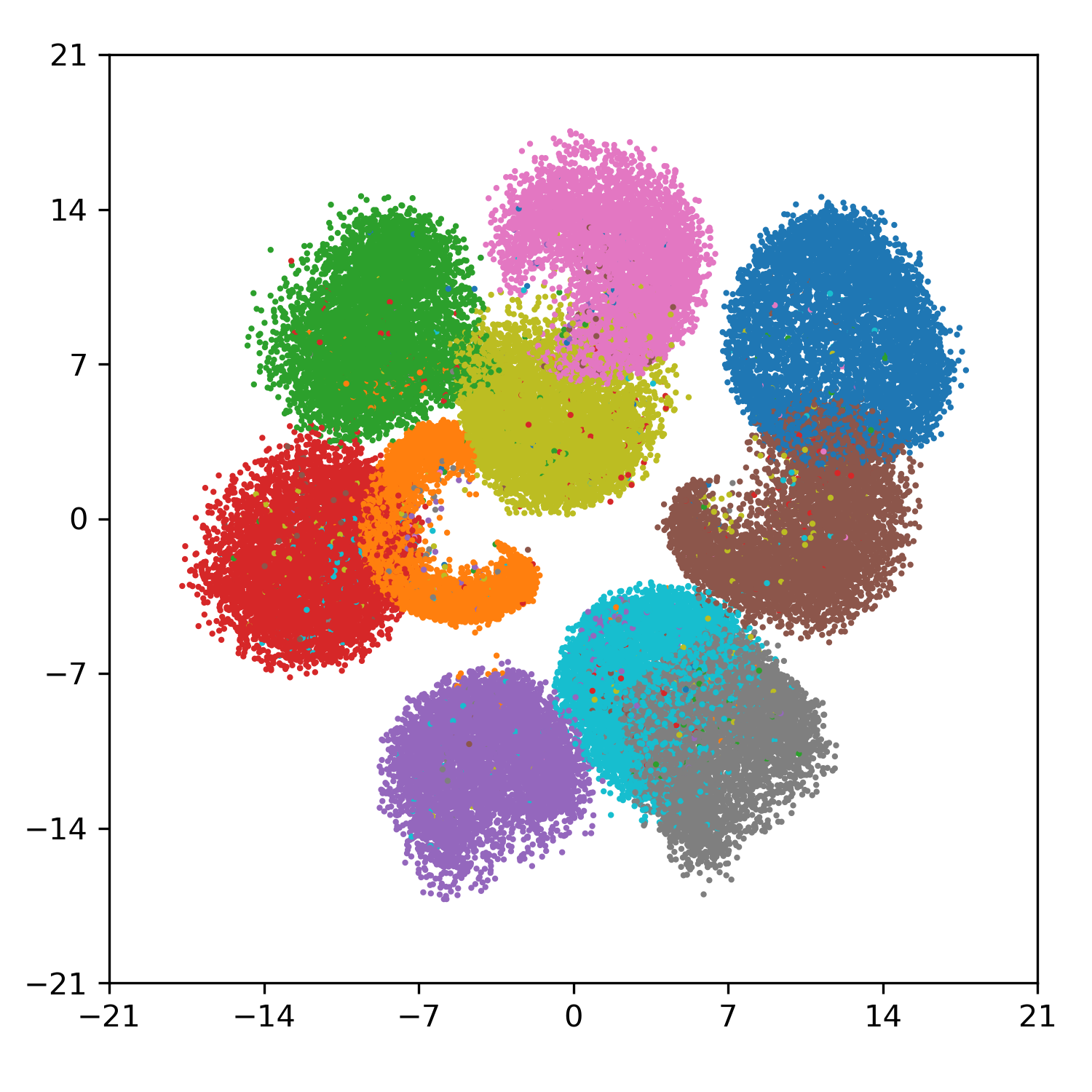}
        \caption*{C+E (PCA, $\alpha=1.83$)}
    \end{subfigure} \hspace{0.02\textwidth}
    \begin{subfigure}[c]{0.225\textwidth}
        \centering
        \includegraphics[width=\textwidth]{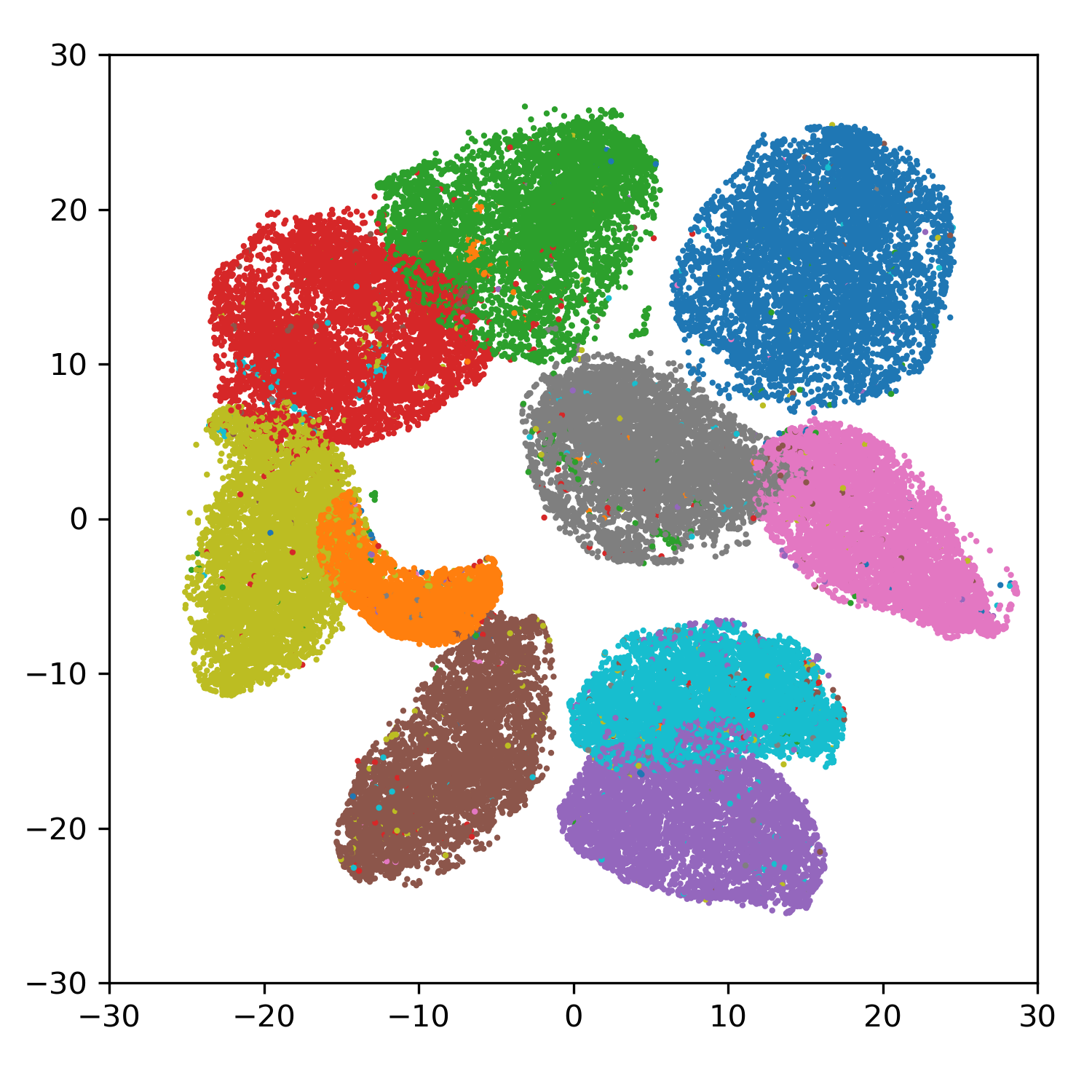}
        \caption*{C+E (TriMap, $\alpha = 3.05$) }
    \end{subfigure} \hspace{0.02\textwidth}
    \begin{subfigure}[c]{0.05\textwidth}
        \centering
        \includegraphics[width=\textwidth]{figs/mnist_colorbar.png}
        \caption*{}
    \end{subfigure}\hfill
    
    \caption{C+E embeddings of MNIST dataset ($n = 60,000$) with the cluster-level embeddings aligned to preserve  Euclidean distances. Points are colored according to their original digit label. These labels were not used by the embedding methods.
}
    \label{fig:mnist_euclidean}
\end{figure}

\begin{table}[!ht]
\centering
\footnotesize
\setlength{\tabcolsep}{2pt}
\begin{tabular}{llccccccccccc}
\toprule
\textbf{Method} & \textbf{Params} 
& \multicolumn{4}{c}{\textbf{Local}} 
& \multicolumn{7}{c}{\textbf{Global}} \\
\cmidrule(lr){3-6} \cmidrule(lr){7-13}
& 
& W-S ${\scriptstyle \uparrow}$ 
& W-NS ${\scriptstyle \downarrow}$ 
& W-SNS ${\scriptstyle \downarrow}$ 
& KNN ${\scriptstyle \uparrow}$
& B-S ${\scriptstyle \uparrow}$  
& T-S ${\scriptstyle \uparrow}$  
& B-NS ${\scriptstyle \downarrow}$ 
& T-NS ${\scriptstyle \downarrow}$ 
& B-SNS ${\scriptstyle \downarrow}$ 
& T-SNS ${\scriptstyle \downarrow}$  
& CP ${\scriptstyle \uparrow}$
 \\
\midrule

C+E (PCA) & $\alpha = 1$ &
$\mathbf{0.72}$ & $0.48$ & $\mathbf{0.36}$ & $0.15$ & $0.43$ & $0.48$ & $\mathbf{0.37}$ & $\mathbf{0.38}$ & $0.33$ & $\mathbf{0.38}$ & $0.31$\\

C+E  (PCA) & $\alpha = 1.83$ &
$0.71$ & $0.50$ & $0.46$ & $0.28$ & $0.39$ & $0.45$ & $0.76$ & $0.84$ & $0.24$ & $\mathbf{0.38}$ & $0.45$ \\

C+E  (TriMap) & $\alpha = 3.05$ &
$0.69$ & $0.57$ & $0.49$ & $0.37$ & $0.33$ & $0.39$ & $1.82$ & $1.91$ & $0.24$ & $0.39$ & $0.41$ \\

t-SNE & $u = 30$ &
$0.67$ & $3.32$ & $0.45$ & $\mathbf{0.39}$ & $0.23$ & $0.33$ & $8.71$ & $8.99$ & $0.25$ & $0.40$ & $0.38$
 \\
 
 UMAP & $q = 15$ &
$0.70$ & $0.76$ & $0.62$ & $\mathbf{0.39}$ & $0.22$ & $0.28$ & $0.44$ & $0.52$ & $\mathbf{0.21}$ & $0.44$ & $0.30$\\

PCA & - &
$0.49$ & $0.73$ & $0.43$ & $0.09$ & $\mathbf{0.55}$ & $0.55$ & $0.60$ & $0.61$ & $0.33$ & $0.40$ & $0.34$ \\

L-Isomap & $q = 15$  &
$0.52$ & $\mathbf{0.47}$ & $0.44$ & $0.11$ & $0.50$ & $\mathbf{0.56}$ & $0.74$ & $0.78$ & $0.32$ & $0.40$ & $\mathbf{0.48}$
 \\

PHATE & $q = 5$ &
$0.56$ & $1.00$ & $0.48$ & $0.22$ & $0.32$ & $0.32$ & $1.00$ & $1.00$ & $0.24$ & $0.44$ & $0.28$ \\

TriMap & - &
$0.67$ & $1.67$ & $0.49$ & $0.34$ & $0.27$ & $0.22$ & $4.47$ & $5.19$ & $0.24$ & $0.51$ & $0.26$\\

\bottomrule
\end{tabular}
\caption{Evaluation of the MNIST dataset against Euclidean distances for $5000$ points selected independently of the points used in alignment. Metrics are reported for within class (W), between class (B), and total (T). S denotes Spearman correlation, NS normalized stress, SNS scale-normalized stress, CP class preservation, and KNN the $30$-NN recall. Arrows indicate whether higher ($\uparrow$) or lower ($\downarrow$) values are better.}
\label{tab:mnist_euc}
\end{table}

In the main text, we embed each cluster using L-Isomap, which preserves geodesic distances, and align the clusters to preserve geodesic distances. The evaluation in the main text is also against geodesic distances. Here, we use the C+E method with the aim of preserving Euclidean distances. The clustering stays the same as in the main text, but embedding is done via PCA. All other methods remain the same and evaluation in \tabref{mnist_euc} is against Euclidean distances. Observe that the metrics exhibit similar trends as when evaluated against geodesic distances, but the overall performance is generally worse for most methods.  

\subsection{Human brain organoid data: Cell-type labels}
\label{sec:brain_cell}

 The within class and between class metrics depend on the ground truth class labels. In \tabref{brain} we evaluated the human brain organoid embeddings using the time point labels. Here, we evaluate using the cell-type labels as ground truth. While overall the metrics are worse, the general trends and observations regarding different methods remain unchanged. 

\begin{table}[!ht]
\centering
\footnotesize
\setlength{\tabcolsep}{2pt}
\begin{tabular}{llccccccccccc}
\toprule
\textbf{Method} & \textbf{Params} 
& \multicolumn{4}{c}{\textbf{Local}} 
& \multicolumn{7}{c}{\textbf{Global}} \\
\cmidrule(lr){3-6} \cmidrule(lr){7-13}
& 
& W-S ${\scriptstyle \uparrow}$ 
& W-NS ${\scriptstyle \downarrow}$ 
& W-SNS ${\scriptstyle \downarrow}$ 
& KNN ${\scriptstyle \uparrow}$
& B-S ${\scriptstyle \uparrow}$  
& T-S ${\scriptstyle \uparrow}$  
& B-NS ${\scriptstyle \downarrow}$ 
& T-NS ${\scriptstyle \downarrow}$ 
& B-SNS ${\scriptstyle \downarrow}$ 
& T-SNS ${\scriptstyle \downarrow}$  
& CP ${\scriptstyle \uparrow}$
 \\
\midrule

C+E (L-Isomap) & $\alpha = 1$ &
$0.71$ & $0.53$ & $\mathbf{0.38}$ & $0.30$ & $0.44$ & $0.88$ & $0.41$ & $0.38$ & $0.23$ & $0.24$ & $0.90$\\

C+E  (L-Isomap) & $\alpha = 1.45$ &
$0.72$ & $0.52$ & $0.41$ & $0.35$ & $\mathbf{0.46}$ & $0.90$ & $\mathbf{0.28}$ & $\mathbf{0.23}$ & $0.21$ & $\mathbf{0.23}$ & $0.92$ \\

C+E  (TriMap) & $\alpha = 1.95$ &
$0.74$ & $0.56$ & $0.45$ & $0.43$ & $0.43$ & $\mathbf{0.91}$ & $0.37$ & $0.40$ & $0.20$ & $\mathbf{0.23}$ & $0.91$ \\

t-SNE & $u = 30$ &
$\mathbf{0.78}$ & $\mathbf{0.51}$ & $0.39$ & $\mathbf{0.51}$ & $0.33$ & $0.68$ & $0.35$ & $0.35$ & $0.23$ & $0.32$ & $0.73$
 \\
 
 UMAP & $q = 15$ &
$0.77$ & $0.93$ & $0.53$ & $0.47$ & $0.19$ & $0.67$ & $0.88$ & $0.88$ & $0.23$ & $0.34$ & $0.64$
\\

PCA & - &
$0.50$ & $0.91$ & $0.47$ & $0.14$ & $0.38$ & $0.85$ & $0.86$ & $0.85$ & $0.24$ & $0.28$ & $0.86$ \\

L-Isomap & $q = 15$  &
$0.57$ & $0.70$ & $0.43$ & $0.18$ & $0.45$ & $0.91$ & $0.52$ & $0.46$ & $0.21$ & $0.27$ & $\mathbf{0.93}$
 \\

PHATE & $q = 5$ &
$0.60$ & $1.00$ & $0.49$ & $0.29$ & $0.23$ & $0.86$ & $1.00$ & $1.00$ & $0.22$ & $0.29$ & $0.85$ \\

TriMap & - &
$0.75$ & $0.75$ & $0.41$ & $0.43$ & $0.42$ & $0.87$ & $0.59$ & $0.55$ & $\mathbf{0.19}$ & $0.26$ & $0.85$\\

\bottomrule
\end{tabular}
\caption{Evaluation of the human brain organoid dataset against geodesic distances for $5000$ points selected independently of the points used in alignment. Cell types are taken as the ground truth class labels. Metrics are reported for within class (W), between class (B), and total (T). S denotes Spearman correlation, NS normalized stress, SNS scale-normalized stress, CP class preservation, and KNN the $30$-NN recall. Arrows indicate whether higher ($\uparrow$) or lower ($\downarrow$) values are better.}
\label{tab:brain_cell}
\end{table}

\subsection{Fashion MNIST}

\begin{figure}[h!]
    \centering
    \captionsetup{justification=centering, singlelinecheck=false, font = footnotesize}
    % ----- Row 1 -----
    \begin{subfigure}[b]{0.18\textwidth}
        \centering
        \includegraphics[width=\textwidth]{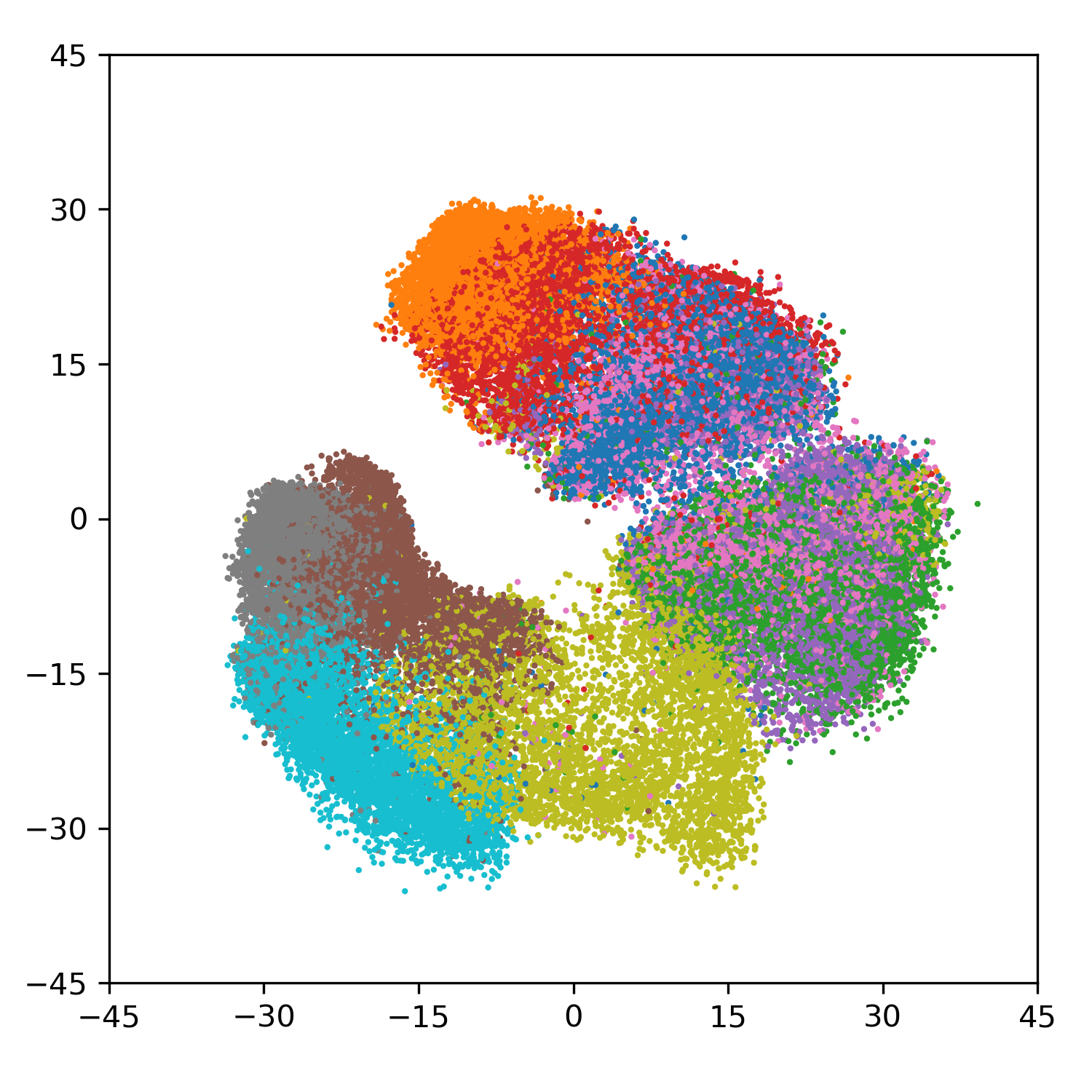}
        \caption*{C+E \\ (L-Isomap, $\alpha=1$)}
    \end{subfigure}\hfill
    \begin{subfigure}[b]{0.18\textwidth}
        \centering
        \includegraphics[width=\textwidth]{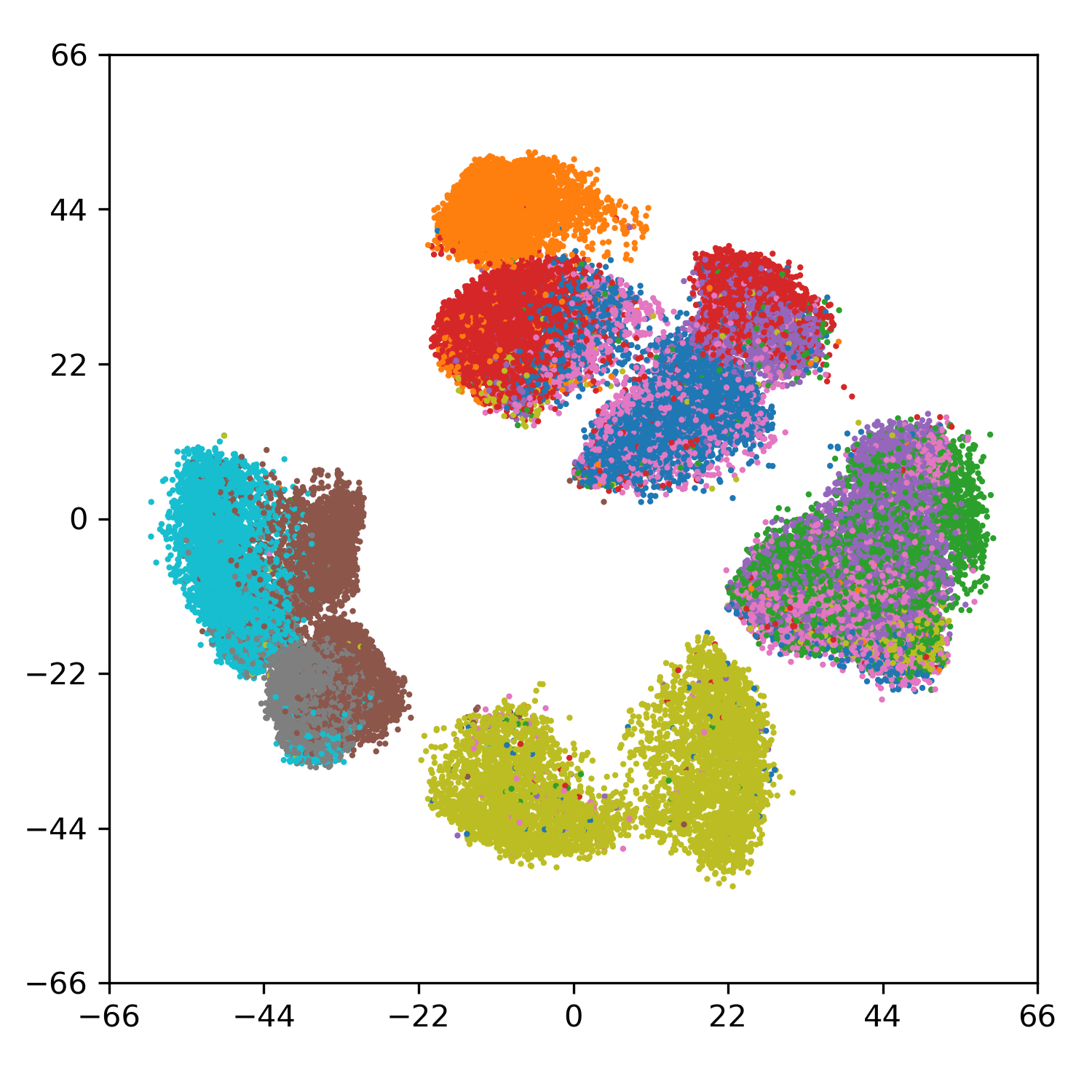}
        \caption*{C+E \\ (L-Isomap, $\alpha=1.70$)}
    \end{subfigure}\hfill
    \begin{subfigure}[b]{0.18\textwidth}
        \centering
        \includegraphics[width=\textwidth]{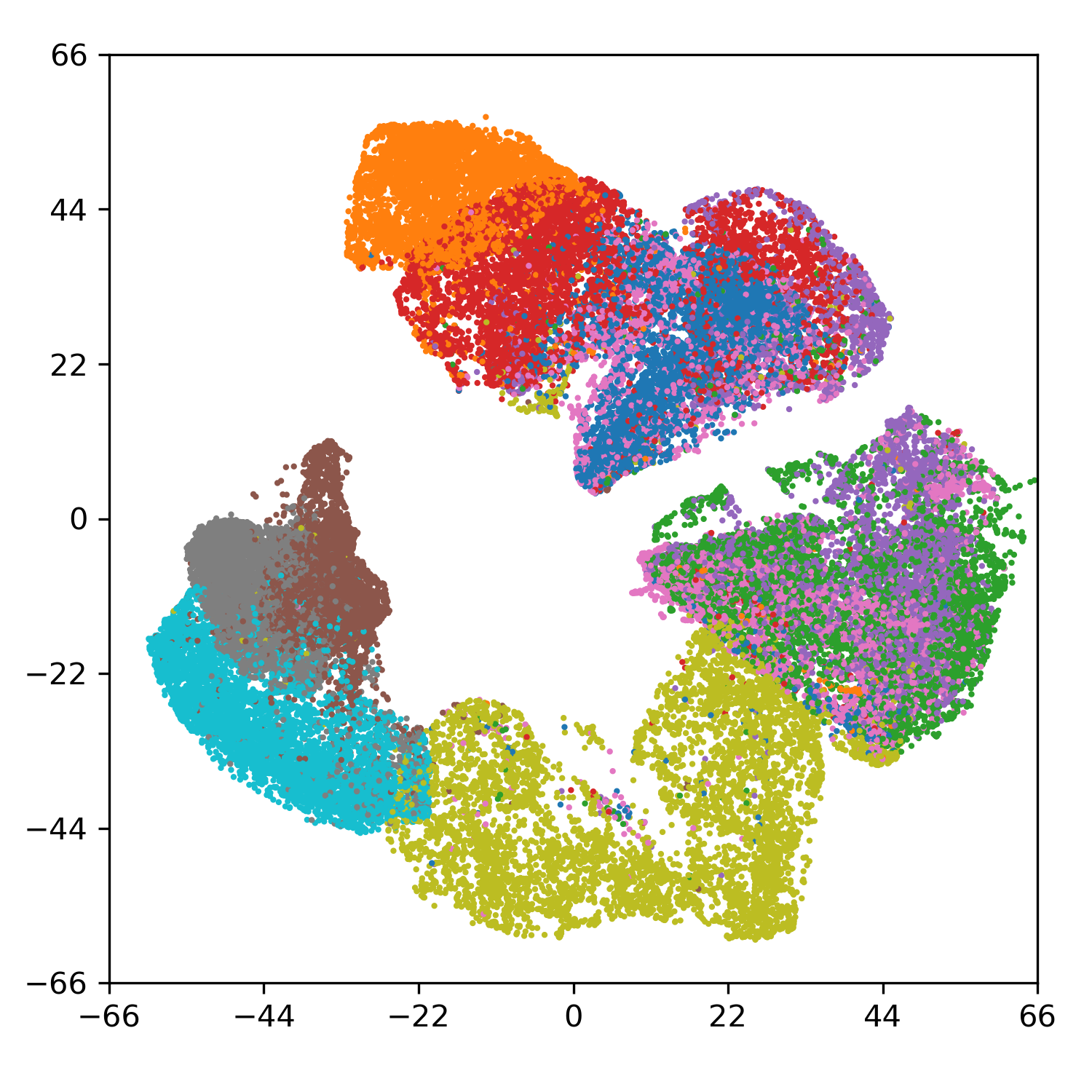}
        \caption*{C+E \\ (TriMap, $\alpha = 1.87$)}
    \end{subfigure}\hfill
    \begin{subfigure}[b]{0.18\textwidth}
        \centering
        \includegraphics[width=\textwidth]{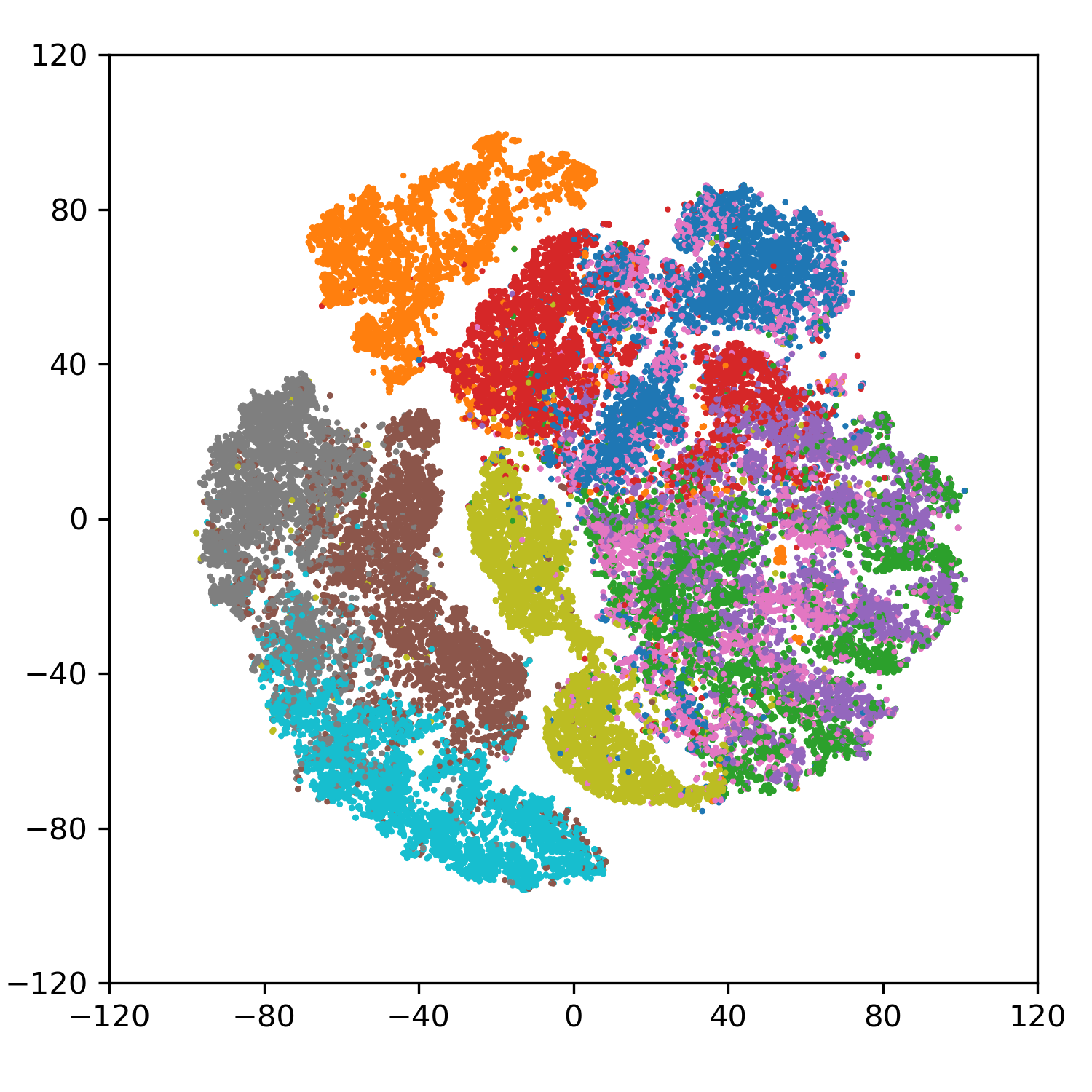}
        \caption*{t-SNE \\  \strut}
    \end{subfigure}\hfill
    \begin{subfigure}[b]{0.18\textwidth}
     \centering
        \includegraphics[width=\textwidth]{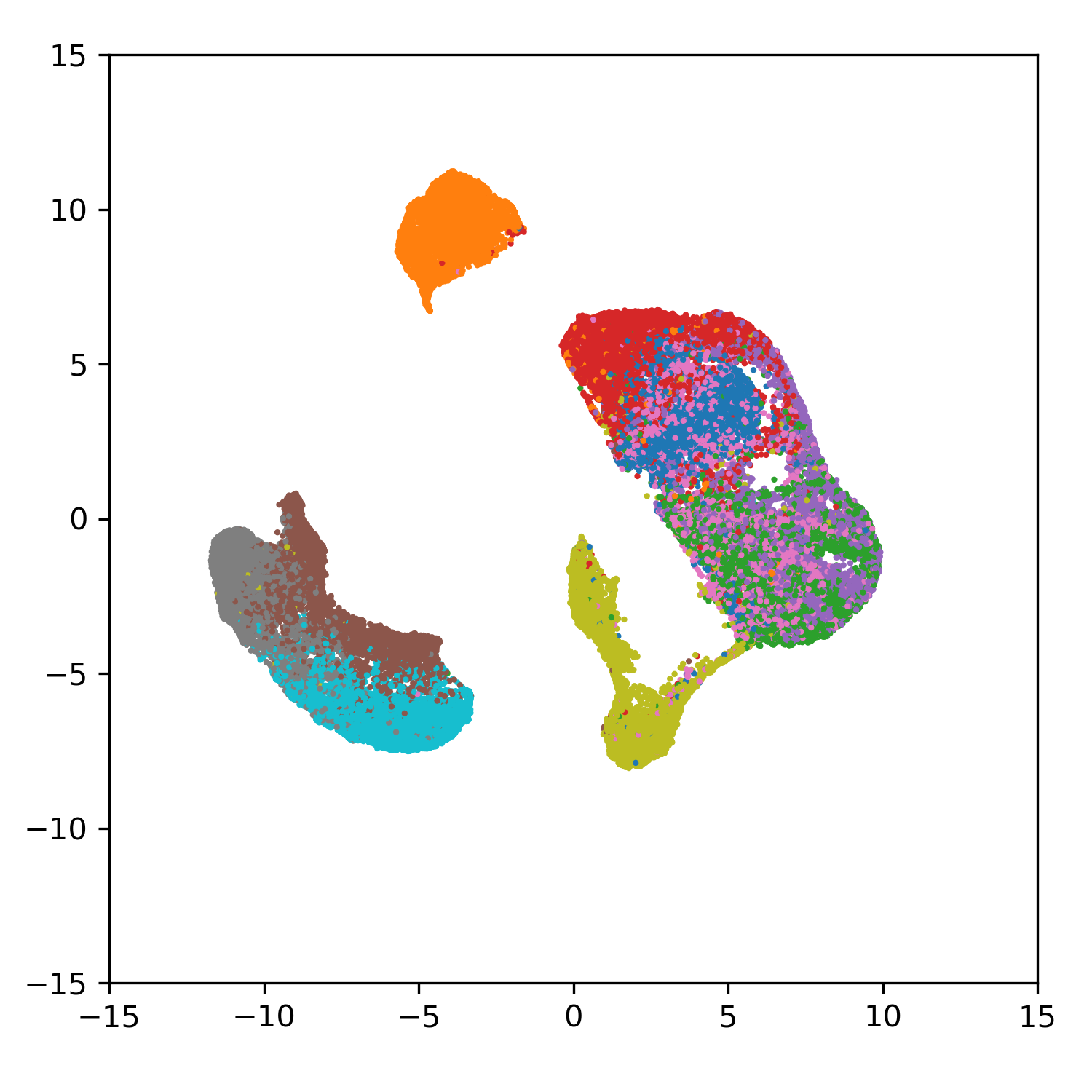}
        \caption*{UMAP \\  \strut}
    \end{subfigure}
    
% ----- Row 2 -----
    \begin{subfigure}[c]{0.18\textwidth}
            \centering
        \includegraphics[width=\textwidth]{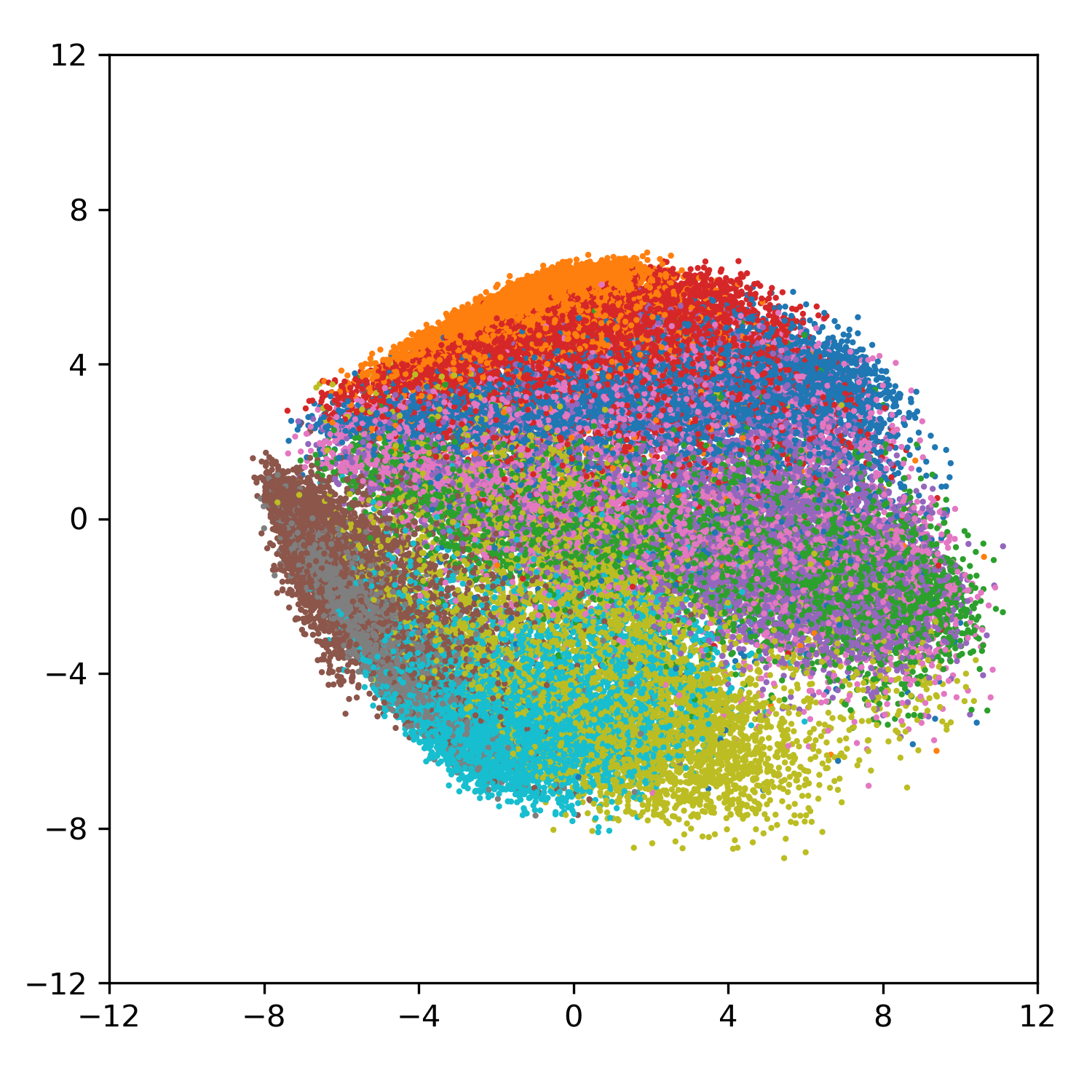}
        \caption*{PCA}
    \end{subfigure}
      \hfill
    \begin{subfigure}[c]{0.18\textwidth}
        \centering
        \includegraphics[width=\textwidth]{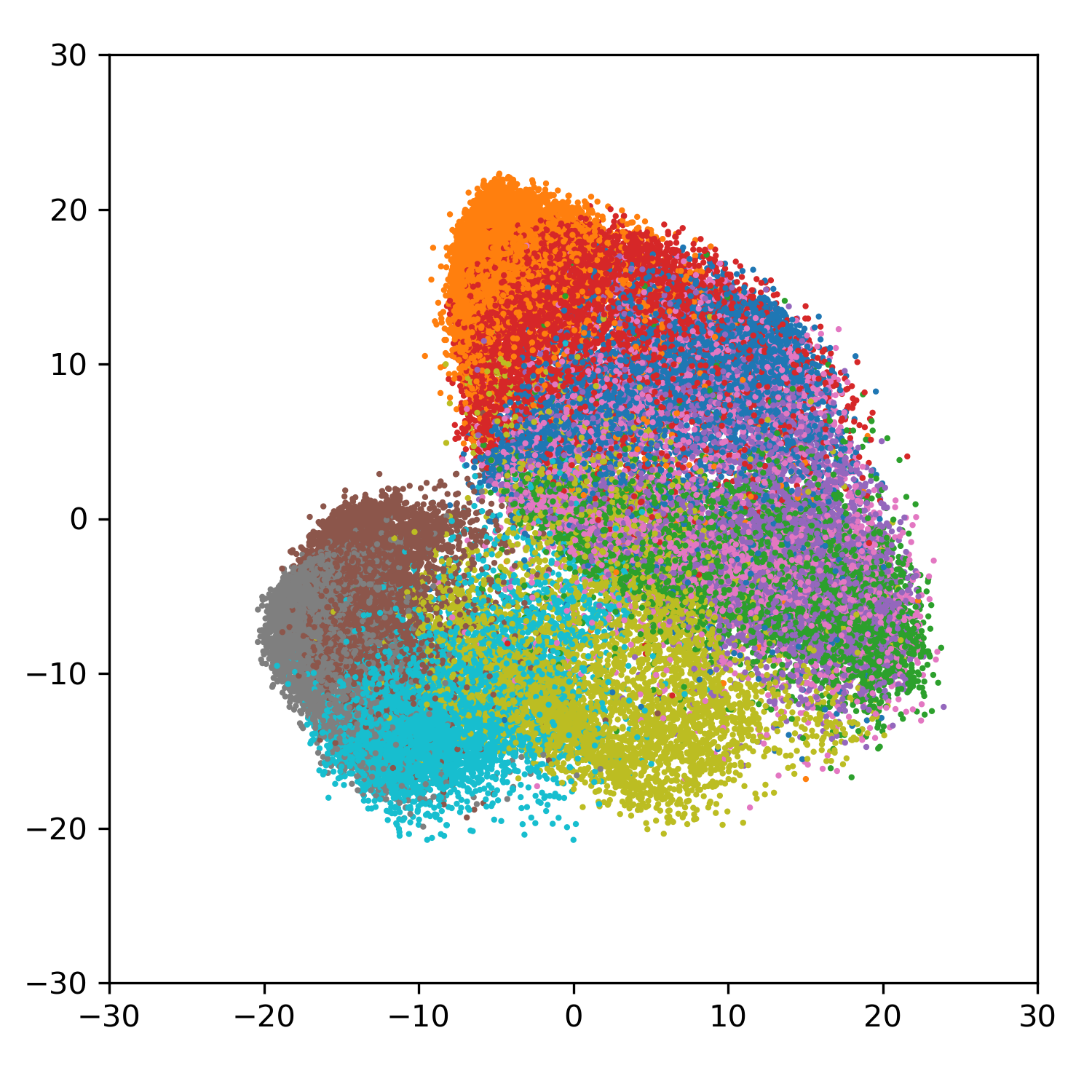}
        \caption*{L-Isomap}
    \end{subfigure}
      \hfill
    \begin{subfigure}[c]{0.18\textwidth}
        \centering
        \includegraphics[width=\textwidth]{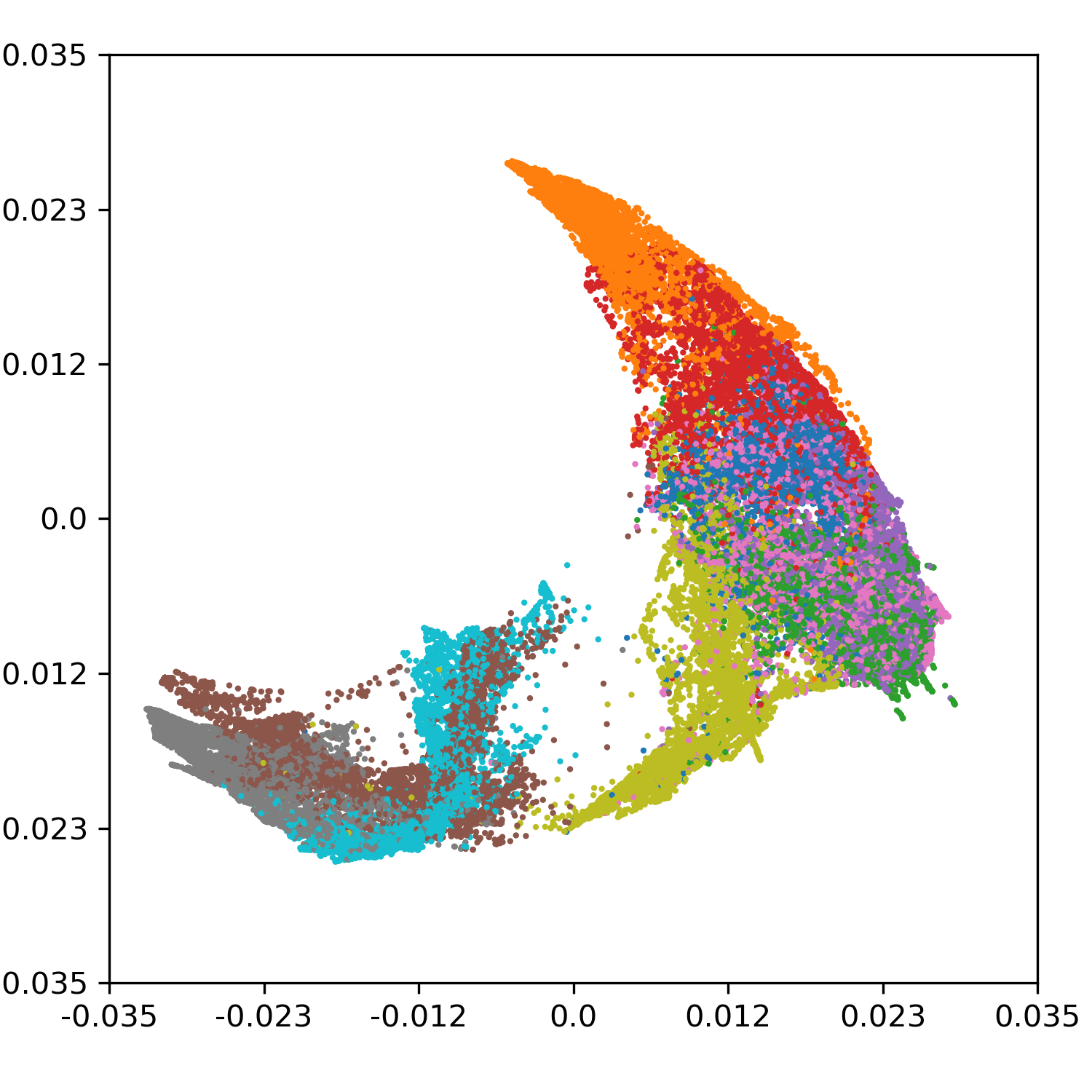}
        \caption*{PHATE}
    \end{subfigure}
     \hfill
    \begin{subfigure}[c]{0.18\textwidth}
        \centering
        \includegraphics[width=\textwidth]{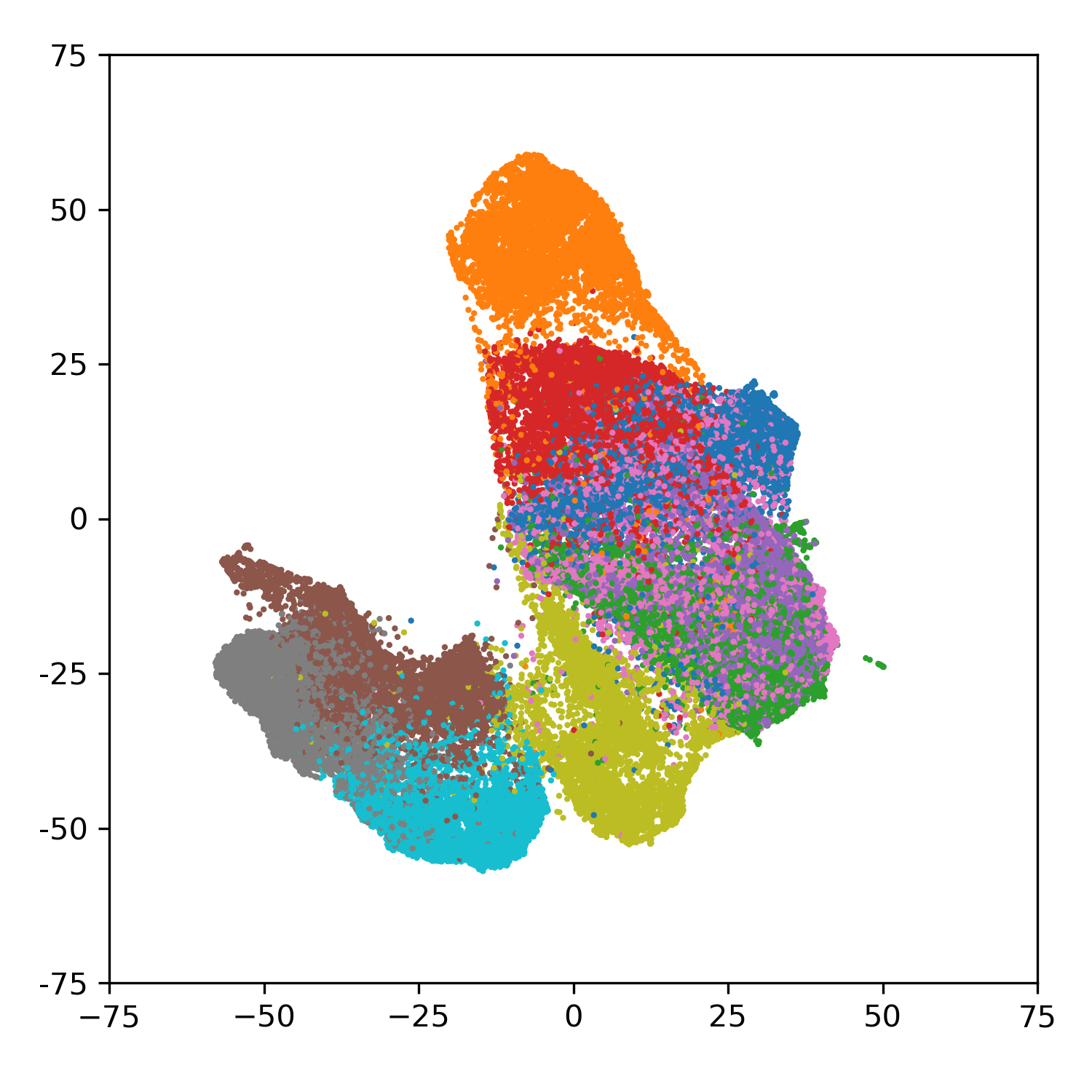}
        \caption*{TriMap}
    \end{subfigure}
     \hfill
    \begin{subfigure}[c]{0.18\textwidth}
        \centering
        \includegraphics[width=.4\textwidth]{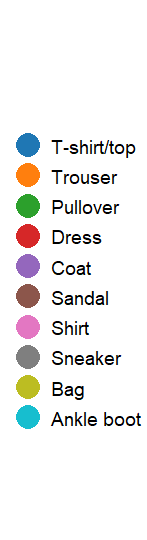}
         \caption*{}
    \end{subfigure}
    \caption{Embedding of Fashion MNIST dataset ($n = 60,000$). Points are colored according to their original labels. These labels were not used by the embedding methods.}
    \label{fig:fmnist}
\end{figure}

\begin{table}[!ht]
\centering
\footnotesize
\setlength{\tabcolsep}{2pt}
\begin{tabular}{llccccccccccc}
\toprule
\textbf{Method} & \textbf{Params} 
& \multicolumn{4}{c}{\textbf{Local}} 
& \multicolumn{7}{c}{\textbf{Global}} \\
\cmidrule(lr){3-6} \cmidrule(lr){7-13}
& 
& W-S ${\scriptstyle \uparrow}$ 
& W-NS ${\scriptstyle \downarrow}$ 
& W-SNS ${\scriptstyle \downarrow}$ 
& KNN ${\scriptstyle \uparrow}$
& B-S ${\scriptstyle \uparrow}$  
& T-S ${\scriptstyle \uparrow}$  
& B-NS ${\scriptstyle \downarrow}$ 
& T-NS ${\scriptstyle \downarrow}$ 
& B-SNS ${\scriptstyle \downarrow}$ 
& T-SNS ${\scriptstyle \downarrow}$  
& CP ${\scriptstyle \uparrow}$
 \\
\midrule

C+E (L-Isomap) & $\alpha = 1$ &
$0.79$ & $0.42$ & $\mathbf{0.31}$ & $0.30$ & $0.73$ & $\mathbf{0.85}$ & $\mathbf{0.24}$ & $\mathbf{0.24}$ & $\mathbf{0.20}$ & $\mathbf{0.24}$ & $0.88$ \\

C+E  (L-Isomap) & $\alpha = 1.70$ &
$0.75$ & $0.47$ & $0.41$ & $0.39$ & $0.65$ & $0.80$ & $0.62$ & $0.67$ & $0.22$ & $0.27$ & $\mathbf{0.91}$ \\

C+E  (TriMap) & $\alpha = 1.87$ &
$0.70$ & $0.52$ & $0.39$ & $0.37$ & $0.64$ & $0.84$ & $0.74$ & $0.79$ & $0.22$ & $0.25$ & $0.88$\\

t-SNE & $u = 30$ &
$0.77$ & $1.34$ & $0.33$ & $\mathbf{0.46}$ & $0.57$ & $0.74$ & $1.62$ & $1.67$ & $0.25$ & $0.29$ & $0.80$
 \\

UMAP & $q = 15$ &
$\mathbf{0.80}$ & $0.85$ & $0.37$ & $0.43$ & $0.56$ & $0.75$ & $0.72$ & $0.72$ & $0.22$ & $0.31$ & $0.83$\\

PCA & - &
$0.63$ & $0.81$ & $0.40$ & $0.18$ & $0.63$ & $0.74$ & $0.79$ & $0.79$ & $0.27$ & $0.30$ & $0.84$ \\

L-Isomap & $q = 15$  &
$0.72$ & $0.59$ & $0.34$ & $0.21$ & $\mathbf{0.74}$ & $0.84$ & $0.47$ & $0.47$ & $0.21$ & $0.25$ & $0.84$
 \\

PHATE & $q = 5$ &
$0.71$ & $1.00$ & $0.38$ & $0.27$ & $0.55$ & $0.74$ & $1.00$ & $1.00$ & $0.25$ & $0.36$ & $0.80$ \\

TriMap & - &
$0.76$ & $\mathbf{0.37}$ & $0.33$ & $0.36$ & $0.63$ & $0.77$ & $0.45$ & $0.51$ & $0.22$ & $0.30$ & $0.80$\\

\bottomrule
\end{tabular}
\caption{Evaluation of the Fashion MNIST dataset against geodesic distances for $5000$ points selected independently of the points used in alignment. Metrics are reported for within class (W), between class (B), and total (T). S denotes Spearman correlation, NS normalized stress, SNS scale-normalized stress, CP class preservation, and KNN the $30$-NN recall. Arrows indicate whether higher ($\uparrow$) or lower ($\downarrow$) values are better.}
\label{tab:fmnist}
\end{table}

\begin{figure}[ht!]
    \centering
    \begin{subfigure}[c]{0.4\textwidth}
        \centering
        \includegraphics[width=\textwidth]{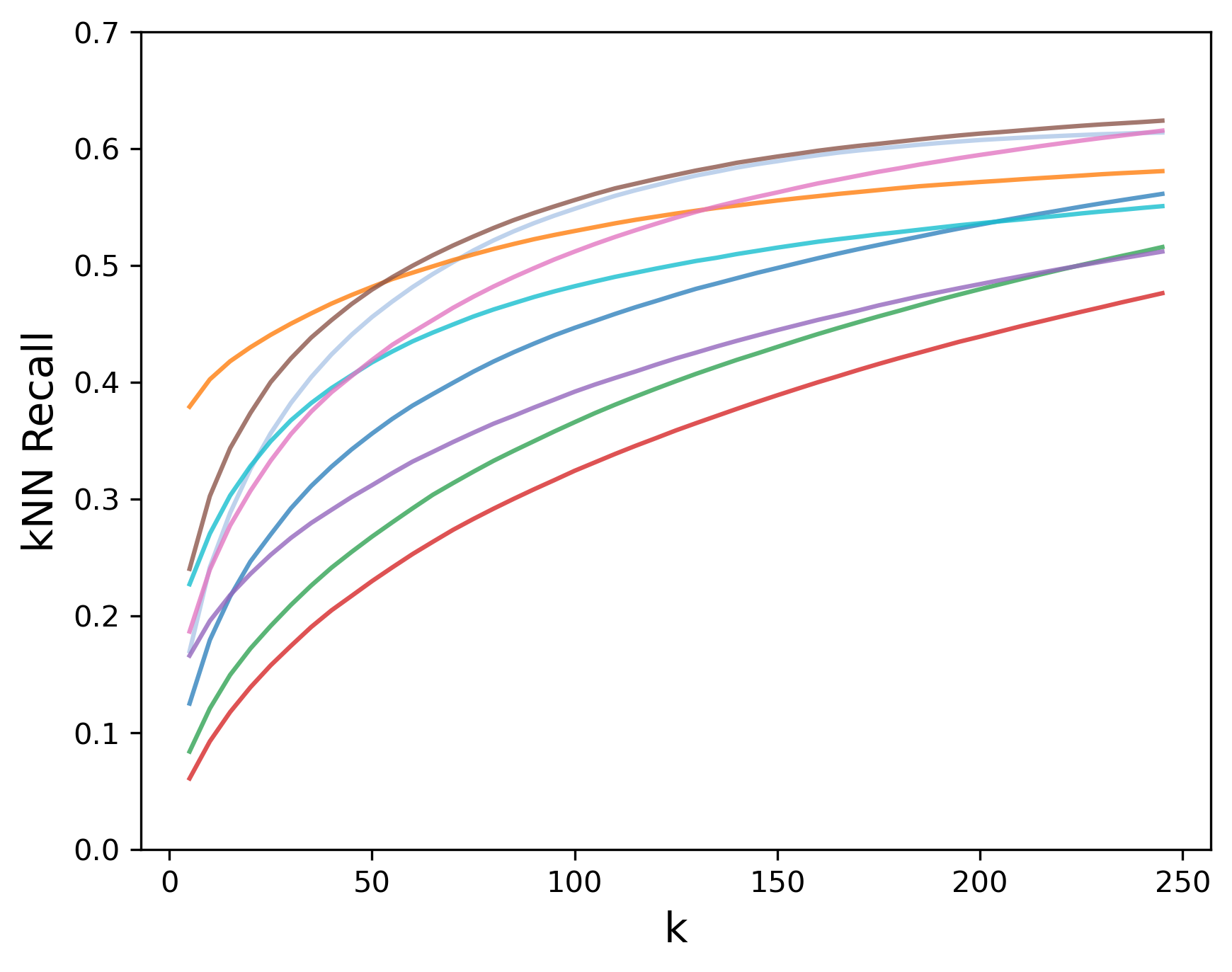}
    \end{subfigure}
    \hspace{0.03\textwidth}
    \begin{subfigure}[c]{0.2\textwidth}
        \centering
        \includegraphics[width=\textwidth]{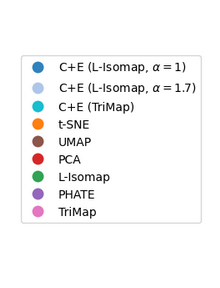}
    \end{subfigure}
    \caption{$k$NN recall for the FMNIST dataset for various values of $k$ for all methods compared.}
    \label{fig:knn_fmnist}
\end{figure}

As an additional example, we use the Fashion MNIST dataset \cite{xiao2017}.  We first flatten the images and then use PCA to reduce the dimension of each vector to $50$. Then, clustering is done via the Leiden algorithm ($q = 15, \gamma = 315$) and ten clusters are obtained with Rand index of $0.899$ with respect to the ground truth labels. We embed using both L-Isomap and TriMap and align cluster-level embeddings to preserve geodesic distances.  

C+E organizes the clusters along an arc with a similar arrangement to the L-Isomap embedding, producing an embedding that preserves global structure, while also providing some local information. In the bottom left half of the C+E embedding (L-Isomap, $\alpha = 1.70$), the shoes are grouped together and the bags are grouped together. The clustering step of C+E forms two shoe clusters: one consisting of flat shoes (sneakers and flat sandals), and one consisting of shoes with high heels (ankle boots and sandals with a heel).  It also forms two bag clusters, distinguished by whether the bag handle is visible or not. In the embedding, these two distinct bag clusters are apparent. Additionally, in the embedding of each shoe cluster there is some organization by shoe type and the two clusters are aligned so the sandals in different clusters are near each other. In the upper right half of the C+E embedding, there appear to be two groups containing tops. The first contains pullovers, coats, and long-sleeved shirts. The second contains t-shirts and short-sleeved shirts, which are embedded close to dresses and trousers. Compared to the long-sleeved shirts, the images of these objects all appear as skinny and elongated, which may explain their proximity in the embedding. 

In contrast, UMAP clearly separates the trousers from the dresses and tops, though the bags are also embedded close to the dresses and tops. The global metrics generally favor the C+E embedding over the UMAP embedding, while locally C+E preserves distances better but has worse $k$NN recall for small values of $k$. Likewise, the t-SNE embedding contains distinct clusters, but the spacing between all clusters is approximately equal and the global metrics are poor, while the $k$NN recall is superior to any other method for small values of $k$. All other methods create an embedding that forms a more continuous spectrum and do not capture the cluster structure well. Both globally and locally, these other methods tend to perform worse than C+E.

\end{document}